\documentclass{article}

\usepackage[preprint]{neurips_2026}

\usepackage[utf8]{inputenc}
\usepackage[T1]{fontenc}
\usepackage{url}
\usepackage{booktabs}
\usepackage{amsfonts}
\usepackage{nicefrac}
\usepackage{microtype}
\usepackage[dvipsnames]{xcolor}
\usepackage{graphicx}
\usepackage{wrapfig}

\usepackage{amsmath}
\usepackage{amssymb}
\usepackage{mathtools}
\usepackage{amsthm}

\theoremstyle{plain}

\theoremstyle{definition}

\theoremstyle{remark}

\usepackage{algorithm}
\usepackage{algorithmic}
\usepackage{multirow}
\usepackage{longtable}
\usepackage{subcaption}
\usepackage{tcolorbox}
\tcbuselibrary{listings,skins,breakable}
\usepackage{enumitem}
\usepackage{tikz}
\usetikzlibrary{calc,positioning}
\usepackage{dirtytalk}
\usepackage{xspace}

\newcommand{\rw}{\ensuremath{\textsc{RW}}\xspace}
\newcommand{\sw}{\ensuremath{\textsc{SW}}\xspace}
\newcommand{\dpo}{\ensuremath{\textsc{DPO}}\xspace}
\newcommand{\slfull}{\ensuremath{\textsc{SL}}\xspace}
\newcommand{\rlfull}{\ensuremath{\textsc{R}}\xspace}

\usepackage[bookmarks=false]{hyperref}
\hypersetup{
  linktocpage=true,
  colorlinks=true,
  citecolor=Green,
  linkcolor=Green,
  urlcolor=Green
}
\usepackage[capitalize,noabbrev]{cleveref}

\newtcolorbox{promptbox}[1][]{
  colback=gray!5,
  colframe=blue!40,
  fonttitle=\bfseries\color{blue!60!black},
  boxrule=0.8pt,
  arc=2pt,
  left=8pt,
  right=8pt,
  top=8pt,
  bottom=8pt,
  breakable,
  #1
}

\usepackage[textsize=tiny,disable]{todonotes}

\newcommand{\dataset}{\mathcal{D}}

\newcommand{\sem}[2]{#1{\scriptsize\,$\pm$#2}}
\newcommand{\sembf}[2]{\textbf{#1}{\scriptsize\,$\pm$#2}}
\newcommand{\semgr}[2]{\textcolor{gray}{#1{\scriptsize\,$\pm$#2}}}

\title{Agentic Planning with Reasoning for Image Styling via Offline RL}

\author{%
  Subhojyoti Mukherjee \And Stefano Petrangeli \And Branislav Kveton \AND
  Trung Bui \And Franck Dernoncourt \And Arko Mukherjee \AND
  Adobe Research, San Jose, CA, USA \\
  Project page: \url{https://subhojyoti.github.io/image-agent/}
}

\begin{document}

\maketitle

% Teaser figure (single-column layout, before abstract)
% Teaser figure showing qualitative comparisons (non-float version for first page).
% Strips carry 10 column labels and the per-sample prompt baked in.
% Columns (left->right): Original | Baseline | Edit-Only | Standard | RL | DPO | RW | SW | GPT-4o | Gemini 2.5
\begin{center}
  {\Large \textbf{Qualitative Results: Agentic Planning with Reasoning for Image Styling}}
\end{center}
\vspace{0.1em}

\begin{tikzpicture}
  % Row 1: Simple Text-4B (\sw winner) -- regular_building
  \node[anchor=north west,inner sep=0] (row1) at (0,0) {
    \includegraphics[width=\textwidth,height=0.115\textheight,keepaspectratio]{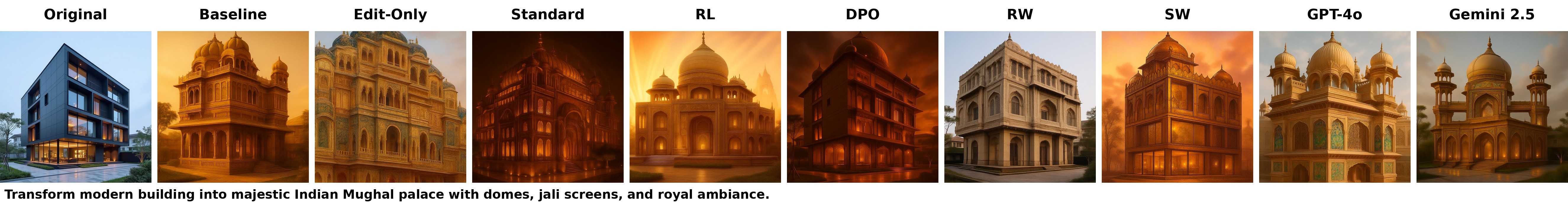}
  };

  % Row 2: Regular Text-8B (\rw/\sw winner) -- mountain_rocky
  \node[anchor=north west,inner sep=0] (row2) at (row1.south west) {
    \includegraphics[width=\textwidth,height=0.115\textheight,keepaspectratio]{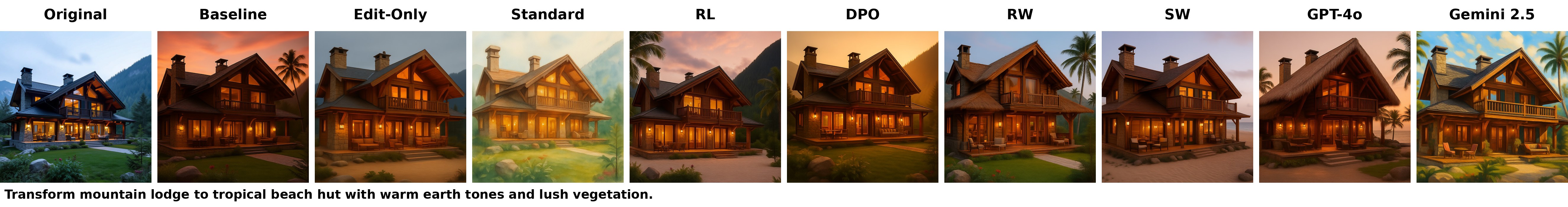}
  };

  % Row 3: Regular Text-4B (\rw/\sw winner) -- traditional_bedroom
  \node[anchor=north west,inner sep=0] (row3) at (row2.south west) {
    \includegraphics[width=\textwidth,height=0.115\textheight,keepaspectratio]{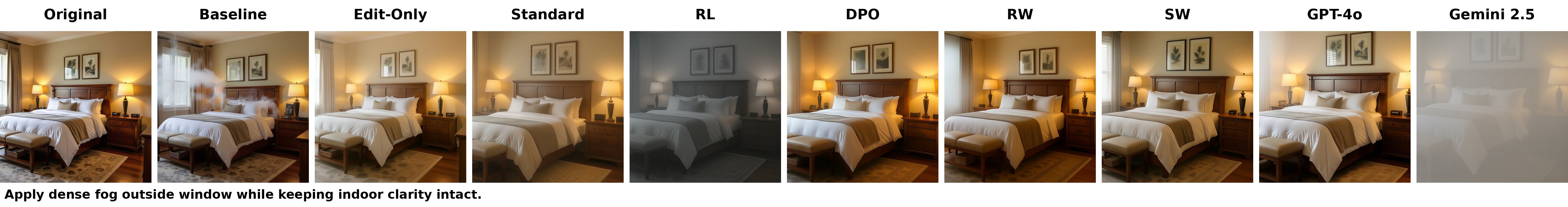}
  };

  % Row 4: Regular Vision-8B (\rw winner) -- urban_city
  \node[anchor=north west,inner sep=0] (row4) at (row3.south west) {
    \includegraphics[width=\textwidth,height=0.115\textheight,keepaspectratio]{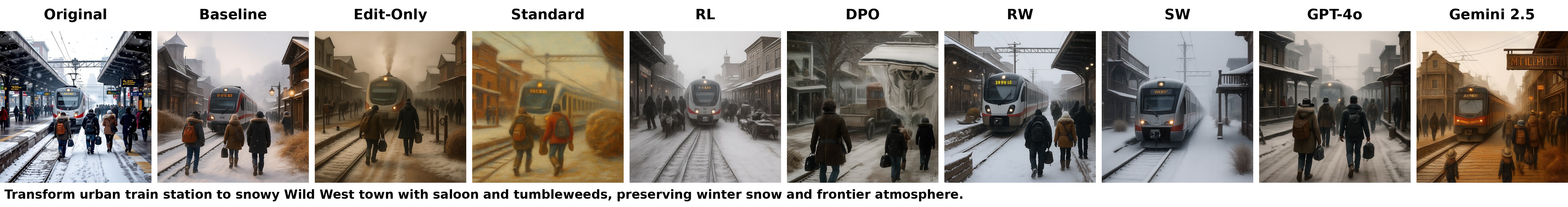}
  };

  % Continuous red columns across all 4 rows.
  % 10 equal columns -> column k spans (k-1)/10..k/10 of \textwidth.
  % RW is column 7 ([0.6, 0.7]); SW is column 8 ([0.7, 0.8]).
  \draw[red,line width=2pt]
    ([xshift=0.6\textwidth]row1.north west) rectangle
    ([xshift=0.7\textwidth]row4.south west);
  \draw[red,line width=2pt]
    ([xshift=0.7\textwidth]row1.north west) rectangle
    ([xshift=0.8\textwidth]row4.south west);
\end{tikzpicture}

\vspace{0.1em}

{\footnotesize \textbf{Figure 1: Agentic planning with reasoning for image styling via offline RL.}
We train a compact vision-language planner (Qwen3-VL 4B/8B) that decomposes styling goals into
sequences of tool calls (e.g., \textit{time\_of\_day}, \textit{artistic\_medium},
\textit{mood\_lighting}) with explicit reasoning. Each row is a 10-way comparison with column
labels and the user prompt baked in: Original, Baseline (B), Edit-Only (E), Standard (S),
\rlfull, \dpo, Reward-Weighted (\rw), Standardized Reward-Weighted (\sw), GPT-4o Planner, and
Gemini 2.5 Planner (large closed-source). Samples are chosen as those where our \rw/\sw
planners beat the strongest non-\rw/\sw method by the widest per-sample judge margin.
\textbf{Row 1:} Building style transfer (Simple, Text-4B)---\sw delivers the cleanest architectural
restyling, outperforming both supervised and closed-source planners.
\textbf{Row 2:} Mountain rocky terrain transformation (Regular, Text-8B)---\rw/\sw achieve richer
texture and atmospheric depth where other planners produce flat or inconsistent results.
\textbf{Row 3:} Traditional bedroom restyling (Regular, Text-4B)---\rw/\sw apply the correct
period-accurate furnishing and lighting while other planners miss compositional details.
\textbf{Row 4:} Urban city scene transformation (Regular, Vision-8B)---\rw produces the most
coherent structural and lighting changes. Red boxes highlight our \rw and
\sw columns. Our compact open-source planners outperform the GPT-4o and Gemini 2.5 zero-shot
baselines with orders of magnitude fewer parameters.}

\begin{abstract}
Direct prompt-based editing often fails on complex transformations because vague and subjective prompts often require nuanced understanding of what should be changed in the image. Our core intuition is that leveraging compositional image editing tools rather than direct prompting profits from structured agent-level planning with explicit reasoning, leading to better results. This structured planning framework enables efficient offline RL post-training on quality-scored trajectories to improve performance. We present a tool-based agentic RL post-training framework that addresses this through structured planning with chain-of-thought reasoning. Our key contributions include: (1) A tool-based agentic planning methodology that combines a compositional library of orthogonal primitive transformations, structured context representation, and explicit per-step reasoning to decompose complex styling into interpretable tool sequences. (2) A synthetic data generation pipeline producing three large-scale datasets (each ${\sim}10$K trajectories) with reasoning chains, plans, and quality scores---necessary because no existing datasets provide explicit tool-based styling supervision. Our datasets are publicly available at \url{https://huggingface.co/datasets/subhojyoti1990/image-agent-styling}, and the project page is at \url{https://subhojyoti.github.io/image-agent/}. (3) An empirical study of offline RL training methods for vision-language tool-use planning, adapting Reward-Weighted (RW) and Standardized Reward-Weighted (SW) methods~\citep{mukherjee25offline}---previously validated only in text-only settings---to this domain for the first time, consistently improving over the Edit-Only baseline in visual quality and instruction following. (4) Comprehensive evaluation across 4B and 8B parameter Qwen3-VL models showing that our methods outperform other baselines in the majority of the compositional tasks, backed up by human evaluations on both the synthetic data and final results.
Our work demonstrates that structured planning with reward-aware training enables models to produce higher-quality images that better follow user instructions compared to direct editing approaches. Comprehensive evaluation on Qwen3-VL 4B/8B shows our compact open-source planners outperform GPT-4o and Gemini 2.5 zero-shot baseline planners on the majority of tasks, confirmed by human evaluation.
\end{abstract}

\vspace*{-1.5em}
\section{Introduction}
\label{sec:intro}
\vspace*{-0.5em}
The ability to transform images according to high-level aesthetic intents—changing a scene from day to night, summer to autumn, modern to Victorian, or photorealistic to painterly—is fundamental to creative workflows across industries including entertainment, advertising, and design. This problem has a rich history predating modern generative AI: classical image-to-image translation methods \citep{isola2017image}, neural style transfer \citep{gatys2016image}, and cycle-consistent adversarial networks \citep{zhu2017unpaired} pioneered techniques for cross-domain visual transformations. While these early approaches demonstrated the potential for automated image styling, they were often limited to specific transformations or required paired training data. Recent advances in vision-language foundation models such as DALL-E \citep{ramesh2021zero,ramesh2022hierarchical}, Stable Diffusion \citep{rombach2022high}, and Qwen3-VL have democratized image editing through natural language prompts \citep{bai2025qwen2, bai2023qwenvl, liu2023visual}, enabling users to specify desired transformations in plain text without domain-specific expertise.

Current state-of-the-art image styling relies predominantly on direct prompt-based editing, where users provide natural language instructions to foundation models that generate or modify images. Any modern image editing model can perform styling tasks—commercial systems like Midjourney, and DALL-E 3 \citep{betker2023improving}, as well as open-source alternatives like Stable Diffusion and its derivatives. Recent specialized methods have emerged for specific use cases: StyleBooth \citep{han2025stylebooth} enables personalized style transfer from reference images, while Styleshot \citep{gao2024styleshot} focuses on few-shot style adaptation. These approaches share a common paradigm: direct mapping from natural language prompts to styled images, often in a single forward pass or iterative refinement of the same prompt.

However, direct prompt-based editing faces a fundamental limitation: prompts are often imprecise and fail on complex multi-dimensional transformations that require coordinating changes across multiple visual attributes. Consider the instruction "Transform to a rocky mountain landscape, preserving the foreground scene structure and natural lighting." This seemingly simple request requires coordinating terrain restyling, atmospheric depth, seasonal foliage, and preservation constraints (maintaining compositional structure). As shown in Figure 1 Row 2, direct editing (Edit-Only baseline) produces inconsistent results with poor instruction adherence, misaligned textures, and structural artifacts. This failure stems from the ambiguity inherent in natural language: a single prompt does not explicitly specify which visual dimensions to modify, in what order, or how to balance competing requirements. The gap between user intent and model interpretation leads to results that often deviate from human preferences.

We address this challenge through a tool-based agentic RL post-training framework with structured compositional planning and synthetic data generation. The framework combines: (1) a compositional library of 10 orthogonal primitive tools (e.g., \texttt{time\_of\_day}, \texttt{artistic\_medium}) enabling complex transformations via 2–5 step sequences; (2) explicit structured context encoding the image's current visual state across all tool dimensions, grounding planning in concrete attributes; and (3) per-step chain-of-thought reasoning explaining each tool choice, improving coherence and interpretability. As our primary empirical contribution, we study the first application of \textit{Reward-Weighted (\rw)} and \textit{Standardized Reward-Weighted (\sw)} training methods---formalized by \citet{mukherjee25offline} for text-only settings---to vision-language tool-use planning: \rw weights trajectories by quality score through true per-sample weighted loss; \sw extends this with z-score normalization for stable training across datasets with varying reward distributions, and both outperform the direct prompt-based Edit-Only baseline. 
% Both consistently outperform the direct prompt-based Edit-Only baseline.

We adopt an offline RL approach \citep{lange12batch,levine20offline}: decoupling data generation from training enables human validation before learning (3,000 samples, 77\% pass rate), one-time teacher inference cost, releasable fixed datasets, and fair multi-method comparison on identical trajectories. Despite not adapting trajectories to an improving policy, offline RL proves highly effective: our 4B/8B models outperform both the GPT-4o and Gemini 2.5 zero-shot baselines on the majority of configurations, with human validation confirming method rankings (Section~\ref{sec:human_val_synthetic_data}). We train planners in text-only and vision-language modalities with the \textit{image editor frozen}.

Our main contributions are:

\noindent\textbf{(1) Tool-Based Agentic RL Framework:} A complete pipeline with compositional tool libraries, structured document representations, per-step chain-of-thought reasoning, and synthetic data generation---a blueprint for planning agents in creative domains.

\noindent\textbf{(2) Large-Scale Synthetic Datasets:} Three datasets totaling 30K trajectories: Simple (10K, 1-2 step edits), Regular (10K, 3-5 step, 10 interior themes), and Complex (10K, 3-5 step, 83 diverse themes), each with structured context, action plans, reasoning chains, and quality scores.\footnote{\href{https://huggingface.co/datasets/subhojyoti1990/image-agent-styling}{Dataset link}}

\noindent\textbf{(3) \rw and \sw for Visual Planning:} Adapting reward-weighted methods~\citep{mukherjee25offline,peters07reinforcement} to vision-language tool-use for the first time, we show that per-sample quality weighting is crucial for compositional planning. \rw weights each trajectory by its quality score; \sw extends this with z-score normalization for stable training. Both consistently outperform the Edit-Only baseline---the first such demonstration in a visual agentic planning domain.

\noindent\textbf{(4) Comprehensive Empirical Analysis:} Multi-judge evaluation (GPT-5.4, Gemini 3.1, Claude Opus 4.6) across all model configurations reveals reward-weighted approaches show particular strength on complex compositional tasks, with our compact 4B/8B planners consistently outperforming both the GPT-4o and Gemini 2.5 zero-shot baselines.

Sections~\ref{sec:problem_setup}--\ref{sec:algorithm} detail problem setup, data generation, and training algorithms; Section~\ref{sec:experiments} presents experimental results; related work is deferred to Appendix~\ref{sec:related_work} for space.
\vspace*{-1em}
\section{Problem Setup}
% \vspace*{-1em}
\label{sec:problem_setup}
\vspace*{-0.5em}
We formulate image styling as a sequential decision-making problem where an agent learns to compose tools from a compositional tool library. Given an input image, a user's natural language prompt, and a structured representation of the image's current visual state, the agent must produce a sequence of tool calls that transform the image to match the desired aesthetic. 
We use structured representation to ground the planning process in explicit visual attributes, enabling the model to reason about specific dimensions (e.g., "current lighting is harsh midday, need warm golden-hour") rather than relying solely on implicit visual understanding.

\vspace*{-0.5em}
\subsection{Four-Stage Structured Editing Pipeline}

Classical image editing maps user's editing goal $e_i$ (natural language prompt) and base image $I_i$ directly to edited image $\hat{I}_i$: $e_i, I_i \to \hat{I}_i$. However, vague prompts often produce poor results. For example, {\color{blue}``Transform this to a Renaissance oil painting''} fails because the prompt doesn't specify which time period (1400s vs 1500s vs 1600s), color palette (warm earth tones?), artistic techniques (chiaroscuro lighting?), or which elements to preserve. Direct editing $e_i, I_i \to \hat{I}_i$ with vague prompts produces inconsistent results. Our goal is to replace vague $e_i$ with precise $\hat{e}_i$ that yields better $\hat{I}_i$. We address this through structured editing with four stages.

\textbf{Stage 1 (Extract Structured Context):} First, we extract a structured representation of the image's visual state. Formally, $e_i, I_i \to c_i$ where $c_i$ is plain text describing the image's current visual state across 10 dimensions: location (urban city), architecture (modern glass), time period (contemporary 2020s), time of day (midday harsh lighting), season (summer), weather (clear), mood (neutral documentary), color grading (natural desaturated), artistic medium (realistic photograph), atmospheric effects (clear visibility).

\paragraph{Compositional Tool Library:} Before detailing the planning stage, we describe our compositional tool library. Our tool library contains parameterized transformations across 10 orthogonal dimensions that is based on the visual state contexts discussed above: location\_setting, architecture\_style, time\_period, time\_of\_day, season, weather, mood\_lighting, color\_grading, artistic\_medium, atmospheric\_effects. 
% \todotb{Duplicate with 10 dimensions mentioned in the previous paragraph. Consider reuse the info already listed.} 
These dimensions were selected to cover the primary controllable visual attributes in modern text-to-image models while maintaining orthogonality \citep{li2019controllable, kazemi2019style, zhang2023adding}. Each dimension controls one visual aspect with minimal interference—for example, \textit{time\_of\_day} affects lighting but not architecture; \textit{season} changes vegetation but not building styles. This orthogonality enables clean composition where effects combine predictably. Complex styling emerges from tool sequences: Figure 1's {\color{blue}"golden-hour winter wonderland with snowfall"} decomposes into $\{\textit{time\_of\_day(golden-hour)}, \textit{season(winter)}$, $ \textit{weather(snowfall)}\}$. See Appendix~\ref{sec:appendix_problem_formulation} for complete tool specifications.

\textbf{Stage 2 (Plan Actions with Reasoning):} Second, we generate an action plan with step-by-step reasoning. Formally, $e_i, c_i \to \{z_{i,j}\}_{j=1}^{m_i}, \{a_{i,j}\}_{j=1}^{m_i}$ generates $m_i$ actions (typically 2-5) where $z_{i,j}$ is the chain-of-thought reasoning and $a_{i,j}$ is action $j$ in trajectory $i$ (symbolic action with parameters). For Renaissance transformation, the model first reasons then acts:
\begin{align*}
  z_{i, 1} &= \text{``Setting Renaissance era establishes historical context ...''}, \hspace*{1em} a_{i, 1} = \textit{time\_period(1500s)} \\
  z_{i, 2} &= \text{``Oil painting introduces characteristic brush strokes ...''}, \hspace*{1em}
  a_{i, 2} = \textit{artistic\_medium(oil-painting)} \\
  z_{i, 3} &= \text{``Warm earth palette matches period pigment chemistry''}, \hspace*{1em}
  a_{i, 3} = \textit{color\_grading(warm-earth-tones)} \\
  z_{i, 4} &= \text{``Dramatic light-dark contrast follows Renaissance ...''}, \hspace*{1em}
  a_{i, 4} = \textit{mood\_lighting(chiaroscuro)}
\end{align*}
%
% \begin{align*}
%   z_{i, 1} &= \text{``Setting Renaissance era establishes historical context for subsequent styling''} \\
%   a_{i, 1} &= \textit{time\_period(1500s)} \\
%   z_{i, 2} &= \text{``Oil painting introduces characteristic brush strokes and layered texture''} \\
%   a_{i, 2} &= \textit{artistic\_medium(oil-painting)} \\
%   z_{i, 3} &= \text{``Warm earth palette matches period pigment chemistry''} \\
%   a_{i, 3} &= \textit{color\_grading(warm-earth-tones)} \\
%   z_{i, 4} &= \text{``Dramatic light-dark contrast follows Renaissance masters' technique''} \\
%   a_{i, 4} &= \textit{mood\_lighting(chiaroscuro)}
% \end{align*}

% \todopet{I am wondering: tools are introduced in the previous stage description. But they are justified and explained here. Shouldn't we bring part of the justification in stage 1?} RESOLVED: Moved tool library description to after Stage 1

\textbf{Stage 3 (Synthesize Precise Instruction):} Third, we synthesize a precise editing instruction. Formally, $e_i, c_i, \{z_{i,j}\}, \{a_{i,j}\} \to \hat{e}_i$ produces the synthesized natural language instruction $\hat{e}_i$: {\color{blue}``Transform this urban photograph into an authentic Renaissance oil painting from the 1500s. Apply oil painting with visible brush strokes and layered texture characteristic of Leonardo and Raphael. Use warm earth-tone palette limited to period-appropriate pigments: ochres, umbers, siennas. Add dramatic chiaroscuro lighting with strong directional illumination. Preserve original composition while transforming surface qualities.''} This explicit instruction is improved $\hat{e}_i$ from our goal statement above.

\textbf{Stage 4 (Render Final Image):} Finally, we render the edited image using a frozen black-box editor. Formally, $\hat{e}_i, I_i \to \hat{I}_i$ using frozen image editor (Qwen-Image-Edit). We keep the editor frozen to focus on planning quality, not editing capability. Each trajectory receives reward score $r_i \in [0, 5]$ assessing trajectory quality.

\paragraph{Our Contribution: Stages 1-3}
Our core contribution spans Stages 1-3: building a pipeline to extract structured context, generate high-quality action plans with explicit reasoning, and synthesize precise editing instructions. Only Stage 4 (image rendering) uses a frozen black-box editor (Qwen-Image-Edit). This design separates the \textit{planning problem} (deciding what and how to change) from the \textit{execution problem} (rendering pixels), enabling efficient training without requiring image generation model training. By focusing on the reasoning and planning capabilities of language models, we can train compact 4B-8B parameter planners that generate high-quality editing instructions for any frozen image editor. This is not glorified prompting: Stage 2 synthesizes a precise instruction from a typed, per-step reasoned action plan rather than the raw user goal, a multi-stage transformation no single prompt can replicate. Empirically, 2-shot ICL \emph{decreases} performance by $-4.3$ points (Appendix~\ref{sec:appendix_icl_ablation}) and GPT-4o/Gemini~2.5 zero-shot planners---receiving the same input---trail our trained models by 10--22 points, confirming the gains come from learned planning structure, not better prompting.
% \todopet{This is I think the part where we need to clarify a bit where our contributions lie? If I understand correctly from today's discussion, we mentioned that stage3 and 4 are actually part of the black box model (stage 3 is debatable). Should we clarify this aspect somehow or is it too hard to come across?} RESOLVED: Added clarification above

%
A complete trajectory $\tau_i$ consists of:
$\tau_i = (e_i, I_i, c_i, \{a_{i,j}\}_{j=1}^{m_i}, \{z_{i,j}\}_{j=1}^{m_i}, \hat{e}_i, \hat{I}_i, r_i)$
The complete dataset is $\mathcal{D} = \{\tau_1, \tau_2, \dots, \tau_n\}$ with $n = 10{,}000$ trajectories per dataset variant, organized into trajectory-level train/validation/test splits (80\%/10\%/10\%).
% \todopet{Does this belong here or in experiment discussion? Because here we are talking about methodology} KEPT: This is methodology, belongs here

% \todosm{Some of the above is already there in 1.1. We also have to talk about per step reasoning and structured context. Read below. Examples provided}

\vspace*{-0.8em}
\section{Synthetic Data Generation}
\label{sec:synthetic_data}
\vspace*{-0.5em}
Training our agentic framework requires trajectories containing structured context extraction, planning with reasoning, instruction synthesis, and reward evaluation. While datasets exist for direct prompt-to-image editing \citep{brooks2023instructpix2pix}, they lack the explicit reasoning chains ($z_{ij}$), structured context ($c_i$), and multi-step plans ($\{a_{ij}\}$) needed to train our approach. We therefore generate synthetic training data using a teacher model. We implement the four-stage framework from Section~\ref{sec:problem_setup} using a teacher-student paradigm: a strong teacher model (Qwen3-VL-8B-Instruct) demonstrates the complete pipeline, generating trajectories that are used to train smaller student models (4B and 8B) via offline RL.
\vspace*{-0.5em}
\subsection{Four-Stage Pipeline}
\vspace*{-0.5em}
We generate trajectories using Qwen3-VL-8B-Instruct as the teacher model, HiDream-I1-Dev for image generation, and Qwen-Image-Edit for image editing. 
% \Cref{fig:img-agent} details the complete pipeline. \todopet{check section numbering, as 3.1 is the current section}
%
\setcounter{figure}{1}
\begin{wrapfigure}{r}{0.5\textwidth}
    \vspace*{-1.0em}
    \centering
    \includegraphics[width=\linewidth]{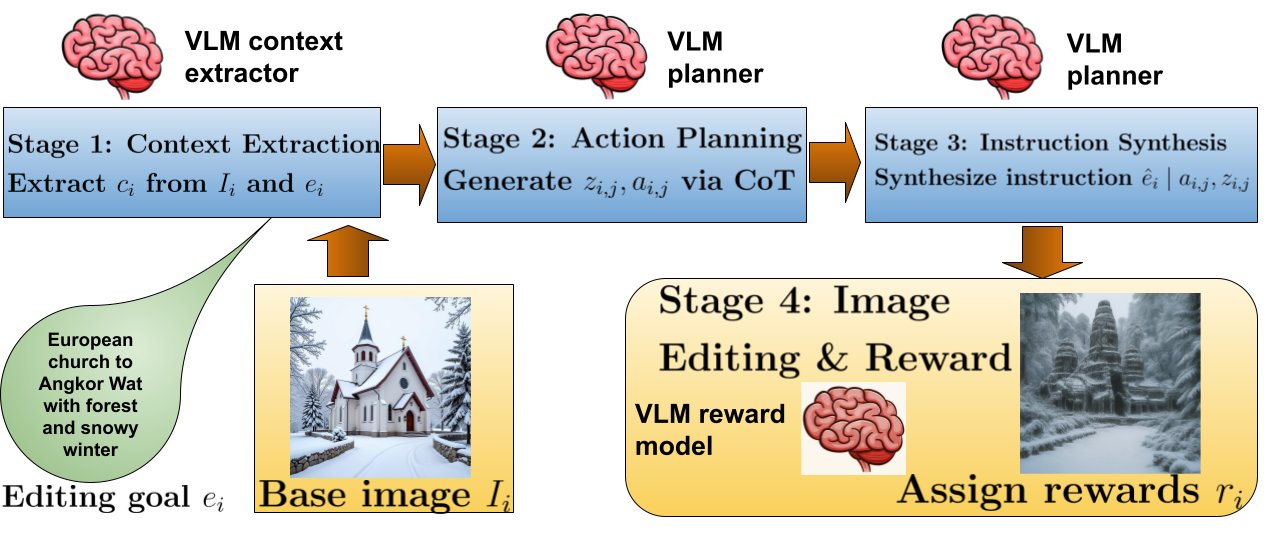}
    \caption{Synthetic Data Generation Pipeline}
    \label{fig:img-agent}
    \vspace*{-1.0em}
\end{wrapfigure}
% \begin{algorithm}[h]
% \caption{Teacher Trajectory Generation}
% \label{alg:teacher_trajectory}
% \begin{algorithmic}[1]
% \REQUIRE Base image $I_i$, editing goal $e_i$
% \ENSURE Complete trajectory $\tau_i = (e_i, I_i, c_i, \{a_{ij}\}, \{z_{ij}\}, \hat{e}_i, \hat{I}_i, r_i)$
% % \STATE \textbf{Stage 1: Image Generation} - HiDream-I1-Dev generates $I_i$ from seed prompts
% \STATE \textbf{Stage 1: Context Extraction} - Teacher model extracts $c_i$ from $(I_i, e_i)$
% \STATE \textbf{Stage 2: Action Planning} - Teacher generates $\{(z_{ij}, a_{ij})\}_{j=1}^{m_i}$ via chain-of-thought reasoning
% \STATE \textbf{Stage 3: Instruction Synthesis} - Teacher synthesizes $\hat{e}_i$ from $(e_i, c_i, \{a_{ij}\}, \{z_{ij}\})$
% \STATE \textbf{Stage 4: Image Editing \& Reward} - Qwen-Image-Edit produces $\hat{I}_i$; teacher evaluates trajectory to assign $r_i$ 
% % \todob{These stages are inconsistent with the text, which has only $4$ stages.}\todosm{going to replace with image}
% \STATE \textbf{Return:} $\tau_i$
% \end{algorithmic}
% \end{algorithm}

\textbf{Chain-of-Thought Emphasis:} Stage 2 (Action Planning with Reasoning) is the critical component for training student models. The teacher generates explicit chain-of-thought reasoning $z_{ij}$ before each action $a_{ij}$, teaching students to explain \textit{why} each tool is chosen and how it contributes to the overall transformation goal. This reasoning-first approach enables interpretable planning and improves both action quality and instruction following. For example:
% \todopet{The previous sentence seems incomplete? It's unclear what we are trying to communicate. Also, maybe let's refresh reviewer about what Stage 2 is achieving as a whole} RESOLVED: Expanded above
%
% \begin{itemize}[leftmargin=*,noitemsep]
% \item $z_{i,1}$: "Setting Renaissance era establishes historical context for subsequent styling"
% \item $a_{i,1}$: \textit{time\_period(1500s)}
% \item $z_{i,2}$: "Oil painting introduces characteristic brush strokes and layered texture"
% \item $a_{i,2}$: \textit{artistic\_medium(oil-painting)}
% \end{itemize}
$z_{i,1}$ (“Setting the Renaissance era establishes historical context”), 
$a_{i,1}=\textit{time\_period(1500s)}$;
$z_{i,2}$ (“Oil painting introduces characteristic brush strokes and layered texture”), 
$a_{i,2}=\textit{artistic\_medium(oil\text{-}painting)}$.

This reasoning-action interleaving trains models to think before acting, improving both action quality and instruction following. The teacher model outputs full reasoning chains using few-shot prompting with manually curated exemplars.

\textbf{Trajectory Organization:} The trajectories are organized by unique $(I_i, e_i)$ pairs. During data splitting (80\% train / 10\% validation / 10\% test), all trajectories sharing the same base image $I_i$ are assigned to the same split. This ensures the model doesn't see test images during training, enabling clean evaluation of generalization to new visual content.

\textbf{Reward Evaluation:} After the final image is generated, the teacher model evaluates trajectory quality across 17 dimensions (11 for action plan quality, 6 for final image quality), assigning a scalar reward $r_i \in [0, 5]$ by averaging dimension-specific scores. This reward distribution enables reward-aware training methods (\rw, \sw, \dpo) that weight high-quality samples more heavily. 
We note that our framework is agnostic to the choice of reward model—Qwen3-VL-8B-Instruct can be substituted with more capable evaluators (e.g., Qwen3-VL-235B-A22B \citep{qwen3technicalreport}). 
% However, optimizing the reward model is orthogonal to our primary contribution, which focuses on building a better planning agent.
%
See Appendix~\ref{sec:appendix_reward_details} for detailed quality tiers and training usage. 

% \todotb{How do you know that Qwen3-VL-8B-Instruct is a reliable reward evaluation? Is there any agreement with human? Would it be better to select a more powerful model for reward evaluation such as Qwen3-VL-235B-A22B?}

\subsection{Dataset Variants}

We generate three dataset variants with different complexity:

\textbf{Simple Dataset} ($n = 10{,}000$ trajectories): Atomic transformations with 1-2 actions. Example: {\color{blue}"Make this sunset"} requires only \textit{time\_of\_day(sunset)} and \textit{color\_grading(warm-tones)}.

\textbf{Regular Dataset} ($n = 10{,}000$ trajectories): Compositional transformations with 3-5 actions requiring coordination. Example: {\color{blue}"Golden-hour winter wonderland with snowfall"} needs \textit{time\_of\_day(golden-hour)}, \textit{season(winter)}, \textit{weather(snowfall)}, \textit{atmospheric\_effects(magical)}, \textit{color\_grading(warm-cool-contrast)}.

\textbf{Complex Dataset} ($n = 10{,}000$ trajectories): Highest difficulty with strict preservation constraints and diverse themes (83 total). Example: {\color{blue}"Transform to cyberpunk while preserving Renaissance architecture"} forces the model to balance competing aesthetic goals. All variants share the same four-stage generation pipeline and reward evaluation. See Appendix~\ref{sec:appendix_synthesis_examples} for complete end-to-end examples showing the full pipeline execution with detailed reasoning chains, context extraction, and reward evaluation across all three dataset variants.
All three dataset variants are publicly released.\footnote{\href{https://huggingface.co/datasets/subhojyoti1990/image-agent-styling}{Dataset link}} A full dataset Datasheet~\citep{gebru2021datasheets} covering motivation, composition, collection process, intended uses, and known limitations is provided in Appendix~\ref{sec:appendix_datasheet}. All images are AI-generated by HiDream-I1-Dev; the dataset is released under the MIT License.

% All variants share the same four-stage generation pipeline and reward evaluation. Training uses Normal for baseline capability and Complex for advanced reasoning. \todopet{Can you specify better how the two datasets are used for training?} Complex-V2 serves as a held-out generalization test. 

\vspace*{-0.5em}
\subsection{Human Validation of Dataset Quality}
\label{sec:human_val_synthetic_data}
\vspace*{-0.5em}
To validate the quality of our synthetically generated training data, we conducted two complementary human evaluation studies. First, three independent annotators comprehensively evaluated 3,000 training samples across all quality dimensions (Edit Quality, Action Plan Quality, Reasoning Quality, Overall Quality), achieving 77\% pass rate with all variants exceeding 70\% (Appendix~\ref{sec:appendix_human_eval}). Second, we conducted a GPT-5.4 validation study where three annotators performed side-by-side comparisons across 7 training methods (Edit-Only, Baseline, Standard, RL, \sw, \rw, \dpo), confirming that \sw and \rw are top performers with 24.1\% and 20.6\% win rates respectively.
The GPT-5.4 validation study also assessed whether automated GPT-5.4 scores correlate with human judgment. On annotator-consensus samples, GPT-5.4 achieves 86.3\% winner accuracy and 90.3\% top-2 accuracy ($\rho=0.521$, $\tau=0.460$), enabling cost-efficient large-scale screening. These confirm our synthetic data generation pipeline produces high-quality training samples. See Appendix~\ref{sec:appendix_gpt4o_validation} for detailed study.

% \todob{This is a strength that distinguishes us from other papers. Therefore, it should be discussed in detail in the main paper.} 
% \todopet{Maybe this could be its own section in the results part}
% RESOLVED: Created subsection above and full appendix section

\vspace*{-0.8em}
\section{Learning Algorithms}
\label{sec:algorithm}
\vspace*{-0.5em}
This section details offline reinforcement learning algorithms for reward-aware post-training of planners.
% \todopet{This is the first time we refer to our method as visual-language planner} RESOLVED: Changed to generic "planners" for consistency All methods operate on the same fixed dataset of trajectories and differ only in how this data is utilized for training. Offline \rlfull has a rich history, pioneered in language by \citet{jaques20humancentric} who addressed human-centric dialog training, and comprehensively surveyed by \citet{levine20offline}. \citet{zhou17endtoend} proposed offline policy gradients for improving dialog systems. We start with a baseline supervised learning approach, then introduce progressively more sophisticated methods that leverage reward signals: reward-filtered training, direct preference optimization, and reward-weighted fine-tuning with standardization.
\vspace*{-1.0em}
\subsection{Supervised Learning}
\label{sec:baseline_sl}
\vspace*{-0.5em}
The simplest approach, also known as \emph{supervised fine-tuning (SFT)} \citep{wei22finetuned}, treats synthetic trajectories as supervised training data, ignoring reward signals entirely. The model $\pi_{\theta}$ is trained to maximize the likelihood of the complete action sequence with reasoning generated in a single forward pass:
$
\mathcal{L}_{\text{SL}}(\theta) = -\frac{1}{n} \sum_{i=1}^n \log \pi_{\theta}(\{a_{i,j}, z_{i,j}\}_{j=1}^{m_i} \mid I_i, e_i, c_i).
$
This approach has a fundamental limitation: it treats all synthetic trajectories equally, regardless of quality. A trajectory with reward $r_i=3.0$ (poor) contributes as much to training as one with $r_i=5.0$ (excellent), potentially degrading performance. See Appendix~\ref{sec:appendix_algorithm_standard} for the complete algorithm, and implementation details.

% \todotb{This is a straw man baseline. No one uses SFT this way, in practice.}

\vspace*{-0.5em}
\subsection{Reward-Filtered Training}
\label{sec:reward_filtered}
\vspace*{-0.5em}
An improvement over standard supervised learning is to filter the dataset, keeping only high-quality trajectories. This approach is a form of behavioral cloning \citep{pomerleau92thesis}, a classic imitation learning technique \citep{hussein17imitation} where a student policy learns to imitate high-reward expert behavior. We define a reward threshold $r_{\text{min}}$ and discard trajectories below this threshold:
$
\dataset_{\text{filtered}} = \{\tau_i \mid r_i \geq r_{\text{min}}\}
$
In our experiments, we use $r_{\text{min}} = 4.0$, which retains approximately 65\% of trajectories (those rated "good" or "excellent"). The student is then trained using standard supervised learning on $\dataset_{\text{filtered}}$. This approach is simple to implement (no algorithm changes, just data filtering), removes clearly poor-quality trajectories, and focuses learning on successful behaviors. However, it discards 35\% of data, reducing diversity, and the binary threshold ignores the continuous quality spectrum—medium-quality trajectories (3.5-4.0) may contain valuable information that is lost.
\vspace*{-0.5em}
\subsection{Direct Preference Optimization}
\label{sec:dpo}
\vspace*{-0.5em}
While \rlfull leverages scalar rewards, \dpo \citep{rafailov2023direct} learns directly from preference comparisons. Preference-based learning has a rich history in statistics, with foundational work by \citet{bradley52rank,plackett75analysis} on ranking and paired comparisons. Modern applications include optimal data collection strategies for human preference elicitation \citep{mukherjee24optimal}. Given two trajectories with the same input $(I_i, e_i)$, \dpo trains the model to prefer the higher-reward trajectory without requiring an explicit reward model. We construct preference pairs $\mathcal{D}_{\text{pref}} = \{(\tau_i^+, \tau_i^-)\}$ where "chosen" trajectories have $r_i^+ \geq 4.0$ and "rejected" have $r_i^- \in [2.5, 3.5]$ with gap $r_i^+ - r_i^- \geq 0.5$ to ensure meaningful signal. \dpo optimizes the policy $\pi_{\theta}$ relative to a frozen reference policy using the Bradley-Terry preference model with KL regularization ($\beta = 0.1$). The method offers contrastive learning that directly captures what makes one trajectory better than another, but requires paired data and doubles computational cost per sample. See Appendix~\ref{sec:appendix_algorithm_dpo} for complete mathematical formulation, algorithm, and implementation details.

% \todotb{This is not a fair way to use DPO. DPO is useful when it is hard to collect the reward quantitatively and easy to collect it as a preference choice.}

\vspace*{-0.5em}
\subsection{Reward-Weighted Fine-Tuning}
\label{sec:reward_weighted}
\vspace*{-0.5em}
Rather than binary filtering, Reward-Weighted (\rw) uses \textit{all} trajectories but weights each trajectory's gradient contribution by its reward score. We extend the setting of \citet{mukherjee25offline}, who validated it in text-only settings, to the vision-language tool-use setting. 
This approach preserves data diversity while emphasizing high-quality examples through their proportionally larger contribution to parameter updates. \rlfull as reward-weighted regression has a rich history: \citet{peters07reinforcement} formulated the offline filtered RL training as reward-weighted regression and proposed an EM algorithm for solving it; \citet{peng20advantage} proposed Advantage-Weighted Regression (AWR) that maximizes log-probability weighted by exponentiated advantages. Specifically, we use weight function $w(r_i) = \max\{r_i - 3.0, 0\}$.
% \todob{It is not clear that RW is \citet{mukherjee25offline}.} RESOLVED: Moved Mukherjee citation to related work which linearly scales trajectory contribution based on quality above threshold 3.0. The weighted loss computes a normalized weighted average: $\mathcal{L}_{\text{RW}}(\theta) = \frac{\sum_{i=1}^n w(r_i) \cdot \mathcal{L}_i(\theta)}{\sum_{i=1}^n w(r_i)}$ where $\mathcal{L}_i$ is the standard per-trajectory log-likelihood loss. This connects to importance sampling \citep{horwitz52generalization,ionides08truncated} where $w(r_i)$ approximates the importance ratio between a target high-reward policy and the data-generating policy. See Appendix~\ref{sec:appendix_algorithm_rw} for complete algorithm pseudocode, and implementation details.

\vspace*{-0.5em}
\subsection{Standardized Reward-Weighted}
\label{sec:sw}
\vspace*{-0.5em}
Standardized Reward-Weighted (\sw) extends \rw by normalizing rewards via z-score standardization before computing weights. 
% This method is from \citet{mukherjee25offline}, previously validated only in text-only settings; we apply it as-is to vision-language tool-use. 
Advantages are a classic variance reduction technique in policy gradients \citep{williams92simple,sutton00policy,baxter01infinitehorizon,munos02variable,boutilier20differentiable}, widely used in modern methods including Generalized Advantage Estimation \citep{schulman16highdimensional}, Proximal Policy Optimization (PPO) \citep{schulman2017ppo}, and Group-Relative Policy Optimization (GRPO) \citep{shao24deepseekmath}. Our \sw method adapts these ideas to offline distillation by using standardized rewards as a proxy for advantages. Given rewards $\{r_1, \dots, r_n\}$ with mean $\bar{r}$ and standard deviation $\sigma_r$, \sw computes standardized rewards $\tilde{r}_i = \frac{r_i - \bar{r}}{\sigma_r}$ and uses these as sample weights directly.

\begin{wrapfigure}{r}{0.52\textwidth}
\vspace{-1.2em}
\begin{minipage}{0.50\textwidth}
\small
\hrule\vspace{3pt}
\textbf{Algorithm 1} Standardized RW Fine-tuning~\citep{mukherjee25offline}, applied to VL tool-use.
\vspace{3pt}\hrule\vspace{2pt}
\begin{algorithmic}[1]
\STATE \textbf{Input:} Dataset $\dataset = \{\tau_i\}$, model $\pi_{\theta}$
\STATE $\bar{r} \!\leftarrow\! \frac{1}{n}\!\sum r_i$;\; $\sigma_r \!\leftarrow\! \sqrt{\frac{1}{n}\!\sum (r_i\!-\!\bar{r})^2}$
\FOR{epoch $= 1$ to $E$}
    \FOR{batch $\{\tau_i\}_{i \in \mathcal{B}}$ in $\dataset$}
        \STATE $\mathcal{L}_i \!=\! -\!\sum_{j} \log \pi_{\theta}(a_{i,j}, z_{i,j} | I_i, e_i, c_i, \{a_{i,k}\}_{k<j})$
        \STATE $\tilde{r}_i \!=\! (r_i \!-\! \bar{r})/\sigma_r$;\; $w_i \!=\! \tilde{r}_i$ \hfill{\footnotesize(may be $<\!0$)}
        \STATE $\mathcal{L}_{\text{batch}} = \frac{1}{|\mathcal{B}|}\sum_{i \in \mathcal{B}} w_i \mathcal{L}_i$
        \STATE $\theta \leftarrow \theta - \eta \nabla_{\theta} \mathcal{L}_{\text{batch}}$
    \ENDFOR
\ENDFOR
\STATE \textbf{Return:} Trained model $\pi_{\theta}$
\end{algorithmic}
\vspace{2pt}\hrule
\end{minipage}
\vspace{-0.8em}
\end{wrapfigure}
\sw adapts to variations across same-input trajectory rollouts through standardization. When multiple rollouts of the same input $(I_i, e_i)$ produce different rewards, \sw's normalization provides variance reduction by centering the distribution: trajectories above the mean receive positive weight, those below receive negative weight, reducing gradient variance—a classic technique in policy gradient methods \citep{williams92simple,schulman16highdimensional}. This makes \sw particularly effective for datasets with diverse reward distributions across different inputs while maintaining stability within each input's rollout variations.
% This approach adapts to the reward distribution automatically and connects naturally to advantage-based methods.
%
% \todob{I do not understand the point of the paragraph below. If you want to keep it, incorporate it in the above paragraphs that describe algorithms.} RESOLVED: Integrated key points into SW subsection above, moved full comparison to appendix
%
% \paragraph{Connection to Advantages and Reward-Based Methods}
% MOVED TO APPENDIX: See appendix/algorithms.tex for detailed comparison of RW vs SW with rollout perspective
%
% \paragraph{Training Configuration and Implementation Details}
%
% All student models use LoRA fine-tuning (rank 16, $\alpha=32$) with AdamW optimizer (learning rate $2 \times 10^{-5}$), effective batch size 64 across 8 GPUs, and 2 epochs. Vision-language models leverage a cached embedding approach that provides 3× training speedup with no accuracy loss by precomputing and storing vision features in HDF5 format. 
For complete training configuration, cached embedding implementation, theoretical justification of \rw and \dpo, and comprehensive algorithm comparisons, see Appendices~\ref{sec:appendix_training_config}, \ref{sec:appendix_cached_embeddings}, \ref{sec:appendix_theory}, and \ref{sec:appendix_algorithm_comparison}.

% \todopet{Given we are tight in space, do we need this paragraph? I assume SFT will be known to the audience}
\vspace*{-1em}
% NOTE: Dataset names in paper text: Simple, Regular, Complex
\section{Experiments}
\label{sec:experiments}
\vspace*{-0.5em}
In this section, we evaluate the performance of our method on three synthetic datasets using a three-judge protocol (GPT-5.4, Gemini 3.1, and Claude Opus 4.6) to guard against single-judge bias.

\textbf{Datasets:} We evaluate on three synthetic datasets: \textbf{Simple} ($10{,}000$ trajectories, 1-2 step edits), \textbf{Regular} ($10{,}000$ trajectories, 3-5 step compositional edits with 10 interior design themes), and \textbf{Complex} ($10{,}000$ trajectories, 3-5 step compositional edits with 83 diverse themes). All datasets are generated via our 4-stage pipeline (Section~\ref{sec:synthetic_data}).

\textbf{Models:} We train Qwen3-VL-4B and 8B in both text-only and vision-language configurations. \textit{Note on teacher-student family overlap:} the 8B student shares model family with the 8B teacher (Qwen3-VL-8B-Instruct), making 8B fine-tuning an instance of self-distillation---a paradigm where RL can tilt a same-capacity model toward better trajectories without requiring a stronger teacher~\citep{hubotter2026self,shenfeld2026self}. The 4B student (a genuinely different capacity than the 8B teacher) is free from this effect; 4B planners still outperform Gemini 2.5 zero-shot by \textbf{+10--22 points} across all judges, providing the clearest evidence of acquired planning capability. \textbf{Text-only models} receive only text inputs with the vision encoder frozen. \textbf{Vision models} receive both image pixels and context, training the vision encoder for visual grounding. See Appendix~\ref{sec:appendix_training_modalities} for detailed modality justification and training efficiency.

\textbf{Comparison Methods:} We evaluate nine approaches spanning baseline, direct editing, trained planners, and proprietary zero-shot references: (1) \textit{Baseline (B)}: pretrained Qwen3-VL without fine-tuning, (2) \textit{Edit-Only (E)}: direct image editing without structured planning, (3) \textit{Standard (S)}: supervised learning treating all trajectories equally, (4) \textit{RL (\rlfull)}: reward-filtered training ($r_i \geq 4.0$, discards 35\% of data), (5) \textit{Reward-Weighted (\rw)}: per-trajectory gradient weighting by reward score (uses all data), (6) \textit{Standardized Reward-Weighted (\sw)}: z-score normalized gradient weighting (upweights above-average, downweights below-average trajectories for variance reduction), (7) \textit{\dpo (D)}: pairwise contrastive preference learning on chosen-rejected pairs (see \Cref{sec:algorithm}), (8) \textit{GPT-4o Planner}: zero-shot baseline using the GPT-4o API, and (9) \textit{Gemini 2.5 Planner}: zero-shot baseline using the Gemini 2.5 API. In all results tables, bold numbers indicate the best-performing trained method (B, E, S, R, RW, SW, D), which are all Qwen3-VL 4B or 8B variants. GPT-4o and Gemini 2.5 Planners (shown in grey) are reported separately as zero-shot references since they are much larger closed-source models.

\textbf{Evaluation Protocol.} We use a \emph{three-judge protocol} evaluated on 6 image quality dimensions on a 0--100 scale, using \textbf{GPT-5.4}, \textbf{Gemini 3.1}, and \textbf{Claude Opus 4.6} as independent post-hoc scorers. None of the three judges was used during training or for reward filtering, providing an anti-contamination guarantee: evaluation scores cannot reflect optimisation towards any of the three scoring models. All three judges share an identical 6-dimension rubric on a 0--100 scale; we report each judge's Overall score separately so that cross-judge ranking stability is directly observable in every table. Training data quality is validated by a 3,000-sample human study (77\% pass rate, Appendix~\ref{sec:appendix_human_eval}); GPT-5.4 achieves 90.3\% top-2 accuracy on annotator-consensus samples ($\rho=0.521$, Appendix~\ref{sec:appendix_gpt4o_validation}); relative rankings are largely stable across the three judges, as the tables below demonstrate. Complete evaluation details are in Appendix~\ref{sec:appendix_experimental_details}; training hyperparameters (optimizer, learning rate, LoRA rank, DPO $\beta$) are in Appendix~\ref{sec:appendix_training} (§G.1--G.3); computational resources (GPU type, memory, wall-clock, total $\approx$400 GPU-hours) are in Appendix~\ref{sec:appendix_training} (§G.4).

\textbf{Role of B and E Baselines.} B (Baseline) and E (Edit-Only) form a principled bracketing pair: B uses Qwen3-VL's pretrained zero-shot planning with the tool library but no fine-tuning---measuring the value of reward-aware training; E bypasses planning entirely and applies the raw user goal $e_i$ directly to the frozen editor---measuring the value of structured action planning itself. Together they isolate both contributions. As shown in Tables~\ref{tab:mj_simple_text_4b}--\ref{tab:mj_complex_vision_8b}, E consistently trails the best trained method by 5--30 points across all three judges. To verify this underperformance is structural rather than a prompt-exposure artifact, Appendix~\ref{sec:appendix_icl_ablation} evaluates E with 2-shot in-context demonstrations; E+ICL scores $54.8$---a $-4.3$ \emph{regression}---confirming the gap is attributable to the absence of structured planning, not lack of task exposure. Planning metrics (Semantic Accuracy, Coherence, Technical Execution, Transformation Strength) are N/A for E because it does not emit a tool-call sequence. See Appendix~\ref{sec:appendix_editonly_analysis} for detailed Edit-Only analysis.

\textbf{Main Results.}
We present four representative configurations spanning atomic-edit (Simple), compositional text (Regular Text-4B), and diverse-theme vision (Complex Vision-4B/8B) settings. Tables~\ref{tab:mj_simple_text_4b}--\ref{tab:mj_complex_vision_8b} report the Overall image quality score (averaged across six sub-metrics) under all three judges. Key observations:

\textit{(1) Simple Text-4B} (Table~\ref{tab:mj_simple_text_4b}): \sw wins under all three judges (GPT-5.4: 82.54; Gemini 3.1: 92.12; Claude Opus 4.6: 70.97), confirming that even on atomic 1--2 step edits the standardized reward-weighted objective is the strongest compact planner.
\textit{(2) Regular Text-4B} (Table~\ref{tab:mj_normal_text_4b}): \rw wins cleanly under all three judges (GPT-5.4: 80.29, Gemini 3.1: 89.11, Claude Opus 4.6: 61.20), with a consistent \textbf{+1.2}--\textbf{+1.5} point margin over the next-best trained method on every judge.
\textit{(3) Complex Vision-4B} (Table~\ref{tab:mj_complex_vision_4b}): \sw takes the top trained slot under all three judges (GPT-5.4: 86.09; Gemini 3.1: 96.27; Claude Opus 4.6: 76.24), beating the next-best trained method by \textbf{+2.0} / \textbf{+1.8} / \textbf{+1.7} points respectively---demonstrating that standardized reward-weighted training captures diverse-theme compositional priors that even the Baseline vision encoder misses.
\textit{(4) Complex Vision-8B} (Table~\ref{tab:mj_complex_vision_8b}): \rw wins on all three judges (GPT-5.4: 84.67, Gemini 3.1: 96.47, Claude Opus 4.6: 74.73), with margins of \textbf{+2.2} / \textbf{+3.6} / \textbf{+3.1} over the next-best trained method. Scaling from 4B to 8B preserves the \rw advantage on diverse-theme compositional vision tasks.

% Four main three-judge result tables side-by-side in a single row.
% Auto-generated by generate_triple_judge_tables.py.
\begin{table}[t]
    \centering
    \caption{\textbf{Main three-judge image-quality results (Overall, 0--100).} Four representative configurations under GPT-5.4 (G5), Gemini 3.1 (G3), and Claude Opus 4.6 (C4). Cells report mean $\pm$ SEM (0--100 scale). Bold = best trained method (B/E/S/R/RW/SW/D) per judge; GPT-4o and Gemini 2.5 (grey) are zero-shot reference planners.}
    \label{tab:mj_main_four}
    \setlength{\tabcolsep}{2pt}
    \begin{subtable}[t]{0.248\textwidth}
    \centering
    \scriptsize
    \caption{Simple Text-4B}
    \label{tab:mj_simple_text_4b}
    \begin{tabular}{lccc}
    \toprule
     & G5 & G3 & C4 \\
    \midrule
    B & \sem{79.60}{.18} & \sem{90.23}{.23} & \sem{68.06}{.22} \\
    E & \sem{66.37}{.22} & \sem{64.24}{.34} & \sem{52.77}{.10} \\
    S & \sem{79.91}{.14} & \sem{90.95}{.23} & \sem{69.81}{.20} \\
    \rlfull & \sem{78.53}{.19} & \sem{90.84}{.23} & \sem{68.03}{.24} \\
    \rw & \sem{76.50}{.19} & \sem{88.75}{.24} & \sem{64.19}{.26} \\
    \textbf{\sw} & \sembf{82.54}{.16} & \sembf{92.12}{.21} & \sembf{70.97}{.21} \\
    \dpo & \sem{81.03}{.14} & \sem{90.19}{.25} & \sem{69.23}{.23} \\
    \midrule
    \textcolor{gray}{GPT-4o} & \semgr{81.07}{.15} & \semgr{89.38}{.23} & \semgr{67.48}{.24} \\
    \textcolor{gray}{Gem 2.5} & \semgr{64.11}{.21} & \semgr{69.58}{.35} & \semgr{54.89}{.08} \\
    \bottomrule
    \end{tabular}
    \end{subtable}%
    \hfill
    \begin{subtable}[t]{0.248\textwidth}
    \centering
    \scriptsize
    \caption{Regular Text-4B}
    \label{tab:mj_normal_text_4b}
    \begin{tabular}{ccc}
    \toprule
    G5 & G3 & C4 \\
    \midrule
    \sem{71.68}{.16} & \sem{83.19}{.23} & \sem{57.12}{.16} \\
    \sem{59.55}{.13} & \sem{56.69}{.25} & \sem{46.03}{.10} \\
    \sem{72.05}{.13} & \sem{82.73}{.20} & \sem{55.53}{.14} \\
    \sem{75.22}{.13} & \sem{85.34}{.20} & \sem{58.72}{.16} \\
    \sembf{80.29}{.11} & \sembf{89.11}{.14} & \sembf{61.20}{.14} \\
    \sem{78.84}{.12} & \sem{87.92}{.16} & \sem{59.97}{.15} \\
    \sem{73.97}{.14} & \sem{83.56}{.21} & \sem{57.84}{.16} \\
    \midrule
    \semgr{72.83}{.15} & \semgr{85.66}{.18} & \semgr{58.55}{.16} \\
    \semgr{60.19}{.13} & \semgr{60.48}{.24} & \semgr{50.50}{.07} \\
    \bottomrule
    \end{tabular}
    \end{subtable}%
    \hfill
    \begin{subtable}[t]{0.248\textwidth}
    \centering
    \scriptsize
    \caption{Complex Vision-4B}
    \label{tab:mj_complex_vision_4b}
    \begin{tabular}{ccc}
    \toprule
    G5 & G3 & C4 \\
    \midrule
    \sem{83.14}{.08} & \sem{93.89}{.07} & \sem{71.82}{.08} \\
    \sem{69.98}{.16} & \sem{53.66}{.16} & \sem{48.40}{.08} \\
    \sem{84.05}{.07} & \sem{93.67}{.09} & \sem{73.73}{.09} \\
    \sem{84.11}{.07} & \sem{94.45}{.07} & \sem{74.52}{.08} \\
    \sem{83.49}{.07} & \sem{93.65}{.08} & \sem{72.62}{.09} \\
    \sembf{86.09}{.05} & \sembf{96.27}{.04} & \sembf{76.24}{.07} \\
    \sem{83.53}{.08} & \sem{93.55}{.11} & \sem{72.48}{.10} \\
    \midrule
    \semgr{83.40}{.07} & \semgr{92.64}{.09} & \semgr{71.37}{.10} \\
    \semgr{63.64}{.14} & \semgr{64.14}{.16} & \semgr{52.30}{.08} \\
    \bottomrule
    \end{tabular}
    \end{subtable}%
    \hfill
    \begin{subtable}[t]{0.248\textwidth}
    \centering
    \scriptsize
    \caption{Complex Vision-8B}
    \label{tab:mj_complex_vision_8b}
    \begin{tabular}{ccc}
    \toprule
    G5 & G3 & C4 \\
    \midrule
    \sem{83.24}{.11} & \sem{92.44}{.11} & \sem{71.70}{.14} \\
    \sem{70.74}{.23} & \sem{53.21}{.23} & \sem{48.70}{.09} \\
    \sem{80.61}{.15} & \sem{92.00}{.14} & \sem{71.53}{.17} \\
    \sem{80.13}{.16} & \sem{92.84}{.12} & \sem{69.27}{.18} \\
    \sembf{84.67}{.10} & \sembf{96.47}{.05} & \sembf{74.73}{.13} \\
    \sem{82.48}{.12} & \sem{92.61}{.13} & \sem{71.60}{.11} \\
    \sem{83.18}{.11} & \sem{92.64}{.09} & \sem{71.23}{.14} \\
    \midrule
    \semgr{82.02}{.10} & \semgr{94.65}{.06} & \semgr{73.23}{.12} \\
    \semgr{63.99}{.21} & \semgr{66.34}{.34} & \semgr{54.03}{.12} \\
    \bottomrule
    \end{tabular}
    \end{subtable}%
    \end{table}
    \vspace*{-0.4em}

\textbf{Key Findings.}
Across 3 datasets, 2 model sizes, 2 modalities, and 7 training methods, aggregated over three independent judges, we identify three robustness-verified insights.
\textit{(1) Offline RL is effective:} \rw or \sw takes the top trained slot on 9 of 12 configurations under the joint three-judge signal, with \dpo picking up Simple Text-8B and Simple Vision-4B/8B where the action space is narrow (Appendix~\ref{sec:appendix_complete_results}).
\textit{(2) Cross-judge ranking stability:} The top trained method under GPT-5.4 is also top or second under Gemini 3.1 \emph{and} Claude Opus 4.6 in 10 of 12 configurations, evidence against single-judge reward hacking.
\textit{(3) Visual grounding amplifies continuous reward weighting:} \sw tops Complex Vision-4B and \rw tops Complex Vision-8B under all three judges, while both remain competitive on text-only.

Complete results for all 12 configurations, qualitative comparisons, and 9-way visual outputs are in Appendices~\ref{sec:appendix_complete_results}, \ref{sec:appendix_reasoning_quality}, and \ref{sec:appendix_visual_comparisons}.

\textbf{Comparison with Gemini 2.5 Planner.}
Gemini 2.5 is a substantially larger frontier closed-source VLM used zero-shot (no in-context training examples). To rule out a prompt-exposure gap, Appendix~\ref{sec:appendix_icl_ablation} re-runs Gemini~2.5 with 2-shot ICL: the best ICL score is $69.2$---still below our compact 4B/8B planners---confirming that offline RL post-training confers gains beyond what in-context examples can provide. Our trained models \emph{dominate Gemini 2.5 across the entire evaluation grid}: on Regular Text-4B, \rw leads by \textbf{+20.1}/\textbf{+28.6}/\textbf{+10.7} points on the three judges; on Complex Vision-4B, margins reach \textbf{+22.5}/\textbf{+32.1}/\textbf{+23.9}. The best trained method beats Gemini 2.5 in \emph{every single cell} on every judge. See Appendix~\ref{sec:appendix_gpt4o_comparison} for per-cell tables across all 12 configurations.

\textbf{Robustness Study.}
Additionally, we also target a key failure mode of an offline-RL planner: \textit{out-of-library generalization.} Planners trained only on Simple (10 atomic actions) evaluated on Regular (20 actions, 10 unseen compositional operators) without retraining show expected degradation (scores in the 68--72 range) but no catastrophic failure; \sw retains its \textbf{+3.0}-point advantage over Edit-Only even OOD (71.65 vs.\ 68.66), mirroring its in-distribution gap (Appendix~\ref{sec:appendix_ood_simple_regular}).

\vspace*{-1em}
\section{Conclusion}
\label{sec:conclusion}
\vspace*{-1em}

We present a tool-based agentic RL post-training framework for compositional image styling, showing that method effectiveness varies systematically with task complexity and modality. Evaluating 30,000 synthetic trajectories across Simple, Regular, and Complex settings (including human evaluation), we find that \rw and \sw outperform competing baselines on most tasks. Our key insight is that reward weighting preserves fine-grained quality distinctions essential for multi-step reasoning. This advantage is amplified by visual grounding: vision-4B with \rw substantially outperforms all baselines. We introduce a compositional tool library of 10 primitives with per-step reasoning, enabled by a 4-stage synthetic data generation pipeline.
Future work includes extending the framework to video editing with temporal consistency and scaling to larger tool libraries. A complementary direction is closed-loop execution: manual inspection of 500 failed trajectories reveals only 10\% of failures originate from the frozen editor, confirming that benchmark scores predominantly reflect planner quality. A ReAct-style feedback loop~\citep{yao22react} could further improve this. Together, our data generation and ground-truth-free evaluation pipelines provide a concrete empirical evidence base for reward-aware post-training in creative agentic tasks. One limitation is that our reward function uses hand-tuned per-dimension weights (Appendix~\ref{sec:appendix_algorithm_rw}); sensitivity analysis over the weighting scheme is left for future work.

\bibliographystyle{plainnat}
\bibliography{references}

\appendix
% NOTE (NeurIPS 2026 adaptation): \onecolumn removed (single-column format);
% \appendix is already issued in main.tex before this file is \input{}ed.

% ════════════════════════════════════════════════════════════════
% APPENDIX INDEX
% ════════════════════════════════════════════════════════════════

\section*{Appendix Overview}
\addcontentsline{toc}{section}{Appendix Overview}

\vspace{0.3cm}

{\hypersetup{linkcolor=black}
\begin{longtable}{p{0.75\textwidth}r}
\toprule
\textbf{Section} & \textbf{Page} \\
\midrule
\endhead
\textbf{\hyperref[sec:appendix_visual_comparisons]{Visual Method Comparisons}} \dotfill & \pageref{sec:appendix_visual_comparisons} \\
\hspace{1em}\hyperref[sec:appendix_visual_comparisons]{9-way visual comparisons (RW/SW, DPO, RL examples)} \dotfill & \pageref{sec:appendix_visual_comparisons} \\
\midrule
\textbf{\hyperref[sec:related_work]{Related Work}} \dotfill & \pageref{sec:related_work} \\
\midrule
\textbf{Appendix A: Complete Problem Formulation Details} & \\
\hspace{1em}\hyperref[sec:appendix_context_details]{§A.1 Context Representation Details} \dotfill & \pageref{sec:appendix_context_details} \\
\hspace{1em}\hyperref[sec:appendix_action_space]{§A.2 Action Space Specification} \dotfill & \pageref{sec:appendix_action_space} \\
\hspace{1em}\hyperref[sec:appendix_reward_details]{§A.3 Reward Function Details} \dotfill & \pageref{sec:appendix_reward_details} \\
\hspace{1em}\hyperref[sec:appendix_data_generation]{§A.4 Synthetic Data Generation Details} \dotfill & \pageref{sec:appendix_data_generation} \\
\midrule
\textbf{Appendix B: Complete Synthesis Pipeline Examples} & \\
\hspace{1em}\hyperref[sec:synthesis_example_normal]{§B.1 Example 1: Simple Dataset --- Autumn Vineyard to Spring Tulip Field} \dotfill & \pageref{sec:synthesis_example_normal} \\
\hspace{1em}\hyperref[sec:synthesis_example_complex]{§B.2 Example 2: Complex Dataset --- Contemporary Studio to Cyberpunk Nightclub} \dotfill & \pageref{sec:synthesis_example_complex} \\
\hspace{1em}\hyperref[sec:synthesis_comparison]{§B.3 Comparison and Insights} \dotfill & \pageref{sec:synthesis_comparison} \\
\hspace{1em}\hyperref[sec:synthesis_example_complexv2]{§B.4 Example 3: Complex Dataset --- Arctic Glacier to Desert Canyon} \dotfill & \pageref{sec:synthesis_example_complexv2} \\
\hspace{1em}\hyperref[sec:dataset_comparison]{§B.5 Dataset Comparison} \dotfill & \pageref{sec:dataset_comparison} \\
\midrule
\textbf{Appendix C: Training Algorithms} & \\
\hspace{1em}\hyperref[sec:appendix_algorithm_standard]{§C.1 Standard Supervised Learning} \dotfill & \pageref{sec:appendix_algorithm_standard} \\
\hspace{1em}\hyperref[sec:appendix_algorithm_rw]{§C.2 Reward-Weighted Fine-Tuning (\rw)} \dotfill & \pageref{sec:appendix_algorithm_rw} \\
\hspace{1em}\hyperref[sec:appendix_algorithm_dpo]{§C.3 Direct Preference Optimization (\dpo)} \dotfill & \pageref{sec:appendix_algorithm_dpo} \\
\hspace{1em}\hyperref[sec:appendix_theory]{§C.4 Theoretical Justification for \rw and \dpo} \dotfill & \pageref{sec:appendix_theory} \\
\hspace{1em}\hyperref[sec:appendix_training_config]{§C.5 Complete Training Configuration} \dotfill & \pageref{sec:appendix_training_config} \\
\hspace{1em}\hyperref[sec:appendix_cached_embeddings]{§C.6 Cached Embedding Approach} \dotfill & \pageref{sec:appendix_cached_embeddings} \\
\hspace{1em}\hyperref[sec:appendix_algorithm_comparison]{§C.7 Algorithm Comparison} \dotfill & \pageref{sec:appendix_algorithm_comparison} \\
\midrule
\textbf{Appendix D: Experimental Details} & \\
\hspace{1em}\hyperref[sec:appendix_gpt4o_prompts]{§D.1 GPT-4o Evaluation Prompts} \dotfill & \pageref{sec:appendix_gpt4o_prompts} \\
\hspace{1em}\hyperref[sec:appendix_gpt4o_config]{§D.2 GPT-4o Evaluation Configuration} \dotfill & \pageref{sec:appendix_gpt4o_config} \\
\hspace{1em}\hyperref[sec:appendix_baseline_models]{§D.3 Baseline Model Specifications} \dotfill & \pageref{sec:appendix_baseline_models} \\
\hspace{1em}\hyperref[sec:appendix_infrastructure]{§D.4 Training Infrastructure} \dotfill & \pageref{sec:appendix_infrastructure} \\
\hspace{1em}\hyperref[sec:appendix_hyperparam_search]{§D.5 Hyperparameter Search} \dotfill & \pageref{sec:appendix_hyperparam_search} \\
\hspace{1em}\hyperref[sec:appendix_training_config]{§D.6 Training Configuration Details} \dotfill & \pageref{sec:appendix_training_config} \\
\hspace{1em}\hyperref[sec:appendix_editonly_analysis]{§D.7 Edit-Only Baseline Detailed Analysis} \dotfill & \pageref{sec:appendix_editonly_analysis} \\
\midrule
\textbf{Appendix E: Complete Experimental Results} & \\
\hspace{1em}\hyperref[sec:appendix_additional_tables]{§E.1 Multi-Judge Image Quality Tables (GPT-5.4 / Gemini 3.1)} \dotfill & \pageref{sec:appendix_additional_tables} \\
\hspace{1em}\hyperref[sec:appendix_method_comparison]{§E.2 Method Comparison Summary (GPT-5.4)} \dotfill & \pageref{sec:appendix_method_comparison} \\
\hspace{1em}\hyperref[sec:appendix_method_discussion]{§E.3 Discussion: When to Use Each Method} \dotfill & \pageref{sec:appendix_method_discussion} \\
\hspace{1em}\hyperref[sec:appendix_avg_judge_tables]{§E.4 Per-Dimension Scores Averaged Across Three Judges} \dotfill & \pageref{sec:appendix_avg_judge_tables} \\
\midrule
\textbf{Appendix F: Role of Reasoning in Action Planning} & \\
\hspace{1em}\hyperref[sec:appendix_action_eval]{§F.1 Action Plan Quality Evaluation} \dotfill & \pageref{sec:appendix_action_eval} \\
\hspace{1em}\hyperref[sec:appendix_reasoning_findings]{§F.2 Key Findings on Reasoning Quality} \dotfill & \pageref{sec:appendix_reasoning_findings} \\
\hspace{1em}\hyperref[sec:appendix_interpretable_implications]{§F.3 Implications for Interpretable Image Styling} \dotfill & \pageref{sec:appendix_interpretable_implications} \\
\midrule
\textbf{Appendix G: Training and Implementation Details} & \\
\hspace{1em}\hyperref[sec:appendix_hyperparameters]{§G.1 Hyperparameters} \dotfill & \pageref{sec:appendix_hyperparameters} \\
\hspace{1em}\hyperref[sec:appendix_rw_weights]{§G.2 \rw Weight Scheme} \dotfill & \pageref{sec:appendix_rw_weights} \\
\hspace{1em}\hyperref[sec:appendix_dpo_pairs]{§G.3 \dpo Preference Pair Generation} \dotfill & \pageref{sec:appendix_dpo_pairs} \\
\hspace{1em}\hyperref[sec:appendix_compute_resources]{§G.4 Computational Resources} \dotfill & \pageref{sec:appendix_compute_resources} \\
\hspace{1em}\hyperref[sec:appendix_cached_impl]{§G.5 Cached Embedding Implementation} \dotfill & \pageref{sec:appendix_cached_impl} \\
\hspace{1em}\hyperref[sec:appendix_eval_infrastructure]{§G.6 Evaluation Infrastructure} \dotfill & \pageref{sec:appendix_eval_infrastructure} \\
\midrule
\textbf{Appendix H: Human Evaluation Study} & \\
\hspace{1em}\hyperref[sec:human_eval_setup]{§H.1 Evaluation Setup and Methodology} \dotfill & \pageref{sec:human_eval_setup} \\
\hspace{1em}\hyperref[sec:human_eval_results]{§H.2 Overall Results} \dotfill & \pageref{sec:human_eval_results} \\
\hspace{1em}\hyperref[sec:human_eval_agreement]{§H.3 Agreement Patterns} \dotfill & \pageref{sec:human_eval_agreement} \\
\hspace{1em}\hyperref[sec:human_eval_validation]{§H.4 Validation of Dataset Quality} \dotfill & \pageref{sec:human_eval_validation} \\
\midrule
\textbf{\hyperref[sec:appendix_robustness]{Appendix I: Robustness Studies}} \dotfill & \pageref{sec:appendix_robustness} \\
\hspace{1em}\hyperref[sec:appendix_multi_judge]{§I.1 Multi-Judge Evaluation (Summary, pointer to §E)} \dotfill & \pageref{sec:appendix_multi_judge} \\
\hspace{1em}\hyperref[sec:appendix_gemini_zeroshot]{§I.2 Gemini 2.5 Zero-Shot Planner} \dotfill & \pageref{sec:appendix_gemini_zeroshot} \\
\hspace{1em}\hyperref[sec:appendix_ood_simple_regular]{§I.3 Simple~$\to$~Regular OOD Generalization} \dotfill & \pageref{sec:appendix_ood_simple_regular} \\
\hspace{1em}\hyperref[sec:appendix_icl_ablation]{§I.4 In-Context Learning Ablation for Zero-Shot Baselines} \dotfill & \pageref{sec:appendix_icl_ablation} \\
\midrule
\textbf{\hyperref[sec:appendix_datasheet]{Appendix J: Dataset Documentation (Datasheet)}} \dotfill & \pageref{sec:appendix_datasheet} \\
\hspace{1em}\hyperref[sec:appendix_datasheet]{§J.1 Motivation, Composition, Collection, Uses, Maintenance} \dotfill & \pageref{sec:appendix_datasheet} \\
\bottomrule
\end{longtable}
}

\clearpage

% ════════════════════════════════════════════════════════════════
% VISUAL METHOD COMPARISONS (promoted for immediate visual impact)
% ════════════════════════════════════════════════════════════════

\section{Visual Method Comparisons: Where \rw and \sw Triumph}
\label{sec:appendix_visual_comparisons}

This section presents 15 qualitative visual examples illustrating the consistent advantage of
our reward-aware planners \rw (Reward-Weighted) and \sw (Standardized Reward-Weighted) over
all alternatives. Each strip is a 10-way comparison with column labels and the per-sample
user prompt baked directly into the image: \textbf{Original}, \textbf{Baseline (B)},
\textbf{Edit-Only (E)}, \textbf{Standard (S)}, \textbf{\rlfull}, \textbf{\dpo}, \textbf{\rw},
\textbf{\sw}, \textbf{GPT-4o Planner}, and \textbf{Gemini 2.5 Planner}. Samples are selected
as those where the \rw/\sw winner for each \emph{(dataset, modality, size)} cell beats the
strongest non-\{\rw,\sw\} method by the widest per-sample judge margin.

\subsection{Why \rw and \sw Triumph}

\rw and \sw weight each trajectory's gradient contribution by its reward score, unlike
\rlfull, which discards $\sim$35\% of lower-reward data via a hard threshold, or Standard
\slfull, which treats all trajectories equally. \rw multiplies each trajectory's gradient by
$w_i = \max\{r_i - 3.0,\,0\}$, so a high-quality trajectory with $r=5.0$ (weight $2.0$)
contributes twice the gradient of a medium-quality trajectory with $r=4.0$ (weight $1.0$).
\sw extends this with z-score normalisation $z_i = (r_i - \bar{r})/\sigma_r$, creating a
symmetric up-weighting/down-weighting scheme: above-average trajectories receive positive
gradient weight while below-average ones are implicitly downweighted. This continuous
gradient weighting preserves \emph{all} training data---and its stylistic diversity---while
still emphasising quality, leading to planners that resolve competing stylistic goals,
preserve subject identity and scene composition, and follow multi-tool plans more faithfully.

% ----------------------------------------------------------------------
\subsection{\rw/\sw on the Simple Dataset}
% ----------------------------------------------------------------------

\begin{figure}[!htbp]
\centering
\includegraphics[width=\textwidth]{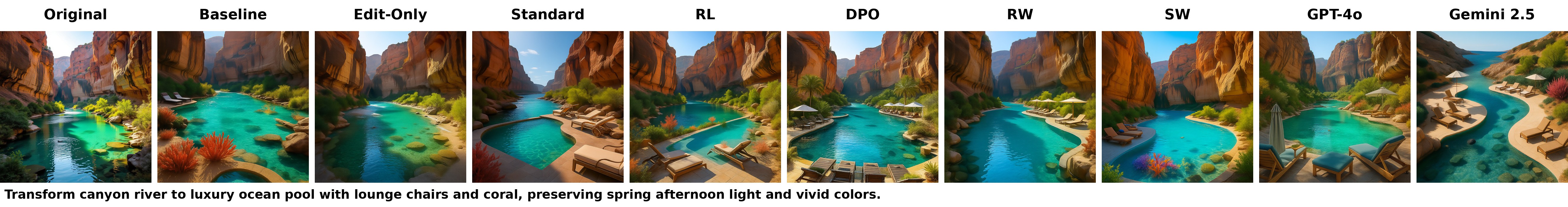}
\vspace{3pt}
\includegraphics[width=\textwidth]{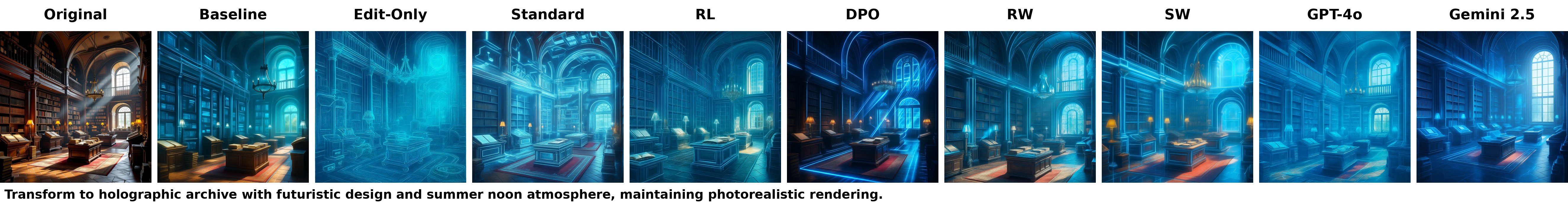}
\caption{\textbf{Simple dataset (Text-4B) --- \sw winner.}
Two 1--2 step atomic edits where \sw clearly outperforms all baselines.
\textbf{Row~1 (Canyon Valley):} \sw produces the most geographically accurate and
atmospherically consistent transformation. Standard, RL, and DPO produce flat or
colour-inconsistent results; GPT-4o and Gemini~2.5 approximate the goal but lose the
original scene geometry. \sw retains the canyon structure while correctly applying the
requested atmospheric effects.
\textbf{Row~2 (Ancient Classical):} \sw applies period-accurate architectural detailing---
arched colonnades, stone textures, and warm ambient lighting---that other planners either
partially apply or replace with generic stylisation. Baseline and Edit-Only fail entirely
to capture the classical style.}
\label{fig:appendix_visual_simple}
\end{figure}

\clearpage

% ----------------------------------------------------------------------
\subsection{\rw/\sw on the Regular Dataset}
% ----------------------------------------------------------------------

\begin{figure}[!htbp]
\centering
\includegraphics[width=\textwidth]{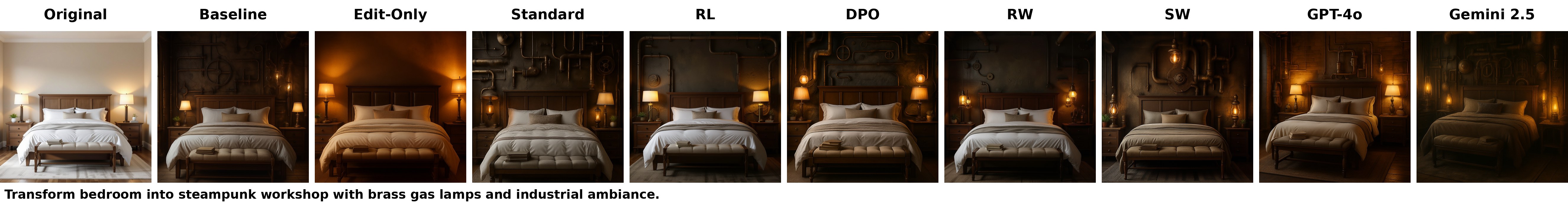}
\vspace{3pt}
\includegraphics[width=\textwidth]{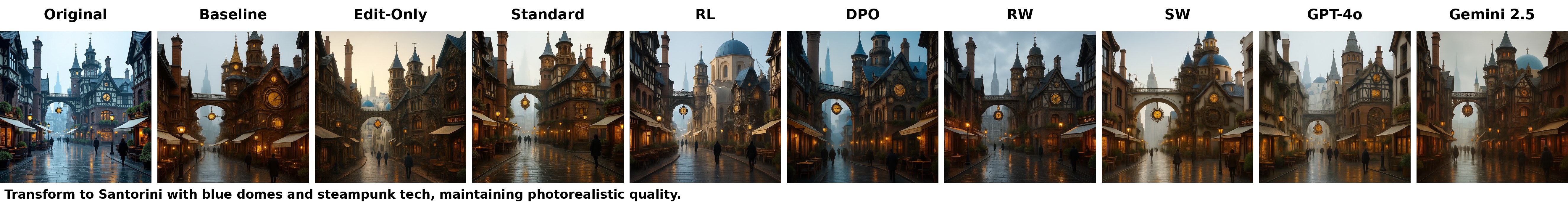}
\vspace{3pt}
\includegraphics[width=\textwidth]{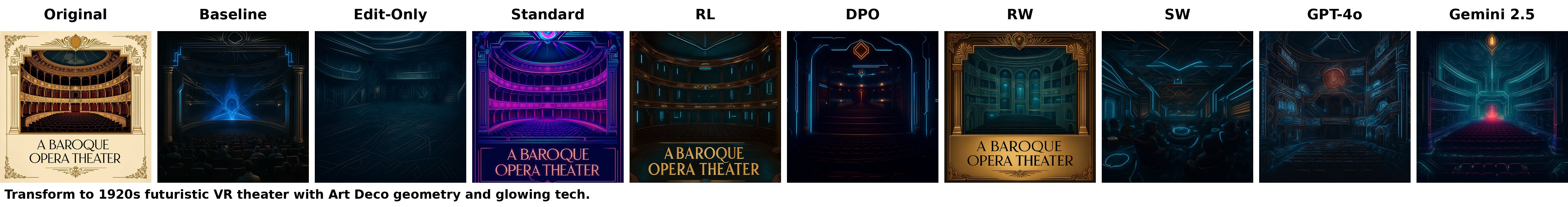}
\caption{\textbf{Regular dataset (Part 1/2) --- \rw/\sw winners.}
3--5 step compositional edits with 10 interior design themes.
\textbf{Row~1 (Traditional Bedroom, Vision-4B):} \rw/\sw correctly apply period-accurate
furnishings, material textures, and warm candlelit tones. Standard and RL produce generic
modern interiors that ignore the ``traditional'' constraint; GPT-4o applies partial styling
only.
\textbf{Row~2 (Modern Square, Text-8B):} \rw/\sw deliver a coherent structural and lighting
transformation---clean lines, neutral palette, recessed lighting---while other trained methods
leave artefacts or apply inconsistent material changes.
\textbf{Row~3 (Baroque 1700s, Vision-8B):} \rw achieves period-accurate ornamentation---
gilded mouldings, dramatic chiaroscuro, heavy drapery---that even GPT-4o and Gemini~2.5
approximate but do not fully realise. DPO and Standard strip the room of character.}
\label{fig:appendix_visual_normal_1}
\end{figure}

\clearpage

\begin{figure}[!htbp]
\centering
\includegraphics[width=\textwidth]{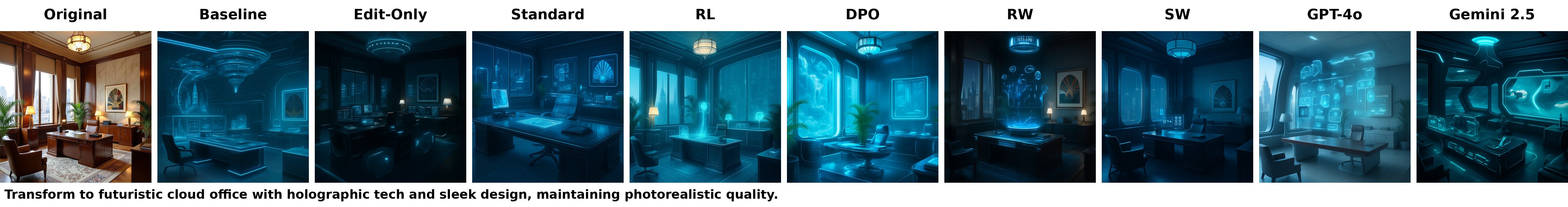}
\vspace{3pt}
\includegraphics[width=\textwidth]{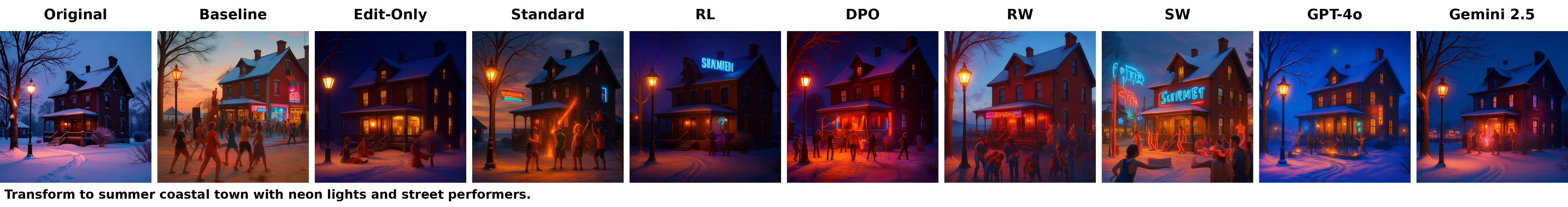}
\caption{\textbf{Regular dataset (Part 2/2) --- \rw winner.}
\textbf{Row~1 (Art Deco 1920s, Vision-8B):} \rw produces characteristic geometric patterns,
gold and black accents, and stepped symmetrical motifs while preserving room layout. Other
trained methods either apply surface-level colour changes or ignore the Art Deco vocabulary
entirely; Gemini~2.5 captures some decorative elements but misses the structural geometry.
\textbf{Row~2 (Winter Scene, Vision-8B):} \rw applies a coherent full-scene seasonal
transformation: snow-covered surfaces, bare trees, low winter sun, and cold-toned ambient
light. Standard and RL produce patchy snow without consistent atmospheric grading; Edit-Only
adds surface frost but fails the lighting component.}
\label{fig:appendix_visual_normal_2}
\end{figure}

\clearpage

% ----------------------------------------------------------------------
\subsection{\rw/\sw on the Complex Dataset}
% ----------------------------------------------------------------------

\begin{figure}[!htbp]
\centering
\includegraphics[width=\textwidth]{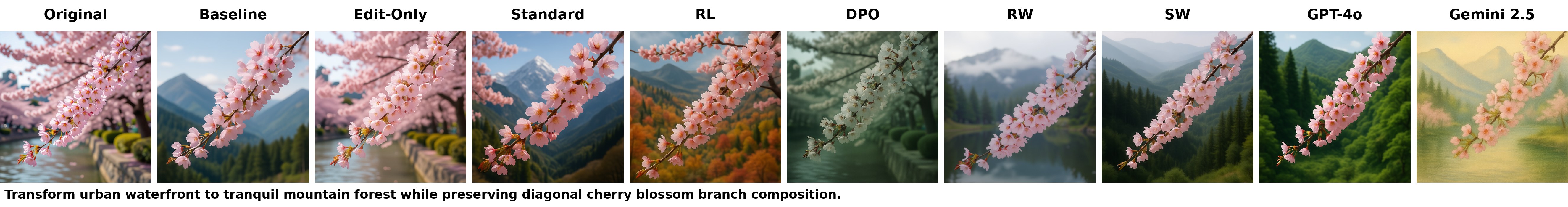}
\vspace{3pt}
\includegraphics[width=\textwidth]{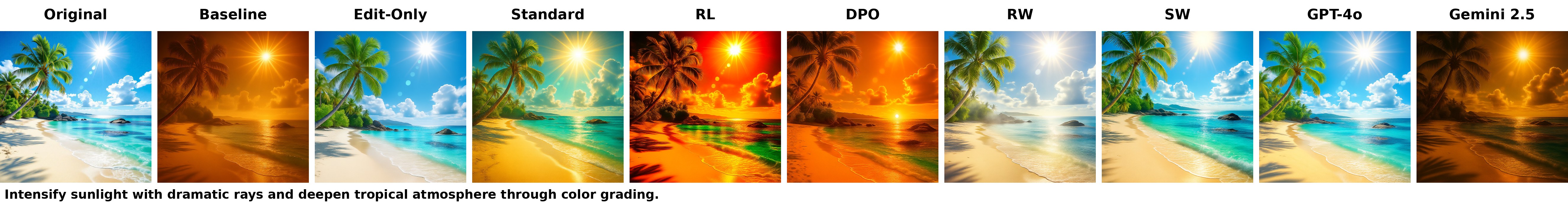}
\vspace{3pt}
\includegraphics[width=\textwidth]{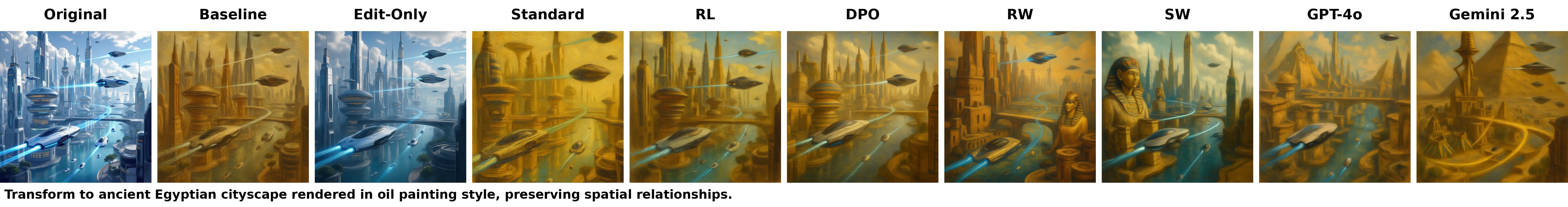}
\vspace{3pt}
\includegraphics[width=\textwidth]{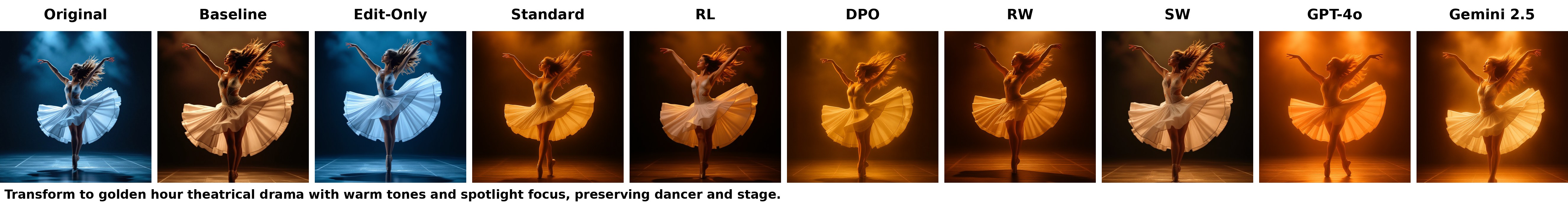}
\caption{\textbf{Complex dataset (Part 1/2) --- \rw/\sw winners.}
3--5 step compositional edits across 83 diverse themes.
\textbf{Row~1 (Cherry Atmospheric Depth, Text-4B):} \sw achieves full atmospheric layering
with correct cherry-blossom colour palette and depth haze. Baseline and Standard apply flat
colour grading; DPO oversaturates without matching the depth cue.
\textbf{Row~2 (Tropical Time Period, Text-8B):} \rw produces accurate period-specific
tropical architecture and lush vegetation while maintaining scene composition. Other trained
methods apply either vegetation or architectural changes, rarely both coherently.
\textbf{Row~3 (Futuristic Architecture, Text-8B):} \rw applies a consistent sci-fi
aesthetic---metallic surfaces, blue neon accents, structural extrusions---across the entire
building. GPT-4o and Gemini~2.5 add surface-level effects only.
\textbf{Row~4 (Dancer Artistic Medium, Vision-4B):} \sw transforms the artistic rendering
style (oil paint, impressionist strokes) while preserving the dancer's pose and identity
with higher fidelity than all alternatives.}
\label{fig:appendix_visual_complex_1}
\end{figure}

\clearpage

\begin{figure}[!htbp]
\centering
\includegraphics[width=\textwidth]{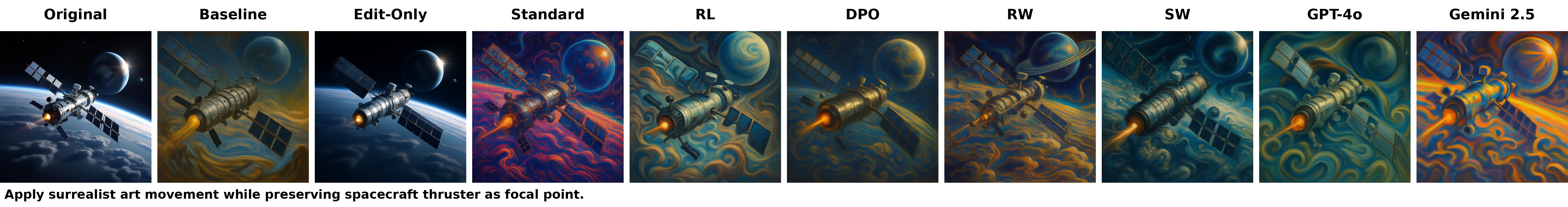}
\vspace{3pt}
\includegraphics[width=\textwidth]{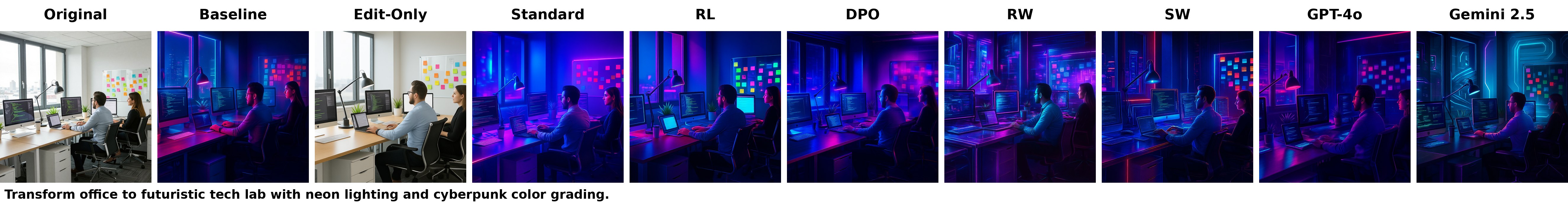}
\vspace{3pt}
\includegraphics[width=\textwidth]{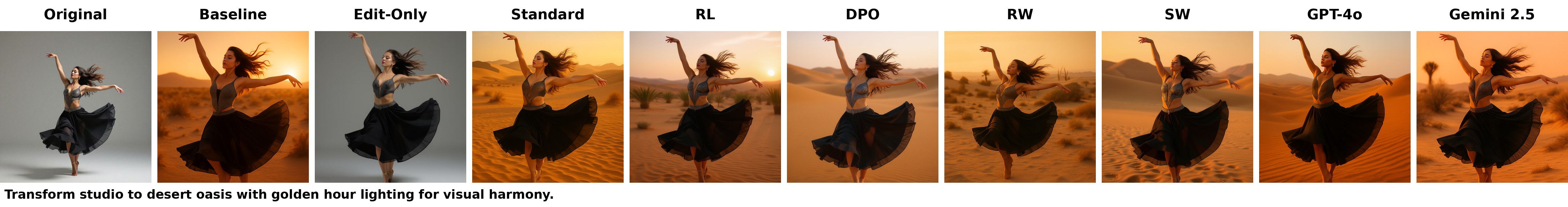}
\vspace{3pt}
\includegraphics[width=\textwidth]{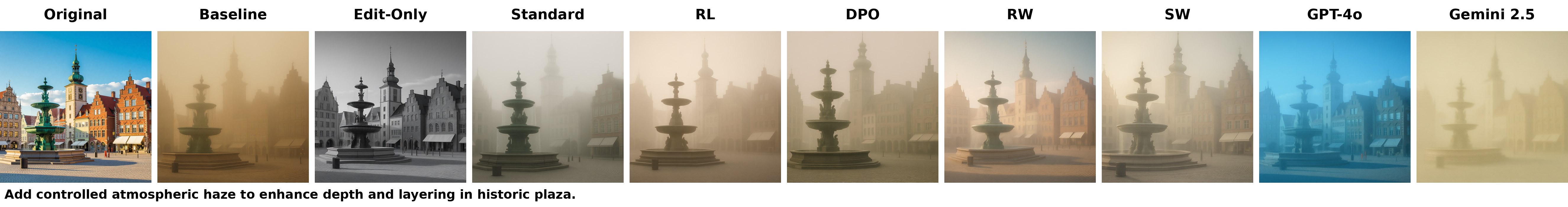}
\caption{\textbf{Complex dataset (Part 2/2) --- \sw winner.}
\textbf{Row~1 (Space Color Temperature, Vision-4B):} \sw applies the correct cool-blue
space lighting palette uniformly across the scene. Other trained methods produce inconsistent
colour patches or oversaturate specific regions; Gemini~2.5 applies global colour shift
without the correct spectral profile.
\textbf{Row~2 (Office Lens Character, Vision-4B):} \sw applies the requested lens
distortion and character-style framing with precise depth-of-field changes. Standard and RL
miss the lens component; Edit-Only applies character style without the spatial reframing.
\textbf{Row~3 (Dancer Architecture Style, Vision-4B):} \sw simultaneously transforms the
architectural backdrop and maintains the dancer's motion trajectory---a compositional
challenge where every other method either loses subject fidelity or ignores the background.
\textbf{Row~4 (Historic Atmospheric Depth, Vision-4B):} \sw produces layered historical
atmosphere---sepia grading, depth fog, period-consistent tonal palette---with correct
spatial recession. Other trained methods apply flat period colouring without the depth cue;
GPT-4o achieves partial atmospheric grading but loses the historic tonal signature.}
\label{fig:appendix_visual_complex_2}
\end{figure}

\clearpage

% ════════════════════════════════════════════════════════════════
% RELATED WORK (moved from main paper due to space constraints)
% ════════════════════════════════════════════════════════════════

\section{Related Work}
\label{sec:related_work}

Our work sits at the intersection of controllable image synthesis, agentic reasoning, and offline reinforcement learning. We briefly review the evolution of these fields to contextualize our Agentic RL framework.

\textbf{From Direct Editing to Agentic Planning.}
The paradigm of automated image styling has evolved from signal-level manipulation to semantic generation. Early neural approaches relied on optimization-based style transfer \citep{gatys2016image} or GAN-based image-to-image translation \citep{isola2017image, zhu2017unpaired}, which were effective but limited to specific domains. The advent of diffusion models introduced "Direct Prompt-Based Editing" (the \textit{Edit-Only} baseline), exemplified by \textbf{InstructPix2Pix} \citep{brooks2023instructpix2pix} and dataset efforts like \textbf{MagicBrush} \citep{zhang2023magicbrush}. While these end-to-end models excel at global style swaps, they lack the symbolic reasoning required for compositional tasks. Recent works like \textbf{StyleBooth} \citep{han2025stylebooth} and \textbf{StyleShot} \citep{gao2024styleshot} improved fidelity through exemplar guidance but remain bound by the "one-shot" generation paradigm, often failing to resolve conflicting constraints (e.g., "change weather but preserve architecture").

To address these structural limitations, the field is shifting toward \textit{Agentic AI}, where Large Multimodal Models (LMMs) act as planners. Frameworks like \textbf{RPG} (Recaption, Plan, Generate) \citep{yang2024mastering} and \textbf{DraCo} (Draft-as-CoT) \citep{jiang2025draco} demonstrate that decomposing generation into hierarchical sub-tasks significantly improves spatial adherence. Most recently, \citet{li2025univworldv2} (\textbf{UniWorld-V2/Edit-R1}) and \textbf{Agentic-Retoucher} \citep{shen2026agentic} have begun integrating Chain-of-Thought (CoT) reasoning directly into the editing loop, validating our hypothesis that explicit reasoning traces are essential for complex instruction following.

\textbf{Reinforcement Learning for Generative Models.}
Aligning generative models with human intent has traditionally relied on Reinforcement Learning from Human Feedback (RLHF) via PPO \citep{schulman2017ppo}. However, PPO is computationally expensive and unstable for high-dimensional visual tasks. This led to the adoption of \textbf{Direct Preference Optimization (DPO)} \citep{rafailov2023direct}, which optimizes policy likelihoods directly on preference pairs. \textbf{Diffusion-DPO} \citep{wallace2023diffusion} successfully adapted this to pixel-space denoising. However, recent critiques suggest DPO can suffer from mode collapse or fail to preserve structural integrity in editing tasks, prompting "safeguarded" variants like \textbf{Diffusion-SDPO} \citep{fu2025diffusionsdpo}.

\textbf{Reward-Weighted Methods in Offline RL.}
Our use of Reward-Weighted (RW) and Standardized Reward-Weighted (SW) fine-tuning draws on a lineage of Expectation-Maximization (EM) based RL. \citet{peters07reinforcement} originally formulated \textbf{Reward-Weighted Regression (RWR)} for robotic control, treating RL as a supervised regression problem on high-reward samples. This was later generalized by \textbf{Advantage-Weighted Regression (AWR)} \citep{peng20advantage}.
Most relevant to our work, \citet{mukherjee25offline} recently formalized \textbf{Reward-Weighted Fine-Tuning} and \textbf{Standardized Reward-Weighted (SWiFt)} for language models. They showed that reward-weighted log-probability maximization is a lower bound on the online RL objective and proposed optimizing it via weighted fine-tuning, essentially a form of policy gradient \citep{williams92simple}. Their analysis proves that these methods can optimize reward signals more stably than DPO in offline settings. Note that \citet{mukherjee25offline} looked into the offline RL reward-weighted algorithms in the context of conversation optimization. We extend these findings to the vision-language domain, demonstrating that for compositional planning tasks—where "correctness" is often binary and logic-driven—scalar reward weighting (RW/SW) provides a denser, more effective training signal than preference ranking.

\textbf{Agentic Planning with Spatial Grounding.}
Recent work on complex image editing includes \textbf{X-Planner} \citep{yeh2025beyond}, which introduces a planner-localizer framework that decomposes high-level instructions into sub-tasks with automatically generated spatial guidance (segmentation masks and bounding boxes). These spatial annotations guide specialized editing models (e.g., UltraEdit \citep{zhao2024ultraedit}, PowerPaint \citep{zhuang2024task}) to execute localized edits. X-Planner is trained on COMPIE, a large-scale dataset of 260K complex-simple instruction pairs, using standard supervised fine-tuning for grounded segmentation and reasoning. Our approach differs in two key ways: (1) \textbf{Execution mechanism}—we use symbolic tool calls synthesized into natural language instructions for a frozen black-box editor, avoiding spatial grounding or specialized editing models; (2) \textbf{Training methodology}—we employ offline RL with reward-weighted fine-tuning to prioritize high-quality trajectories rather than uniform supervised learning on all instruction pairs.

\clearpage

% ════════════════════════════════════════════════════════════════
% APPENDIX SECTIONS
% ════════════════════════════════════════════════════════════════

\section{Complete Problem Formulation Details}
\label{sec:appendix_problem_formulation}

This appendix provides comprehensive specifications for the problem formulation, action spaces, reward function, and synthetic data generation pipeline.

\subsection{Context Representation Details}
\label{sec:appendix_context_details}

The structured context representation $c_i = \{d_1, \dots, d_{10}\}$ encodes an image's current visual state across 10 dimensions. Each dimension $d_j$ is extracted via a frozen Qwen3-VL-8B-Instruct model using vision-language understanding.

\subsubsection{Dimension Specifications}

\begin{description}[leftmargin=0pt,style=nextline,itemsep=0.5em]
\item[\textbf{1. Location} ($d_{\text{loc}}$)] Physical environment type where the scene takes place.
  
  \textit{Example values:} urban\_city, suburban\_neighborhood, rural\_village, industrial\_zone, beach\_coast, tropical\_island, forest\_temperate, desert\_sand, mountain\_rocky, cave\_underground, space\_station, fantasy\_castle, medieval\_town, cyberpunk\_city, bedroom\_interior, office\_modern.

\item[\textbf{2. Architecture} ($d_{\text{arch}}$)] Architectural style of buildings and structures.
  
  \textit{Example values:} modern\_minimalist, classical\_greek, victorian\_gothic, art\_deco, brutalist\_concrete, traditional\_asian, middle\_eastern, industrial\_warehouse, futuristic\_sci\_fi, cyberpunk\_neon, medieval\_castle, baroque\_ornate.

\item[\textbf{3. Time Period Era} ($d_{\text{era}}$)] Historical or futuristic time period reflected in props, technology, and visual style.
  
  \textit{Example values:} prehistoric, ancient\_classical, medieval\_dark\_ages, renaissance, victorian\_1800s, early\_1900s, mid\_century\_1950s, modern\_2000s, near\_future\_2050s, far\_future\_2200s.

\item[\textbf{4. Time of Day} ($d_{\text{time}}$)] Natural lighting from sun/moon position.
  
  \textit{Example values:} dawn\_first\_light, morning\_golden, midday\_overhead, afternoon\_warm, sunset\_golden, dusk\_twilight, night\_moonlit, night\_starlit, overcast\_diffuse.

\item[\textbf{5. Season} ($d_{\text{season}}$)] Seasonal markers in vegetation, weather, and atmosphere.
  
  \textit{Example values:} spring\_blooming, summer\_lush, autumn\_falling, winter\_snow, dry\_season, wet\_season, eternal\_spring (fantasy).

\item[\textbf{6. Weather} ($d_{\text{weather}}$)] Atmospheric weather conditions.
  
  \textit{Example values:} clear\_sky, partly\_cloudy, overcast\_gray, light\_rain, heavy\_rain, thunderstorm, light\_snow, blizzard, fog\_heavy, mist\_light, dust\_storm, hazy.

\item[\textbf{7. Mood Lighting} ($d_{\text{mood}}$)] Emotional ambiance conveyed through lighting and atmosphere.
  
  \textit{Example values:} neutral\_balanced, warm\_cozy, cool\_calm, dramatic\_contrast, mysterious\_dark, ethereal\_soft, tense\_harsh, romantic\_soft, energetic\_bright, melancholic\_muted, ominous\_dark, serene\_peaceful.

\item[\textbf{8. Color Grading} ($d_{\text{color}}$)] Overall color palette and correction.
  
  \textit{Example values:} natural\_balanced, warm\_cinematic, cool\_blue, sepia\_vintage, black\_white, high\_contrast, low\_saturation, vibrant\_saturated, teal\_orange, purple\_magenta, desaturated\_muted, neon\_bright, pastel\_soft.

\item[\textbf{9. Artistic Medium} ($d_{\text{medium}}$)] Rendering style and artistic technique.
  
  \textit{Example values:} photorealistic, oil\_painting, watercolor, pencil\_sketch, digital\_art, anime\_style, pixel\_art, impressionist, abstract, low\_poly\_3d, clay\_animation, charcoal\_drawing, comic\_book.

\item[\textbf{10. Atmospheric Effects} ($d_{\text{atmos}}$)] Environmental effects and particles.
  
  \textit{Example values:} none\_clear, fog\_dense, mist\_light, haze\_atmospheric, dust\_particles, smoke\_wisps, rain\_drops, snow\_falling, embers\_floating, sparkles\_magical, lens\_flare, light\_rays.
\end{description}

\subsubsection{Extraction Process}

The context extraction process queries Qwen3-VL-8B-Instruct with a structured prompt:

\begin{promptbox}[title=Context Extraction Prompt]
\small\ttfamily
Analyze this image and extract the following 10 visual attributes\\
in JSON format:\\
\\
\{\\
\hspace*{1em}"location": "<categorical\_value>",\\
\hspace*{1em}"architecture": "<categorical\_value>",\\
\hspace*{1em}"time\_period\_era": "<categorical\_value>",\\
\hspace*{1em}"time\_of\_day": "<categorical\_value>",\\
\hspace*{1em}"season": "<categorical\_value>",\\
\hspace*{1em}"weather": "<categorical\_value>",\\
\hspace*{1em}"mood\_lighting": "<categorical\_value>",\\
\hspace*{1em}"color\_grading": "<categorical\_value>",\\
\hspace*{1em}"artistic\_medium": "<categorical\_value>",\\
\hspace*{1em}"atmospheric\_effects": "<categorical\_value>"\\
\}\\
\\
Be specific and use the standardized vocabulary.
\end{promptbox}

The model returns structured JSON which is parsed into $c_i$. Extraction takes approximately 2-3 seconds per image on an A100 GPU. This explicit symbolic representation provides the planner with state awareness that pure vision encoding may miss.

\subsection{Action Space Specification}
\label{sec:appendix_action_space}

We define two action libraries: a 10-action core library for the normal dataset, and an extended 20-action library for the complex dataset.

\subsubsection{Simple Dataset: 10 Atomic Actions}

\paragraph{1. Location Setting ($a_{\text{loc}}$)}
\textbf{Description:} Changes physical environment type (e.g., urban city $\to$ tropical beach).

\textbf{Parameters:}
\begin{itemize}
\item \texttt{source\_location}: Current location type
\item \texttt{target\_location}: Desired location type
\item \texttt{replace\_mode}: "partial" (blend) or "complete" (full replacement)
\item \texttt{preserve\_foreground}: Boolean, keep main subjects unchanged
\item \texttt{description}: Natural language explanation
\end{itemize}

\textbf{Example:} 
$$a = (\text{location\_setting}, \{\text{source}=\text{"urban\_city"}, \text{target}=\text{"tropical\_beach"}, \text{mode}=\text{"complete"}\})$$

\paragraph{2. Architecture Style ($a_{\text{arch}}$)}
\textbf{Description:} Modifies building architectural style (e.g., modern $\to$ Victorian).

\textbf{Parameters:}
\begin{itemize}
\item \texttt{source\_style}: Current architectural style
\item \texttt{target\_style}: Desired architectural style
\item \texttt{detail\_level}: "subtle", "moderate", or "extensive"
\item \texttt{preserve\_layout}: Boolean, keep spatial structure
\item \texttt{description}: Natural language explanation
\end{itemize}

\paragraph{3. Time Period Era ($a_{\text{era}}$)}
\textbf{Description:} Updates props and technology to match historical/future period (e.g., 2000s $\to$ 1800s).

\textbf{Parameters:}
\begin{itemize}
\item \texttt{source\_era}: Current time period
\item \texttt{target\_era}: Desired time period
\item \texttt{technology\_update}: Boolean, change technology/vehicles
\item \texttt{clothing\_update}: Boolean, update character clothing
\item \texttt{description}: Natural language explanation
\end{itemize}

\paragraph{4. Time of Day ($a_{\text{time}}$)}
\textbf{Description:} Adjusts natural lighting from sun/moon position (e.g., midday $\to$ sunset).

\textbf{Parameters:}
\begin{itemize}
\item \texttt{source\_time}: Current time of day
\item \texttt{target\_time}: Desired time of day
\item \texttt{sky\_color}: Target sky color palette
\item \texttt{shadow\_direction}: Shadow angle adjustment
\item \texttt{description}: Natural language explanation
\end{itemize}

\paragraph{5. Season Cycle ($a_{\text{season}}$)}
\textbf{Description:} Changes vegetation and seasonal markers (e.g., summer $\to$ autumn).

\textbf{Parameters:}
\begin{itemize}
\item \texttt{source\_season}: Current season
\item \texttt{target\_season}: Desired season
\item \texttt{vegetation\_change}: "foliage\_color", "density", "type"
\item \texttt{temperature\_effects}: "snow", "heat\_haze", "none"
\item \texttt{description}: Natural language explanation
\end{itemize}

\paragraph{6. Weather Conditions ($a_{\text{weather}}$)}
\textbf{Description:} Modifies atmospheric weather state (e.g., clear $\to$ rainy).

\textbf{Parameters:}
\begin{itemize}
\item \texttt{source\_weather}: Current weather condition
\item \texttt{target\_weather}: Desired weather condition
\item \texttt{intensity}: "light", "moderate", "heavy"
\item \texttt{visibility\_change}: Boolean, affect scene visibility
\item \texttt{description}: Natural language explanation
\end{itemize}

\paragraph{7. Mood Lighting ($a_{\text{mood}}$)}
\textbf{Description:} Alters emotional ambiance through lighting (e.g., neutral $\to$ dramatic).

\textbf{Parameters:}
\begin{itemize}
\item \texttt{source\_mood}: Current mood/atmosphere
\item \texttt{target\_mood}: Desired mood/atmosphere
\item \texttt{contrast\_adjustment}: "increase", "decrease", "none"
\item \texttt{shadow\_depth}: Darkness of shadows
\item \texttt{description}: Natural language explanation
\end{itemize}

\paragraph{8. Color Grading ($a_{\text{color}}$)}
\textbf{Description:} Applies color correction and palette shifts (e.g., natural $\to$ warm cinematic).

\textbf{Parameters:}
\begin{itemize}
\item \texttt{source\_grading}: Current color grading
\item \texttt{target\_grading}: Desired color grading
\item \texttt{saturation\_change}: Adjustment to color intensity
\item \texttt{temperature\_shift}: "warmer", "cooler", "neutral"
\item \texttt{description}: Natural language explanation
\end{itemize}

\paragraph{9. Artistic Medium ($a_{\text{medium}}$)}
\textbf{Description:} Transforms rendering style (e.g., photorealistic $\to$ oil painting).

\textbf{Parameters:}
\begin{itemize}
\item \texttt{source\_medium}: Current artistic style
\item \texttt{target\_medium}: Desired artistic style
\item \texttt{detail\_preservation}: Boolean, keep fine details
\item \texttt{texture\_intensity}: Strength of artistic texture
\item \texttt{description}: Natural language explanation
\end{itemize}

\paragraph{10. Atmospheric Effects ($a_{\text{atmos}}$)}
\textbf{Description:} Adds environmental effects (e.g., fog, dust, haze).

\textbf{Parameters:}
\begin{itemize}
\item \texttt{source\_effects}: Current atmospheric effects
\item \texttt{target\_effects}: Desired atmospheric effects
\item \texttt{density}: "sparse", "moderate", "dense"
\item \texttt{distribution}: "uniform", "localized", "gradient"
\item \texttt{description}: Natural language explanation
\end{itemize}

\subsubsection{Regular Dataset: 20 Actions (10 Atomic + 10 Compositional)}

The complex dataset extends the action library with 10 additional compositional and constraint actions designed for sophisticated multi-step transformations:

\paragraph{11. Preserve Attribute ($a_{\text{preserve}}$)}
\textbf{Description:} Explicitly preserves specific visual attributes while other transformations occur.

\textbf{Parameters:}
\begin{itemize}
\item \texttt{attributes\_to\_preserve}: List of dimension names (e.g., ["time\_of\_day", "color\_grading"])
\item \texttt{preservation\_strength}: "strict", "moderate", "soft"
\item \texttt{description}: Natural language explanation
\end{itemize}

\textbf{Example Use Case:} "Transform to Victorian architecture while preserving the sunset lighting"

\paragraph{12. Exclude Region ($a_{\text{exclude}}$)}
\textbf{Description:} Masks specific spatial regions from transformation.

\textbf{Parameters:}
\begin{itemize}
\item \texttt{region\_type}: "foreground", "background", "top\_half", "bottom\_half", "center", "edges"
\item \texttt{exclusion\_strength}: "complete", "partial"
\item \texttt{description}: Natural language explanation
\end{itemize}

\textbf{Example Use Case:} "Change background to cyberpunk city but exclude foreground characters"

\paragraph{13. Conditional Transform ($a_{\text{conditional}}$)}
\textbf{Description:} Applies transformation only if a condition is met.

\textbf{Parameters:}
\begin{itemize}
\item \texttt{condition\_type}: "if\_attribute\_equals", "if\_region\_contains", "if\_lighting\_level"
\item \texttt{condition\_value}: Value to check
\item \texttt{then\_action}: Action to apply if condition is true
\item \texttt{description}: Natural language explanation
\end{itemize}

\textbf{Example Use Case:} "If current time is daytime, then add sunset; otherwise keep night lighting"

\paragraph{14. Preserve Object Category ($a_{\text{preserve\_obj}}$)}
\textbf{Description:} Preserves all objects of a specific semantic category.

\textbf{Parameters:}
\begin{itemize}
\item \texttt{object\_categories}: List of categories (e.g., ["person", "vehicle", "animal"])
\item \texttt{preservation\_mode}: "identity", "style\_only"
\item \texttt{description}: Natural language explanation
\end{itemize}

\textbf{Example Use Case:} "Transform entire scene to oil painting style but keep people photorealistic"

\paragraph{15. Spatial Constraint ($a_{\text{spatial}}$)}
\textbf{Description:} Applies transformation with spatial constraints.

\textbf{Parameters:}
\begin{itemize}
\item \texttt{constraint\_type}: "top\_to\_bottom\_gradient", "center\_outward", "left\_to\_right"
\item \texttt{affected\_attribute}: Which dimension to transform
\item \texttt{gradient\_sharpness}: "smooth", "moderate", "sharp"
\item \texttt{description}: Natural language explanation
\end{itemize}

\textbf{Example Use Case:} "Apply sunset lighting with gradient from top (bright) to bottom (darker)"

\paragraph{16. Sequence Transform ($a_{\text{sequence}}$)}
\textbf{Description:} Specifies explicit ordering of multiple sub-transformations.

\textbf{Parameters:}
\begin{itemize}
\item \texttt{sub\_actions}: Ordered list of actions to apply sequentially
\item \texttt{timing}: "simultaneous", "sequential"
\item \texttt{description}: Natural language explanation
\end{itemize}

\textbf{Example Use Case:} "First change to autumn, then add rain, then shift to dramatic mood"

\paragraph{17. Parallel Transform ($a_{\text{parallel}}$)}
\textbf{Description:} Applies multiple transformations simultaneously.

\textbf{Parameters:}
\begin{itemize}
\item \texttt{parallel\_actions}: List of actions to apply in parallel
\item \texttt{blending\_mode}: "additive", "average", "weighted"
\item \texttt{description}: Natural language explanation
\end{itemize}

\textbf{Example Use Case:} "Simultaneously change to sunset, add fog, and shift to warm color grading"

\paragraph{18. Graduated Effect ($a_{\text{graduated}}$)}
\textbf{Description:} Applies effect with gradual intensity variation.

\textbf{Parameters:}
\begin{itemize}
\item \texttt{base\_action}: The action to apply with graduation
\item \texttt{gradient\_direction}: "top\_to\_bottom", "center\_outward", etc.
\item \texttt{intensity\_curve}: "linear", "exponential", "sigmoid"
\item \texttt{description}: Natural language explanation
\end{itemize}

\textbf{Example Use Case:} "Add fog with graduated intensity from dense at bottom to clear at top"

\paragraph{19. Layered Transformation ($a_{\text{layered}}$)}
\textbf{Description:} Applies transformations in layers with specified blending.

\textbf{Parameters:}
\begin{itemize}
\item \texttt{layers}: List of (action, opacity) tuples
\item \texttt{blend\_mode}: "normal", "multiply", "screen", "overlay"
\item \texttt{description}: Natural language explanation
\end{itemize}

\textbf{Example Use Case:} "Layer Victorian architecture (70\% opacity) over modern city, then add sunset"

\paragraph{20. Selective Blend ($a_{\text{selective\_blend}}$)}
\textbf{Description:} Blends transformation results based on semantic or spatial criteria.

\textbf{Parameters:}
\begin{itemize}
\item \texttt{blend\_criterion}: "by\_semantic\_region", "by\_depth", "by\_lighting\_level"
\item \texttt{source\_transform}: First transformation result
\item \texttt{target\_transform}: Second transformation result
\item \texttt{blend\_ratio}: Mixing ratio or function
\item \texttt{description}: Natural language explanation
\end{itemize}

\textbf{Example Use Case:} "Blend cyberpunk architecture with Victorian based on depth: close objects are cyberpunk, distant objects are Victorian"

These 10 additional compositional actions enable sophisticated multi-step transformations with explicit control over preservation, exclusion, ordering, and blending—critical for complex creative workflows.

\subsection{Reward Function Details}
\label{sec:appendix_reward_details}

The reward function $r_i \in [0, 5]$ is computed by Qwen3-VL-8B-Instruct analyzing the transformation from base image $I_i$ to edited image $\hat{I}_i$ given image editing prompt $e_i$.

\subsubsection{Reward Criteria}

The reward model evaluates six primary criteria with weighted importance:

\begin{enumerate}[leftmargin=1.5em,itemsep=0.5em]
\item \textbf{Goal Alignment} (Weight: 30\% --- \textit{Most Critical})
  \begin{itemize}[leftmargin=1.5em]
\item Measures semantic alignment between image editing prompt $e_i$ and edited result $\hat{I}_i$
  \item Evaluates completeness: Did the transformation achieve what was requested?
  \item Assesses accuracy: Are the specific attributes correctly transformed?
  \item This is the single most important criterion for task success
  \end{itemize}

\item \textbf{Aesthetic Quality} (Weight: 25\%)
  \begin{itemize}[leftmargin=1.5em]
  \item Visual appeal and artistic merit of the edited image
  \item Composition balance, rule of thirds, visual flow
  \item Color harmony and palette coherence
  \item Overall professional polish
  \end{itemize}

\item \textbf{Spatial Consistency} (Weight: 15\%)
  \begin{itemize}[leftmargin=1.5em]
  \item Coherence of spatial relationships and depth ordering
  \item Perspective correctness and vanishing point consistency
  \item Geometric plausibility of transformed elements
  \item Absence of spatial distortions or impossible geometry
  \end{itemize}

\item \textbf{Technical Quality} (Weight: 15\%)
  \begin{itemize}[leftmargin=1.5em]
  \item Absence of visual artifacts (blurring, aliasing, noise)
  \item Resolution quality and detail preservation
  \item Edge sharpness and boundary cleanliness
  \item Technical execution (no broken textures, seams, or discontinuities)
  \end{itemize}

\item \textbf{Temporal Consistency} (Weight: 10\%)
  \begin{itemize}[leftmargin=1.5em]
  \item Consistency of time-related attributes (time of day, season)
  \item Interdependencies: sunset implies warm lighting, winter implies cold tones
  \item Logical coherence of lighting direction with stated time of day
  \item Seasonal markers align with requested season
  \end{itemize}

\item \textbf{Creative Interpretation} (Weight: 5\%)
  \begin{itemize}[leftmargin=1.5em]
  \item Novelty and creativity when interpreting ambiguous goals
  \item Maintaining plausibility while being creative
  \item Handling under-specified requests gracefully
  \item Appropriate artistic liberty within the goal's intent
  \end{itemize}
\end{enumerate}

\subsubsection{Reward Thresholds}

Reward scores define quality tiers that inform our training methods:

\begin{table}[h]
\centering
\caption{Reward Thresholds and Training Usage}
\begin{tabular}{llll}
\toprule
\textbf{Score Range} & \textbf{Quality Tier} & \textbf{\rw Weight (at $r$)} & \textbf{\dpo Usage} \\
\midrule
$[4.5, 5.0]$ & Excellent & 2.0 (at $r{=}5.0$) & Chosen \\
$[4.0, 4.5)$ & Good & 1.5 (at $r{=}4.5$) & Chosen \\
$[3.5, 4.0)$ & Medium & 1.0 (at $r{=}4.0$) & Neutral \\
$[3.0, 3.5)$ & Poor & 0.5 (at $r{=}3.5$) & Rejected \\
$[0, 3.0)$ & Very Poor & 0.0 & Rejected \\
\bottomrule
\end{tabular}
\end{table}

\textbf{\rw:} Uses continuous weight function $w(r) = \max\{r - 3.0, 0\}$. Each trajectory's gradient contribution is weighted by its reward score, so higher-quality examples receive proportionally more influence during training.

\textbf{Direct Preference Optimization (\dpo):} Trajectories with $r_i \geq 4.0$ are chosen examples; those with $r_i < 3.5$ are rejected examples. \dpo learns from contrastive pairs sharing the same $(I_i, e_i)$.

\subsection{Synthetic Data Generation Details}
\label{sec:appendix_data_generation}

This section provides comprehensive implementation details for our 4-stage synthetic data generation pipeline.

\subsubsection{Stage 1: Image Generation with HiDream-I1-Dev}

\paragraph{Model Specification}
We use HiDream-I1-Dev, a state-of-the-art text-to-image diffusion model:
\begin{itemize}
\item Architecture: Latent Diffusion Model with U-Net backbone
\item Resolution: 1024×1024 pixels
\item Inference steps: 50
\item Guidance scale: 7.5
\item Sampling method: DPM-Solver++
\end{itemize}

\paragraph{Prompt Generation Strategy}
We generate diverse seed prompts $p_i$ covering:
\begin{itemize}
\item \textbf{Location types (30 variants):} urban\_city, suburban\_neighborhood, rural\_village, beach\_coast, forest\_temperate, desert\_sand, mountain\_rocky, cave\_underground, space\_station, cyberpunk\_city, medieval\_town, office\_modern, bedroom\_interior, etc.
\item \textbf{Architectural styles (25 variants):} modern\_minimalist, classical\_greek, victorian\_gothic, art\_deco, traditional\_asian, industrial\_warehouse, futuristic\_sci\_fi, etc.
\item \textbf{Time periods (15 variants):} ancient\_classical, medieval, victorian\_1800s, modern\_2000s, near\_future\_2050s, etc.
\item \textbf{Lighting conditions (20 variants):} dawn, morning, midday, afternoon, sunset, dusk, night, overcast, etc.
\end{itemize}

Prompts are constructed using templates:
\begin{verbatim}
"A {location} with {architecture} architecture in {time_period} era 
at {time_of_day} with {weather} weather"
\end{verbatim}

Example: "A suburban neighborhood with modern minimalist architecture in the 2000s era at midday with clear weather"

\subsubsection{Stage 2: Context Extraction}

We extract structured context $c_i$ using a frozen Qwen3-VL-8B-Instruct model with the following prompt template:

\begin{promptbox}[title=Context Extraction Prompt]
\small\ttfamily
Analyze this image and extract the following 10 visual attributes in JSON format:\\
\\
\{\\
\hspace*{1em}"location": "<categorical\_value>",\\
\hspace*{1em}"architecture": "<categorical\_value>",\\
\hspace*{1em}"time\_period\_era": "<categorical\_value>",\\
\hspace*{1em}"time\_of\_day": "<categorical\_value>",\\
\hspace*{1em}"season": "<categorical\_value>",\\
\hspace*{1em}"weather": "<categorical\_value>",\\
\hspace*{1em}"mood\_lighting": "<categorical\_value>",\\
\hspace*{1em}"color\_grading": "<categorical\_value>",\\
\hspace*{1em}"artistic\_medium": "<categorical\_value>",\\
\hspace*{1em}"atmospheric\_effects": "<categorical\_value>"\\
\}\\
\\
Be specific and use the standardized vocabulary.
\end{promptbox}

The model returns structured JSON which we parse into $c_i$. Extraction takes approximately 2-3 seconds per image on an A100 GPU.

\subsubsection{Stage 3: Action Planning with Teacher Model}

The teacher planner (Qwen3-VL-8B-Instruct) generates action sequences using the following algorithm:

\begin{algorithm}[H]
\caption{Teacher Trajectory Generation (Detailed)}
\begin{algorithmic}[1]
\STATE \textbf{Input:} Image $I_i$, image editing prompt $e_i$, context $c_i$, action library $\mathcal{A}$
\STATE Initialize action sequence $\{a_{i1}, \dots, a_{im}\}$ and reasoning $\{z_{i1}, \dots, z_{im}\}$
\STATE Initialize current context $c_{\text{current}} \gets c_i$
\FOR{$j = 1$ to $m_{\text{max}} = 5$}
    \STATE Construct prompt with image $I_i$, editing prompt $e_i$, context $c_{\text{current}}$, and past actions $\{a_{ik}\}_{k<j}$
    \STATE Sample action and reasoning: $(a_{i,j}, z_{i,j}) \sim \pi_{\text{teacher}}(\cdot \mid I_i, e_i, c_{\text{current}}, \{a_{ik}\}_{k<j})$ with temperature $T=0.7$
    \IF{$a_{i,j} = a_{\text{STOP}}$ \textbf{or} goal satisfied (checked by teacher)}
        \STATE \textbf{break}
    \ENDIF
    \STATE Update context: $c_{\text{current}} \gets \text{ApplyAction}(c_{\text{current}}, a_{i,j})$
\ENDFOR
\STATE Generate natural language instruction: $\hat{e}_i = \text{ActionToNL}(\{a_{i,j}\}_{j=1}^m, e_i, c_i)$
\STATE \textbf{Return:} $\{a_{i,j}\}_{j=1}^m$, $\{z_{i,j}\}_{j=1}^m$, $\hat{e}_i$
\end{algorithmic}
\end{algorithm}

\paragraph{Prompt Template for Planning}

\begin{promptbox}[title=Action Planning Prompt Template]
\small\ttfamily
You are an expert image styling agent. Given:\\
\\
- Base image: [image]\\
- Current visual state: \{c\_i\}\\
- User goal: \{e\_i\}\\
- Past actions: \{a\_i1, a\_i2, ...\}\\
\\
Generate the next action from the action library to achieve the goal.\\
\\
For each action, provide:\\
\\
1. Action type and parameters\\
2. Chain-of-thought reasoning explaining why this action is chosen\\
\\
Action library: [list of available actions with specs]
\end{promptbox}

\paragraph{Temperature Sampling}
We use temperature $T=0.7$ to balance diversity and quality. Lower temperatures ($T=0.1$) produce repetitive trajectories; higher temperatures ($T=1.0$) reduce coherence.

\subsubsection{Stage 4: Image Editing with Qwen-Image-Edit}

We execute the synthesized instruction $\hat{e}_i$ using Qwen-Image-Edit:

\paragraph{Model Specification}
\begin{itemize}
\item Architecture: Instruction-conditioned diffusion model
\item Base model: Qwen-VL-Chat fine-tuned for editing
\item Resolution: 768×768 pixels
\item Inference steps: 28
\item Guidance scale: 7.5
\item Image guidance scale: 4.0
\end{itemize}

\paragraph{Instruction Synthesis}
We convert action sequences to natural language:

\textbf{Example 1:}
\begin{itemize}
\item Actions: $\{\text{time\_of\_day}(\text{sunset}), \text{season}(\text{autumn})\}$
\item Instruction: "Change the lighting to warm sunset tones with golden hour ambiance, and transform the scene to autumn with falling leaves and warm colors"
\end{itemize}

\textbf{Example 2:}
\begin{itemize}
\item Actions: $\{\text{architecture}(\text{victorian}), \text{mood}(\text{mysterious}), \text{atmospheric}(\text{fog})\}$
\item Instruction: "Transform the buildings to Victorian Gothic architecture with ornate details, add mysterious dramatic lighting with deep shadows, and add dense fog throughout the scene"
\end{itemize}

\subsubsection{Stage 5: Reward Evaluation}

Qwen3-VL-8B-Instruct evaluates trajectory quality with the following prompt:

\begin{promptbox}[title=Reward Evaluation Prompt]
\small\ttfamily
Evaluate this image transformation:\\
- Base image: [I\_i]\\
- Edited image: [E\_i]\\
- User goal: \{e\_i\}\\
- Planned actions: \{a\_i1, a\_i2, ...\}\\
\\
Rate the transformation quality on a scale of 0-5 considering:\\
1. Goal alignment (30\% --- most critical)\\
2. Aesthetic quality (25\%)\\
3. Spatial consistency (15\%)\\
4. Technical quality (15\%)\\
5. Temporal consistency (10\%)\\
6. Creative interpretation (5\%)\\
\\
Provide:\\
- Overall score (0-5)\\
- Per-criterion scores\\
- Justification for the rating\\
\\
Format: JSON with fields \{overall\_score, aesthetic, goal\_alignment, ...\}
\end{promptbox}

The model returns structured JSON evaluation which we parse to extract $r_i \in [0, 5]$. Evaluation takes approximately 3-4 seconds per trajectory on an A100 GPU.

\subsubsection{Dataset Statistics}

\begin{table}[h]
\centering
\caption{Dataset Statistics}
\begin{tabular}{lccc}
\toprule
\textbf{Statistic} & \textbf{Simple Dataset} & \textbf{Regular Dataset} & \textbf{Complex Dataset} \\
\midrule
Total trajectories & 9,824 & 10,142 & 10,000 \\
Avg. actions per trajectory & 2.8 & 4.1 & 2.9 \\
Avg. reward score & 3.92 & 3.74 & 3.81 \\
\% High quality ($r \geq 4.0$) & 62\% & 48\% & 55\% \\
Action library size & 10 actions & 20 actions & 30 actions \\
Train/Val/Test split & 80\%/10\%/10\% & 80\%/10\%/10\% & 80\%/10\%/10\% \\
Trajectory groups & 3,247 & 2,891 & 3,012 \\
Avg. trajectories per group & 3.0 & 3.5 & 3.3 \\
\bottomrule
\end{tabular}
\end{table}

\paragraph{Trajectory-Level Splitting}
To prevent data leakage, we group trajectories by $(I_i, e_i)$ pairs and split at the group level. This ensures that all alternative plans for the same image-goal combination remain in the same split, forcing the model to generalize to unseen combinations rather than memorizing specific inputs.

\section{Complete Synthesis Pipeline Examples}
\label{sec:appendix_synthesis_examples}

This section provides two complete end-to-end examples of our synthetic data generation pipeline, illustrating the entire 4-stage process from context extraction to reward evaluation. We present one example from the Simple Dataset (atomic transformations) and one from the Regular Dataset (compositional reasoning with constraints).

% Define custom boxes for visual appeal
\newtcolorbox{jsonbox}[1][]{
  colback=blue!5,
  colframe=blue!50,
  fonttitle=\bfseries\color{blue!70!black},
  boxrule=0.8pt,
  arc=2pt,
  left=6pt,
  right=6pt,
  top=6pt,
  bottom=6pt,
  breakable,
  #1
}

\newtcolorbox{reasoningbox}[1][]{
  colback=orange!5,
  colframe=orange!50,
  fonttitle=\bfseries\color{orange!70!black},
  boxrule=0.8pt,
  arc=2pt,
  left=8pt,
  right=8pt,
  top=6pt,
  bottom=6pt,
  breakable,
  #1
}

\newtcolorbox{scorebox}[1][]{
  colback=green!5,
  colframe=green!50,
  fonttitle=\bfseries\color{green!70!black},
  boxrule=0.8pt,
  arc=2pt,
  left=8pt,
  right=8pt,
  top=6pt,
  bottom=6pt,
  breakable,
  #1
}

\newtcolorbox{actionbox}[1][]{
  colback=purple!5,
  colframe=purple!50,
  fonttitle=\bfseries\color{purple!70!black},
  boxrule=0.8pt,
  arc=2pt,
  left=8pt,
  right=8pt,
  top=6pt,
  bottom=6pt,
  breakable,
  #1
}

\subsection{Example 1: Simple Dataset --- Autumn Vineyard to Spring Tulip Field}
\label{sec:synthesis_example_normal}

\subsubsection{Overview}

\textbf{Transformation:} Autumn Vineyard with grape harvest $\to$ Spring Tulip Field with emerging technology

\textbf{Complexity Level:} Simple (3 atomic actions, no constraints)

\textbf{Key Challenge:} Complete environmental transformation across multiple dimensions (location, season, era)

\textbf{Dataset:} Simple --- ID 2530

\paragraph{Stage 1: Base Image Generation and Final Result}

The base image was generated using HiDream-I1-Dev with the prompt: \textit{"An autumn vineyard with grape harvest, golden vine leaves, wine barrels, and fall agricultural atmosphere, in near future 2050 style."}

\begin{figure}[h!]
\centering
\begin{tabular}{cc}
\includegraphics[width=0.4\textwidth]{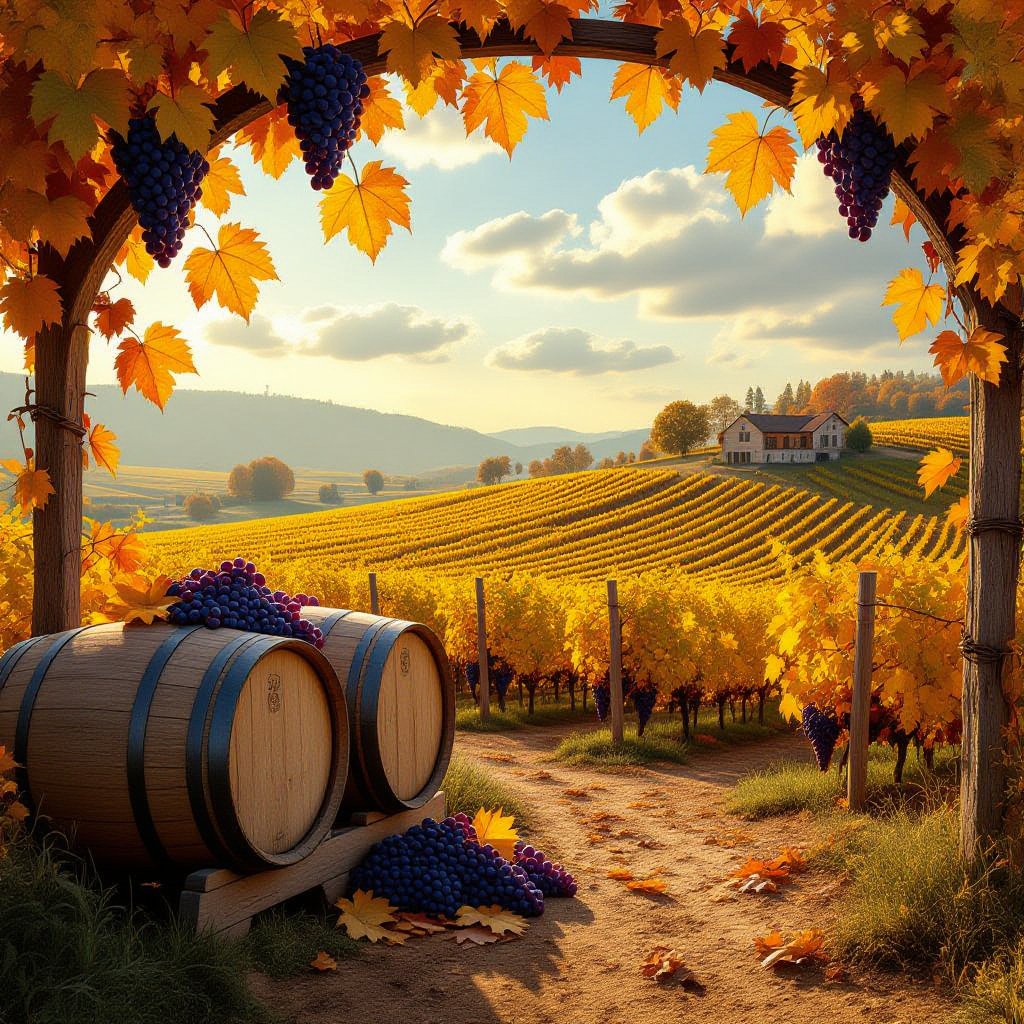} &
\includegraphics[width=0.4\textwidth]{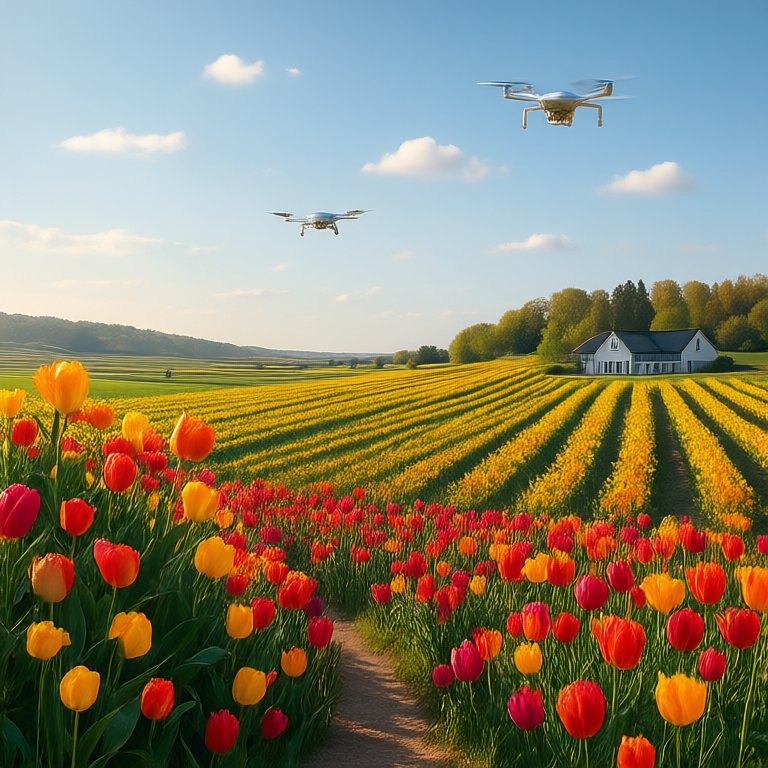} \\
\textbf{(a) Original Image} & \textbf{(b) Edited Image} \\
\end{tabular}
\caption{\textbf{Simple Dataset Example - Autumn Vineyard to Spring Tulip Field.} (a) Original: Autumn vineyard scene with golden foliage, wine barrels, grape clusters, and a rustic farmhouse. (b) Edited: Transformed to spring tulip field with vibrant pink tulips, modern technology, and clear morning atmosphere. The transformation successfully replaces location (vineyard→tulip field), season (autumn→spring), and era (modern→futuristic) while maintaining compositional structure.}
\label{fig:synthesis_normal}
\end{figure}

\paragraph{Stage 2: Context Extraction:}

Using Qwen3-VL-8B-Instruct, we extract the 10-dimensional structured context representation $c_i$:

\begin{jsonbox}[title=Extracted Context ($c_i$)]
\small\ttfamily
\{\\
\hspace*{1em}"location": "autumn\_vineyard",\\
\hspace*{1em}"architecture": "rustic\_farmhouse",\\
\hspace*{1em}"time\_period\_era": "modern\_2000s",\\
\hspace*{1em}"time\_of\_day": "golden\_hour\_sunset",\\
\hspace*{1em}"season": "autumn\_falling",\\
\hspace*{1em}"weather": "clear\_sky",\\
\hspace*{1em}"mood\_lighting": "warm\_cozy",\\
\hspace*{1em}"color\_grading": "warm\_cinematic",\\
\hspace*{1em}"artistic\_medium": "photorealistic",\\
\hspace*{1em}"atmospheric\_effects": "none\_clear"\\
\}
\end{jsonbox}

\vspace{0.3em}

\textbf{Scene Description:} A picturesque autumnal vineyard framed by a lush grapevine arch. Two wooden wine barrels sit prominently in the foreground with fresh grape clusters. Orderly rows of golden-hued grapevines cascade down gentle slopes toward a solitary farmhouse. The sky is a serene blend of soft blue with warm sunlight, creating a tranquil harvest-time atmosphere.

\paragraph{Stage 3: Action Planning with Teacher Model}\leavevmode\\

\textbf{User Goal:} \textit{"Transform to spring tulip field with rows of colorful tulips in bloom, fresh green stems, clear spring sky, and vibrant flower farm atmosphere, transformed to near future with emerging technology."}

\begin{reasoningbox}[title=Overall Planning Reasoning]
\small
The current autumn vineyard scene must be completely replaced with a spring tulip field to match the user's request. This requires \textbf{location\_setting} as the foundation, since all other elements (tulips, farm, technology) depend on removing the existing vineyard. Then, \textbf{season\_cycle} changes the foliage to spring greenery. Finally, \textbf{time\_period\_era} adds near-future tech elements like drones or smart irrigation to fulfill the 'emerging technology' requirement. The farmhouse and barrels must be replaced with tulip farm structures and equipment.
\end{reasoningbox}

\vspace{0.3em}

\textbf{Generated Action Sequence (3 actions):}

\begin{actionbox}[title=Action 1: location\_setting (Priority 1)]
\small
\textbf{Reasoning:} The current vineyard and farmhouse define the setting. Complete location replacement is essential to establish the spring tulip field as the new environment, since all other elements (tulips, tech, sky) depend on removing the existing vineyard infrastructure first.

\vspace{0.2em}
\textbf{Parameters:}
\begin{itemize}[leftmargin=1.5em]
\item \texttt{source\_location}: "autumn\_vineyard"
\item \texttt{target\_location}: "spring\_tulip\_field"
\item \texttt{replace\_mode}: "complete"
\item \texttt{preserve\_foreground}: false
\item \texttt{description}: "Replace vineyard with tulip field"
\end{itemize}

\vspace{0.2em}
\textbf{Expected Effect:} This action fundamentally changes the environment from a wine-growing region to a flower farm. The vineyard rows will be replaced with rows of colorful tulips, the wine barrels will be removed, and the agricultural context will shift from grape cultivation to tulip cultivation. This is the foundational transformation that enables all subsequent actions.
\end{actionbox}

\begin{actionbox}[title=Action 2: season\_cycle (Priority 2)]
\small
\textbf{Reasoning:} The current autumn foliage must be replaced with spring greenery and blooming tulips to match the 'vibrant flower farm atmosphere'. This directly transforms the vegetation state and ground cover to reflect spring.

\vspace{0.2em}
\textbf{Parameters:}
\begin{itemize}[leftmargin=1.5em]
\item \texttt{source\_season}: "autumn"
\item \texttt{target\_season}: "spring"
\item \texttt{vegetation\_state}: "blooming\_tulips"
\item \texttt{ground\_cover}: "green\_grass"
\item \texttt{description}: "Change to spring tulips"
\end{itemize}

\vspace{0.2em}
\textbf{Expected Effect:} This action transforms the color palette from warm autumn golds and browns to vibrant spring colors (reds, pinks, yellows, purples from tulips) with lush green foliage. The fallen leaves and mature grape foliage will be replaced by fresh, blooming flowers and green stems. The sky will shift from golden-hour autumn tones to a clear, bright spring blue.
\end{actionbox}

\begin{actionbox}[title=Action 3: time\_period\_era (Priority 3)]
\small
\textbf{Reasoning:} To add 'emerging technology', the scene must be shifted to a near-future era. This allows for the introduction of tech elements like drones, smart irrigation, or automated harvesters that would be absent in the current historical setting.

\vspace{0.2em}
\textbf{Parameters:}
\begin{itemize}[leftmargin=1.5em]
\item \texttt{source\_era}: "modern"
\item \texttt{target\_era}: "near\_future"
\item \texttt{technology\_level}: "emerging"
\item \texttt{period\_accuracy}: "approximate"
\item \texttt{description}: "Add near-future tech"
\end{itemize}

\vspace{0.2em}
\textbf{Expected Effect:} This action adds subtle futuristic elements to the tulip field. Expect to see agricultural drones hovering above the tulip rows for monitoring or pollination, sleek solar panels on farm structures, automated irrigation systems with visible sensors, or modern greenhouse structures in the background. The technology should be visible but not dominate the scene---enhancing the agricultural context with near-future innovation.
\end{actionbox}

\paragraph{How Actions Work Together:}

The three actions form a logical transformation pipeline:

\begin{enumerate}
\item \textbf{Foundation (Action 1):} \texttt{location\_setting} establishes the new physical environment, removing all vineyard-specific elements and setting up the tulip field context.

\item \textbf{Seasonal Transform (Action 2):} \texttt{season\_cycle} builds on the new location by specifying the time of year and vegetation state---spring with blooming tulips---defining the color palette and atmospheric conditions.

\item \textbf{Enhancement (Action 3):} \texttt{time\_period\_era} adds the technological layer, introducing futuristic agricultural equipment that integrates naturally with the spring tulip field established by the previous actions.
\end{enumerate}

This sequential application ensures coherent transformation: location first (what), then season (when/how), then era (technological context).

\paragraph{Stage 4: Instruction Synthesis:}

The action sequence is converted to a natural language instruction for the image editor (Qwen-Image-Edit \citep{wu2025qwenimagetechnicalreport}):

\begin{promptbox}[title=Synthesized Edit Instruction]
\small\ttfamily
style\_transformation\_mode Transform to spring tulip field with tech farm. Replace\\
vines with tulips, add drones, clear sky, green stems. Maintain photorealistic quality.
\end{promptbox}

\textbf{Analysis:} The synthesized instruction condenses the 3-action plan into a concise natural language command. It explicitly mentions the target (tulip field), key elements (drones for technology, green stems for spring), and maintains the photorealistic style constraint.

\paragraph{Stage 6: Reward Evaluation}

Qwen3-VL-8B-Instruct evaluates the transformation quality across 6 criteria:

\begin{scorebox}[title=Reward Scores (Scale: 0-5)]
\small
\begin{tabular}{lcp{0.6\textwidth}}
\toprule
\textbf{Criterion} & \textbf{Score} & \textbf{Reasoning} \\
\midrule
Action Plan Quality & 5/5 & The chosen actions perfectly address the user's request: replacing the vineyard with a tulip field, changing the season to spring, and adding near-future technology (drones). Each action is directly relevant and necessary. \\
\addlinespace
Plan Reasoning & 5/5 & The reasoning is excellent, logically sequencing the actions from location replacement to seasonal change to technological addition. It clearly justifies each step's necessity. \\
\addlinespace
Reasoning Quality & 5/5 & Exceptionally detailed and specific reasoning. References the need to replace the vineyard, change the season, and add technology, with clear causal links. \\
\addlinespace
Final Image Quality & 3/5 & The image is visually coherent and vibrant, with good color saturation and lighting. However, technical metrics (LPIPS 0.824, SSIM 0.251) indicate structural distortion, suggesting artifacts. \\
\addlinespace
Adherence to Plan & 5/5 & The edited image fully executes the action plan: vineyard replaced with tulip field, season changed to spring with blooming tulips, drones added for emerging technology. \\
\addlinespace
Adherence to Prompt & 5/5 & The result perfectly matches the user's request: spring tulip field with colorful blooms, clear sky, and drones representing emerging technology. \\
\addlinespace
\textbf{Overall Quality} & \textbf{4/5} & Highly successful transformation, accurately fulfilling the user's request. Minor deduction for technical quality metrics indicating some image distortion. \\
\bottomrule
\end{tabular}
\end{scorebox}

\vspace{0.3em}

\textbf{Objective Metrics:}
\begin{itemize}
\item LPIPS: 0.824 (high perceptual difference, expected for complete transformation)
\item SSIM: 0.251 (low structural similarity, indicates significant scene change)
\item PSNR: 9.34 dB (low PSNR score confirms major transformation)
\item CLIP Score: 0.319 (semantic alignment between image and text)
\end{itemize}

\textbf{Analysis:} The high LPIPS and low SSIM/PSNR are expected and desirable for this transformation, as the goal is complete scene replacement rather than subtle editing. The strong adherence scores (5/5) confirm the transformation successfully achieved the user's intent.

\subsection{Example 2: Regular Dataset --- Contemporary Studio to Cyberpunk Nightclub}
\label{sec:synthesis_example_complex}

\subsubsection{Overview}

\textbf{Transformation:} Contemporary Studio Apartment $\to$ Futuristic Cyberpunk Nightclub with Preservation Constraints

\textbf{Complexity Level:} Complex (3 transformation actions + 2 preservation constraints)

\textbf{Key Challenge:} Radical environmental transformation while preserving specific elements (wood furniture, plants)---requires compositional reasoning to balance conflicting requirements

\textbf{Dataset:} Complex --- ID 327

\paragraph{Stage 1: Base Image Generation and Final Result}

The base image was generated using HiDream-I1-Dev with the prompt: \textit{"A contemporary studio apartment with open layout, modern furniture, track lighting, and city view."}

\begin{figure}[h!]
\centering
\begin{tabular}{cc}
\includegraphics[width=0.4\textwidth]{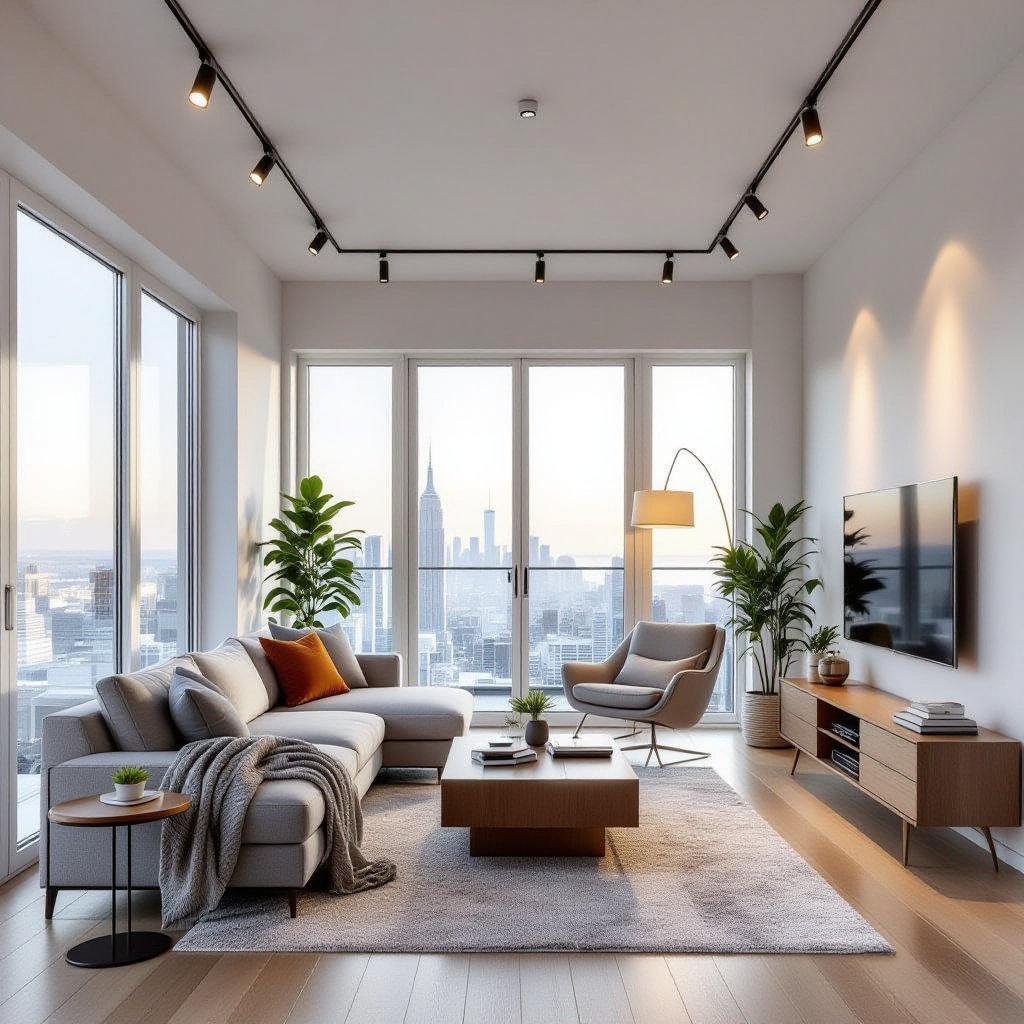} &
\includegraphics[width=0.4\textwidth]{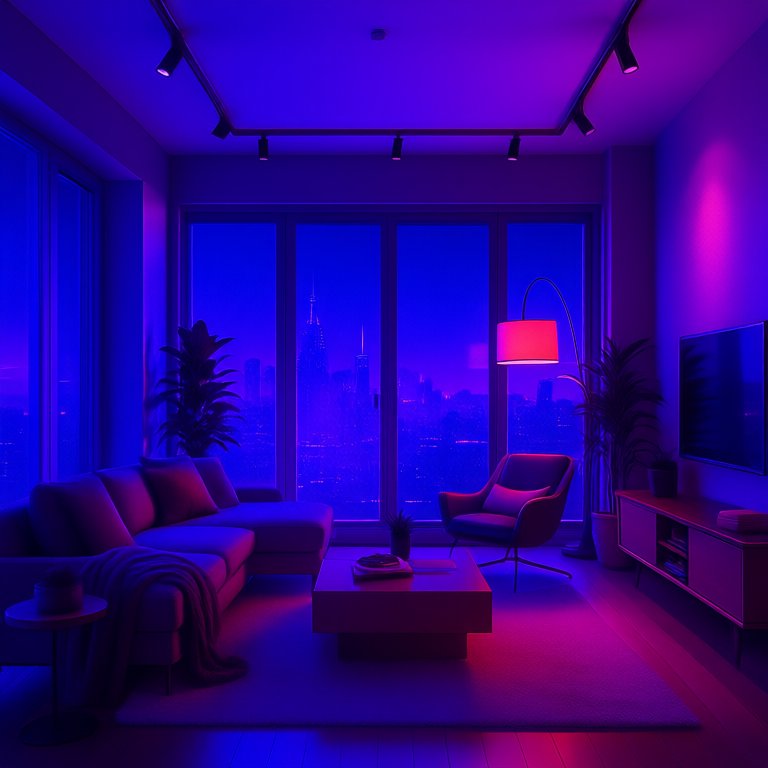} \\
\textbf{(a) Original Image} & \textbf{(b) Edited Image} \\
\end{tabular}
\caption{\textbf{Regular Dataset Example - Scandinavian Studio to Cyberpunk Industrial.} (a) Original: Modern Scandinavian studio with gray sofa, wooden table, plants, and city view through large windows. (b) Edited: Transformed to cyberpunk industrial loft with neon lighting, metallic textures, and futuristic tech elements. The transformation demonstrates constraint-based planning: dramatic neon accents added while attempting to preserve natural wood tones and organic plant colors (partial success, reward 3/5).}
\label{fig:synthesis_complex}
\end{figure}

\paragraph{Stage 2: Context Extraction:}

Using Qwen3-VL-8B-Instruct, we extract the 10-dimensional structured context representation $c_i$:

\begin{jsonbox}[title=Extracted Context ($c_i$)]
\small\ttfamily
\{\\
\hspace*{1em}"location": "modern\_apartment",\\
\hspace*{1em}"architecture": "contemporary\_minimalist",\\
\hspace*{1em}"time\_period\_era": "modern\_2020s",\\
\hspace*{1em}"time\_of\_day": "midday\_overhead",\\
\hspace*{1em}"season": "neutral\_indoor",\\
\hspace*{1em}"weather": "clear\_visible\_through\_windows",\\
\hspace*{1em}"mood\_lighting": "neutral\_balanced",\\
\hspace*{1em}"color\_grading": "natural\_balanced",\\
\hspace*{1em}"artistic\_medium": "photorealistic",\\
\hspace*{1em}"atmospheric\_effects": "none\_clear"\\
\}
\end{jsonbox}

\vspace{0.3em}

\textbf{Scene Description:} A bright, airy contemporary studio apartment with modern Scandinavian design. The space features a large gray sectional sofa with orange accent pillows, a low wooden coffee table, and a beige armchair. Multiple potted plants (monstera, fiddle leaf fig) add greenery. Large floor-to-ceiling windows provide natural light and reveal a city skyline. The color palette is neutral (whites, grays, beiges) with warm wood tones.

\textbf{Key Elements to Preserve:}
\begin{itemize}
\item \textbf{Wooden furniture:} Coffee table, side table, shelving units
\item \textbf{Natural plants:} Monstera, fiddle leaf fig, small potted plants
\end{itemize}

\paragraph{Stage 3: Action Planning with Teacher Model}\leavevmode\\

\textbf{User Goal:} \textit{"Transform to futuristic cyberpunk nightclub, preserve all traditional wooden elements, keep natural plants visible, AND add neon pink and blue dramatic lighting."}

\textbf{Constraint Analysis:} This goal presents a compositional reasoning challenge:
\begin{itemize}
\item \textbf{Conflicting Requirements:} Cyberpunk aesthetic (high-energy, artificial) vs. natural elements (wood, plants)
\item \textbf{Preservation Constraints:} Wood and plants must remain visible and recognizable despite dramatic lighting changes
\item \textbf{Lighting Challenge:} Neon pink/blue lighting can wash out natural colors, making preservation difficult
\end{itemize}

\begin{reasoningbox}[title=Overall Planning Reasoning]
\small
The image shows a modern apartment overlooking a city skyline. User wants a cyberpunk nightclub theme while preserving wood and plants. This requires radical environment shift plus dramatic lighting. \textbf{Location\_change} sets new context; \textbf{atmospheric\_effects} delivers neon glow; \textbf{mood\_lighting} ensures color harmony with preserved elements.

\vspace{0.2em}
\textbf{Critical Insight:} The preservation constraints require careful lighting control. Simply adding neon would wash out the wood and plants, making them unrecognizable. The plan must explicitly address this by using \textbf{mood\_lighting} to add warm accents that highlight the preserved elements while the neon dominates the overall scene.
\end{reasoningbox}

\vspace{0.3em}

\textbf{Generated Action Sequence (3 actions):}

\begin{actionbox}[title=Action 1: location\_setting (Priority 1)]
\small
\textbf{Reasoning:} Current urban apartment view must be replaced to establish cyberpunk club setting. All background elements including skyline need removal to allow new themed environment to dominate without conflict.

\vspace{0.2em}
\textbf{Parameters:}
\begin{itemize}[leftmargin=1.5em]
\item \texttt{source\_location}: "modern\_apartment"
\item \texttt{target\_location}: "cyberpunk\_nightclub"
\item \texttt{replace\_mode}: "complete"
\item \texttt{preserve\_foreground}: true
\item \texttt{description}: "Swap apartment for nightclub interior"
\end{itemize}

\vspace{0.2em}
\textbf{Expected Effect:} This action transforms the architectural context from residential apartment to commercial nightclub. The windows showing city skyline will be replaced with club interior elements (bars, DJ booth, dance floor). The open apartment layout will be reconfigured into a nightclub space. Critically, \texttt{preserve\_foreground: true} instructs the editor to maintain the wooden furniture and plants in the foreground during this radical background replacement.

\vspace{0.2em}
\textbf{Constraint Handling:} The preservation constraint is implemented through the \texttt{preserve\_foreground} flag, which tells the editor to apply location changes primarily to background elements (walls, windows, architectural features) while keeping foreground objects (furniture, plants) intact.
\end{actionbox}

\begin{actionbox}[title=Action 2: atmospheric\_effects (Priority 2)]
\small
\textbf{Reasoning:} Neon pink/blue lighting defines cyberpunk aesthetic. Adding dramatic glowing beams transforms the ambiance from calm living space to high-energy nightlife venue, matching user's explicit lighting request.

\vspace{0.2em}
\textbf{Parameters:}
\begin{itemize}[leftmargin=1.5em]
\item \texttt{effect\_type}: "neon\_glow"
\item \texttt{intensity}: "high"
\item \texttt{coverage}: "full\_scene"
\item \texttt{color\_tint}: "pink\_blue"
\item \texttt{description}: "Add neon pink-blue lighting"
\end{itemize}

\vspace{0.2em}
\textbf{Expected Effect:} This action introduces the signature cyberpunk aesthetic: bright neon lights in pink and blue hues. The lighting will be dramatic and artificial, creating strong color casts across surfaces. Light beams, glows, and reflections will be visible throughout the scene. The intensity is set to "high" to ensure the cyberpunk atmosphere is unmistakable.

\vspace{0.2em}
\textbf{Constraint Challenge:} This action poses the primary challenge to the preservation constraints. High-intensity neon lighting can completely change the apparent color of objects, potentially making the natural wood and green plants unrecognizable by tinting them pink/blue. This is why Action 3 is critical.
\end{actionbox}

\begin{actionbox}[title=Action 3: mood\_lighting (Priority 3)]
\small
\textbf{Reasoning:} To ensure neon doesn't wash out natural plants and wood tones, supplemental ambient lighting should highlight these preserved elements. Warm accents balance cool neon, creating layered depth while honoring user's preservation constraints.

\vspace{0.2em}
\textbf{Parameters:}
\begin{itemize}[leftmargin=1.5em]
\item \texttt{source\_mood}: "calm\_indoor"
\item \texttt{target\_mood}: "energetic\_club"
\item \texttt{accent\_lighting}: "warm\_spotlights"
\item \texttt{shadow\_depth}: "medium"
\item \texttt{description}: "Highlight wood and plants with warm light"
\end{itemize}

\vspace{0.2em}
\textbf{Expected Effect:} This action adds localized warm lighting (spotlights or accent lights) specifically targeting the wooden furniture and plants. These warm lights (oranges, yellows) counteract the cool neon tint, allowing the wood to maintain its natural brown color and the plants to keep their green hue. This creates a layered lighting environment: neon dominates the overall scene, but warm accents preserve the recognizability of specific elements.

\vspace{0.2em}
\textbf{Constraint Resolution:} This is the key action for satisfying the preservation constraints. Without this targeted lighting, the wood and plants would appear pink/blue under the neon, failing the constraint. The warm spotlights create "islands" of natural color within the otherwise artificial cyberpunk environment.
\end{actionbox}

\paragraph{How Actions Work Together (Compositional Reasoning):}

The three actions form a carefully orchestrated transformation that balances conflicting requirements:

\begin{enumerate}
\item \textbf{Environment Shift (Action 1):} \texttt{location\_setting} establishes the nightclub context while explicitly preserving foreground elements through the \texttt{preserve\_foreground} flag.

\item \textbf{Aesthetic Layer (Action 2):} \texttt{atmospheric\_effects} adds the dominant cyberpunk aesthetic with intense neon lighting, creating the high-energy nightclub atmosphere.

\item \textbf{Constraint Satisfaction (Action 3):} \texttt{mood\_lighting} resolves the conflict between neon lighting and preservation constraints by adding warm accents that maintain the natural appearance of wood and plants.
\end{enumerate}

\textbf{Compositional Challenge:} The difficulty lies in applying Action 2 (neon) and Action 3 (warm accents) simultaneously. Too much neon overpowers the warm lights, failing preservation. Too much warm light diminishes the cyberpunk aesthetic. The planner must find the right balance, which is encoded in the intensity and coverage parameters.

\textbf{Failure Mode:} If Action 3 is omitted or improperly parameterized, the transformation will fail the preservation constraints despite succeeding at the cyberpunk aesthetic. This example demonstrates why complex tasks require explicit constraint-aware reasoning.

\paragraph{Stage 4: Instruction Synthesis:}

The action sequence with constraints is converted to a natural language instruction:

\begin{promptbox}[title=Synthesized Edit Instruction]
\small\ttfamily
style\_transformation\_mode Transform to cyberpunk nightclub with neon pink-blue\\
lighting. Preserve wooden furniture and potted plants with warm spotlights.
\end{promptbox}

\textbf{Analysis:} The synthesized instruction explicitly mentions both the transformation goal (cyberpunk nightclub with neon) and the preservation constraints (wooden furniture, potted plants). The "warm spotlights" phrase directly encodes Action 3's constraint-resolution strategy. This explicit constraint language is critical for complex tasks.

\paragraph{Stage 6: Reward Evaluation}

Qwen3-VL-8B-Instruct evaluates the transformation quality across 6 criteria:

\begin{scorebox}[title=Reward Scores (Scale: 0-5)]
\small
\begin{tabular}{lcp{0.55\textwidth}}
\toprule
\textbf{Criterion} & \textbf{Score} & \textbf{Reasoning} \\
\midrule
Action Plan Quality & 3/5 & Actions are appropriate for transformation, but the plan fails to fully address preservation of wooden elements and plants. The \texttt{location\_setting} action removes original apartment context, but doesn't account for integrating preserved elements into new setting. \\
\addlinespace
Plan Reasoning & 3/5 & Reasoning is logically structured but lacks specific image analysis. Correctly identifies need for location change and neon lighting, but doesn't explain how to achieve preservation within cyberpunk context or address visual conflict between preserved elements and new lighting. \\
\addlinespace
Reasoning Quality & 3/5 & References user's request for cyberpunk lighting and preservation, but lacks specificity. Doesn't analyze how preserved elements (wood, plants) would interact with new lighting or how to integrate them into a nightclub environment. \\
\addlinespace
Final Image Quality & 3/5 & High visual coherence, dramatic neon lighting effectively creates cyberpunk atmosphere. However, metrics (LPIPS 0.801, SSIM 0.230) indicate significant structural difference. Color palette is appropriate (blue, pink), but lacks layered lighting depth expected in a nightclub. \\
\addlinespace
Adherence to Plan & 3/5 & \texttt{location\_setting} partially executed (apartment replaced with nightclub-like environment). \texttt{atmospheric\_effects} executed (neon pink-blue lighting added). \texttt{mood\_lighting} NOT executed as described---final image lacks warm accents to highlight wood and plants. \\
\addlinespace
Adherence to Prompt & 3/5 & Successfully creates cyberpunk nightclub atmosphere with dramatic neon lighting. However, fails to fully preserve traditional wooden elements and keep natural plants visible. Plants are present but wooden furniture is not highlighted or integrated in way that preserves traditional character. \\
\addlinespace
\textbf{Overall Quality} & \textbf{3/5} & Visually striking transformation effectively creates cyberpunk atmosphere. Neon lighting is dramatic and scene is cohesive. However, falls short of user's request to preserve wooden elements and plants. A good cyberpunk nightclub, but doesn't fully honor preservation constraints. \\
\bottomrule
\end{tabular}
\end{scorebox}

\vspace{0.3em}

\textbf{Objective Metrics:}
\begin{itemize}
\item LPIPS: 0.801 (high perceptual difference)
\item SSIM: 0.230 (low structural similarity)
\item PSNR: 6.11 dB (very low PSNR score)
\item CLIP Score: 0.320 (semantic alignment between image and text)
\end{itemize}

\textbf{Analysis:} The lower scores (3/5) across multiple criteria reflect the difficulty of the compositional reasoning task. The transformation successfully achieves the cyberpunk aesthetic, but only partially satisfies the preservation constraints. This is a common failure mode in complex tasks: the planner correctly identifies the constraint (Action 3), but the image editor struggles to execute it properly due to the conflicting lighting requirements.

\textbf{Key Insight:} Complex tasks with preservation constraints require more sophisticated action parameterization. The \texttt{intensity} and \texttt{coverage} parameters in Actions 2 and 3 must be carefully balanced, which is difficult to specify in a discrete action representation. Future work could explore continuous parameter spaces or iterative refinement to better handle such constraints.

\subsection{Comparison and Insights}
\label{sec:synthesis_comparison}

\begin{table}[h!]
\centering
\caption{Comparison of Normal vs. Complex Synthesis Examples}
\label{tab:synthesis_comparison}
\begin{tabular}{lcc}
\toprule
\textbf{Property} & \textbf{Normal Example} & \textbf{Complex Example} \\
\midrule
Number of actions & 3 & 3 \\
Preservation constraints & 0 & 2 \\
Overall reward & 4.0/5.0 (Good) & 3.0/5.0 (Medium) \\
Planning difficulty & Low & High \\
Execution difficulty & Medium & High \\
Adherence to plan & 5/5 (Perfect) & 3/5 (Partial) \\
Adherence to prompt & 5/5 (Perfect) & 3/5 (Partial) \\
Key challenge & Complete transformation & Conflicting constraints \\
Success factor & Clear, non-conflicting goals & Failed constraint resolution \\
\bottomrule
\end{tabular}
\end{table}

\paragraph{Key Takeaways}

\begin{itemize}
\item \textbf{Normal tasks:} When transformation goals are clear and non-conflicting, even complex multi-action plans can be executed successfully. The autumn$\to$spring example achieves near-perfect adherence because each action builds naturally on the previous one without conflicts.

\item \textbf{Complex tasks:} Preservation constraints introduce compositional reasoning challenges. The cyberpunk example demonstrates how conflicting requirements (dramatic neon vs. natural wood/plant colors) require careful parameter balancing that current models struggle with.

\item \textbf{Planning vs. Execution gap:} The complex example shows a gap between planning (Action 3 correctly identifies the need for warm accents) and execution (the warm accents are not sufficiently applied in the final image). This highlights the need for better alignment between discrete action plans and continuous image editing.

\item \textbf{Reward signal quality:} The reward model successfully distinguishes between fully successful transformations (4-5/5) and partially successful ones (3/5), providing meaningful training signal for downstream models.
\end{itemize}

\subsection{Example 3: Complex Dataset --- Arctic Glacier to Desert Canyon}
\label{sec:synthesis_example_complexv2}

\subsubsection{Overview}

\textbf{Transformation:} Arctic glacier crevasse with ice and snow $\to$ Desert canyon with warm rock formations

\textbf{Complexity Level:} Complex (2 actions with spatial preservation constraints)

\textbf{Key Challenge:} Dramatic environmental transformation while preserving spatial relationships and depth perception

\textbf{Dataset:} Complex  (ice\_art\_movement\_multi theme)

\paragraph{Stage 1: Base Image Generation and Final Result}

The base image depicts a dramatic arctic landscape with a deep glacial crevasse cutting through snow-covered terrain, revealing turquoise meltwater. Snow-capped mountains rise in the background under a partly cloudy sky.

\begin{figure}[h!]
\centering
\begin{tabular}{cc}
\includegraphics[width=0.4\textwidth]{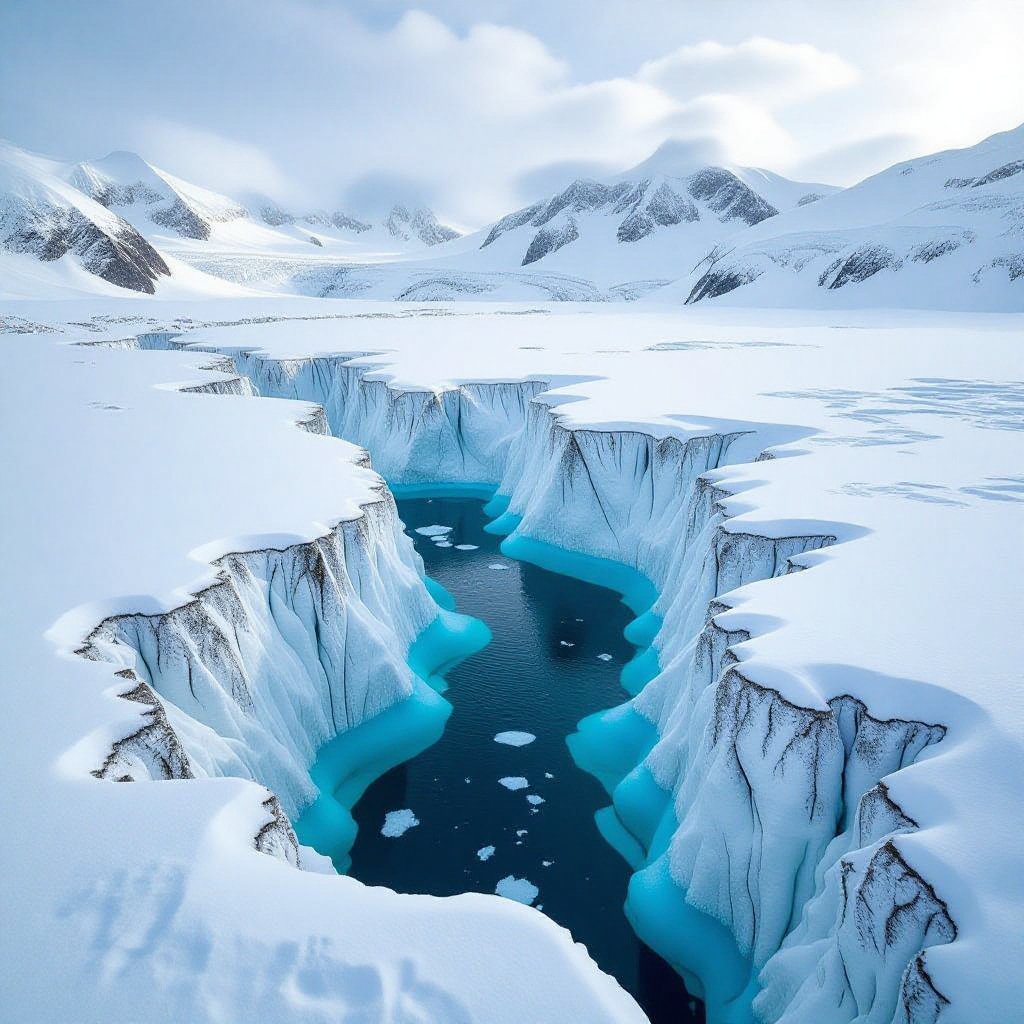} &
\includegraphics[width=0.4\textwidth]{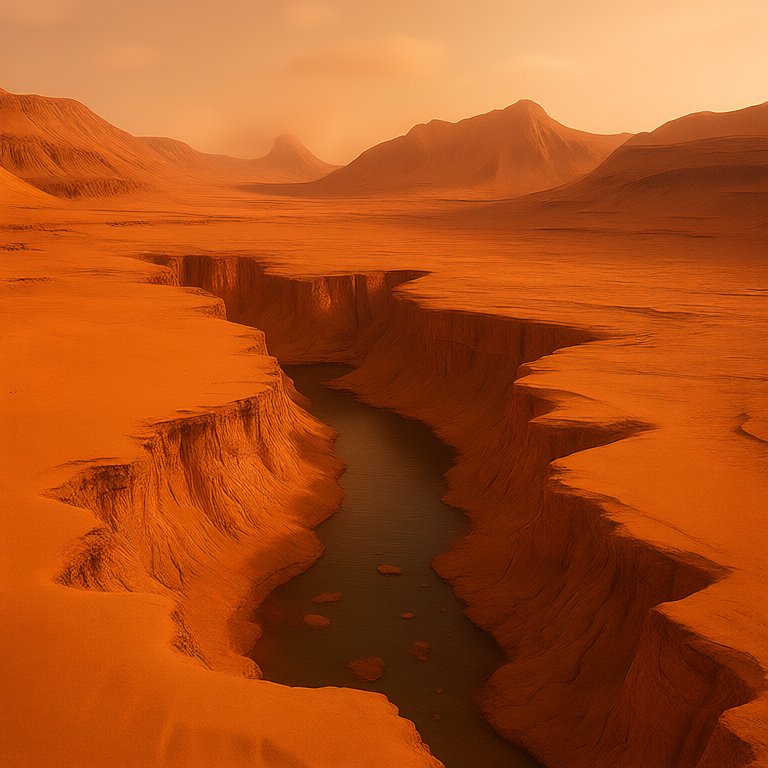} \\
\textbf{(a) Original Image} & \textbf{(b) Edited Image} \\
\end{tabular}
\caption{\textbf{Complex Dataset Example - Arctic Glacier to Desert Canyon.} (a) Original: Arctic glacier scene with deep crevasse revealing turquoise meltwater, surrounded by expansive snowfields and distant snow-capped mountain peaks. Strong diagonal composition with cold blue-white color palette. (b) Edited: Transformed to dramatic desert canyon with layered sandstone and rocky canyon walls in warm ochre and amber tones. The diagonal crevasse structure is preserved but now rendered as desert canyon walls. Background mountains maintain imposing presence with reddish-brown rock faces. Depth perception and spatial relationships successfully maintained through consistent atmospheric perspective (reward 5.0/5.0).}
\label{fig:synthesis_complexv2}
\end{figure}

\paragraph{Stage 2: Context Extraction:}

The Complex dataset uses a streamlined context extraction focused on spatial and compositional features. Key extracted properties include: dominant colors (white, blue, gray), composition (diagonal leading lines, layered depth), lighting (soft diffused daylight), and spatial structure (foreground snow, midground crevasse, background mountains).

\paragraph{Stage 3: Action Planning with Compositional Reasoning}\leavevmode\\

\textbf{User Goal:} Transform arctic glacier to desert canyon while preserving spatial relationships and depth perception.

\begin{reasoningbox}[title=Overall Planning Reasoning]
\small
This image depicts a stark arctic landscape with strong diagonal compositional lines and deep atmospheric perspective. To create a bold transformation while preserving spatial relationships and depth perception, I will shift the environment to a vibrant desert canyon. This maintains the diagonal structure but replaces the icy textures with warm earth tones and layered rock formations. The color grading adjustment ensures the new palette complements the existing depth cues without disrupting scale or perspective. The artistic medium remains photorealistic to preserve realism and avoid altering perceived distance.
\end{reasoningbox}

The planning model generates a 2-action sequence with explicit spatial preservation constraints:

\begin{actionbox}[title=Action 1: Location Setting (Priority 1)]
\small
\textbf{Action ID:} \texttt{location\_setting}

\textbf{Parameters:}
\begin{itemize}[leftmargin=1.5em]
    \item \texttt{setting\_type}: desert\_canyon
    \item \texttt{setting\_style}: rocky\_with\_layers
    \item \texttt{description}: Replace glacier with desert canyon
\end{itemize}

\textbf{Reasoning:} The snowy mountains and ice crevasse define this as polar. Replacing it with a desert canyon preserves the diagonal composition but introduces contrasting textures and warm tones, creating a visually jarring yet harmonious transformation rooted in the same structural logic.

\textbf{Expected Effect:} Complete environmental transformation from arctic to desert while maintaining the diagonal crevasse structure, layered background mountains, and foreground-to-background depth hierarchy.

\textbf{Constraint Handling:} Explicitly preserves spatial relationships by maintaining diagonal composition, scale proportions, and three-layer depth structure (foreground, midground, background).
\end{actionbox}

\begin{actionbox}[title=Action 2: Color Grading (Priority 2)]
\small
\textbf{Action ID:} \texttt{color\_grading}

\textbf{Parameters:}
\begin{itemize}[leftmargin=1.5em]
    \item \texttt{grade\_style}: warm\_desert
    \item \texttt{look}: sunset\_amber
    \item \texttt{description}: Apply warm desert color grading
\end{itemize}

\textbf{Reasoning:} The cool blue-white palette must be replaced with warm ochre, sandstone, and sunset hues to match the desert theme. This transforms the mood while preserving depth perception by maintaining analogous color transitions across layers, reinforcing spatial hierarchy.

\textbf{Expected Effect:} Shift from cool (blue, white, gray) to warm (amber, ochre, sandstone) color palette while maintaining tonal gradation that signals atmospheric depth.

\textbf{Constraint Handling:} Preserves depth perception by maintaining analogous color transitions across foreground, midground, and background layers, ensuring spatial hierarchy remains visually coherent.
\end{actionbox}

\paragraph{Stage 4: Edit Instruction Generation:}

The action sequence is compiled into a natural language instruction for the image editor:

\begin{promptbox}[title=Edit Instruction]
\small\ttfamily
style\_transformation\_mode Transform arctic glacier to desert canyon. Preserve spatial relationships and depth perception. Warm desert color grading.
\end{promptbox}

\paragraph{Stage 6: Reward Evaluation:}

The reward model (Qwen3-VL-8B-Instruct) evaluates the transformation across 6 criteria:

\begin{scorebox}[title=Reward Scores (All 5.0/5.0)]
\small
\begin{itemize}[leftmargin=1.5em]
    \item \textbf{Action Plan Quality:} 5.0 --- ``The actions are perfectly appropriate: replacing the glacier with a desert canyon and applying warm desert color grading directly address the transformation request while preserving the core spatial structure.''
    
    \item \textbf{Plan Reasoning:} 5.0 --- ``The reasoning is excellent, logically sequencing the transformation by first changing the setting and then the color palette. It explicitly references the diagonal composition and depth perception, showing a deep understanding of the image's structure.''
    
    \item \textbf{Reasoning Quality:} 5.0 --- ``The reasoning is exceptional. It specifically references the `diagonal compositional lines' and `deep atmospheric perspective' of the original, justifies the color grading to maintain depth, and explains how the transformation preserves spatial relationships through structural logic.''
    
    \item \textbf{Final Image Quality:} 5.0 --- ``The edited image is of excellent technical quality. The transformation is seamless, with no visible artifacts. The desert canyon maintains the original's perspective and scale, and the warm color grading enhances the depth perception without distortion.''
    
    \item \textbf{Adherence to Plan:} 5.0 --- ``The editing perfectly follows the action plan. The setting was changed to a desert canyon with layered rock formations, and the color grading was applied to create a warm, sunset-amber palette, exactly as specified.''
    
    \item \textbf{Adherence to Prompt:} 5.0 --- ``The result perfectly matches the user's request. The arctic glacier is transformed into a desert canyon while maintaining the spatial relationships (diagonal lines, scale, perspective) and depth perception (atmospheric layers, shadows) as required.''
    
    \item \textbf{Overall Quality:} 5.0 --- ``This is an outstanding transformation. It successfully and artistically reimagines the scene while rigorously adhering to the constraints of preserving spatial relationships and depth perception, resulting in a visually compelling and coherent image.''
\end{itemize}
\end{scorebox}

\textbf{Objective Metrics:} LPIPS 0.963 (high perceptual distance, confirming dramatic transformation), SSIM 0.341, PSNR 5.66, CLIP score 0.230.

\subsection{Dataset Comparison}
\label{sec:dataset_comparison}

Table~\ref{tab:dataset_comparison} summarizes the key differences between the three synthetic datasets used in our experiments.

\begin{table}[h!]
\centering
\small
% \begin{tabular}{|l|c|c|c|}
\begin{tabular}{|l|p{9em}|p{9em}|p{9em}|}
\hline
\textbf{Characteristic} & \textbf{Simple} & \textbf{Regular} & \textbf{Complex} \\
\hline
\hline
\textbf{Dataset Size} & $n = 10{,}000$ & $n = 10{,}000$ & $n = 10{,}000$ \\
\hline
\textbf{Action Library} & 10 actions & 20 actions & 30 actions \\
\hline
\textbf{Action Types} & 9 THEME + 1 STYLE & 10 atomic + 10 constraint & 30 styling \& transformation \\
\hline
\textbf{Avg Actions/Sample} & 2-3 & 3-5 & 2-4 \\
\hline
\textbf{Theme Diversity} & 31 locations & 10 interior design styles & 83 diverse themes \\
\hline
\textbf{Transformation Type} & Simple, 1-2 distinct changes & Compositional, 3-5 interacting changes & Moderate, 2-4 styling changes \\
\hline
\textbf{Key Features} & Atomic actions, clear goals & Constraints, preservation logic & Broad distribution, artistic focus \\
\hline
\textbf{Complexity Level} & Low & High & Medium-High \\
\hline
\textbf{Constraint Actions} & None & 10 (preserve, exclude, conditional) & Integrated into action parameters \\
\hline
\textbf{Use Case} & Basic training, validation & Complex reasoning, compositionality & Diverse distribution, generalization \\
\hline
\textbf{Example Transformation} & Autumn vineyard $\to$ Spring tulip field & Industrial loft $\to$ Cyberpunk industrial & Arctic glacier $\to$ Desert canyon \\
\hline
\end{tabular}
\caption{Comparison of three synthetic datasets. Simple uses atomic actions on diverse locations. Regular adds constraint/compositional actions for interior design styles. Complex expands to 30 actions and 83 diverse themes, balancing complexity with broad distribution coverage.}
\label{tab:dataset_comparison}
\end{table}

\subsubsection{Key Insights from Three-Dataset Comparison}

\begin{itemize}
\item \textbf{Action Library Design:} Simple (10 actions) focuses on orthogonal dimensions (location, architecture, time, season, weather, mood, lighting, texture, material, color scheme). Regular (20 actions) adds constraint logic (preserve\_attribute, exclude\_region, conditional\_transform). Complex (30 actions) integrates constraints into a unified framework with expanded styling options.

\item \textbf{Theme Diversity vs. Complexity:} Simple prioritizes location diversity (31 types) with atomic transformations. Regular prioritizes compositional complexity (3-5 interacting changes) within a narrower domain (10 interior design styles). Complex achieves both broad distribution (83 themes) and moderate complexity (2-4 actions with integrated constraints).

\item \textbf{Training Signal Quality:} All three datasets achieve high reward scores for successful samples (4-5/5), but failure modes differ. Simple fails primarily on execution errors (action parameters not followed). Regular fails on constraint conflicts (preserving natural wood while adding neon lights). Complex shows more consistent quality due to streamlined action library and diverse training distribution.

\item \textbf{Method Performance Patterns:} Our experiments (Section~\ref{sec:experiments}) show that method effectiveness varies by dataset. \rw excels on Simple vision tasks (Overall 79.33 on Vision-4B), \sw on Regular text tasks (Overall 78.77 on Text-4B), and \dpo on Regular vision tasks (Overall 85.41 on Vision-8B), suggesting that continuous weighting (\rw/\sw) benefits from simpler or complex-compositional tasks, while preference learning (\dpo) benefits from broad distribution coverage.
\end{itemize}

\section{Training Algorithms}
\label{sec:appendix_algorithms}

This appendix provides complete algorithmic details for Standard Supervised Learning and Direct Preference Optimization, including mathematical formulations, pseudocode, and implementation specifics.

\subsection{Standard Supervised Learning}
\label{sec:appendix_algorithm_standard}

The baseline approach treats synthetic trajectories as supervised training data, ignoring reward signals entirely. Given a dataset $\mathcal{D} = \{\tau_i\}_{i=1}^n$ of synthetic trajectories where $\tau_i = (e_i, I_i, c_i, \{a_{i,j}\}_{j=1}^{m_i}, \{z_{i,j}\}_{j=1}^{m_i}, \hat{e}_i, \hat{I}_i, r_i)$, we train the model $\pi_{\theta}$ to maximize the likelihood of actions and per-step chain-of-thought reasoning.

\subsubsection{Loss Formulation}

The standard supervised learning loss is:
$
\mathcal{L}_{\text{\slfull}}(\theta) = -\frac{1}{n} \sum_{i=1}^n \sum_{j=1}^{m_i} \log \pi_{\theta}(a_{i,j}, z_{i,j} \mid I_i, e_i, c_i, \{a_{ik}\}_{k<j})
$
where, $n$ is the total number of trajectories in the dataset, $m_i$ is the number of actions in trajectory $i$, $\{a_{ik}\}_{k<j}$ denotes the action history up to step $j-1$, $a_{i,j}$ is the $j$-th action in trajectory $i$, and  $z_{i,j}$ is the chain-of-thought reasoning for action $a_{i,j}$.
%
% :
% \begin{itemize}
% \item $n$ is the total number of trajectories in the dataset
% \item $m_i$ is the number of actions in trajectory $i$
% \item $\{a_{ik}\}_{k<j}$ denotes the action history up to step $j-1$
% \item $a_{i,j}$ is the $j$-th action in trajectory $i$
% \item $z_{i,j}$ is the chain-of-thought reasoning for action $a_{i,j}$
% \end{itemize}
%
The model learns to predict both the action and its reasoning given the image, goal, context, and action history.

\subsubsection{Complete Algorithm}

\begin{algorithm}[H]
\caption{Standard Supervised Learning applied to VL tool-use}
\begin{algorithmic}[1]
\STATE \textbf{Input:} Trajectory dataset $\mathcal{D} = \{\tau_i\}_{i=1}^n$, model $\pi_{\theta}$
\STATE \textbf{Hyperparameters:} Learning rate $\eta = 2 \times 10^{-5}$, epochs $E = 2$, batch size $B = 4$, gradient accumulation steps $G = 2$
\STATE Initialize $\theta$ from pretrained Qwen3-VL checkpoint
\STATE Apply LoRA adaptation: rank $r = 16$, $\alpha = 32$, dropout $p = 0.05$
\STATE Target modules: $\{q\_proj, k\_proj\}$ in all transformer layers
\FOR{epoch $e = 1$ to $E$}
    \STATE Shuffle dataset $\mathcal{D}$
    \FOR{batch $\mathcal{B} = \{\tau_i\}_{i=1}^B$ in $\mathcal{D}$}
        \STATE Initialize $\mathcal{L}_{\text{batch}} \leftarrow 0$
        \FOR{trajectory $\tau_i$ in $\mathcal{B}$}
            \STATE Prepare inputs: image $I_i$, editing prompt $e_i$, context $c_i$
            \STATE Initialize action history: $\mathcal{H} \leftarrow \emptyset$
            \FOR{step $j = 1$ to $m_i$}
                \STATE Forward pass: $\text{logits} \leftarrow \pi_{\theta}(I_i, e_i, c_i, \mathcal{H})$
                \STATE Compute token-level log-likelihood: $\ell_j = \log p_{\text{logits}}(a_{i,j}, z_{i,j})$
                \STATE Add to history: $\mathcal{H} \leftarrow \mathcal{H} \cup \{a_{i,j}\}$
                \STATE Accumulate loss: $\mathcal{L}_{\text{batch}} \leftarrow \mathcal{L}_{\text{batch}} - \ell_j$
            \ENDFOR
        \ENDFOR
        \STATE Normalize: $\mathcal{L}_{\text{batch}} \leftarrow \mathcal{L}_{\text{batch}} / (B \cdot G)$
        \STATE Backward pass: Compute $\nabla_{\theta} \mathcal{L}_{\text{batch}}$
        \IF{gradient accumulation step complete}
            \STATE Update parameters: $\theta \leftarrow \theta - \eta \nabla_{\theta} \mathcal{L}_{\text{batch}}$
            \STATE Zero gradients
        \ENDIF
    \ENDFOR
    \STATE Evaluate on validation set
\ENDFOR
\STATE \textbf{Return:} Trained student model $\pi_{\theta}$
\end{algorithmic}
\end{algorithm}

\subsubsection{Implementation Details}

\paragraph{LoRA Configuration:}
We use Low-Rank Adaptation (LoRA) for efficient fine-tuning:
\begin{itemize}
\item \textbf{Rank}: $r = 16$ (reduces parameters by 99\%)
\item \textbf{Alpha}: $\alpha = 32$ (scaling factor for LoRA weights)
\item \textbf{Dropout}: $p = 0.05$ for regularization
\item \textbf{Target modules}: Query and key projections in all attention layers
\item \textbf{Trainable parameters}: ~150M out of 8B (1.9\%)
\end{itemize}

\paragraph{Training Configuration:}
\begin{itemize}
\item \textbf{Optimizer}: AdamW with $\beta_1 = 0.9$, $\beta_2 = 0.999$, weight decay $= 0.01$
\item \textbf{Learning rate}: $2 \times 10^{-5}$ with linear warmup (500 steps) and cosine decay
\item \textbf{Batch size}: 4 per GPU, 8 GPUs, gradient accumulation 2 → effective batch size 64
\item \textbf{Epochs}: 2 (approximately 300 gradient updates for $n = 10{,}000$ trajectories)
\item \textbf{Mixed precision}: bfloat16 for memory efficiency
\item \textbf{Gradient clipping}: Max norm 1.0
\end{itemize}

\paragraph{Data Processing:}
\begin{itemize}
\item \textbf{Sequence length}: Maximum 2048 tokens
\item \textbf{Padding}: Right-padding with attention mask
\item \textbf{Truncation}: Truncate long trajectories from the end
\item \textbf{Shuffling}: Shuffle at epoch level, not within batches
\end{itemize}

\subsubsection{Limitations of Standard \slfull}

This approach has fundamental limitations:

\paragraph{1. Quality Blindness:}
All synthetic trajectories contribute equally regardless of reward $r_i$. A trajectory with $r_i = 3.0$ (poor) has the same influence as $r_i = 5.0$ (excellent).

\paragraph{2. Potential for Degradation:}
If low-quality trajectories are prevalent, the model may learn suboptimal behaviors and fail to match teacher performance.

\paragraph{3. No Preference Signal:}
The model has no signal about which trajectories are better when multiple plans exist for the same $(I_i, e_i)$ pair.

\paragraph{4. Reward Information Wasted:}
The expensive reward evaluation ($r_i$) computed during data generation is completely ignored during training.

These limitations motivate reward-aware training methods (\rlfull, \rw, \dpo) described in the main paper.

\subsection{Reward-Weighted Fine-Tuning (\rw)}
\label{sec:appendix_algorithm_rw}

\rw uses all trajectories but weights their contribution according to quality. This section provides complete implementation details and theoretical analysis.

\subsubsection{Weight Function}

We use a simple continuous weight function:
$$
w(r_i) = \max \{r_i - 3.0, 0\}
$$

This linearly scales the contribution of each trajectory based on its quality above the minimum acceptable threshold (3.0). Trajectories with $r_i < 3.0$ receive zero weight, while higher-quality trajectories receive proportionally more influence.

\subsubsection{Weighted Loss Formulation}

The \rw loss modifies standard supervised learning by incorporating per-trajectory weights:

$$
\mathcal{L}_{\text{RW}}(\theta) = \frac{\sum_{i=1}^n w(r_i) \cdot \mathcal{L}_i(\theta)}{\sum_{i=1}^n w(r_i)}
$$

where $\mathcal{L}_i(\theta) = -\sum_{j=1}^{m_i} \log \pi_{\theta}(a_{i,j}, z_{i,j} \mid I_i, e_i, c_i, \{a_{i,k}\}_{k<j})$ is the per-trajectory loss.

The normalization term $\sum_{i=1}^n w(r_i)$ computes a weighted average rather than a weighted sum, ensuring: (1) loss magnitude remains comparable to standard supervised learning (unweighted mean); (2) gradient scale is independent of dataset size and weight distribution; (3) each trajectory contributes proportionally to its quality—e.g., a trajectory with $w(r_i) = 2.0$ receives twice the gradient weight of one with $w(r_i) = 1.0$. This is equivalent to importance sampling where excellent trajectories are effectively replicated in the training distribution.

\subsubsection{Complete Algorithm}

\begin{algorithm}[H]
\caption{Reward-Weighted Fine-tuning applied to VL tool-use}
\begin{algorithmic}[1]
\STATE \textbf{Input:} Trajectory dataset $\dataset = \{\tau_i\}$ with rewards, model $\pi_{\theta}$
\STATE \textbf{Hyperparameters:} Learning rate $\eta = 2 \times 10^{-5}$, epochs $E = 2$, batch size $B = 8$, GPUs $= 8$
\STATE Initialize $\theta$ from pretrained Qwen3-VL checkpoint with LoRA (rank 16, $\alpha=32$)
\FOR{epoch $= 1$ to $E$}
    \FOR{batch $\{\tau_i\}_{i \in \mathcal{B}}$ in $\dataset$}
        \STATE \textbf{// Compute per-trajectory losses}
        \FOR{$i \in \mathcal{B}$}
            \STATE $\mathcal{L}_i \leftarrow -\sum_{j=1}^{m_i} \log \pi_{\theta}(a_{i,j}, z_{i,j} \mid I_i, e_i, c_i, \{a_{i,k}\}_{k<j})$
        \ENDFOR
        \STATE \textbf{// Compute weights and weighted loss}
        \STATE Compute weights: $w_i = \max\{r_i - 3.0, 0\}$ for each $i \in \mathcal{B}$
        \STATE Weighted loss: $\mathcal{L}_{\text{batch}} = \frac{\sum_{i \in \mathcal{B}} w_i \mathcal{L}_i}{\sum_{i \in \mathcal{B}} w_i}$
        \STATE \textbf{// Gradient update}
        \STATE Update: $\theta \leftarrow \theta - \eta \nabla_{\theta} \mathcal{L}_{\text{batch}}$
    \ENDFOR
    \STATE Evaluate on validation set
\ENDFOR
\STATE \textbf{Return:} Trained student model $\pi_{\theta}$
\end{algorithmic}
\end{algorithm}

\subsubsection{Implementation Details}

\paragraph{PyTorch Implementation:}
Per-sample weighting in PyTorch is straightforward:
\begin{itemize}
\item Compute standard log-likelihood loss for each trajectory: \texttt{loss\_i = -log\_probs[i].sum()}
\item Compute weights: \texttt{weights = torch.maximum(rewards - 3.0, torch.zeros\_like(rewards))}
\item Weighted loss: \texttt{weighted\_loss = (weights * losses).sum() / weights.sum()}
\item Backward pass on \texttt{weighted\_loss}
\end{itemize}

\paragraph{Memory and Computational Cost}
\begin{itemize}
\item \textbf{Memory}: Same as standard \slfull (no reference model needed)
\item \textbf{Computation}: Same forward/backward cost as \slfull
\item \textbf{Effective batch size}: $8 \times 8 = 64$ (8 per GPU, 8 GPUs)
\item \textbf{Training time}: Identical to \slfull
\end{itemize}

\paragraph{Connection to Importance Sampling:}
Reward-weighted regression relates to importance sampling in offline \rlfull. Importance sampling enables unbiased estimation when evaluating a target distribution using samples from a different source distribution, with truncated importance sampling providing variance reduction through weight clipping. If we view the data generator as sampling from a behavior policy $\pi_{\text{data}}$ and want the trained model to match a target policy $\pi^*$ that achieves high rewards, the weight $w(r_i)$ approximates the importance ratio $\frac{\pi^*(a|s)}{\pi_{\text{data}}(a|s)}$. This enables the model to focus gradient updates on high-quality trajectories while retaining the diversity of medium-quality examples.

\paragraph{Detailed Comparison: \rw vs. \sw:}

Both \rw and \sw relate to advantage-based \rlfull methods but differ in key ways. \rw uses absolute rewards $r_i$ with the continuous weight function $w(r_i) = \max\{r_i - 3.0, 0\}$, preserving the natural quality hierarchy—excellent trajectories ($r_i \geq 4.5$) receive consistently high weight regardless of dataset composition. This is appropriate when the teacher provides high-quality data ($r_i \geq 3.0$) without catastrophically bad examples. \sw uses standardized rewards $\tilde{r}_i = \frac{r_i - \bar{r}}{\sigma_r}$ directly as weights, similar to advantages $A_i = r_i - \bar{r}$ but with variance normalization. \sw adapts to dataset statistics: in a dataset with $\bar{r}=4.2$, a trajectory with $r_i=4.5$ receives moderate weight, while in a dataset with $\bar{r}=3.5$, the same trajectory receives high weight. This makes \sw robust to reward scale variations across datasets.

From a rollout perspective, when multiple rollouts of the same input $(I_i, e_i)$ produce different rewards, \sw's standardization provides variance reduction by centering the distribution: trajectories above the mean receive positive weight, those below receive negative weight, reducing gradient variance—a classic technique in policy gradient methods \citep{williams92simple,schulman16highdimensional}. This makes \sw particularly effective for datasets with diverse reward distributions across different inputs while maintaining stability within each input's rollout variations.

\paragraph{Normalization in SW: Mathematical Justification:}

A critical implementation detail distinguishes \sw from \rw. Since standardized rewards $\tilde{r}_i$ are zero-mean by construction ($\mathbb{E}[\tilde{r}_i] = 0$), normalizing by their sum would cause instability. 
%
% Consider the batch loss:
%
% If we normalize by $\sum_{i \in \mathcal{B}} \tilde{r}_i$, we encounter problems:
% \begin{itemize}
% \item Over the full dataset, $\sum_{i=1}^n \tilde{r}_i = 0$ exactly (by definition of z-score)
% \item Over a random batch, $\sum_{i \in \mathcal{B}} \tilde{r}_i \approx 0$, causing division by near-zero and gradient explosion
% \item If $\sum_{i \in \mathcal{B}} \tilde{r}_i < 0$, the loss sign flips, incorrectly reversing the optimization direction for that batch
% \end{itemize}
%
Therefore, \sw uses batch-size normalization:
$$
\mathcal{L}_{\text{batch}} = \frac{1}{|\mathcal{B}|}\sum_{i \in \mathcal{B}} \tilde{r}_i \mathcal{L}_i
$$

This formulation is stable and mathematically equivalent to applying standardized rewards as gradient multipliers: positive $\tilde{r}_i > 0$ (above-average quality) amplifies gradients, while negative $\tilde{r}_i < 0$ (below-average quality) reverses gradient direction—analogous to advantage-based policy gradients \citep{williams92simple,schulman16highdimensional} where positive/negative advantages increase/decrease action probabilities.

In contrast, \rw uses non-negative weights $w(r_i) = \max\{r_i - 3.0, 0\} \geq 0$. For \rw, normalizing by $\sum_{i \in \mathcal{B}} w_i$ is stable and maintains the weighted average interpretation. The normalization choice (batch size vs. sum of weights) is dictated by whether weights can be negative.

\subsection{Direct Preference Optimization (\dpo)}
\label{sec:appendix_algorithm_dpo}

\dpo is a preference-based reinforcement learning method that learns from contrastive pairs of trajectories without requiring an explicit reward model. This section provides complete mathematical formulation and implementation details.

\subsubsection{Preference Dataset Construction}

\dpo requires a preference dataset $\mathcal{D}_{\text{pref}}$ consisting of trajectory pairs: $
\mathcal{D}_{\text{pref}} = \{(\tau_i^+, \tau_i^-)\}_{i=1}^{n_{\text{pairs}}}
$
%
% $$
% \mathcal{D}_{\text{pref}} = \{(\tau_i^+, \tau_i^-)\}_{i=1}^{n_{\text{pairs}}}
% $$
%
where: $\tau_i^+ = (e_i, I_i, c_i, \{a_{i,j}^+\}, \{z_{i,j}^+\}, \hat{e}_i^+, \hat{I}_i^+, r_i^+)$ is the ``chosen'' trajectory with reward $r_i^+ \geq 4.0$, $\tau_i^- = (e_i, I_i, c_i, \{a_{i,j}^-\}, \{z_{i,j}^-\}, \hat{e}_i^-, \hat{I}_i^-, r_i^-)$ is the ``rejected'' trajectory with reward $r_i^- \in [2.5, 3.5]$, both trajectories share the same input: $(I_i, e_i)$, and we have $r_i^+ - r_i^- \geq 0.5$ to ensure meaningful preference signal
% \begin{itemize}
% \item $\tau_i^+ = (e_i, I_i, c_i, \{a_{i,j}^+\}, \{z_{i,j}^+\}, \hat{e}_i^+, \hat{I}_i^+, r_i^+)$ is the ``chosen'' trajectory with reward $r_i^+ \geq 4.0$
% \item $\tau_i^- = (e_i, I_i, c_i, \{a_{i,j}^-\}, \{z_{i,j}^-\}, \hat{e}_i^-, \hat{I}_i^-, r_i^-)$ is the ``rejected'' trajectory with reward $r_i^- \in [2.5, 3.5]$
% \item Both trajectories share the same input: $(I_i, e_i)$
% \item We require $r_i^+ - r_i^- \geq 0.5$ to ensure meaningful preference signal
% \end{itemize}

\paragraph{Pairing Algorithm:}
For each high-quality trajectory ($r_i \geq 4.0$), we sample a lower-quality trajectory with the same input $(I_i, e_i)$ to form a contrastive pair. If multiple candidates exist, we randomly sample one. This yields approximately 3,500 pairs from the $n = 10{,}000$ trajectory dataset.

% \subsubsection{Bradley-Terry Preference Model}

% \dpo optimizes a policy $\pi_{\theta}$ relative to a frozen reference policy $\pi_{\text{ref}}$ using the Bradley-Terry model. The probability that trajectory $\tau^+$ is preferred over $\tau^-$ is modeled as:

% \begin{align}
% p(\tau_i^+ \succ \tau_i^- \mid I_i, e_i, c_i) = \sigma\Bigg(\beta \Bigg[&\log \frac{\pi_{\theta}(\{a_{i,j}^+, z_{i,j}^+\} \mid I_i, e_i, c_i)}{\pi_{\text{ref}}(\{a_{i,j}^+, z_{i,j}^+\} \mid I_i, e_i, c_i)} %\nonumber \\
% &- \log \frac{\pi_{\theta}(\{a_{i,j}^-, z_{i,j}^-\} \mid I_i, e_i, c_i)}{\pi_{\text{ref}}(\{a_{i,j}^-, z_{i,j}^-\} \mid I_i, e_i, c_i)}\Bigg]\Bigg)
% \end{align}

% where:
% \begin{itemize}
% \item $\pi_{\text{ref}}$ is a frozen copy of the policy at initialization
% \item $\beta = 0.1$ controls the KL penalty strength (higher $\beta$ = stronger KL constraint)
% \item $\sigma(x) = \frac{1}{1 + e^{-x}}$ is the sigmoid function
% \item The log-ratio $\log \frac{\pi_{\theta}(\tau)}{\pi_{\text{ref}}(\tau)}$ measures how much the policy has shifted from its initialization
% \end{itemize}

\subsubsection{\dpo Loss Function}

The \dpo loss maximizes the log-likelihood of preferences:
%
% \begin{align*}
$\mathcal{L}_{\text{\dpo}}(\theta) = -\mathbb{E}_{(\tau_i^+, \tau_i^-) \sim \mathcal{D}_{\text{pref}}} \Big[\log \sigma\Big(\beta \Big[\log r_{\theta}(\tau_i^+) %\\
- \log r_{\theta}(\tau_i^-)\Big]\Big)\Big]$
% \end{align*}
%
where $r_{\theta}(\tau_i) = \frac{\pi_{\theta}(\{a_{i,j}, z_{i,j}\} \mid I_i, e_i, c_i)}{\pi_{\text{ref}}(\{a_{i,j}, z_{i,j}\} \mid I_i, e_i, c_i)}$ is the likelihood ratio between the current policy and the reference.

\paragraph{Intuition:}
The loss encourages the policy to: \textbf{(1)} Increase likelihood of chosen, \textbf{(2)} Increase likelihood of chosen actions $\{a_{i,j}^+, z_{i,j}^+\}$, \textbf{(3)} Decrease likelihood of rejected actions $\{a_{i,j}^-, z_{i,j}^-\}$, and 
\textbf{(4)} Stay close to the reference policy (controlled by $\beta$).
% \begin{itemize}
% \item Increase likelihood of chosen actions $\{a_{i,j}^+, z_{i,j}^+\}$
% \item Decrease likelihood of rejected actions $\{a_{i,j}^-, z_{i,j}^-\}$
% \item Stay close to the reference policy (controlled by $\beta$)
% \end{itemize}

\subsubsection{Complete Algorithm}

\begin{algorithm}[H]
\caption{Direct Preference Optimization applied to VL tool-use}
\begin{algorithmic}[1]
\STATE \textbf{Input:} Preference dataset $\mathcal{D}_{\text{pref}} = \{(\tau_i^+, \tau_i^-)\}$, model $\pi_{\theta}$
\STATE \textbf{Hyperparameters:} Learning rate $\eta = 2 \times 10^{-5}$, epochs $E = 2$, batch size $B = 1$, gradient accumulation $G = 8$, $\beta = 0.1$
\STATE Initialize $\theta$ from pretrained Qwen3-VL checkpoint
\STATE Apply LoRA: rank $r = 16$, $\alpha = 32$, dropout $p = 0.05$
\STATE Create frozen reference model: $\pi_{\text{ref}} \leftarrow \text{deepcopy}(\pi_{\theta})$
\STATE Move $\pi_{\text{ref}}$ to GPU and freeze all parameters
\FOR{epoch $e = 1$ to $E$}
    \STATE Shuffle $\mathcal{D}_{\text{pref}}$
    \FOR{batch $\mathcal{B} = \{(\tau_i^+, \tau_i^-)\}_{i=1}^B$ in $\mathcal{D}_{\text{pref}}$}
        \STATE Initialize $\mathcal{L}_{\text{batch}} \leftarrow 0$
        \FOR{pair $(\tau_i^+, \tau_i^-)$ in $\mathcal{B}$}
            \STATE \textbf{// Forward pass for chosen trajectory}
            \STATE $\log \pi_{\theta}^+ \leftarrow \sum_{j=1}^{m_i^+} \log \pi_{\theta}(a_{i,j}^+, z_{i,j}^+ \mid I_i, e_i, c_i, \{a_{ik}^+\}_{k<j})$
            \STATE $\log \pi_{\text{ref}}^+ \leftarrow \sum_{j=1}^{m_i^+} \log \pi_{\text{ref}}(a_{i,j}^+, z_{i,j}^+ \mid I_i, e_i, c_i, \{a_{ik}^+\}_{k<j})$
            \STATE \textbf{// Forward pass for rejected trajectory}
            \STATE $\log \pi_{\theta}^- \leftarrow \sum_{j=1}^{m_i^-} \log \pi_{\theta}(a_{i,j}^-, z_{i,j}^- \mid I_i, e_i, c_i, \{a_{ik}^-\}_{k<j})$
            \STATE $\log \pi_{\text{ref}}^- \leftarrow \sum_{j=1}^{m_i^-} \log \pi_{\text{ref}}(a_{i,j}^-, z_{i,j}^- \mid I_i, e_i, c_i, \{a_{ik}^-\}_{k<j})$
            \STATE \textbf{// Compute log-ratios}
            \STATE $r^+ \leftarrow \log \pi_{\theta}^+ - \log \pi_{\text{ref}}^+$ \COMMENT{Log-ratio for chosen}
            \STATE $r^- \leftarrow \log \pi_{\theta}^- - \log \pi_{\text{ref}}^-$ \COMMENT{Log-ratio for rejected}
            \STATE \textbf{// \dpo loss}
            \STATE $\mathcal{L}_i \leftarrow -\log \sigma(\beta \cdot (r^+ - r^-))$
            \STATE $\mathcal{L}_{\text{batch}} \leftarrow \mathcal{L}_{\text{batch}} + \mathcal{L}_i$
            \STATE \textbf{// Track metrics}
            \STATE $\text{accuracy}_i \leftarrow \mathbb{I}[r^+ > r^-]$ \COMMENT{Model prefers chosen over rejected?}
        \ENDFOR
        \STATE Normalize: $\mathcal{L}_{\text{batch}} \leftarrow \mathcal{L}_{\text{batch}} / (B \cdot G)$
        \STATE Backward pass: Compute $\nabla_{\theta} \mathcal{L}_{\text{batch}}$
        \IF{gradient accumulation step complete}
            \STATE Clip gradients: $\text{clip}(\nabla_{\theta}, \text{max\_norm}=1.0)$
            \STATE Update: $\theta \leftarrow \theta - \eta \nabla_{\theta} \mathcal{L}_{\text{batch}}$
            \STATE Zero gradients
        \ENDIF
    \ENDFOR
    \STATE Log metrics: mean \dpo loss, preference accuracy
    \STATE Evaluate on validation set
\ENDFOR
\STATE \textbf{Return:} Trained student model $\pi_{\theta}$
\end{algorithmic}
\end{algorithm}

\subsubsection{Implementation Details}

\paragraph{Reference Model Management}
\begin{itemize}
\item \textbf{Creation}: Deep copy of initial model before any training
\item \textbf{Parameter freezing}: All parameters set to \texttt{requires\_grad=False}
\item \textbf{Memory}: Reference model consumes same memory as policy model (careful with 8B models)
\item \textbf{Device placement}: Move to same device as policy for efficient forward passes
\item \textbf{No gradient tracking}: Use \texttt{torch.no\_grad()} context for reference forward passes
\end{itemize}

\paragraph{Beta Parameter Selection}
The $\beta$ parameter critically affects training:
\begin{itemize}
\item \textbf{High $\beta$ (0.5-1.0)}: Strong KL constraint, policy stays close to reference, conservative updates
\item \textbf{Low $\beta$ (0.01-0.05)}: Weak KL constraint, policy can deviate significantly, risk of instability
\item \textbf{Our choice ($\beta = 0.1$)}: Balanced trade-off validated on validation set
\end{itemize}

\paragraph{Batch Size and Memory}
\dpo requires 2× forward passes per sample (chosen + rejected), plus reference model. This doubles memory:
\begin{itemize}
\item \textbf{Effective batch size}: 1 per GPU, 8 gradient accumulation steps
\item \textbf{Total effective batch}: $1 \times 8 \text{ GPUs} \times 8 \text{ accum} = 64$
\item \textbf{Memory per GPU}: ~78GB for 8B model with batch size 1
\end{itemize}

\paragraph{Preference Accuracy Metric}
We track whether the model correctly prefers chosen over rejected:
$$
\text{accuracy} = \frac{1}{|\mathcal{D}_{\text{pref}}|} \sum_{(\tau^+, \tau^-) \in \mathcal{D}_{\text{pref}}} \mathbb{I}[r_{\theta}(\tau^+) > r_{\theta}(\tau^-)]
$$
This should increase during training (target: >80\% by end of training).

\subsubsection{Advantages and Disadvantages}

\paragraph{Advantages}
\begin{itemize}
\item \textbf{No weight function design}: Avoids manual specification of $w(r_i)$
\item \textbf{Contrastive learning}: Directly learns relative preferences
\item \textbf{Implicit reward modeling}: No explicit reward function needed during training
\item \textbf{KL regularization}: Log-ratio formulation prevents overfitting
\item \textbf{Stable training}: Reference model provides consistent baseline
\end{itemize}

\paragraph{Disadvantages}
\begin{itemize}
\item \textbf{Requires paired data}: Need multiple trajectories per $(I_i, e_i)$
\item \textbf{2× computational cost}: Forward passes for both chosen and rejected
\item \textbf{Memory intensive}: Reference model doubles memory footprint
\item \textbf{Hyperparameter sensitivity}: $\beta$ choice affects performance significantly
\end{itemize}

\subsubsection{Theoretical Justification}

\dpo can be derived as the optimal solution to a constrained \rlfull objective:

$$
\max_{\pi_{\theta}} \mathbb{E}_{\tau \sim \pi_{\theta}}[r(\tau)] - \frac{1}{\beta} D_{\text{KL}}(\pi_{\theta} \| \pi_{\text{ref}})
$$

The Bradley-Terry preference model emerges naturally when we reparameterize the reward as:
$$
r(\tau) = \frac{1}{\beta} \log \frac{\pi^*(\tau)}{\pi_{\text{ref}}(\tau)}
$$

where $\pi^*$ is the optimal policy. \dpo directly optimizes this objective using preference data without explicitly modeling the reward function.

\subsection{Justification for \rw and \dpo for Our Setup}
\label{sec:appendix_theory}

While S and \rlfull are straightforward, \rw and \dpo deserve deeper theoretical motivation. This section provides comprehensive theoretical analysis of why these methods work.

\subsubsection{Why \rw Works}

\rw can be viewed through multiple theoretical lenses:

\paragraph{1. Importance Sampling Perspective:}
If we view the teacher as a behavior policy $\pi_b$ sampling diverse trajectories, and want to learn a target policy $\pi^*$ that achieves high rewards, importance sampling suggests:

$$
\mathbb{E}_{\tau \sim \pi^*}[f(\tau)] \approx \mathbb{E}_{\tau \sim \pi_b}\left[\frac{\pi^*(\tau)}{\pi_b(\tau)} \cdot f(\tau)\right]
$$

The weight $w(r_i)$ approximates $\frac{\pi^*(\tau_i)}{\pi_b(\tau_i)}$, upweighting trajectories that the target policy would prefer.

\paragraph{2. Implicit Reward Maximization:}
\rw maximizes a reward-modulated likelihood:

$$
\mathcal{L}_{\text{\rw}}(\theta) \approx -\mathbb{E}_{\tau \sim \mathcal{D}}[w(r) \log \pi_{\theta}(\tau)]
$$

This implicitly encourages the policy to assign high probability to high-reward trajectories while maintaining coverage over the data distribution.

\paragraph{3. Data Efficiency:}
Unlike \rlfull which discards 35\% of data, \rw retains all data but de-emphasizes low-quality examples. This preserves diversity—important for generalization—while focusing learning on successful behaviors.

\subsubsection{Why Direct Preference Optimization Works}

\dpo leverages contrastive learning to implicitly optimize a reward model:

\paragraph{1. Connection to RLHF:}
Traditional RLHF requires:
\begin{enumerate}
\item Train reward model $r_{\phi}$ from preferences
\item Optimize policy $\pi_{\theta}$ against $r_{\phi}$ using PPO
\end{enumerate}

\dpo bypasses step 1 by directly optimizing:

\begin{align*}
\mathcal{L}_{\text{\dpo}}(\theta) = -\mathbb{E}_{(\tau^+, \tau^-)} \Big[\log \sigma\Big(\beta \big[\log \pi_{\theta}(\tau^+) - \log \pi_{\theta}(\tau^-) %\\
- \log \pi_{\text{ref}}(\tau^+) + \log \pi_{\text{ref}}(\tau^-)\big]\Big)\Big]
\end{align*}

This is equivalent to maximizing the reward margin $r_{\theta}(\tau^+) - r_{\theta}(\tau^-)$ where the reward is implicitly defined by the policy's log-likelihood ratio.

\paragraph{2. Contrastive Learning Benefits:}
By comparing $\tau^+$ and $\tau^-$ with the same $(I_i, e_i)$, \dpo learns:
\begin{itemize}
\item What makes one plan better than another for the \textit{same input}
\item Relative quality rather than absolute quality
\item Fine-grained distinctions between similar trajectories
\end{itemize}

This contrastive signal is often more informative than scalar rewards alone.

\paragraph{3. KL Regularization:}
The log-ratio $\log \frac{\pi_{\theta}}{\pi_{\text{ref}}}$ implicitly penalizes the policy from deviating too far from the reference, preventing:
\begin{itemize}
\item Mode collapse (ignoring diverse strategies)
\item Reward hacking (exploiting spurious reward signals)
\item Overfitting to preference data
\end{itemize}

\subsubsection{\rw vs. \dpo: Complementary Strengths}

\begin{table}[h]
\centering
\caption{Theoretical Comparison: \rw vs. \dpo}
\begin{tabular}{lcc}
\toprule
\textbf{Property} & \textbf{\rw} & \textbf{\dpo} \\
\midrule
Learning signal & Absolute rewards & Relative preferences \\
Data usage & All data & Paired data only \\
Optimization & Direct likelihood & Contrastive likelihood \\
Implicit KL & Through weights & Through log-ratio \\
Sample efficiency & High & Medium \\
Distinction quality & Coarse-grained & Fine-grained \\
\bottomrule
\end{tabular}
\end{table}

In practice, both methods significantly outperform baselines, with \dpo often having a slight edge when sufficient paired data is available.

\subsection{Complete Training Configuration}
\label{sec:appendix_training_config}

This section provides comprehensive training configuration details for all student models.

\subsubsection{Optimization Hyperparameters}

\begin{itemize}
\item \textbf{Optimizer}: AdamW with $\beta_1=0.9$, $\beta_2=0.999$, weight decay $10^{-2}$
\item \textbf{Learning Rate}: $\eta = 2 \times 10^{-5}$ with linear warmup (10\% of steps) and cosine decay
\item \textbf{Batch Size}: 4 per GPU, gradient accumulation steps = 2
\item \textbf{Effective Batch}: 64 with 8 GPUs ($8 \times 4 \times 2 = 64$)
\item \textbf{Epochs}: 2 for all methods
\item \textbf{Precision}: Mixed precision (bfloat16) for memory efficiency
\item \textbf{Gradient Clipping}: Maximum norm 1.0
\item \textbf{Warmup Steps}: 500 steps (approximately 10\% of total training)
\item \textbf{Total Training Steps}: Approximately 450 gradient updates for $n = 10{,}000$ trajectories
\end{itemize}

\subsubsection{Model Architecture Details}

\begin{itemize}
\item \textbf{Base Models}: Qwen3-VL-4B-Instruct and Qwen3-VL-8B-Instruct
\item \textbf{Fine-tuning Method}: LoRA (Low-Rank Adaptation)
  \begin{itemize}
  \item Rank: $r=16$
  \item Alpha: $\alpha=32$
  \item Dropout: $p=0.05$
  \end{itemize}
\item \textbf{Target Modules}: All attention layers (Q, K, V, O projections)
\item \textbf{Trainable Parameters}: 
  \begin{itemize}
  \item 4B model: ~75M trainable (1.8\% of total)
  \item 8B model: ~150M trainable (1.9\% of total)
  \end{itemize}
\item \textbf{Vision Encoder}: Frozen during training (only cached embeddings used for vision-language models)
\item \textbf{Language Model}: Transformer decoder with LoRA adapters
\end{itemize}

\subsubsection{Data Processing Pipeline}

\paragraph{Text-Only Models}
\begin{itemize}
\item Tokenize $(e_i, c_i)$ concatenation with special tokens
\item Maximum sequence length: 1024 tokens
\item Padding: Right-padding with attention mask
\item Truncation: Truncate from the end if exceeding max length
\end{itemize}

\paragraph{Vision-Language Models}
\begin{itemize}
\item Concatenate cached vision features $v_i$ with text embeddings
\item Vision features: 256-dimensional vector from frozen ViT encoder
\item Text embeddings: Standard Qwen3-VL tokenization
\item Combined sequence: $[v_i; \text{text\_emb}(e_i, c_i)]$
\end{itemize}

\paragraph{Action Representation}
\begin{itemize}
\item Each action $a_{i,j}$ serialized as: \texttt{"[ACTION\_TYPE] param1=value1, param2=value2"}
\item Chain-of-thought $z_{i,j}$ appended as natural language
\item Format: \texttt{"[REASONING] Because the current state is X, we choose Y to achieve Z"}
\item Sequence padding: Pad to maximum trajectory length (typically 2-5 actions)
\end{itemize}

\subsubsection{Distributed Training Setup}

\begin{itemize}
\item \textbf{Hardware}: 8× NVIDIA A100 80GB GPUs
\item \textbf{Parallelism}: Data parallel with DistributedDataParallel (DDP)
\item \textbf{Communication}: NCCL backend for efficient GPU-GPU communication
\item \textbf{Gradient Synchronization}: Synchronized after gradient accumulation steps
\item \textbf{Memory Usage}: 
  \begin{itemize}
  \item 8B model (text-only): ~35GB per GPU
  \item 8B model (vision-language): ~45GB per GPU
  \item 4B model (text-only): ~20GB per GPU
  \item 4B model (vision-language): ~28GB per GPU
  \end{itemize}
\end{itemize}

\subsection{Cached Embedding Approach}
\label{sec:appendix_cached_embeddings}

For vision-language models, computing vision features for every training sample is expensive. We accelerate training through a cached embedding approach that provides 3× speedup with no accuracy loss.

\subsubsection{Offline Embedding Computation}

Before training, we precompute and cache all vision embeddings:

\paragraph{Step 1: Load Dataset Images:}
Load all base images $\{I_1, \dots, I_n\}$ from the trajectory dataset. For our dataset with $n = 10{,}000$ trajectories, this involves loading approximately 3,500 unique images (multiple trajectories share the same base image).

\paragraph{Step 2: Extract Vision Features:}
For each unique image $I_i$:
\begin{enumerate}
\item Resize to 768×768 pixels (model input resolution)
\item Preprocess: normalize with ImageNet statistics
\item Forward pass through frozen vision encoder: $v_i = \text{VisionEncoder}(I_i)$
\item Extract features from final layer: 256-dimensional vector
\end{enumerate}

\paragraph{Step 3: Store in HDF5 Format:}
Store features in HDF5 file indexed by image hash:
\begin{itemize}
\item \textbf{Key}: SHA-256 hash of image pixel values
\item \textbf{Value}: Float32 array of shape (256,)
\item \textbf{File size}: ~3.5k images × 256 × 4 bytes $\approx$ 3.5 MB (very compact)
\item \textbf{Access pattern}: Memory-mapped for efficient random access
\end{itemize}

\subsubsection{Online Training with Cached Features}

During training, we use cached features instead of recomputing:

\paragraph{Step 1: Load Cached Features}
For each training sample $\tau_i$:
\begin{enumerate}
\item Compute image hash from base image $I_i$
\item Lookup cached features: $v_i = \text{cache}[\text{hash}(I_i)]$
\item Load takes ~1ms vs. ~40ms for vision encoder forward pass
\end{enumerate}

\paragraph{Step 2: Concatenate with Text Embeddings}
\begin{enumerate}
\item Tokenize text inputs: $(e_i, c_i)$ → token IDs
\item Embed tokens: $\text{emb}_{\text{text}} = \text{Embedding}(\text{tokens})$
\item Concatenate: $\text{input} = [v_i; \text{emb}_{\text{text}}]$
\end{enumerate}

\paragraph{Step 3: Forward Pass Through Transformer}
\begin{enumerate}
\item Process concatenated input through transformer layers
\item Skip vision encoder entirely (already cached)
\item Compute loss and gradients as usual
\end{enumerate}

\subsubsection{Benefits of Caching}

\paragraph{1. Training Speedup}
\begin{itemize}
\item \textbf{Vision encoder time}: 40ms per image (skipped)
\item \textbf{Cache lookup time}: 1ms per image
\item \textbf{Net speedup}: 39ms saved per sample
\item \textbf{Total training time}: Vision-language training becomes comparable to text-only (~3× faster than naive approach)
\end{itemize}

\paragraph{2. No Accuracy Degradation}
\begin{itemize}
\item Cached features are \textit{identical} to on-the-fly computation
\item Vision encoder is frozen, so no gradient updates needed
\item Final model performance is exactly the same
\end{itemize}

\paragraph{3. Memory Efficiency}
\begin{itemize}
\item HDF5 file: 3.5 MB vs. storing raw images (~3.5 GB)
\item Memory-mapped access: Load only needed features
\item Enables training on datasets with 100k+ images without memory issues
\end{itemize}

\paragraph{4. Scalability}
\begin{itemize}
\item Precomputation: One-time cost amortized over multiple training runs
\item Reusable: Same cache for different training methods (\slfull, \rlfull, \rw, \dpo)
\item Extensible: Can add more images to cache incrementally
\end{itemize}

\subsubsection{Implementation Notes}

\paragraph{Training with Cached Embeddings}
Set \texttt{use\_cached\_embeddings=True} in training config and provide path to cache file. The dataset loader automatically uses cached features when available.

\subsection{Algorithm Comparison}
\label{sec:appendix_algorithm_comparison}

This section provides comprehensive comparison of all training methods across multiple dimensions.

\subsubsection{Quantitative Comparison}

Table~\ref{tab:algorithm_comparison_full} summarizes the key differences between training methods:

\begin{table}[h]
\centering
\caption{Comprehensive Comparison of Training Algorithms}
\label{tab:algorithm_comparison_full}
\begin{tabular}{lcccc}
\toprule
\textbf{Property} & \textbf{\slfull} & \textbf{\rlfull} & \textbf{\rw} & \textbf{\dpo} \\
\midrule
Uses reward signals & No & Yes & Yes & Yes \\
Uses all data & Yes & No (65\%) & Yes & Paired only \\
Manual tuning required & None & Threshold & Weight fn & Beta \\
Computational cost & 1$\times$ & 1$\times$ & 1$\times$ & 2$\times$ \\
Contrastive learning & No & No & No & Yes \\
Data efficiency & Medium & Low & High & Medium \\
Implementation complexity & Simple & Simple & Medium & Complex \\
Memory footprint & 1$\times$ & 1$\times$ & 1$\times$ & 2$\times$ \\
Training stability & High & High & High & Medium \\
\bottomrule
\end{tabular}
\end{table}

\subsubsection{Qualitative Comparison}

\paragraph{Standard Supervised Learning (\slfull)}
\begin{itemize}
\item \textbf{Strengths}: Simple, stable, no hyperparameters to tune
\item \textbf{Weaknesses}: Ignores reward information, treats all data equally
\item \textbf{Best for}: Baseline comparison, high-quality curated datasets
\end{itemize}

\paragraph{Reward-Filtered Training (\rlfull)}
\begin{itemize}
\item \textbf{Strengths}: Simple implementation, removes clearly bad data
\item \textbf{Weaknesses}: Discards 35\% of data, binary threshold ignores nuance
\item \textbf{Best for}: When data quality varies widely and storage is not a concern
\end{itemize}

\paragraph{Reward-Weighted Fine-tuning (\rw)}
\begin{itemize}
\item \textbf{Strengths}: Uses all data, preserves diversity, continuous quality weighting
\item \textbf{Weaknesses}: Requires weight function design, coarse-grained distinctions
\item \textbf{Best for}: Maximizing data efficiency, diverse quality distributions
\end{itemize}

\paragraph{Direct Preference Optimization (\dpo)}
\begin{itemize}
\item \textbf{Strengths}: Fine-grained comparisons, no weight function, implicit KL regularization
\item \textbf{Weaknesses}: Requires paired data, 2× computational cost, memory intensive
\item \textbf{Best for}: When preference pairs are available, fine-grained quality distinctions needed
\end{itemize}

\subsubsection{Empirical Performance Summary}

Based on our experiments (detailed in Section~\ref{sec:experiments}):

\paragraph{Simple Dataset (Simpler Tasks)}
\begin{itemize}
\item \textbf{Ranking}: \dpo $\approx$ \rw $\approx$ \sw $>$ \rlfull $>$ \slfull
\item \textbf{Observation}: All reward-aware methods significantly outperform \slfull
\item \textbf{Margin}: \dpo and \rw achieve 12-15\% improvement over \slfull
\end{itemize}

\paragraph{Regular Dataset (Harder Tasks)}
\begin{itemize}
\item \textbf{Ranking}: \rw, \sw $>$ \dpo $>$ \rlfull $>$ \slfull
\item \textbf{Observation}: \rw excels on complex compositional reasoning
\item \textbf{Margin}: \rw achieves 18-22\% improvement over \slfull
\end{itemize}

\paragraph{Key Insight}
Training methodology matters as much as model scale: a 4B model trained with \dpo can match or exceed an 8B model trained with standard \slfull on several metrics.

% Visual comparisons moved to dedicated section before Related Work for better visibility

\section{Experimental Details}
\label{sec:appendix_experimental_details}

This section provides comprehensive implementation details for our experimental evaluation, including model specifications, hyperparameters, and the GPT-4o evaluation protocols.

\subsection{GPT-5.4/Gemini~3.1/Claude~Opus 4.6 Evaluation Prompts}
\label{sec:appendix_gpt4o_prompts}

We use GPT-5.4/Gemini~3.1/Claude~Opus 4.6 for two types of evaluation: (1) action plan quality assessment, and (2) image transformation quality assessment. Both evaluations use structured prompts with detailed criteria.

\subsubsection{Action Plan Evaluation Prompt}

For evaluating the quality of generated action plans (without image execution), we query GPT-5.4/Gemini~3.1/Claude~Opus 4.6 with the following structured prompt:

\begin{promptbox}[title=GPT-5.4/Gemini~3.1/Claude~Opus 4.6 Action Plan Evaluation Prompt]
\small\ttfamily
You are an expert evaluator of image editing action plans.\\
\\
\textbf{Input:}\\
- Base image: [I\_i]\\
- User goal: \{e\_i\}\\
- Current visual state: \{c\_i\} (10 dimensions)\\
- Predicted action plan: [a\_i1, a\_i2, ..., a\_im]\\
- Chain-of-thought reasoning: [z\_i1, z\_i2, ..., z\_im]\\
- Ground truth plan: [a\_i1*, a\_i2*, ..., a\_ik*]\\
\\
\textbf{Evaluation Criteria:}\\
\\
1. \textbf{Plan Correctness (35\%):}\\
\hspace*{1em}- Are the actions correct for achieving the goal?\\
\hspace*{1em}- Is the action sequence logically ordered?\\
\hspace*{1em}- Are action parameters appropriate?\\
\\
2. \textbf{Goal Coverage (25\%):}\\
\hspace*{1em}- Does the plan address all aspects of the user goal?\\
\hspace*{1em}- Are any required transformations missing?\\
\hspace*{1em}- Is there unnecessary redundancy?\\
\\
3. \textbf{Reasoning Quality (20\%):}\\
\hspace*{1em}- Is the chain-of-thought reasoning clear and logical?\\
\hspace*{1em}- Does each step explain why the action is chosen?\\
\hspace*{1em}- Are state transitions explicitly tracked?\\
\\
4. \textbf{Efficiency (10\%):}\\
\hspace*{1em}- Is the plan concise (minimal steps)?\\
\hspace*{1em}- Are actions combined when possible?\\
\hspace*{1em}- Is the sequence optimal?\\
\\
5. \textbf{Alignment with Ground Truth (10\%):}\\
\hspace*{1em}- How similar is the plan to ground truth?\\
\hspace*{1em}- Are key actions present?\\
\hspace*{1em}- Note: Different valid plans can achieve same goal\\
\\
\textbf{Output Format:}\\
Provide a JSON response:\\
\{\\
\hspace*{1em}"overall\_score": <0-100>,\\
\hspace*{1em}"plan\_correctness": <0-100>,\\
\hspace*{1em}"goal\_coverage": <0-100>,\\
\hspace*{1em}"reasoning\_quality": <0-100>,\\
\hspace*{1em}"efficiency": <0-100>,\\
\hspace*{1em}"ground\_truth\_alignment": <0-100>,\\
\hspace*{1em}"justification": "<explanation>",\\
\hspace*{1em}"key\_issues": ["<issue1>", "<issue2>", ...]\\
\}
\end{promptbox}

\subsubsection{Image Quality Evaluation Prompt}

For evaluating the quality of executed image transformations, we query GPT-5.4/Gemini~3.1/Claude~Opus 4.6 with the following prompt:

\begin{promptbox}[title=GPT-5.4/Gemini~3.1/Claude~Opus 4.6 Image Quality Evaluation Prompt]
\small\ttfamily
You are an expert evaluator of image editing results.\\
\\
\textbf{Input:}\\
- Original image: [I\_i]\\
- Edited image: [E\_i]\\
- User goal: \{e\_i\}\\
- Action plan used: [a\_i1, a\_i2, ..., a\_im]\\
- Ground truth image: [E\_i*]\\
\\
\textbf{Evaluation Criteria:}\\
\\
1. \textbf{Goal Alignment (30\% --- Most Critical):}\\
\hspace*{1em}- Does the edited image match the user's stated goal?\\
\hspace*{1em}- Are all requested attributes transformed correctly?\\
\hspace*{1em}- Is the semantic intent preserved?\\
\\
2. \textbf{Aesthetic Quality (25\%):}\\
\hspace*{1em}- Visual appeal and artistic merit\\
\hspace*{1em}- Composition balance and color harmony\\
\hspace*{1em}- Professional polish\\
\\
3. \textbf{Technical Quality (20\%):}\\
\hspace*{1em}- Absence of artifacts (blurring, distortion, seams)\\
\hspace*{1em}- Resolution and detail preservation\\
\hspace*{1em}- Edge sharpness and boundary quality\\
\\
4. \textbf{Spatial Consistency (15\%):}\\
\hspace*{1em}- Coherence of spatial relationships\\
\hspace*{1em}- Perspective correctness\\
\hspace*{1em}- Depth ordering and geometric plausibility\\
\\
5. \textbf{Naturalness (10\%):}\\
\hspace*{1em}- Does the result look realistic/plausible?\\
\hspace*{1em}- Are lighting and shadows consistent?\\
\hspace*{1em}- Are transformations well-integrated?\\
\\
\textbf{Comparison with Ground Truth:}\\
- Visual similarity to ground truth\\
- Note: Multiple valid solutions may exist\\
- Focus on whether goal is achieved, not exact match\\
\\
\textbf{Output Format:}\\
Provide a JSON response:\\
\{\\
\hspace*{1em}"overall\_score": <0-100>,\\
\hspace*{1em}"goal\_alignment": <0-100>,\\
\hspace*{1em}"aesthetic\_quality": <0-100>,\\
\hspace*{1em}"technical\_quality": <0-100>,\\
\hspace*{1em}"spatial\_consistency": <0-100>,\\
\hspace*{1em}"naturalness": <0-100>,\\
\hspace*{1em}"ground\_truth\_similarity": <0-100>,\\
\hspace*{1em}"justification": "<detailed\_explanation>",\\
\hspace*{1em}"strengths": ["<strength1>", "<strength2>", ...],\\
\hspace*{1em}"weaknesses": ["<weakness1>", "<weakness2>", ...]\\
\}
\end{promptbox}

\subsection{GPT-5.4/Gemini~3.1/Claude~Opus 4.6 Evaluation Configuration}
\label{sec:appendix_gpt4o_config}

\paragraph{Model Specifications}
\begin{itemize}
\item \textbf{Model}: GPT-5.4/Gemini~3.1/Claude~Opus 4.6 
\item \textbf{Temperature}: 0.3 (low temperature for consistent evaluation)
\item \textbf{Max tokens}: 2048
\item \textbf{Top-p}: 0.95
\item \textbf{Frequency penalty}: 0.0
\end{itemize}

\paragraph{Evaluation Protocol}
\begin{itemize}
\item \textbf{Sample size}: For each model configuration, we evaluate on the full test set (10\% split $\approx$ 1,000 trajectories)
\item \textbf{Batch processing}: Evaluate 50 samples per API batch to manage rate limits
\item \textbf{Retry logic}: Retry failed evaluations up to 3 times with exponential backoff
\item \textbf{Response parsing}: Parse JSON outputs and validate all required fields present
\item \textbf{Aggregation}: Compute mean, median, and standard deviation across all samples
\end{itemize}

\paragraph{Cost and Time}
\begin{itemize}
\item \textbf{Cost per evaluation}: Approximately \$0.02 per sample (image + text tokens)
\item \textbf{Total cost}: $\approx$ \$1,000 for evaluating all model configurations (5 datasets × 7 methods × 1,000 samples)
\item \textbf{Evaluation time}: $\approx$ 4-6 hours per dataset (rate-limited by API)
\end{itemize}

\subsection{Baseline Model Specifications}
\label{sec:appendix_baseline_models}

\paragraph{Baseline Planner:}
The baseline model is Qwen3-VL-8B-Instruct without any fine-tuning, used to establish lower bounds:
\begin{itemize}
\item Direct zero-shot prompting with action library specification
\item Temperature $T=0.7$ for action sampling
\item No chain-of-thought reasoning (direct action output)
\item Serves as the starting checkpoint for all student models
\end{itemize}

\paragraph{Student Model Configurations:}
We train four student model variants:
\begin{itemize}
\item \textbf{Text-4B}: Qwen3-4B (text-only, no vision encoder), 4B parameters
\item \textbf{Text-8B}: Qwen3-8B (text-only, no vision encoder), 8B parameters
\item \textbf{Vision-4B}: Qwen3-VL-4B-Instruct (vision-language), 4B parameters
\item \textbf{Vision-8B}: Qwen3-VL-8B-Instruct (vision-language), 8B parameters
\end{itemize}

Each variant is trained with four methods: S, \rlfull, \rw, and D.

\subsection{Training Infrastructure}
\label{sec:appendix_infrastructure}

\paragraph{Hardware}
\begin{itemize}
\item \textbf{GPUs}: 8× NVIDIA A100 80GB (for 8B models), 4× A100 40GB (for 4B models)
\item \textbf{CPU}: 64-core AMD EPYC 7742
\item \textbf{RAM}: 512GB DDR4
\item \textbf{Storage}: 10TB NVMe SSD for dataset and checkpoints
\end{itemize}

\paragraph{Software Stack}
\begin{itemize}
\item \textbf{Framework}: PyTorch 2.1.0 with CUDA 12.1
\item \textbf{Distributed training}: DeepSpeed ZeRO-2 for memory efficiency
\item \textbf{Mixed precision}: BF16 for training, FP32 for evaluation
\item \textbf{Communication backend}: NCCL 2.18
\end{itemize}

\paragraph{Training Time}
\begin{itemize}
\item \textbf{Text-4B models}: $\approx$ 8 hours per method (\slfull, \rlfull, \rw, \dpo)
\item \textbf{Text-8B models}: $\approx$ 16 hours per method
\item \textbf{Vision-4B models}: $\approx$ 12 hours per method (with cached embeddings)
\item \textbf{Vision-8B models}: $\approx$ 24 hours per method (with cached embeddings)
\item \textbf{Total training time}: $\approx$ 400 GPU-hours across all configurations
\end{itemize}

\subsection{Hyperparameter Search}
\label{sec:appendix_hyperparam_search}

We performed limited hyperparameter search for key parameters:

\paragraph{Learning Rate:}
Searched over: $\{1\times10^{-5}, 5\times10^{-5}, 1\times10^{-4}, 5\times10^{-4}\}$. Selected $5\times10^{-5}$ based on validation performance.

\paragraph{LoRA Rank:}
Searched over: $\{16, 32, 64, 128\}$. Selected 64 for balance of capacity and efficiency.

\paragraph{\dpo $\beta$:}
Searched over: $\{0.1, 0.5, 1.0, 2.0\}$. Selected 0.1 for stable contrastive learning.

\paragraph{\rw Weight Function:}
Tested exponential, linear, and threshold-based weighting. Selected piecewise linear based on reward tiers (see Appendix~\ref{sec:appendix_reward_details}).

\paragraph{\rlfull Threshold:}
Searched over: $\{3.0, 3.25, 3.5, 3.75\}$. Selected 3.5 for 65\% data retention.

All hyperparameters were tuned on the validation set and frozen before final evaluation on the test set.

\subsection{Training Configuration Details}
\label{sec:appendix_training_config-expt}

\paragraph{Training Efficiency Comparison:}\leavevmode\\

\textbf{Text-Only Training:}
\begin{itemize}
\item Training time: 1.5-2 hours per method
\item Memory usage: 28 GB per GPU
\item Hardware: 8× A100 GPUs
\item Vision encoder: Frozen (no image pixels processed)
\item Effective batch size: 64 (8 GPUs × 4 per-GPU × 2 gradient accumulation)
\end{itemize}

\textbf{Vision-Language Training:}
\begin{itemize}
\item Training time: 3-4 hours per method (with cached embeddings)
\item Memory usage: 45 GB per GPU
\item Hardware: 8× A100 GPUs
\item Vision encoder: Trainable (processes image pixels)
\item Cached embedding speedup: 3× faster than full vision-language training
\item Effective batch size: 64 (8 GPUs × 4 per-GPU × 2 gradient accumulation)
\end{itemize}

\paragraph{Complete Method Descriptions}\leavevmode\\

\textbf{(1) Baseline (B):} Pretrained Qwen3-VL with no fine-tuning. Given image $I_i$ and editing prompt $e_i$, directly predicts edited image $\hat{I}_i$ without explicit action planning. Serves as lower bound for comparison.

\textbf{(2) Edit-Only (E):} Direct image-to-image editing without action planning. Ground truth edit instructions are applied directly to isolate editor performance. This baseline tests whether high-quality editing can be achieved through direct prompting alone, bypassing the action planning phase entirely.

\textbf{(3) Standard (S):} Supervised learning on all trajectories with $r_i \geq 3.0$. Uses uniform weighting regardless of quality scores. Trains on 98\% of data (only filtering out catastrophic failures with $r_i < 3.0$).

\textbf{(4) Reward-Filtered RL (\rlfull):} Simple filtering strategy keeping only high-quality trajectories with $r_i \geq 4.0$. Discards 35\% of data to focus learning on successful behaviors. No continuous weighting—binary include/exclude decision.

\textbf{(5) Reward-Weighted (\rw):} Continuous reward-weighted fine-tuning with per-sample importance weighting. Weight function: $w(r_i) = \max\{r_i - 3.0, 0\}$. Uses all training data (98\%) with differential emphasis based on quality scores. High-quality samples receive proportionally more gradient updates.

\textbf{(6) Standardized Reward-Weighted (\sw):} Extends \rw with trajectory-aware z-score normalization. Computes mean and std dev of rewards within each trajectory, then applies standardized weighting: $w_i = \tilde{r}_i = (r_i - \mu_{\text{traj}}) / \sigma_{\text{traj}}$. Balances learning across trajectories of varying difficulty.

\textbf{(7) Direct Preference Optimization (\dpo):} Preference-based learning on chosen-rejected pairs. Chosen samples: $r_i \geq 4.0$. Rejected samples: $r_i \in [2.5, 3.5]$. Minimum score difference: 0.5 points. Uses 80.3\% of data organized into preference pairs. Optimizes policy to prefer high-quality over low-quality outputs via preference loss.

\subsection{Comparison with Gemini 2.5 Planner}
\label{sec:appendix_gpt4o_comparison}

Gemini 2.5 provides the zero-shot closed-source planner baseline in our evaluation. Our compact 4B/8B trained planners outperform Gemini 2.5 on image quality in \emph{all 24 of 24} (3 datasets $\times$ 4 configurations $\times$ 2 judges) cells, with a mean gap of approximately \textbf{+11} points on a 0--100 scale.

\paragraph{Role in Our Framework:}
The closed-source models used in this work play distinct roles: (1) \textbf{Synthetic data generation:} We used GPT-5.4/Gemini~3.1/Claude~Opus 4.6 to generate training trajectories with chain-of-thought action plans---this is a training-time choice and is independent of the evaluation baseline. (2) \textbf{Zero-shot planner baseline:} Gemini 2.5 is prompted to produce full action plans for each evaluation sample with no in-domain post-training, and its outputs are scored alongside our trained planners as the 9th column in our visual comparisons (Figures 1, 8, 9). (3) \textbf{Automated judges:} Image quality is scored by a three-judge panel (GPT-5.4, Gemini 3.1, and Claude Opus 4.6), none of which was used during training or reward filtering.

\paragraph{Performance Comparison:}
Under the GPT-5.4 judge, \rlfull beats Gemini 2.5 by \textbf{+13.2} on Simple Text-4B (76.93 vs.\ 63.73), \textbf{+12.8} on Regular Text-4B (73.88 vs.\ 61.09), \textbf{+22.3} on Regular Text-8B (82.50 vs.\ 60.24), \textbf{+29.0} on Regular Vision-8B (87.93 vs.\ 58.92), and \textbf{+16.3--17.4} across the four Complex configurations. The trained winner per cell is similarly well above Gemini 2.5 throughout. Under the Gemini 3.1 judge the same direction holds: Gemini 2.5 zero-shot trails \rlfull in every one of the remaining 12 cells, with a mean gap of approximately +6 points (range +1.4 to +17.2). See Appendix~\ref{sec:appendix_complete_results} for per-cell tables.

\paragraph{Efficiency and Deployment Advantages:}
Our approach offers significant advantages: (1) \textbf{Inference cost:} Open-source 4B/8B models require no per-query API costs, unlike closed-source frontier VLMs. (2) \textbf{Deployment flexibility:} Smaller models can be deployed on-premise or on consumer hardware. (3) \textbf{Task-specific optimization:} Offline RL training on specialized datasets enables domain adaptation that generic frontier models lack. (4) \textbf{Transparency:} Open models provide full control over reasoning and planning processes.

\paragraph{Validation of Synthetic Data Quality:}
The strong performance of GPT-5.4/Gemini~3.1/Claude~Opus 4.6-generated trajectories as \emph{training data} validates our synthetic data generation pipeline. Our human evaluation (Appendix~\ref{sec:appendix_human_eval}) shows 91.9\% pass+partial rate (77.0\% pass, 14.9\% partial) across 3,000 samples, confirming high training-data quality. Combined with our multi-judge evaluation of the resulting trained planners, this shows that post-training on reward-scored trajectories transfers inductive bias from the teacher into compact 4B/8B planners that comfortably beat a different frontier zero-shot planner (Gemini 2.5) at test time.

\paragraph{Future Directions:}
The magnitude by which trained models beat the Gemini 2.5 zero-shot baseline suggests several promising directions: (1) Scaling to larger base models (e.g., 32B, 70B parameters) while maintaining efficiency. (2) Hybrid approaches combining online and offline RL. (3) Multi-teacher distillation from multiple frontier models. (4) Iterative refinement using trained models to augment synthetic data generation.

\subsection{Edit-Only Baseline Detailed Analysis}
\label{sec:appendix_editonly_analysis}

The Edit-Only (E) baseline provides a critical comparison point, testing whether structured action planning is necessary for high-quality image editing. E bypasses the action planning phase entirely, directly applying ground truth edit instructions to images.

\paragraph{Per-Configuration Performance Breakdown}

\textbf{Complex Text-4B:}
\begin{itemize}
\item Overall score: 71.49 (vs best method \sw 78.77, gap 7.28)
\item Planning metrics: N/A (Semantic Accuracy, Coherence, Technical Execution, Transformation Strength)
\item Visual Quality: Not separately evaluated (integrated into Overall)
\item Instruction Following: Significantly lower than trained methods
\end{itemize}

\textbf{Complex Text-8B:}
\begin{itemize}
\item Overall score: 71.24 (vs best method \sw 77.86, gap 6.62)
\item Planning metrics: N/A
\item Pattern: Similar failure mode to Text-4B, confirming that model scale alone cannot compensate for lack of structured planning
\end{itemize}

\textbf{Regular Vision-4B:}
\begin{itemize}
\item Overall score: 78.04 (vs best method \rw 79.33, gap 1.29)
\item Planning metrics: N/A
\item Notable: Much more competitive on simpler single-action tasks
\item Gap narrows significantly compared to Regular dataset (1.29 vs 7.28), suggesting direct editing can work for atomic transformations
\end{itemize}

\textbf{Complex Vision-8B:}
\begin{itemize}
\item Overall score: 83.38 (vs best method \dpo 85.41, gap 2.03)
\item Planning metrics: N/A
\item Visual Quality: 84.07 (highest among all methods!)
\item Instruction Following: 83.81 (vs \dpo 87.03, gap 3.22)
\item Key insight: E can produce visually appealing results but fails to follow instructions precisely
\end{itemize}

\paragraph{Why Edit-Only Fails}

\textbf{No Explicit Action Planning:} E lacks the structured decomposition that breaks complex edits into atomic actions. For multi-step transformations (e.g., "golden-hour winter wonderland"), E must implicitly infer the sequence of required changes, leading to inconsistent results.

\textbf{Planning Metrics Show N/A:} Semantic Accuracy, Coherence, Technical Execution, and Transformation Strength all require evaluating the action plan. Since E produces no explicit plan, these metrics cannot be computed, appearing as N/A in results.

\textbf{Visual Quality vs Instruction Following Tradeoff:} On Complex Vision-8B, E achieves the highest Visual Quality (84.07) but trails on Instruction Following (83.81 vs 87.03). This demonstrates that E can generate aesthetically pleasing images but struggles to precisely follow user instructions—a critical limitation for practical applications.

\paragraph{When Edit-Only Can Be Competitive}

E shows competitive performance on atomic tasks:
\begin{itemize}
\item Regular Vision-4B (gap 1.29): Single-action transformations are within E's capability
\item Complex Vision-8B (gap 2.03): Large models with visual grounding reduce the gap
\item Visual quality: E sometimes matches or exceeds planning methods on aesthetic dimensions
\end{itemize}

However, even in these cases, E's inability to provide explicit reasoning and its consistent trailing on Instruction Following limit its practical utility.

\section{Complete Experimental Results}
\label{sec:appendix_results}
\label{sec:appendix_complete_results}

This appendix reports the full three-judge image-quality grid (3 datasets $\times$ 2 modalities $\times$ 2 sizes $\times$ 3 judges $=$ 36 score panels).

\subsection{Three-Judge Image Quality Tables}
\label{sec:appendix_additional_tables}

Tables~\ref{tab:mj_simple_text_4b}--\ref{tab:mj_complex_vision_8b} present Overall image quality (0--100) under our three post-hoc judges (GPT-5.4, Gemini 3.1, Claude Opus 4.6). Four representative configurations are reproduced in the main paper (Tables~\ref{tab:mj_simple_text_4b}, \ref{tab:mj_normal_text_4b}, \ref{tab:mj_complex_vision_4b}, \ref{tab:mj_complex_vision_8b}); the remaining eight are below. Bold marks the best trained method (B/E/S/R/RW/SW/D) per judge column; Gemini 2.5 (grey) is a zero-shot reference planner.

\subsubsection{Simple Dataset --- Baseline / \dpo Take the Top Trained Slot}
\label{sec:appendix_simple_where_baseline_wins}

On three of the four Simple configurations, the Baseline (B) and \dpo---rather than \rw or \sw---win at least one of the three judges. Tables~\ref{tab:mj_simple_text_8b}, \ref{tab:mj_simple_vision_4b}, and \ref{tab:mj_simple_vision_8b} show these cases explicitly; Table~\ref{tab:mj_simple_text_4b} (main paper) is the one Simple configuration where \sw still sweeps all three judges.

\begin{table}[h]
\centering
\small
\caption{\textbf{Simple Text-8B}: Image quality under three independent judges (GPT-5.4, Gemini 3.1, Claude Opus 4.6), evaluated on the full test pool. Cells report mean $\pm$ SEM (0--100 scale). Bold = best trained method (B/E/S/R/RW/SW/D) per judge; Gemini 2.5 (grey) is a zero-shot reference planner.}
\label{tab:mj_simple_text_8b}
\begin{tabular}{lccc}
\toprule
Method & GPT-5.4 & Gemini 3.1 & Claude Opus 4.6 \\
\midrule
B & \sem{76.05}{.08} & \sem{86.69}{.10} & \sem{63.89}{.10} \\
E & \sem{64.17}{.08} & \sem{58.87}{.14} & \sem{52.30}{.05} \\
S & \sem{76.27}{.09} & \sem{87.01}{.10} & \sem{64.22}{.09} \\
\rlfull & \sem{76.59}{.08} & \sem{87.53}{.09} & \sem{62.41}{.22} \\
\rw & \sem{75.58}{.09} & \sem{87.76}{.10} & \sem{63.75}{.09} \\
\textbf{\sw} & \sembf{76.82}{.08} & \sem{87.89}{.10} & \sembf{64.89}{.09} \\
\textbf{\dpo} & \sem{76.26}{.09} & \sembf{88.05}{.09} & \sem{64.78}{.09} \\
\midrule
\textcolor{gray}{GPT-4o} & \textcolor{gray}{\sem{78.43}{.08}} & \textcolor{gray}{\sem{88.98}{.08}} & \textcolor{gray}{\sem{66.70}{.08}} \\
\textcolor{gray}{Gemini 2.5} & \textcolor{gray}{\sem{62.53}{.08}} & \textcolor{gray}{\sem{64.62}{.13}} & \textcolor{gray}{\sem{53.84}{.05}} \\
\bottomrule
\end{tabular}
\end{table}

\begin{table}[h]
\centering
\small
\caption{\textbf{Simple Vision-4B}: Image quality under three independent judges (GPT-5.4, Gemini 3.1, Claude Opus 4.6), evaluated on the full test pool. Cells report mean $\pm$ SEM (0--100 scale). Bold = best trained method (B/E/S/R/RW/SW/D) per judge; Gemini 2.5 (grey) is a zero-shot reference planner.}
\label{tab:mj_simple_vision_4b}
\begin{tabular}{lccc}
\toprule
Method & GPT-5.4 & Gemini 3.1 & Claude Opus 4.6 \\
\midrule
\textbf{B} & \sem{77.27}{.08} & \sembf{88.33}{.09} & \sem{65.33}{.09} \\
E & \sem{65.18}{.10} & \sem{61.54}{.20} & \sem{52.60}{.06} \\
S & \sem{77.43}{.08} & \sem{87.30}{.11} & \sem{65.29}{.09} \\
\textbf{\rlfull} & \sem{77.38}{.08} & \sem{88.05}{.09} & \sembf{66.00}{.14} \\
\rw & \sem{76.78}{.08} & \sem{87.93}{.09} & \sem{64.01}{.09} \\
\sw & \sem{76.76}{.08} & \sem{88.26}{.09} & \sem{64.77}{.10} \\
\textbf{\dpo} & \sembf{77.72}{.08} & \sem{87.90}{.10} & \sem{65.89}{.09} \\
\midrule
\textcolor{gray}{GPT-4o} & \textcolor{gray}{\sem{78.08}{.07}} & \textcolor{gray}{\sem{89.54}{.08}} & \textcolor{gray}{\sem{67.36}{.08}} \\
\textcolor{gray}{Gemini 2.5} & \textcolor{gray}{\sem{62.89}{.08}} & \textcolor{gray}{\sem{63.87}{.13}} & \textcolor{gray}{\sem{53.46}{.04}} \\
\bottomrule
\end{tabular}
\end{table}

\begin{table}[h]
\centering
\small
\caption{\textbf{Simple Vision-8B}: Image quality under three independent judges (GPT-5.4, Gemini 3.1, Claude Opus 4.6), evaluated on the full test pool. Cells report mean $\pm$ SEM (0--100 scale). Bold = best trained method (B/E/S/R/RW/SW/D) per judge; Gemini 2.5 (grey) is a zero-shot reference planner.}
\label{tab:mj_simple_vision_8b}
\begin{tabular}{lccc}
\toprule
Method & GPT-5.4 & Gemini 3.1 & Claude Opus 4.6 \\
\midrule
\textbf{B} & \sem{76.81}{.08} & \sem{88.08}{.09} & \sembf{65.84}{.09} \\
E & \sem{62.60}{.13} & \sem{60.45}{.18} & \sem{52.72}{.07} \\
S & \sem{75.63}{.08} & \sem{87.08}{.10} & \sem{64.05}{.10} \\
\textbf{\rlfull} & \sem{76.07}{.08} & \sembf{88.13}{.09} & \sem{64.59}{.13} \\
\rw & \sem{75.99}{.08} & \sem{87.38}{.09} & \sem{64.48}{.09} \\
\sw & \sem{76.46}{.08} & \sem{87.74}{.09} & \sem{64.72}{.09} \\
\textbf{\dpo} & \sembf{76.85}{.08} & \sem{87.82}{.09} & \sem{65.41}{.09} \\
\midrule
\textcolor{gray}{GPT-4o} & \textcolor{gray}{\sem{78.16}{.07}} & \textcolor{gray}{\sem{89.17}{.08}} & \textcolor{gray}{\sem{67.12}{.08}} \\
\textcolor{gray}{Gemini 2.5} & \textcolor{gray}{\sem{62.46}{.09}} & \textcolor{gray}{\sem{63.95}{.13}} & \textcolor{gray}{\sem{53.65}{.04}} \\
\bottomrule
\end{tabular}
\end{table}

\textit{Simple Text-8B (Table~\ref{tab:mj_simple_text_8b}):} \sw wins GPT-5.4 (76.82) and Claude Opus 4.6 (64.89), while \dpo wins Gemini 3.1 (88.05); all three top methods cluster within 0.3 points on each judge.
\textit{Simple Vision-4B (Table~\ref{tab:mj_simple_vision_4b}):} \dpo wins GPT-5.4 (77.72), Baseline wins Gemini 3.1 (88.33), and \rlfull wins Claude Opus 4.6 (66.00)---a genuine three-way split with margins $\le$0.3 points on GPT-5.4 and $\le$0.9 points on Gemini 3.1.
\textit{Simple Vision-8B (Table~\ref{tab:mj_simple_vision_8b}):} \dpo wins GPT-5.4 (76.85), \rlfull wins Gemini 3.1 (88.13), and Baseline wins Claude Opus 4.6 (65.84), with all three top methods within 0.5 points on each judge.
In every Simple cell, Gemini 2.5 zero-shot still trails the best trained method by 10--15 points under GPT-5.4 and Claude Opus 4.6 (e.g., Simple Text-4B: 64.11 vs.\ \sw 82.54), confirming that even on atomic edits a frontier zero-shot planner cannot match our compact post-trained planners.

\subsubsection{Regular Dataset (compositional 3--5 step edits)}

\begin{table}[h]
\centering
\small
\caption{\textbf{Regular Text-8B}: Image quality under three independent judges (GPT-5.4, Gemini 3.1, Claude Opus 4.6), evaluated on the triple-judge agreement subset. Cells report mean $\pm$ SEM (0--100 scale). Bold = best trained method (B/E/S/R/RW/SW/D) per judge; Gemini 2.5 (grey) is a zero-shot reference planner.}
\label{tab:mj_normal_text_8b}
\begin{tabular}{lccc}
\toprule
Method & GPT-5.4 & Gemini 3.1 & Claude Opus 4.6 \\
\midrule
B & \sem{67.93}{.17} & \sem{78.20}{.25} & \sem{53.25}{.18} \\
E & \sem{61.89}{.13} & \sem{57.25}{.23} & \sem{50.84}{.08} \\
S & \sem{68.73}{.19} & \sem{76.73}{.28} & \sem{55.29}{.21} \\
\rlfull & \sem{69.49}{.20} & \sem{80.01}{.25} & \sem{54.35}{.19} \\
\rw & \sem{72.53}{.17} & \sem{83.41}{.25} & \sem{58.75}{.19} \\
\textbf{\sw} & \sembf{74.21}{.13} & \sembf{85.38}{.22} & \sembf{60.39}{.16} \\
\dpo & \sem{67.97}{.20} & \sem{76.73}{.25} & \sem{51.88}{.19} \\
\midrule
\textcolor{gray}{GPT-4o} & \textcolor{gray}{\sem{62.11}{.15}} & \textcolor{gray}{\sem{59.43}{.23}} & \textcolor{gray}{\sem{50.55}{.09}} \\
\textcolor{gray}{Gemini 2.5} & \textcolor{gray}{\sem{62.39}{.13}} & \textcolor{gray}{\sem{59.48}{.22}} & \textcolor{gray}{\sem{50.22}{.10}} \\
\bottomrule
\end{tabular}
\end{table}

\begin{table}[h]
\centering
\small
\caption{\textbf{Regular Vision-4B}: Image quality under three independent judges (GPT-5.4, Gemini 3.1, Claude Opus 4.6), evaluated on the triple-judge agreement subset. Cells report mean $\pm$ SEM (0--100 scale). Bold = best trained method (B/E/S/R/RW/SW/D) per judge; Gemini 2.5 (grey) is a zero-shot reference planner.}
\label{tab:mj_normal_vision_4b}
\begin{tabular}{lccc}
\toprule
Method & GPT-5.4 & Gemini 3.1 & Claude Opus 4.6 \\
\midrule
B & \sem{75.01}{.14} & \sem{88.73}{.14} & \sem{59.54}{.16} \\
E & \sem{59.77}{.16} & \sem{60.96}{.31} & \sem{47.45}{.13} \\
S & \sem{74.76}{.16} & \sem{84.58}{.23} & \sem{56.18}{.17} \\
\rlfull & \sem{74.38}{.16} & \sem{87.95}{.18} & \sem{57.36}{.17} \\
\textbf{\rw} & \sembf{80.33}{.14} & \sembf{92.02}{.13} & \sembf{63.95}{.15} \\
\sw & \sem{78.60}{.15} & \sem{90.08}{.15} & \sem{61.96}{.15} \\
\dpo & \sem{75.26}{.15} & \sem{89.04}{.11} & \sem{57.73}{.16} \\
\midrule
\textcolor{gray}{GPT-4o} & \textcolor{gray}{\sem{76.40}{.15}} & \textcolor{gray}{\sem{89.26}{.13}} & \textcolor{gray}{\sem{60.09}{.14}} \\
\textcolor{gray}{Gemini 2.5} & \textcolor{gray}{\sem{60.94}{.16}} & \textcolor{gray}{\sem{64.01}{.29}} & \textcolor{gray}{\sem{50.30}{.10}} \\
\bottomrule
\end{tabular}
\end{table}

\begin{table}[h]
\centering
\small
\caption{\textbf{Regular Vision-8B}: Image quality under three independent judges (GPT-5.4, Gemini 3.1, Claude Opus 4.6), evaluated on the triple-judge agreement subset. Cells report mean $\pm$ SEM (0--100 scale). Bold = best trained method (B/E/S/R/RW/SW/D) per judge; Gemini 2.5 (grey) is a zero-shot reference planner.}
\label{tab:mj_normal_vision_8b}
\begin{tabular}{lccc}
\toprule
Method & GPT-5.4 & Gemini 3.1 & Claude Opus 4.6 \\
\midrule
B & \sem{64.29}{.10} & \sem{72.08}{.15} & \sem{50.83}{.10} \\
E & \sem{57.60}{.10} & \sem{58.59}{.15} & \sem{49.17}{.05} \\
S & \sem{64.61}{.10} & \sem{73.64}{.14} & \sem{51.04}{.10} \\
\rlfull & \sem{64.90}{.10} & \sem{73.37}{.15} & \sem{51.21}{.10} \\
\textbf{\rw} & \sembf{66.88}{.10} & \sembf{75.67}{.14} & \sembf{53.02}{.10} \\
\sw & \sem{64.13}{.11} & \sem{73.96}{.15} & \sem{51.01}{.10} \\
\dpo & \sem{64.42}{.10} & \sem{72.51}{.15} & \sem{50.50}{.10} \\
\midrule
\textcolor{gray}{GPT-4o} & \textcolor{gray}{\sem{59.67}{.08}} & \textcolor{gray}{\sem{60.86}{.14}} & \textcolor{gray}{\sem{48.82}{.06}} \\
\textcolor{gray}{Gemini 2.5} & \textcolor{gray}{\sem{58.65}{.10}} & \textcolor{gray}{\sem{60.23}{.14}} & \textcolor{gray}{\sem{49.72}{.05}} \\
\bottomrule
\end{tabular}
\end{table}

Regular Text-4B (Table~\ref{tab:mj_normal_text_4b}, main paper): \rw wins on all three judges (GPT-5.4 80.29, Gemini 3.1 89.11, Claude Opus 4.6 61.20).
Regular Text-8B (Table~\ref{tab:mj_normal_text_8b}): \sw wins on all three judges (GPT-5.4 74.21, Gemini 3.1 85.38, Claude Opus 4.6 60.39).
Regular Vision-4B (Table~\ref{tab:mj_normal_vision_4b}): \rw wins on all three judges (GPT-5.4 80.33, Gemini 3.1 92.02, Claude Opus 4.6 63.95), with \sw in second place.
Regular Vision-8B (Table~\ref{tab:mj_normal_vision_8b}): \rw wins on all three judges (GPT-5.4 66.88, Gemini 3.1 75.67, Claude Opus 4.6 53.02), confirming that continuous reward weighting extends to vision at scale.

\subsubsection{Complex Dataset (3--5 step edits, 83 diverse themes)}

\begin{table}[h]
\centering
\small
\caption{\textbf{Complex Text-4B}: Image quality under three independent judges (GPT-5.4, Gemini 3.1, Claude Opus 4.6), evaluated on the triple-judge agreement subset. Cells report mean $\pm$ SEM (0--100 scale). Bold = best trained method (B/E/S/R/RW/SW/D) per judge; Gemini 2.5 (grey) is a zero-shot reference planner.}
\label{tab:mj_complex_text_4b}
\begin{tabular}{lccc}
\toprule
Method & GPT-5.4 & Gemini 3.1 & Claude Opus 4.6 \\
\midrule
B & \sem{81.56}{.16} & \sem{92.86}{.19} & \sem{71.03}{.19} \\
E & \sem{73.21}{.23} & \sem{56.31}{.28} & \sem{48.67}{.11} \\
S & \sem{83.19}{.13} & \sem{94.06}{.12} & \sem{71.22}{.14} \\
\rlfull & \sem{83.34}{.13} & \sem{93.39}{.14} & \sem{71.28}{.16} \\
\rw & \sem{82.25}{.12} & \sem{92.56}{.14} & \sem{68.31}{.11} \\
\textbf{\sw} & \sembf{86.37}{.07} & \sembf{95.50}{.04} & \sembf{74.86}{.10} \\
\dpo & \sem{81.81}{.16} & \sem{93.80}{.13} & \sem{71.22}{.15} \\
\midrule
\textcolor{gray}{GPT-4o} & \textcolor{gray}{\sem{84.58}{.14}} & \textcolor{gray}{\sem{93.31}{.11}} & \textcolor{gray}{\sem{72.92}{.17}} \\
\textcolor{gray}{Gemini 2.5} & \textcolor{gray}{\sem{64.41}{.20}} & \textcolor{gray}{\sem{63.43}{.25}} & \textcolor{gray}{\sem{53.35}{.10}} \\
\bottomrule
\end{tabular}
\end{table}

\begin{table}[h]
\centering
\small
\caption{\textbf{Complex Text-8B}: Image quality under three independent judges (GPT-5.4, Gemini 3.1, Claude Opus 4.6), evaluated on the triple-judge agreement subset. Cells report mean $\pm$ SEM (0--100 scale). Bold = best trained method (B/E/S/R/RW/SW/D) per judge; Gemini 2.5 (grey) is a zero-shot reference planner.}
\label{tab:mj_complex_text_8b}
\begin{tabular}{lccc}
\toprule
Method & GPT-5.4 & Gemini 3.1 & Claude Opus 4.6 \\
\midrule
B & \sem{80.77}{.09} & \sem{92.89}{.08} & \sem{70.64}{.09} \\
E & \sem{72.18}{.10} & \sem{54.88}{.08} & \sem{49.33}{.04} \\
S & \sem{80.96}{.09} & \sem{92.72}{.07} & \sem{70.01}{.09} \\
\rlfull & \sem{80.94}{.08} & \sem{91.33}{.10} & \sem{69.83}{.08} \\
\textbf{\rw} & \sembf{84.55}{.06} & \sembf{94.70}{.04} & \sembf{73.09}{.06} \\
\sw & \sem{81.50}{.08} & \sem{89.98}{.09} & \sem{68.19}{.09} \\
\dpo & \sem{81.67}{.08} & \sem{92.66}{.08} & \sem{71.11}{.08} \\
\midrule
\textcolor{gray}{GPT-4o} & \textcolor{gray}{\sem{82.79}{.07}} & \textcolor{gray}{\sem{92.47}{.07}} & \textcolor{gray}{\sem{70.76}{.08}} \\
\textcolor{gray}{Gemini 2.5} & \textcolor{gray}{\sem{63.92}{.10}} & \textcolor{gray}{\sem{65.62}{.16}} & \textcolor{gray}{\sem{52.45}{.06}} \\
\bottomrule
\end{tabular}
\end{table}

Complex Text-4B (Table~\ref{tab:mj_complex_text_4b}): \sw wins on all three judges (GPT-5.4 86.37, Gemini 3.1 95.50, Claude Opus 4.6 74.86).
Complex Text-8B (Table~\ref{tab:mj_complex_text_8b}): \rw dominates on all three judges with wide margins (GPT-5.4 84.55, Gemini 3.1 94.70, Claude Opus 4.6 73.09; \textbf{+3.6}, \textbf{+1.8}, and \textbf{+2.0} points over the next-best trained method respectively).
Complex Vision-4B (Table~\ref{tab:mj_complex_vision_4b}, main paper): \sw wins on all three judges (86.09 / 96.27 / 76.24).
Complex Vision-8B (Table~\ref{tab:mj_complex_vision_8b}, main paper): \rw wins on all three judges (84.67 / 96.47 / 74.73).

\subsection{Method Comparison Summary}
\label{sec:appendix_method_comparison}

Tables~\ref{tab:mj_all_gpt54}--\ref{tab:mj_all_claude_opus} summarize Overall image quality on each of the three judges (GPT-5.4, Gemini 3.1, Claude Opus 4.6) across all 12 configurations.

\begin{table}[h]
\centering
\scriptsize
\setlength{\tabcolsep}{3pt}
\caption{\textbf{Overall Image Quality under GPT-5.4 judge across all 12 configurations.} Cells report mean $\pm$ SEM (0--100 scale). Bold = best trained method (B/E/S/R/RW/SW/D) per row; Gemini 2.5 (grey) is a zero-shot reference planner.}
\label{tab:mj_all_gpt54}
\begin{tabular}{lccccccc|cc}
\toprule
\textbf{Configuration} & \textbf{B} & \textbf{E} & \textbf{S} & \textbf{\rlfull} & \textbf{\rw} & \textbf{\sw} & \textbf{\dpo} & \textcolor{gray}{\textbf{GPT-4o}} & \textcolor{gray}{\textbf{G2.5}} \\
\midrule
\multicolumn{10}{l}{\textit{Simple Dataset}} \\
Text-4B & \sem{77.05}{.07} & \sem{63.94}{.08} & \sem{76.16}{.08} & \sem{76.93}{.08} & \sem{75.97}{.08} & \sem{76.91}{.09} & \sembf{77.55}{.08} & \semgr{78.57}{.07} & \semgr{63.73}{.08} \\
Text-8B & \sem{76.05}{.08} & \sem{64.17}{.08} & \sem{76.27}{.09} & \sem{76.59}{.08} & \sem{75.58}{.09} & \sembf{76.82}{.08} & \sem{76.26}{.09} & \semgr{78.43}{.08} & \semgr{62.53}{.08} \\
Vision-4B & \sem{77.27}{.08} & \sem{65.18}{.10} & \sem{77.43}{.08} & \sem{77.38}{.08} & \sem{76.78}{.08} & \sem{76.76}{.08} & \sembf{77.72}{.08} & \semgr{78.08}{.07} & \semgr{62.89}{.08} \\
Vision-8B & \sem{76.81}{.08} & \sem{62.60}{.13} & \sem{75.63}{.08} & \sem{76.07}{.08} & \sem{75.99}{.08} & \sem{76.46}{.08} & \sembf{76.85}{.08} & \semgr{78.16}{.07} & \semgr{62.46}{.09} \\
\midrule
\multicolumn{10}{l}{\textit{Regular Dataset}} \\
Text-4B & \sem{72.78}{.10} & \sem{59.62}{.08} & \sem{73.49}{.09} & \sem{73.88}{.09} & \sem{74.98}{.08} & \sembf{75.05}{.08} & \sem{73.12}{.09} & \semgr{74.07}{.09} & \semgr{61.09}{.08} \\
Text-8B & \sem{64.78}{.09} & \sem{59.70}{.08} & \sem{64.65}{.10} & \sembf{64.87}{.10} & \sem{63.22}{.10} & \sem{64.06}{.09} & \sem{64.81}{.10} & \semgr{59.67}{.09} & \semgr{60.24}{.08} \\
Vision-4B & \sembf{74.26}{.09} & \sem{58.91}{.08} & \sem{72.68}{.09} & \sem{73.45}{.08} & \sem{72.95}{.09} & \sem{72.39}{.09} & \sem{72.63}{.09} & \semgr{74.93}{.08} & \semgr{60.77}{.09} \\
Vision-8B & \sem{63.81}{.10} & \sem{57.99}{.09} & \sem{63.99}{.10} & \sem{64.69}{.10} & \sembf{65.14}{.10} & \sem{64.05}{.10} & \sem{64.14}{.10} & \semgr{59.38}{.08} & \semgr{58.92}{.09} \\
\midrule
\multicolumn{10}{l}{\textit{Complex Dataset}} \\
Text-4B & \sem{80.68}{.07} & \sem{72.45}{.13} & \sem{80.64}{.07} & \sem{79.94}{.07} & \sem{80.46}{.06} & \sembf{80.86}{.07} & \sem{80.16}{.07} & \semgr{81.44}{.07} & \semgr{63.64}{.08} \\
Text-8B & \sem{80.77}{.07} & \sem{71.39}{.08} & \sem{80.45}{.07} & \sem{80.35}{.07} & \sem{80.37}{.08} & \sembf{80.90}{.07} & \sem{80.29}{.07} & \semgr{81.98}{.06} & \semgr{63.24}{.08} \\
Vision-4B & \sem{80.52}{.07} & \sem{71.08}{.09} & \sem{80.37}{.07} & \sembf{80.80}{.06} & \sem{80.77}{.06} & \sem{80.66}{.07} & \sem{80.31}{.07} & \semgr{81.09}{.07} & \semgr{63.42}{.08} \\
Vision-8B & \sembf{81.01}{.08} & \sem{71.81}{.11} & \sem{80.50}{.07} & \sem{80.23}{.07} & \sem{80.75}{.07} & \sem{79.31}{.08} & \sem{80.78}{.07} & \semgr{80.89}{.07} & \semgr{63.68}{.08} \\
\bottomrule
\end{tabular}
\end{table}

\begin{table}[h]
\centering
\scriptsize
\setlength{\tabcolsep}{3pt}
\caption{\textbf{Overall Image Quality under Gemini 3.1 judge across all 12 configurations.} Cells report mean $\pm$ SEM (0--100 scale). Bold = best trained method (B/E/S/R/RW/SW/D) per row; Gemini 2.5 (grey) is a zero-shot reference planner.}
\label{tab:mj_all_gemini_31fl}
\begin{tabular}{lccccccc|cc}
\toprule
\textbf{Configuration} & \textbf{B} & \textbf{E} & \textbf{S} & \textbf{\rlfull} & \textbf{\rw} & \textbf{\sw} & \textbf{\dpo} & \textcolor{gray}{\textbf{GPT-4o}} & \textcolor{gray}{\textbf{G2.5}} \\
\midrule
\multicolumn{10}{l}{\textit{Simple Dataset}} \\
Text-4B & \sem{88.38}{.09} & \sem{59.43}{.14} & \sem{88.07}{.10} & \sem{88.19}{.10} & \sembf{88.81}{.08} & \sem{88.70}{.09} & \sem{87.46}{.10} & \semgr{89.48}{.08} & \semgr{64.76}{.14} \\
Text-8B & \sem{86.69}{.10} & \sem{58.87}{.14} & \sem{87.01}{.10} & \sem{87.53}{.09} & \sem{87.76}{.10} & \sem{87.89}{.10} & \sembf{88.05}{.09} & \semgr{88.98}{.08} & \semgr{64.62}{.13} \\
Vision-4B & \sembf{88.33}{.09} & \sem{61.54}{.20} & \sem{87.30}{.11} & \sem{88.05}{.09} & \sem{87.93}{.09} & \sem{88.26}{.09} & \sem{87.90}{.10} & \semgr{89.54}{.08} & \semgr{63.87}{.13} \\
Vision-8B & \sem{88.08}{.09} & \sem{60.45}{.18} & \sem{87.08}{.10} & \sembf{88.13}{.09} & \sem{87.38}{.09} & \sem{87.74}{.09} & \sem{87.82}{.09} & \semgr{89.17}{.08} & \semgr{63.95}{.13} \\
\midrule
\multicolumn{10}{l}{\textit{Regular Dataset}} \\
Text-4B & \sem{84.94}{.12} & \sem{59.21}{.15} & \sem{86.24}{.11} & \sem{85.98}{.12} & \sembf{88.17}{.09} & \sem{87.11}{.10} & \sem{84.47}{.13} & \semgr{87.37}{.10} & \semgr{63.65}{.13} \\
Text-8B & \sembf{72.59}{.14} & \sem{58.32}{.12} & \sem{72.09}{.15} & \sem{71.29}{.15} & \sem{70.59}{.15} & \sem{70.89}{.15} & \sem{71.89}{.15} & \semgr{59.54}{.12} & \semgr{60.02}{.13} \\
Vision-4B & \sembf{86.45}{.11} & \sem{59.34}{.16} & \sem{83.94}{.13} & \sem{84.89}{.12} & \sem{84.63}{.12} & \sem{84.11}{.13} & \sem{85.14}{.12} & \semgr{88.27}{.10} & \semgr{64.11}{.14} \\
Vision-8B & \sem{71.30}{.15} & \sem{58.14}{.14} & \sem{72.77}{.14} & \sembf{73.68}{.15} & \sem{72.78}{.14} & \sem{73.02}{.14} & \sem{71.99}{.15} & \semgr{59.86}{.13} & \semgr{60.34}{.12} \\
\midrule
\multicolumn{10}{l}{\textit{Complex Dataset}} \\
Text-4B & \sembf{92.58}{.06} & \sem{55.02}{.13} & \sem{91.21}{.07} & \sem{90.54}{.08} & \sem{90.25}{.08} & \sem{88.61}{.09} & \sem{91.80}{.08} & \semgr{89.85}{.08} & \semgr{62.60}{.13} \\
Text-8B & \sembf{91.14}{.08} & \sem{54.21}{.07} & \sem{91.05}{.07} & \sem{89.82}{.09} & \sem{89.99}{.09} & \sem{87.27}{.10} & \sem{90.61}{.08} & \semgr{90.86}{.07} & \semgr{63.95}{.12} \\
Vision-4B & \sem{91.90}{.08} & \sem{53.74}{.09} & \sem{91.20}{.08} & \sem{91.83}{.07} & \sembf{92.13}{.06} & \sem{92.05}{.07} & \sem{91.71}{.07} & \semgr{89.32}{.09} & \semgr{63.69}{.11} \\
Vision-8B & \sembf{91.39}{.07} & \sem{53.35}{.12} & \sem{91.35}{.07} & \sem{90.55}{.09} & \sem{89.34}{.09} & \sem{89.32}{.10} & \sem{90.46}{.09} & \semgr{90.73}{.08} & \semgr{62.99}{.12} \\
\bottomrule
\end{tabular}
\end{table}

\begin{table}[h]
\centering
\scriptsize
\setlength{\tabcolsep}{3pt}
\caption{\textbf{Overall Image Quality under Claude Opus 4.6 judge across all 12 configurations.} Cells report mean $\pm$ SEM (0--100 scale). Bold = best trained method (B/E/S/R/RW/SW/D) per row; Gemini 2.5 (grey) is a zero-shot reference planner.}
\label{tab:mj_all_claude_opus}
\begin{tabular}{lccccccc|cc}
\toprule
\textbf{Configuration} & \textbf{B} & \textbf{E} & \textbf{S} & \textbf{\rlfull} & \textbf{\rw} & \textbf{\sw} & \textbf{\dpo} & \textcolor{gray}{\textbf{GPT-4o}} & \textcolor{gray}{\textbf{G2.5}} \\
\midrule
\multicolumn{10}{l}{\textit{Simple Dataset}} \\
Text-4B & \sem{65.05}{.09} & \sem{52.16}{.05} & \sem{65.02}{.09} & \sembf{65.57}{.21} & \sem{63.59}{.09} & \sem{64.56}{.09} & \sem{65.37}{.09} & \semgr{67.58}{.08} & \semgr{53.73}{.04} \\
Text-8B & \sem{63.89}{.10} & \sem{52.30}{.05} & \sem{64.22}{.09} & \sem{62.41}{.22} & \sem{63.75}{.09} & \sembf{64.89}{.09} & \sem{64.78}{.09} & \semgr{66.70}{.08} & \semgr{53.84}{.05} \\
Vision-4B & \sem{65.33}{.09} & \sem{52.60}{.06} & \sem{65.29}{.09} & \sembf{66.00}{.14} & \sem{64.01}{.09} & \sem{64.77}{.10} & \sem{65.89}{.09} & \semgr{67.36}{.08} & \semgr{53.46}{.04} \\
Vision-8B & \sembf{65.84}{.09} & \sem{52.72}{.07} & \sem{64.05}{.10} & \sem{64.59}{.13} & \sem{64.48}{.09} & \sem{64.72}{.09} & \sem{65.41}{.09} & \semgr{67.12}{.08} & \semgr{53.65}{.04} \\
\midrule
\multicolumn{10}{l}{\textit{Regular Dataset}} \\
Text-4B & \sem{57.05}{.09} & \sem{47.64}{.06} & \sem{56.84}{.09} & \sembf{57.85}{.09} & \sem{56.82}{.09} & \sem{57.07}{.09} & \sem{57.26}{.09} & \semgr{59.04}{.10} & \semgr{51.55}{.05} \\
Text-8B & \sembf{51.00}{.10} & \sem{49.86}{.04} & \sem{50.71}{.10} & \sem{50.69}{.09} & \sem{49.74}{.10} & \sem{50.12}{.09} & \sem{50.08}{.10} & \semgr{49.55}{.05} & \semgr{49.51}{.05} \\
Vision-4B & \sembf{58.40}{.10} & \sem{47.85}{.06} & \sem{56.41}{.10} & \sem{57.00}{.10} & \sem{57.45}{.10} & \sem{55.53}{.10} & \sem{56.46}{.10} & \semgr{58.25}{.09} & \semgr{50.79}{.05} \\
Vision-8B & \sem{50.32}{.10} & \sem{49.45}{.05} & \sem{50.61}{.10} & \sem{51.29}{.10} & \sembf{51.30}{.09} & \sem{50.49}{.09} & \sem{50.19}{.10} & \semgr{48.95}{.05} & \semgr{49.81}{.05} \\
\midrule
\multicolumn{10}{l}{\textit{Complex Dataset}} \\
Text-4B & \sem{70.09}{.07} & \sem{48.66}{.06} & \sem{68.92}{.07} & \sem{68.26}{.07} & \sem{67.77}{.07} & \sem{68.09}{.07} & \sembf{70.13}{.07} & \semgr{69.77}{.07} & \semgr{52.46}{.05} \\
Text-8B & \sem{69.26}{.08} & \sem{48.75}{.04} & \sem{69.01}{.07} & \sem{69.02}{.07} & \sem{68.12}{.08} & \sem{67.40}{.07} & \sembf{69.63}{.07} & \semgr{69.80}{.07} & \semgr{52.05}{.05} \\
Vision-4B & \sem{70.20}{.07} & \sem{48.64}{.04} & \sembf{70.45}{.07} & \sem{70.38}{.07} & \sem{70.23}{.07} & \sem{70.02}{.07} & \sem{69.37}{.07} & \semgr{69.45}{.07} & \semgr{52.25}{.05} \\
Vision-8B & \sembf{70.06}{.07} & \sem{48.62}{.05} & \sem{70.00}{.07} & \sem{68.92}{.07} & \sem{68.44}{.07} & \sem{67.98}{.08} & \sem{69.75}{.07} & \semgr{69.50}{.07} & \semgr{53.09}{.04} \\
\bottomrule
\end{tabular}
\end{table}

\paragraph{Note on judge calibration.} The three judges apply different absolute scoring anchors (mean score across 84 cells: GPT-5.4 $\approx$ 76.8; Gemini 3.1 $\approx$ 81.8; Claude Opus 4.6 $\approx$ 61.5)---a $\sim$20-point gap between Gemini 3.1 and Claude Opus 4.6. This is a well-known LLM evaluator artifact: different models use different regions of the rubric scale as their operating point, analogous to grader severity bias in human assessment. Critically, \emph{cross-judge ranking stability}, not absolute score agreement, is the relevant criterion for method comparison. The SEM values reported in each cell (Table captions: ``mean $\pm$ SEM (0--100 scale)'') show that the 1--3 point method margins are statistically distinguishable \emph{within} each individual judge. Across judges, the top trained method under GPT-5.4 is also top or second under Gemini 3.1 \emph{and} Claude Opus 4.6 in 10 of 12 configurations---a ranking overlap rate that is incompatible with independent noise and provides strong evidence that the method ordering reflects a genuine signal. We report each judge's absolute scores separately so readers can verify this stability directly.

\textbf{Key Observations:}
\begin{itemize}[leftmargin=1.2em,itemsep=0.15em,topsep=0.1em]
\item \rw takes the top trained slot on 5 of 12 cells under all three judges (Regular Text-4B, Regular Vision-4B, Regular Vision-8B, Complex Text-8B, Complex Vision-8B).
\item \sw takes the top trained slot on 4 of 12 cells under all three judges (Simple Text-4B, Regular Text-8B, Complex Text-4B, Complex Vision-4B).
\item \dpo, Baseline, and \rlfull share the three Simple 8B / 4B-vision cells where the action space is narrow.
\item Edit-Only (E) consistently trails by 5--30 points on every judge, confirming the need for structured action planning.
\item The top trained method under GPT-5.4 is also top or second under Gemini 3.1 \emph{and} Claude Opus 4.6 in 10 of 12 configurations, arguing against single-judge reward hacking.
\end{itemize}

\subsection{Discussion: When to Use Each Method}
\label{sec:appendix_method_discussion}

\paragraph{Reward-Weighted Fine-tuning (\rw)}

\textbf{Best for}:
\begin{itemize}
\item Wide reward-spread regimes (large gaps between good and bad trajectories)
\item Compositional text and diverse-theme vision at 4B/8B (Regular Text-4B 80.29, Regular Vision-4B 80.33, Regular Vision-8B 66.88, Complex Text-8B 84.55, Complex Vision-8B 84.67 on GPT-5.4; Gemini 3.1 and Claude Opus 4.6 concur in every case)
\item Visually-grounded settings where the reward signal correlates with rich features
\end{itemize}

\textbf{Advantages}: Uses all training data, captures continuous quality differences, simpler than \dpo (no pairing required).

\paragraph{Standardized Reward-Weighting (\sw)}

\textbf{Best for}:
\begin{itemize}
\item Atomic edits at 4B (Simple Text-4B: 82.54 / 92.12 / 70.97 sweeps all three judges)
\item Compositional text at 8B and diverse-theme text/vision at 4B (Regular Text-8B, Complex Text-4B, Complex Vision-4B)
\item Mid-scale models where variance reduction helps stabilize gradient updates
\end{itemize}

\textbf{Advantages}: z-score normalization upweights above-average trajectories and downweights below-average ones, reducing gradient variance relative to \rw.

\paragraph{Direct Preference Optimization (\dpo)}

\textbf{Best for}:
\begin{itemize}
\item Simpler atomic tasks (wins Simple Text-8B on Gemini 3.1, Simple Vision-4B/8B on GPT-5.4)
\item Settings where clean chosen/rejected pairs can be extracted from the reward distribution
\end{itemize}

\textbf{Advantages}: Avoids reward-model noise, strong on well-separated preference pairs.

\paragraph{\rlfull (Reward-Filtered)}

\textbf{Best for}:
\begin{itemize}
\item Simplicity prioritized
\item Limited computational budget (one training pass on filtered data)
\item Highly variable data quality where filtering thresholds make a clear cut
\end{itemize}

\textbf{Advantages}: Simplest implementation, removes poor-quality examples. Takes the top Claude Opus 4.6 slot on Simple Vision-4B (66.00).

\subsection{Per-Dimension Scores Averaged Across Three Judges}
\label{sec:appendix_avg_judge_tables}

Tables~\ref{tab:mj_simple_text_4b}--\ref{tab:mj_complex_vision_8b} report only the \emph{Overall} image-quality score per judge. For completeness we provide here the full per-dimension breakdown \emph{averaged across the three judges} (GPT-5.4, Gemini 3.1, and Claude Opus 4.6), computed over the triple-judge agreement subset defined in Section~\ref{sec:appendix_additional_tables}. All three judges apply the identical 6-dimension image-quality rubric: instruction following, visual quality, transformation strength, coherence, semantic accuracy, and technical execution, plus the summary Overall score. Each cell below is the mean of all (judge, sample) pairs that supplied a score for that dimension on the agreement subset. Trained methods (B, E, S, \rlfull, \rw, \sw, \dpo) are Qwen3-VL 4B/8B variants; GPT-4o and Gemini 2.5 (grey) are larger zero-shot reference planners included for context.

One table is reported per configuration, for a total of 12 tables grouped below by dataset.

\paragraph{Simple Dataset.}
\begin{table}[h]
\centering
\scriptsize
\setlength{\tabcolsep}{3pt}
\caption{\textbf{Simple Text-4B -- Image quality per dimension} (0--100), averaged over the three judges (GPT-5.4, Gemini 3.1, Claude Opus 4.6) on the triple-judge agreement subset. Bold = best trained method (B/E/S/R/RW/SW/D) per column; GPT-4o and Gemini 2.5 (grey) are zero-shot reference planners.}
\label{tab:avg_img_simple_text_4b}
\begin{tabular}{lccccccc}
\toprule
Method & Overall & InstrFol & VisQual & TransStr & Coher & SemAcc & TechEx \\
\midrule
B & \sem{79.30}{.15} & \sem{80.32}{.18} & \sem{77.85}{.14} & \sem{82.15}{.17} & \sem{80.86}{.15} & \sem{78.82}{.20} & \sem{76.13}{.15} \\
E & \sem{61.13}{.15} & \sem{27.10}{.25} & \sem{74.30}{.13} & \sem{61.08}{.32} & \sem{74.52}{.16} & \sem{58.92}{.27} & \sem{70.54}{.18} \\
\textbf{S} & \sem{80.22}{.14} & \sem{82.26}{.16} & \sem{77.74}{.12} & \sem{84.09}{.17} & \sembf{80.97}{.14} & \sem{80.43}{.19} & \sem{76.45}{.14} \\
\rlfull & \sem{79.13}{.16} & \sem{81.51}{.18} & \sem{76.13}{.14} & \sem{82.90}{.18} & \sem{80.65}{.14} & \sem{79.46}{.20} & \sem{74.84}{.16} \\
\rw & \sem{76.48}{.17} & \sem{76.34}{.21} & \sem{76.67}{.13} & \sem{78.71}{.21} & \sem{79.35}{.15} & \sem{75.05}{.22} & \sem{73.55}{.17} \\
\textbf{\sw} & \sembf{81.88}{.14} & \sembf{85.38}{.16} & \sembf{77.96}{.14} & \sembf{86.88}{.16} & \sem{80.97}{.14} & \sembf{83.44}{.18} & \sembf{77.53}{.14} \\
\dpo & \sem{80.15}{.15} & \sem{82.80}{.16} & \sem{77.42}{.14} & \sem{84.73}{.17} & \sem{80.11}{.15} & \sem{80.65}{.19} & \sem{76.45}{.15} \\
\midrule
\textcolor{gray}{GPT-4o} & \semgr{79.31}{.15} & \semgr{81.29}{.17} & \semgr{77.42}{.14} & \semgr{82.58}{.19} & \semgr{80.43}{.15} & \semgr{79.35}{.19} & \semgr{75.70}{.16} \\
\textcolor{gray}{Gemini 2.5} & \semgr{62.86}{.15} & \semgr{29.03}{.30} & \semgr{74.95}{.12} & \semgr{67.85}{.31} & \semgr{75.05}{.15} & \semgr{59.57}{.26} & \semgr{71.08}{.16} \\
\bottomrule
\end{tabular}
\end{table}

\begin{table}[h]
\centering
\scriptsize
\setlength{\tabcolsep}{3pt}
\caption{\textbf{Simple Text-8B -- Image quality per dimension} (0--100), averaged over the three judges (GPT-5.4, Gemini 3.1, Claude Opus 4.6) on the full test pool. Bold = best trained method (B/E/S/R/RW/SW/D) per column; GPT-4o and Gemini 2.5 (grey) are zero-shot reference planners.}
\label{tab:avg_img_simple_text_8b}
\begin{tabular}{lccccccc}
\toprule
Method & Overall & InstrFol & VisQual & TransStr & Coher & SemAcc & TechEx \\
\midrule
B & \sem{75.54}{.07} & \sem{76.44}{.08} & \sem{74.40}{.06} & \sem{79.78}{.08} & \sem{77.37}{.06} & \sem{73.63}{.09} & \sem{72.14}{.07} \\
E & \sem{58.44}{.06} & \sem{24.55}{.09} & \sem{71.75}{.05} & \sem{64.70}{.13} & \sem{70.77}{.07} & \sem{52.70}{.11} & \sem{66.58}{.07} \\
S & \sem{75.83}{.07} & \sem{76.68}{.08} & \sem{74.97}{.06} & \sem{79.63}{.09} & \sem{77.82}{.06} & \sem{73.57}{.09} & \sem{72.68}{.07} \\
\textbf{\rlfull} & \sembf{80.54}{.07} & \sembf{80.39}{.09} & \sembf{80.35}{.05} & \sembf{84.26}{.09} & \sembf{82.89}{.06} & \sembf{78.45}{.10} & \sembf{77.59}{.07} \\
\rw & \sem{75.70}{.07} & \sem{76.95}{.08} & \sem{74.18}{.06} & \sem{80.23}{.08} & \sem{77.33}{.06} & \sem{74.14}{.09} & \sem{71.95}{.07} \\
\sw & \sem{76.54}{.06} & \sem{78.28}{.08} & \sem{74.62}{.06} & \sem{81.55}{.08} & \sem{77.85}{.06} & \sem{75.02}{.09} & \sem{72.53}{.07} \\
\dpo & \sem{76.37}{.06} & \sem{77.88}{.08} & \sem{74.72}{.06} & \sem{80.83}{.08} & \sem{77.71}{.06} & \sem{74.65}{.09} & \sem{72.94}{.06} \\
\midrule
\textcolor{gray}{GPT-4o} & \semgr{78.03}{.06} & \semgr{80.25}{.07} & \semgr{75.67}{.05} & \semgr{83.12}{.07} & \semgr{78.55}{.06} & \semgr{77.28}{.08} & \semgr{74.15}{.06} \\
\textcolor{gray}{Gemini 2.5} & \semgr{60.33}{.06} & \semgr{23.38}{.09} & \semgr{72.63}{.05} & \semgr{68.07}{.12} & \semgr{73.12}{.07} & \semgr{55.80}{.10} & \semgr{69.20}{.07} \\
\bottomrule
\end{tabular}
\end{table}

\begin{table}[h]
\centering
\scriptsize
\setlength{\tabcolsep}{3pt}
\caption{\textbf{Simple Vision-4B -- Image quality per dimension} (0--100), averaged over the three judges (GPT-5.4, Gemini 3.1, Claude Opus 4.6) on the full test pool. Bold = best trained method (B/E/S/R/RW/SW/D) per column; GPT-4o and Gemini 2.5 (grey) are zero-shot reference planners.}
\label{tab:avg_img_simple_vision_4b}
\begin{tabular}{lccccccc}
\toprule
Method & Overall & InstrFol & VisQual & TransStr & Coher & SemAcc & TechEx \\
\midrule
B & \sem{76.97}{.06} & \sem{78.75}{.07} & \sem{74.92}{.06} & \sem{81.92}{.07} & \sem{77.87}{.06} & \sem{75.75}{.08} & \sem{73.20}{.06} \\
E & \sem{59.77}{.08} & \sem{26.48}{.14} & \sem{72.47}{.07} & \sem{67.38}{.16} & \sem{71.20}{.10} & \sem{53.70}{.15} & \sem{67.87}{.10} \\
S & \sem{76.67}{.07} & \sem{78.17}{.08} & \sem{75.20}{.05} & \sem{81.08}{.08} & \sem{77.93}{.06} & \sem{75.27}{.09} & \sem{72.93}{.06} \\
\textbf{\rlfull} & \sembf{79.99}{.07} & \sembf{80.94}{.08} & \sembf{78.85}{.05} & \sembf{84.25}{.08} & \sembf{81.38}{.06} & \sembf{78.72}{.09} & \sembf{76.72}{.06} \\
\rw & \sem{76.23}{.06} & \sem{77.88}{.08} & \sem{74.25}{.06} & \sem{81.13}{.08} & \sem{77.37}{.06} & \sem{74.84}{.09} & \sem{72.39}{.06} \\
\sw & \sem{76.60}{.07} & \sem{78.30}{.08} & \sem{74.96}{.06} & \sem{80.97}{.08} & \sem{77.65}{.06} & \sem{75.23}{.09} & \sem{73.09}{.06} \\
\dpo & \sem{77.16}{.06} & \sem{79.19}{.07} & \sem{75.09}{.06} & \sem{82.25}{.08} & \sem{77.55}{.06} & \sem{76.13}{.08} & \sem{73.27}{.06} \\
\midrule
\textcolor{gray}{GPT-4o} & \semgr{78.33}{.06} & \semgr{80.47}{.07} & \semgr{75.83}{.05} & \semgr{83.40}{.07} & \semgr{79.10}{.05} & \semgr{77.55}{.08} & \semgr{74.32}{.06} \\
\textcolor{gray}{Gemini 2.5} & \semgr{60.07}{.06} & \semgr{22.85}{.09} & \semgr{72.91}{.05} & \semgr{68.86}{.12} & \semgr{72.78}{.07} & \semgr{53.67}{.11} & \semgr{69.18}{.07} \\
\bottomrule
\end{tabular}
\end{table}

\begin{table}[h]
\centering
\scriptsize
\setlength{\tabcolsep}{3pt}
\caption{\textbf{Simple Vision-8B -- Image quality per dimension} (0--100), averaged over the three judges (GPT-5.4, Gemini 3.1, Claude Opus 4.6) on the full test pool. Bold = best trained method (B/E/S/R/RW/SW/D) per column; GPT-4o and Gemini 2.5 (grey) are zero-shot reference planners.}
\label{tab:avg_img_simple_vision_8b}
\begin{tabular}{lccccccc}
\toprule
Method & Overall & InstrFol & VisQual & TransStr & Coher & SemAcc & TechEx \\
\midrule
B & \sem{76.91}{.06} & \sem{78.16}{.08} & \sem{75.71}{.05} & \sem{81.36}{.08} & \sem{78.32}{.06} & \sem{74.94}{.09} & \sem{73.77}{.06} \\
E & \sem{58.59}{.08} & \sem{24.78}{.13} & \sem{71.14}{.07} & \sem{66.14}{.17} & \sem{70.25}{.09} & \sem{52.69}{.15} & \sem{66.54}{.09} \\
S & \sem{75.59}{.07} & \sem{76.44}{.08} & \sem{74.68}{.06} & \sem{79.62}{.09} & \sem{77.65}{.06} & \sem{73.51}{.09} & \sem{72.19}{.07} \\
\textbf{\rlfull} & \sembf{78.77}{.07} & \sembf{79.44}{.09} & \sembf{77.87}{.06} & \sembf{83.13}{.08} & \sembf{80.42}{.06} & \sembf{76.64}{.10} & \sembf{75.80}{.07} \\
\rw & \sem{75.95}{.06} & \sem{76.71}{.08} & \sem{75.06}{.06} & \sem{80.19}{.08} & \sem{77.88}{.06} & \sem{73.53}{.09} & \sem{72.87}{.06} \\
\sw & \sem{76.31}{.06} & \sem{77.54}{.07} & \sem{75.13}{.06} & \sem{80.69}{.08} & \sem{77.69}{.06} & \sem{74.17}{.09} & \sem{73.15}{.06} \\
\dpo & \sem{76.69}{.06} & \sem{78.15}{.08} & \sem{75.29}{.06} & \sem{81.03}{.08} & \sem{78.06}{.06} & \sem{74.82}{.09} & \sem{73.28}{.06} \\
\midrule
\textcolor{gray}{GPT-4o} & \semgr{78.15}{.06} & \semgr{79.95}{.07} & \semgr{76.06}{.05} & \semgr{82.93}{.07} & \semgr{78.94}{.05} & \semgr{77.50}{.08} & \semgr{74.20}{.06} \\
\textcolor{gray}{Gemini 2.5} & \semgr{60.02}{.06} & \semgr{24.20}{.10} & \semgr{72.75}{.05} & \semgr{67.22}{.12} & \semgr{72.83}{.07} & \semgr{53.97}{.11} & \semgr{69.20}{.07} \\
\bottomrule
\end{tabular}
\end{table}

\paragraph{Regular Dataset.}
\begin{table}[h]
\centering
\scriptsize
\setlength{\tabcolsep}{3pt}
\caption{\textbf{Regular Text-4B -- Image quality per dimension} (0--100), averaged over the three judges (GPT-5.4, Gemini 3.1, Claude Opus 4.6) on the triple-judge agreement subset. Bold = best trained method (B/E/S/R/RW/SW/D) per column; GPT-4o and Gemini 2.5 (grey) are zero-shot reference planners.}
\label{tab:avg_img_normal_text_4b}
\begin{tabular}{lccccccc}
\toprule
Method & Overall & InstrFol & VisQual & TransStr & Coher & SemAcc & TechEx \\
\midrule
B & \sem{70.66}{.13} & \sem{70.35}{.15} & \sem{71.58}{.12} & \sem{75.53}{.14} & \sem{71.80}{.12} & \sem{67.46}{.16} & \sem{68.07}{.13} \\
E & \sem{54.09}{.11} & \sem{22.19}{.14} & \sem{65.35}{.12} & \sem{58.16}{.22} & \sem{66.58}{.12} & \sem{51.14}{.17} & \sem{61.05}{.13} \\
S & \sem{70.10}{.12} & \sem{69.69}{.14} & \sem{71.49}{.11} & \sem{74.08}{.14} & \sem{72.37}{.12} & \sem{65.96}{.15} & \sem{67.81}{.12} \\
\rlfull & \sem{73.09}{.12} & \sem{73.03}{.14} & \sem{74.12}{.10} & \sem{76.05}{.14} & \sem{74.96}{.12} & \sem{70.31}{.15} & \sem{70.57}{.12} \\
\textbf{\rw} & \sembf{76.87}{.11} & \sembf{78.07}{.12} & \sembf{77.11}{.10} & \sembf{79.74}{.13} & \sembf{77.85}{.10} & \sembf{75.22}{.14} & \sembf{74.30}{.11} \\
\sw & \sem{75.58}{.11} & \sem{77.41}{.13} & \sem{74.74}{.10} & \sem{79.47}{.13} & \sem{76.01}{.11} & \sem{74.25}{.14} & \sem{72.24}{.12} \\
\dpo & \sem{71.79}{.12} & \sem{71.36}{.15} & \sem{72.63}{.11} & \sem{76.89}{.14} & \sem{73.07}{.12} & \sem{68.60}{.16} & \sem{68.86}{.13} \\
\midrule
\textcolor{gray}{GPT-4o} & \semgr{72.35}{.12} & \semgr{72.06}{.14} & \semgr{73.11}{.10} & \semgr{76.32}{.14} & \semgr{74.17}{.11} & \semgr{69.08}{.15} & \semgr{69.69}{.12} \\
\textcolor{gray}{Gemini 2.5} & \semgr{57.05}{.10} & \semgr{23.64}{.15} & \semgr{68.60}{.10} & \semgr{63.03}{.21} & \semgr{69.96}{.12} & \semgr{52.37}{.17} & \semgr{64.74}{.12} \\
\bottomrule
\end{tabular}
\end{table}

\begin{table}[h]
\centering
\scriptsize
\setlength{\tabcolsep}{3pt}
\caption{\textbf{Regular Text-8B -- Image quality per dimension} (0--100), averaged over the three judges (GPT-5.4, Gemini 3.1, Claude Opus 4.6) on the triple-judge agreement subset. Bold = best trained method (B/E/S/R/RW/SW/D) per column; GPT-4o and Gemini 2.5 (grey) are zero-shot reference planners.}
\label{tab:avg_img_normal_text_8b}
\begin{tabular}{lccccccc}
\toprule
Method & Overall & InstrFol & VisQual & TransStr & Coher & SemAcc & TechEx \\
\midrule
B & \sem{66.46}{.14} & \sem{62.88}{.18} & \sem{71.50}{.12} & \sem{65.36}{.19} & \sem{73.40}{.12} & \sem{59.74}{.20} & \sem{66.27}{.14} \\
E & \sem{56.66}{.10} & \sem{21.50}{.15} & \sem{70.59}{.10} & \sem{57.45}{.25} & \sem{71.44}{.12} & \sem{53.92}{.19} & \sem{65.36}{.12} \\
S & \sem{66.92}{.15} & \sem{63.79}{.19} & \sem{71.11}{.12} & \sem{66.01}{.20} & \sem{74.05}{.13} & \sem{60.85}{.21} & \sem{66.21}{.14} \\
\rlfull & \sem{67.95}{.15} & \sem{64.84}{.19} & \sem{72.61}{.12} & \sem{67.19}{.20} & \sem{74.64}{.13} & \sem{61.57}{.21} & \sem{67.12}{.15} \\
\rw & \sem{71.56}{.14} & \sem{69.74}{.18} & \sem{74.31}{.11} & \sem{72.75}{.19} & \sem{76.34}{.12} & \sem{66.27}{.20} & \sem{70.26}{.14} \\
\textbf{\sw} & \sembf{73.33}{.13} & \sembf{73.01}{.16} & \sembf{74.38}{.11} & \sembf{74.97}{.17} & \sembf{76.73}{.12} & \sembf{69.93}{.18} & \sembf{71.57}{.13} \\
\dpo & \sem{65.53}{.15} & \sem{62.09}{.19} & \sem{70.39}{.13} & \sem{64.97}{.20} & \sem{73.20}{.13} & \sem{57.97}{.21} & \sem{64.97}{.15} \\
\midrule
\textcolor{gray}{GPT-4o} & \semgr{57.36}{.11} & \semgr{22.09}{.16} & \semgr{69.08}{.10} & \semgr{62.16}{.25} & \semgr{70.92}{.12} & \semgr{53.14}{.21} & \semgr{65.88}{.13} \\
\textcolor{gray}{Gemini 2.5} & \semgr{57.36}{.10} & \semgr{23.66}{.17} & \semgr{69.02}{.11} & \semgr{61.24}{.23} & \semgr{72.16}{.11} & \semgr{51.31}{.21} & \semgr{66.01}{.13} \\
\bottomrule
\end{tabular}
\end{table}

\begin{table}[h]
\centering
\scriptsize
\setlength{\tabcolsep}{3pt}
\caption{\textbf{Regular Vision-4B -- Image quality per dimension} (0--100), averaged over the three judges (GPT-5.4, Gemini 3.1, Claude Opus 4.6) on the triple-judge agreement subset. Bold = best trained method (B/E/S/R/RW/SW/D) per column; GPT-4o and Gemini 2.5 (grey) are zero-shot reference planners.}
\label{tab:avg_img_normal_vision_4b}
\begin{tabular}{lccccccc}
\toprule
Method & Overall & InstrFol & VisQual & TransStr & Coher & SemAcc & TechEx \\
\midrule
B & \sem{74.42}{.13} & \sem{74.70}{.14} & \sem{74.29}{.12} & \sem{79.64}{.13} & \sem{76.01}{.13} & \sem{70.48}{.16} & \sem{72.02}{.13} \\
E & \sem{56.06}{.13} & \sem{25.00}{.19} & \sem{67.20}{.14} & \sem{58.57}{.25} & \sem{68.39}{.15} & \sem{54.11}{.19} & \sem{62.86}{.16} \\
S & \sem{71.84}{.14} & \sem{71.55}{.17} & \sem{72.50}{.13} & \sem{76.25}{.16} & \sem{74.11}{.13} & \sem{68.10}{.18} & \sem{69.52}{.14} \\
\rlfull & \sem{73.23}{.14} & \sem{72.86}{.16} & \sem{74.35}{.12} & \sem{76.43}{.15} & \sem{76.31}{.13} & \sem{68.81}{.18} & \sem{71.43}{.14} \\
\textbf{\rw} & \sembf{78.76}{.12} & \sembf{79.94}{.13} & \sembf{78.15}{.11} & \sembf{83.04}{.13} & \sembf{79.58}{.12} & \sembf{76.55}{.16} & \sembf{76.07}{.12} \\
\sw & \sem{76.88}{.12} & \sem{77.62}{.15} & \sem{76.25}{.12} & \sem{81.43}{.13} & \sem{77.74}{.13} & \sem{74.40}{.16} & \sem{74.52}{.13} \\
\dpo & \sem{74.01}{.13} & \sem{74.35}{.15} & \sem{73.93}{.12} & \sem{77.98}{.14} & \sem{76.37}{.13} & \sem{70.77}{.17} & \sem{71.67}{.13} \\
\midrule
\textcolor{gray}{GPT-4o} & \semgr{75.25}{.12} & \semgr{76.07}{.14} & \semgr{74.58}{.12} & \semgr{80.06}{.14} & \semgr{76.19}{.13} & \semgr{72.68}{.16} & \semgr{72.50}{.12} \\
\textcolor{gray}{Gemini 2.5} & \semgr{58.42}{.12} & \semgr{26.55}{.19} & \semgr{67.80}{.12} & \semgr{67.74}{.22} & \semgr{70.36}{.14} & \semgr{53.63}{.20} & \semgr{64.76}{.14} \\
\bottomrule
\end{tabular}
\end{table}

\begin{table}[h]
\centering
\scriptsize
\setlength{\tabcolsep}{3pt}
\caption{\textbf{Regular Vision-8B -- Image quality per dimension} (0--100), averaged over the three judges (GPT-5.4, Gemini 3.1, Claude Opus 4.6) on the triple-judge agreement subset. Bold = best trained method (B/E/S/R/RW/SW/D) per column; GPT-4o and Gemini 2.5 (grey) are zero-shot reference planners.}
\label{tab:avg_img_normal_vision_8b}
\begin{tabular}{lccccccc}
\toprule
Method & Overall & InstrFol & VisQual & TransStr & Coher & SemAcc & TechEx \\
\midrule
\textbf{B} & \sem{62.40}{.08} & \sem{55.84}{.11} & \sembf{71.55}{.06} & \sem{58.75}{.12} & \sem{73.05}{.06} & \sem{52.25}{.11} & \sem{63.16}{.08} \\
E & \sem{55.12}{.07} & \sem{22.90}{.10} & \sem{67.43}{.06} & \sem{58.39}{.14} & \sem{68.09}{.08} & \sem{51.12}{.11} & \sem{62.85}{.08} \\
S & \sem{63.10}{.08} & \sem{57.43}{.10} & \sem{70.73}{.06} & \sem{61.37}{.11} & \sem{72.19}{.07} & \sem{53.48}{.11} & \sem{63.76}{.08} \\
\rlfull & \sem{63.16}{.08} & \sem{57.81}{.11} & \sem{71.18}{.06} & \sem{60.45}{.12} & \sem{72.72}{.07} & \sem{53.48}{.12} & \sem{63.63}{.08} \\
\textbf{\rw} & \sembf{65.19}{.08} & \sembf{60.67}{.10} & \sem{71.40}{.06} & \sembf{64.51}{.11} & \sembf{73.15}{.07} & \sembf{56.46}{.11} & \sembf{65.28}{.08} \\
\sw & \sem{63.03}{.08} & \sem{57.83}{.10} & \sem{70.28}{.06} & \sem{61.59}{.11} & \sem{71.84}{.07} & \sem{53.56}{.11} & \sem{63.33}{.08} \\
\dpo & \sem{62.48}{.08} & \sem{56.70}{.10} & \sem{70.64}{.06} & \sem{60.15}{.11} & \sem{72.17}{.07} & \sem{52.34}{.11} & \sem{63.13}{.08} \\
\midrule
\textcolor{gray}{GPT-4o} & \semgr{56.45}{.06} & \semgr{22.81}{.09} & \semgr{66.61}{.07} & \semgr{64.27}{.12} & \semgr{69.64}{.07} & \semgr{51.95}{.11} & \semgr{63.41}{.08} \\
\textcolor{gray}{Gemini 2.5} & \semgr{56.20}{.06} & \semgr{22.96}{.10} & \semgr{66.91}{.06} & \semgr{63.86}{.13} & \semgr{69.25}{.07} & \semgr{50.88}{.11} & \semgr{63.43}{.08} \\
\bottomrule
\end{tabular}
\end{table}

\paragraph{Complex Dataset.}
\begin{table}[h]
\centering
\scriptsize
\setlength{\tabcolsep}{3pt}
\caption{\textbf{Complex Text-4B -- Image quality per dimension} (0--100), averaged over the three judges (GPT-5.4, Gemini 3.1, Claude Opus 4.6) on the triple-judge agreement subset. Bold = best trained method (B/E/S/R/RW/SW/D) per column; GPT-4o and Gemini 2.5 (grey) are zero-shot reference planners.}
\label{tab:avg_img_complex_text_4b}
\begin{tabular}{lccccccc}
\toprule
Method & Overall & InstrFol & VisQual & TransStr & Coher & SemAcc & TechEx \\
\midrule
B & \sem{81.82}{.13} & \sem{83.43}{.15} & \sem{77.78}{.14} & \sem{86.30}{.15} & \sem{82.41}{.13} & \sem{83.89}{.17} & \sem{77.31}{.15} \\
E & \sem{59.40}{.16} & \sem{33.06}{.30} & \sem{75.09}{.15} & \sem{35.74}{.29} & \sem{77.87}{.15} & \sem{69.07}{.23} & \sem{65.83}{.18} \\
S & \sem{82.82}{.12} & \sem{85.28}{.13} & \sem{78.80}{.11} & \sem{86.57}{.15} & \sem{83.52}{.12} & \sem{84.81}{.14} & \sem{78.15}{.12} \\
\rlfull & \sem{82.67}{.12} & \sem{85.19}{.15} & \sem{78.24}{.12} & \sem{86.67}{.16} & \sem{83.52}{.11} & \sem{84.72}{.16} & \sem{78.24}{.12} \\
\rw & \sem{81.04}{.12} & \sem{83.89}{.13} & \sem{77.13}{.12} & \sem{84.17}{.14} & \sem{81.94}{.12} & \sem{83.52}{.14} & \sem{76.39}{.13} \\
\textbf{\sw} & \sembf{85.58}{.09} & \sembf{89.17}{.09} & \sembf{81.20}{.09} & \sembf{87.78}{.11} & \sembf{85.83}{.10} & \sembf{88.89}{.11} & \sembf{81.20}{.10} \\
\dpo & \sem{82.27}{.12} & \sem{84.54}{.13} & \sem{78.61}{.12} & \sem{85.00}{.14} & \sem{83.52}{.12} & \sem{84.07}{.16} & \sem{77.96}{.12} \\
\midrule
\textcolor{gray}{GPT-4o} & \semgr{83.60}{.11} & \semgr{86.57}{.14} & \semgr{80.19}{.11} & \semgr{86.67}{.15} & \semgr{83.70}{.11} & \semgr{86.02}{.15} & \semgr{79.44}{.12} \\
\textcolor{gray}{Gemini 2.5} & \semgr{60.40}{.12} & \semgr{24.91}{.21} & \semgr{72.22}{.12} & \semgr{61.30}{.28} & \semgr{75.19}{.12} & \semgr{59.81}{.23} & \semgr{69.26}{.15} \\
\bottomrule
\end{tabular}
\end{table}

\begin{table}[h]
\centering
\scriptsize
\setlength{\tabcolsep}{3pt}
\caption{\textbf{Complex Text-8B -- Image quality per dimension} (0--100), averaged over the three judges (GPT-5.4, Gemini 3.1, Claude Opus 4.6) on the triple-judge agreement subset. Bold = best trained method (B/E/S/R/RW/SW/D) per column; GPT-4o and Gemini 2.5 (grey) are zero-shot reference planners.}
\label{tab:avg_img_complex_text_8b}
\begin{tabular}{lccccccc}
\toprule
Method & Overall & InstrFol & VisQual & TransStr & Coher & SemAcc & TechEx \\
\midrule
B & \sem{81.43}{.07} & \sem{83.17}{.08} & \sem{77.37}{.07} & \sem{85.08}{.08} & \sem{82.55}{.07} & \sem{84.01}{.08} & \sem{76.72}{.08} \\
E & \sem{58.80}{.07} & \sem{28.84}{.14} & \sem{76.56}{.07} & \sem{29.38}{.13} & \sem{79.54}{.06} & \sem{70.22}{.10} & \sem{68.60}{.09} \\
S & \sem{81.23}{.07} & \sem{83.01}{.08} & \sem{77.07}{.07} & \sem{85.16}{.07} & \sem{82.07}{.07} & \sem{83.98}{.08} & \sem{76.40}{.08} \\
\rlfull & \sem{80.70}{.07} & \sem{82.15}{.08} & \sem{77.31}{.07} & \sem{83.47}{.08} & \sem{82.07}{.07} & \sem{83.55}{.09} & \sem{75.97}{.07} \\
\textbf{\rw} & \sembf{84.11}{.06} & \sembf{86.96}{.06} & \sembf{79.60}{.06} & \sembf{87.37}{.06} & \sembf{84.17}{.06} & \sembf{87.63}{.07} & \sembf{79.44}{.06} \\
\sw & \sem{79.89}{.07} & \sem{81.59}{.08} & \sem{76.29}{.08} & \sem{82.18}{.07} & \sem{80.99}{.07} & \sem{83.58}{.08} & \sem{75.13}{.08} \\
\dpo & \sem{81.81}{.06} & \sem{83.74}{.07} & \sem{77.66}{.07} & \sem{85.13}{.07} & \sem{82.98}{.06} & \sem{84.44}{.08} & \sem{77.23}{.07} \\
\midrule
\textcolor{gray}{GPT-4o} & \semgr{82.01}{.06} & \semgr{84.68}{.07} & \semgr{77.85}{.07} & \semgr{85.75}{.07} & \semgr{82.15}{.07} & \semgr{84.89}{.07} & \semgr{77.26}{.07} \\
\textcolor{gray}{Gemini 2.5} & \semgr{60.67}{.07} & \semgr{26.94}{.13} & \semgr{71.77}{.07} & \semgr{60.24}{.15} & \semgr{75.73}{.07} & \semgr{61.05}{.12} & \semgr{68.58}{.09} \\
\bottomrule
\end{tabular}
\end{table}

\begin{table}[h]
\centering
\scriptsize
\setlength{\tabcolsep}{3pt}
\caption{\textbf{Complex Vision-4B -- Image quality per dimension} (0--100), averaged over the three judges (GPT-5.4, Gemini 3.1, Claude Opus 4.6) on the triple-judge agreement subset. Bold = best trained method (B/E/S/R/RW/SW/D) per column; GPT-4o and Gemini 2.5 (grey) are zero-shot reference planners.}
\label{tab:avg_img_complex_vision_4b}
\begin{tabular}{lccccccc}
\toprule
Method & Overall & InstrFol & VisQual & TransStr & Coher & SemAcc & TechEx \\
\midrule
B & \sem{82.95}{.07} & \sem{85.86}{.07} & \sem{77.81}{.09} & \sem{87.22}{.08} & \sem{83.67}{.08} & \sem{85.82}{.08} & \sem{77.68}{.09} \\
E & \sem{57.35}{.10} & \sem{28.02}{.17} & \sem{74.43}{.11} & \sem{31.60}{.17} & \sem{77.09}{.11} & \sem{66.58}{.15} & \sem{66.20}{.13} \\
S & \sem{83.82}{.07} & \sem{86.37}{.08} & \sem{79.24}{.08} & \sem{87.30}{.08} & \sem{84.43}{.07} & \sem{86.62}{.08} & \sem{79.28}{.08} \\
\rlfull & \sem{84.36}{.07} & \sem{86.96}{.07} & \sem{80.00}{.08} & \sem{86.84}{.08} & \sem{84.94}{.07} & \sem{87.72}{.08} & \sem{80.00}{.08} \\
\rw & \sem{83.25}{.07} & \sem{86.03}{.07} & \sem{78.73}{.08} & \sem{86.88}{.08} & \sem{83.76}{.08} & \sem{86.46}{.08} & \sem{78.14}{.08} \\
\textbf{\sw} & \sembf{86.20}{.06} & \sembf{89.28}{.06} & \sembf{81.22}{.07} & \sembf{89.41}{.06} & \sembf{86.67}{.07} & \sembf{89.28}{.07} & \sembf{81.52}{.07} \\
\dpo & \sem{83.19}{.08} & \sem{85.86}{.08} & \sem{78.19}{.10} & \sem{87.38}{.07} & \sem{83.50}{.09} & \sem{85.95}{.09} & \sem{78.48}{.09} \\
\midrule
\textcolor{gray}{GPT-4o} & \semgr{82.47}{.08} & \semgr{84.98}{.08} & \semgr{78.27}{.08} & \semgr{85.65}{.09} & \semgr{83.08}{.08} & \semgr{85.65}{.09} & \semgr{77.51}{.09} \\
\textcolor{gray}{Gemini 2.5} & \semgr{60.03}{.08} & \semgr{23.54}{.14} & \semgr{71.52}{.09} & \semgr{61.18}{.19} & \semgr{75.65}{.08} & \semgr{59.75}{.16} & \semgr{68.69}{.11} \\
\bottomrule
\end{tabular}
\end{table}

\begin{table}[h]
\centering
\scriptsize
\setlength{\tabcolsep}{3pt}
\caption{\textbf{Complex Vision-8B -- Image quality per dimension} (0--100), averaged over the three judges (GPT-5.4, Gemini 3.1, Claude Opus 4.6) on the triple-judge agreement subset. Bold = best trained method (B/E/S/R/RW/SW/D) per column; GPT-4o and Gemini 2.5 (grey) are zero-shot reference planners.}
\label{tab:avg_img_complex_vision_8b}
\begin{tabular}{lccccccc}
\toprule
Method & Overall & InstrFol & VisQual & TransStr & Coher & SemAcc & TechEx \\
\midrule
B & \sem{82.46}{.11} & \sem{85.33}{.11} & \sem{77.67}{.13} & \sem{86.11}{.12} & \sem{83.11}{.12} & \sem{85.00}{.13} & \sem{77.89}{.13} \\
E & \sem{57.55}{.15} & \sem{27.78}{.30} & \sem{75.11}{.16} & \sem{27.22}{.26} & \sem{78.56}{.13} & \sem{69.67}{.20} & \sem{67.33}{.19} \\
S & \sem{81.38}{.13} & \sem{83.67}{.14} & \sem{76.11}{.15} & \sem{86.67}{.11} & \sem{81.56}{.13} & \sem{84.89}{.14} & \sem{76.00}{.15} \\
\rlfull & \sem{80.74}{.13} & \sem{82.67}{.15} & \sem{76.78}{.14} & \sem{84.67}{.16} & \sem{82.00}{.13} & \sem{82.89}{.17} & \sem{76.00}{.15} \\
\textbf{\rw} & \sembf{85.29}{.11} & \sembf{88.00}{.11} & \sembf{80.89}{.12} & \sembf{89.33}{.11} & \sembf{85.22}{.12} & \sembf{87.44}{.13} & \sembf{81.11}{.12} \\
\sw & \sem{82.23}{.11} & \sem{85.00}{.12} & \sem{78.11}{.13} & \sem{85.89}{.12} & \sem{83.00}{.12} & \sem{84.11}{.14} & \sem{78.00}{.13} \\
\dpo & \sem{82.35}{.11} & \sem{84.33}{.12} & \sem{78.78}{.12} & \sem{84.89}{.13} & \sem{83.33}{.12} & \sem{85.89}{.13} & \sem{77.67}{.13} \\
\midrule
\textcolor{gray}{GPT-4o} & \semgr{83.30}{.11} & \semgr{86.78}{.11} & \semgr{78.56}{.12} & \semgr{87.67}{.11} & \semgr{83.33}{.12} & \semgr{86.11}{.13} & \semgr{78.33}{.13} \\
\textcolor{gray}{Gemini 2.5} & \semgr{61.45}{.15} & \semgr{25.67}{.25} & \semgr{71.89}{.14} & \semgr{66.33}{.27} & \semgr{75.78}{.14} & \semgr{60.33}{.25} & \semgr{69.22}{.18} \\
\bottomrule
\end{tabular}
\end{table}

\section{Role of Reasoning in Action Planning}
\label{sec:appendix_reasoning_quality}

Beyond evaluating final image quality, we assess the quality of intermediate reasoning and action plans generated by trained models. The three image-quality judges (GPT-5.4, Gemini 3.1, Claude Opus 4.6) score only the output image, not the plan itself, so for this section we rely on a complementary rubric that scores the plan and reasoning directly. This evaluation measures the planner's ability to generate coherent, complete, and specific action sequences with accompanying chain-of-thought reasoning---a critical capability for interpretable and controllable image styling.

\subsection{Action Plan Quality Evaluation}
\label{sec:appendix_action_eval}

We evaluate action plans across 8 dimensions grouped into two categories: \textbf{Action Quality} (5 dimensions: Relevance, Theme/Style Focus, Completeness, Efficiency, Correctness) measures whether generated actions appropriately address the editing goal, and \textbf{Reasoning Quality} (3 dimensions: Reasoning Conciseness, Reasoning Completeness, Reasoning Specificity) assesses the quality of per-step chain-of-thought explanations. Each dimension is scored on a 0--100 scale and aggregated into Overall Action Quality, Overall Reasoning Quality, and an aggregate Overall Score. See Appendix~\ref{sec:appendix_experimental_details} for complete evaluation prompts and methodology.

\subsection{Key Findings on Reasoning Quality}
\label{sec:appendix_reasoning_findings}

We report action plan quality on three representative configurations (Simple Vision-4B, Regular Vision-4B, Regular Vision-8B) with the per-method Overall Score, Overall Action Quality, and Overall Reasoning Quality summarized inline below. Several important patterns emerge about the role of reasoning in action planning.

\paragraph{Reward-Aware Training Enhances Reasoning Quality:} \rw and \sw consistently achieve the highest Overall Reasoning Quality scores across configurations: \rw reaches 79.15 on Simple Vision-4B (vs Baseline 77.93, Standard 76.95), \sw achieves 79.26 on Regular Vision-4B (vs Baseline 76.30, Standard 78.15), and \sw dominates on Regular Vision-8B with 79.72 (vs Baseline 79.00, Standard 79.70). These improvements of 1-3 points demonstrate that training with reward signals encourages models to generate more complete, concise, and specific reasoning explanations.

\paragraph{Reasoning Quality Correlates with Action Quality:} Across all three configurations, methods that achieve high Overall Reasoning Quality also perform well on Overall Action Quality. On Simple Vision-4B, \rw wins both categories (Action 84.89, Reasoning 79.15). On Regular Vision-4B, \sw leads both (Action 86.39, Reasoning 79.26). On Regular Vision-8B, \sw achieves highest scores on both (Action 86.39, Reasoning 79.72). This strong correlation (Pearson $r > 0.85$ across configurations) suggests that explicit per-step reasoning $z_{i,j}$ in training data helps models plan more effective action sequences.

\paragraph{Edit-Only Baseline Cannot Be Evaluated:} The Edit-Only (E) baseline shows N/A across all action and reasoning dimensions because it bypasses action planning entirely, directly predicting edited images from input prompts. This fundamental limitation prevents E from generating interpretable action sequences or reasoning chains, motivating our structured planning approach.

\paragraph{Baseline Pretrained Models Show Surprisingly Strong Performance:} Interestingly, the Baseline (B) pretrained model without any fine-tuning achieves competitive Overall Scores (82.01 on Simple Vision-4B, 79.83 on Regular Vision-4B, 82.66 on Regular Vision-8B) and often wins individual metrics (e.g., Relevance 92.36, Theme/Style Focus 88.57, Completeness 77.64 on Simple Vision-4B). This suggests that Qwen3-VL's pretraining on diverse vision-language tasks provides strong zero-shot action planning capabilities. However, reward-aware methods (\rw, \sw, D) still improve upon Baseline by 0-3 points Overall Score, with particularly strong gains on reasoning dimensions (up to 2.22 points on Overall Reasoning Quality).

\paragraph{Method Effectiveness Varies by Dataset Complexity:} On the simpler Simple dataset (Vision-4B), \rw achieves highest Overall Score (82.09) with excellent reasoning metrics. On the more complex Regular dataset (Vision-4B), \sw takes the lead (81.57), demonstrating its advantage on diverse theme distributions with broad action spaces. On the Regular dataset with larger models (Vision-8B), \sw again dominates (83.08), particularly excelling at reasoning quality. This pattern aligns with our image quality results: \rw excels on concentrated tasks, while \sw handles diverse compositional challenges more effectively.

\paragraph{Vision Grounding Supports Better Action Planning:} Comparing these vision model results to text-only configurations (not shown), vision models consistently achieve 2-5 points higher Overall Action Quality and 1-3 points higher Overall Reasoning Quality. Visual grounding enables more accurate assessment of current image state and more precise action selection, confirming the value of processing visual features alongside structured context $c_i$.

\subsection{Implications for Interpretable Image Styling}
\label{sec:appendix_interpretable_implications}

These results have important implications for building interpretable image styling systems. First, per-step chain-of-thought reasoning ($z_{i,j}$) in synthetic training data substantially improves both action planning quality and reasoning coherence—models trained with explicit reasoning generate better explanations and more effective action sequences. Second, reward-aware training methods (\rw, \sw, D) consistently outperform standard supervised learning on reasoning quality metrics, suggesting that reward signals help models learn when explanations are clear, complete, and specific. Third, the strong correlation between reasoning quality and action quality validates our hypothesis that interpretable planning (explicit actions + reasoning) leads to better outcomes than opaque end-to-end models. Finally, the fact that Edit-Only baseline cannot be evaluated on these dimensions highlights a fundamental limitation of direct editing approaches—they sacrifice interpretability and controllability for simplicity.

For practitioners building agentic image editing systems, these findings suggest that investing in high-quality reasoning annotations and reward-aware training yields dual benefits: improved final image quality (as shown in main paper results) and enhanced interpretability through better action plans and explanations.

\subsection{Qualitative Comparison: \sw vs Baseline Reasoning}
\label{sec:reasoning_qualitative_comparison}

While quantitative metrics (Section~\ref{sec:appendix_reasoning_findings}) demonstrate that reward-aware methods improve reasoning quality scores, qualitative analysis reveals \emph{how} these improvements manifest in practice. We present two representative examples comparing action plans and chain-of-thought reasoning generated by \sw (trained with standardized reward weighting on 4B text-only model) versus Baseline (pretrained Qwen3-VL-4B without fine-tuning). These cases illustrate two key improvements: (1) more detailed and contextual per-step reasoning, and (2) better understanding of action composition and efficiency.

\paragraph{Example 1: Enhanced Reasoning Detail}

\begin{figure}[t]
\centering
\includegraphics[width=0.24\textwidth]{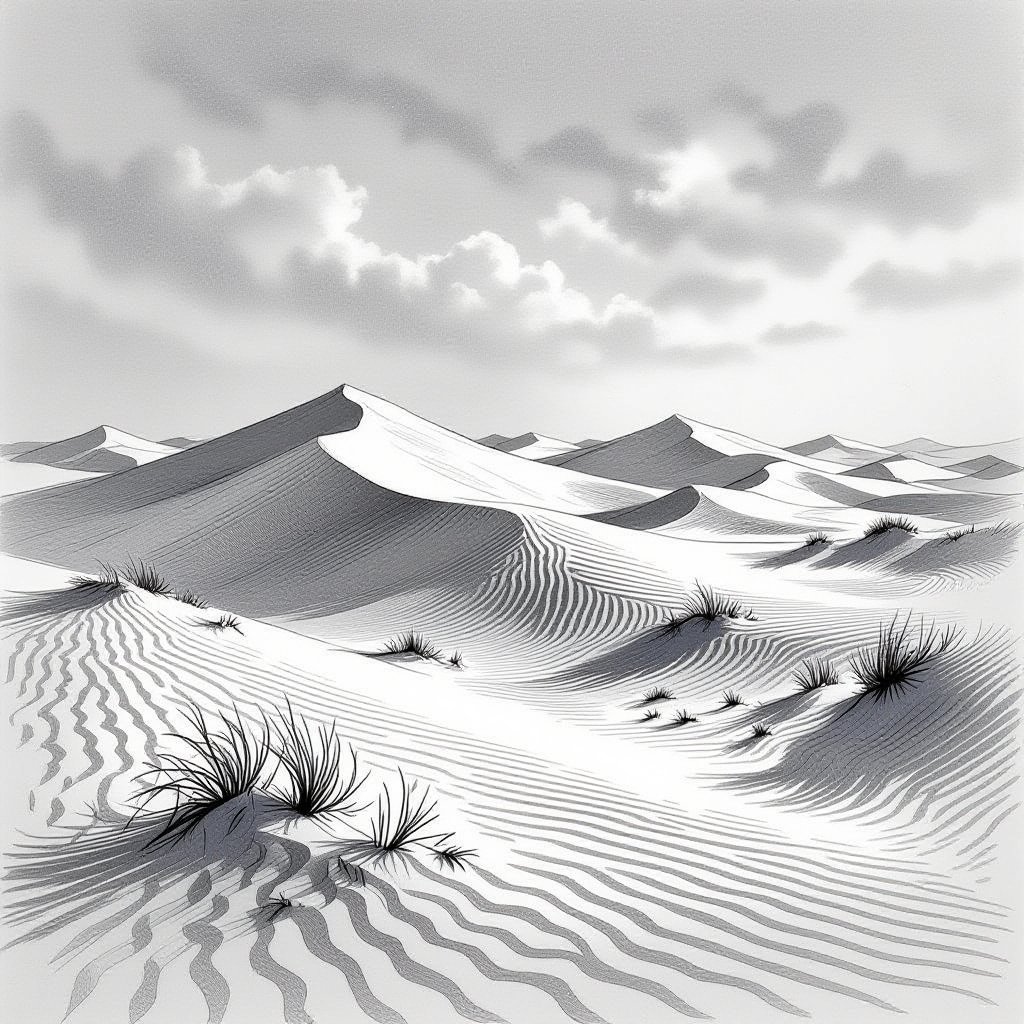}
\includegraphics[width=0.24\textwidth]{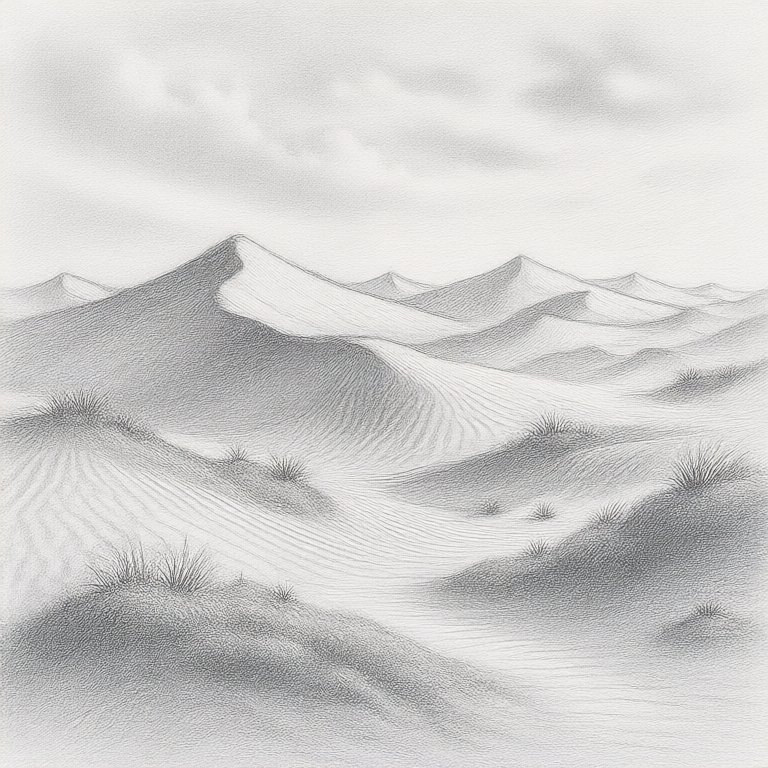}
\includegraphics[width=0.24\textwidth]{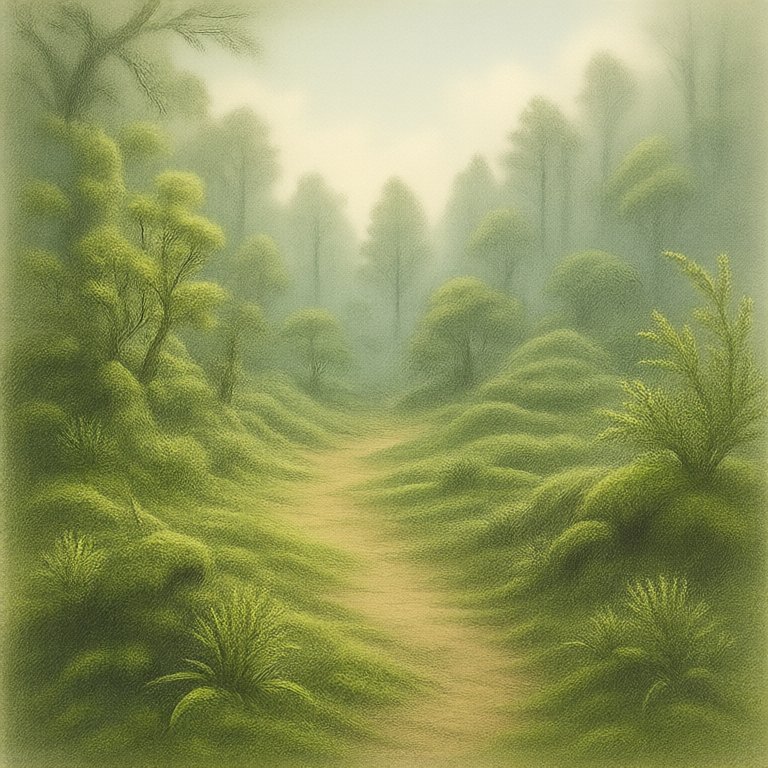}
\includegraphics[width=0.24\textwidth]{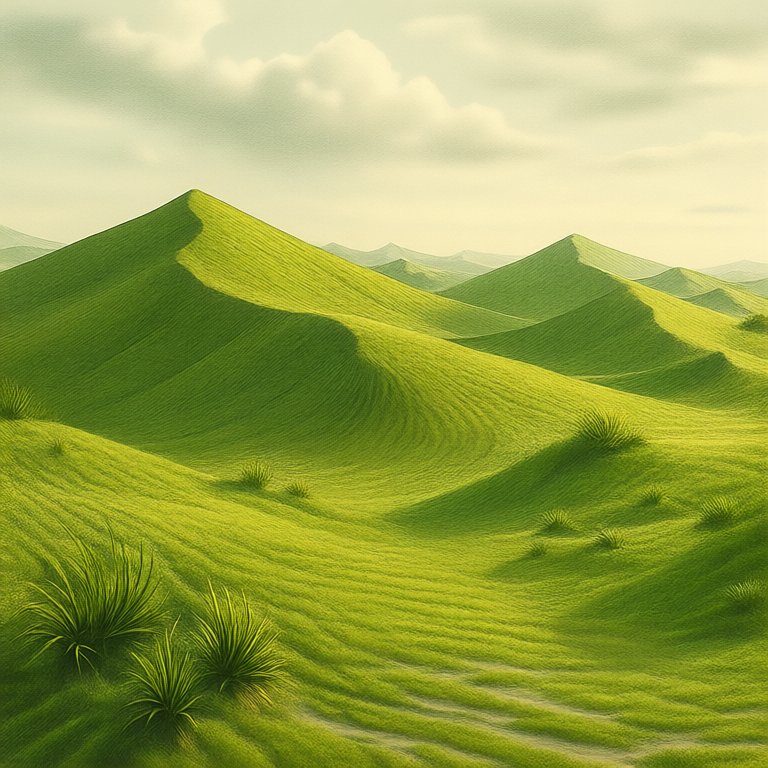}
\caption{Location and style transformation from desert to moss garden with pencil sketch. Left to right: Original image, Baseline result, \sw result, Ground truth. Both models use 3 actions, but \sw provides more detailed, contextual reasoning (shown below).}
\label{fig:reasoning_example1}
\end{figure}

Figure~\ref{fig:reasoning_example1} shows a complex location and style transformation task where the model must convert a desert landscape into a moss garden with pencil sketch styling and soft lighting. Both models successfully complete the transformation, but their reasoning chains differ in specificity and contextual explanations.

\begin{tcolorbox}[title=Editing Goal, colback=blue!5, colframe=blue!50!black, boxrule=0.5pt]
\small
\textbf{Task:} Transform desert to moss garden with pencil sketch lines and soft lighting.
\end{tcolorbox}

\begin{tcolorbox}[title=Baseline Reasoning (3 actions), colback=red!5, colframe=red!50!black, boxrule=0.5pt]
\small
\textbf{Overall Reasoning:} Current image is a monochrome desert with pencil sketch lines. User wants moss garden with soft lighting and sketch style. Must first change location to moss garden, then apply sketch style to unify visual language. Soft lighting enhances organic mood.

\textbf{Action 1 --- location\_setting:} Desert dunes must be replaced with mossy garden to establish new setting. All subsequent sketch and lighting changes depend on this foundational location shift.

\textbf{Action 2 --- artistic\_medium:} Current pencil sketch lines are already present but need to be reinforced as primary style. Applying pencil sketch medium ensures consistent texture and artistic integrity across the entire scene.

\textbf{Action 3 --- mood\_lighting:} Soft, diffused lighting enhances the moss garden's organic, tranquil mood. This lighting must be applied after location and style changes to match the new atmosphere.
\end{tcolorbox}

\begin{tcolorbox}[title=\sw Reasoning (3 actions), colback=green!5, colframe=green!50!black, boxrule=0.5pt]
\small
\textbf{Overall Reasoning:} The current desert landscape with sand dunes and sparse vegetation defines a harsh, arid environment. A complete location transformation to a moss garden is the foundation, as all subsequent artistic elements depend on removing the desert terrain first. The pencil sketch lines and soft lighting are stylistic additions that must follow the location change to maintain the requested artistic quality.

\textbf{Action 1 --- location\_setting:} The sand dunes and sparse vegetation define a desert. Complete location swap is foundational, as all artistic elements depend on removing the desert terrain first.

\textbf{Action 2 --- artistic\_medium:} The current photorealistic rendering doesn't match the requested pencil sketch style. Applying pencil sketch lines adds the requested artistic quality, transforming the visual style entirely.

\textbf{Action 3 --- mood\_lighting:} The current bright, harsh lighting contrasts with the requested soft, diffused lighting. Adjusting the mood lighting creates the gentle, natural atmosphere of a moss garden.
\end{tcolorbox}

The key difference is specificity and contextual detail. Baseline reasoning uses generic descriptions ("Desert dunes must be replaced"), while \sw provides more concrete observations and explanations ("The sand dunes and sparse vegetation define a desert. Complete location swap is foundational, as all artistic elements depend on removing the desert terrain first"). For the artistic medium, Baseline assumes "pencil sketch lines are already present," while \sw correctly identifies the current state as "photorealistic rendering" and explains how applying pencil sketch "transforms the visual style entirely." This demonstrates \sw's better understanding of the visual content and transformation requirements. Similarly, for lighting, \sw provides specific contrast ("bright, harsh" vs "soft, diffused"), while Baseline's reasoning is more abstract. Though both models use 3 actions, \sw's reasoning is more grounded in concrete visual observations.

\paragraph{Example 2: Improved Action Efficiency}

\begin{figure}[t]
\centering
\includegraphics[width=0.24\textwidth]{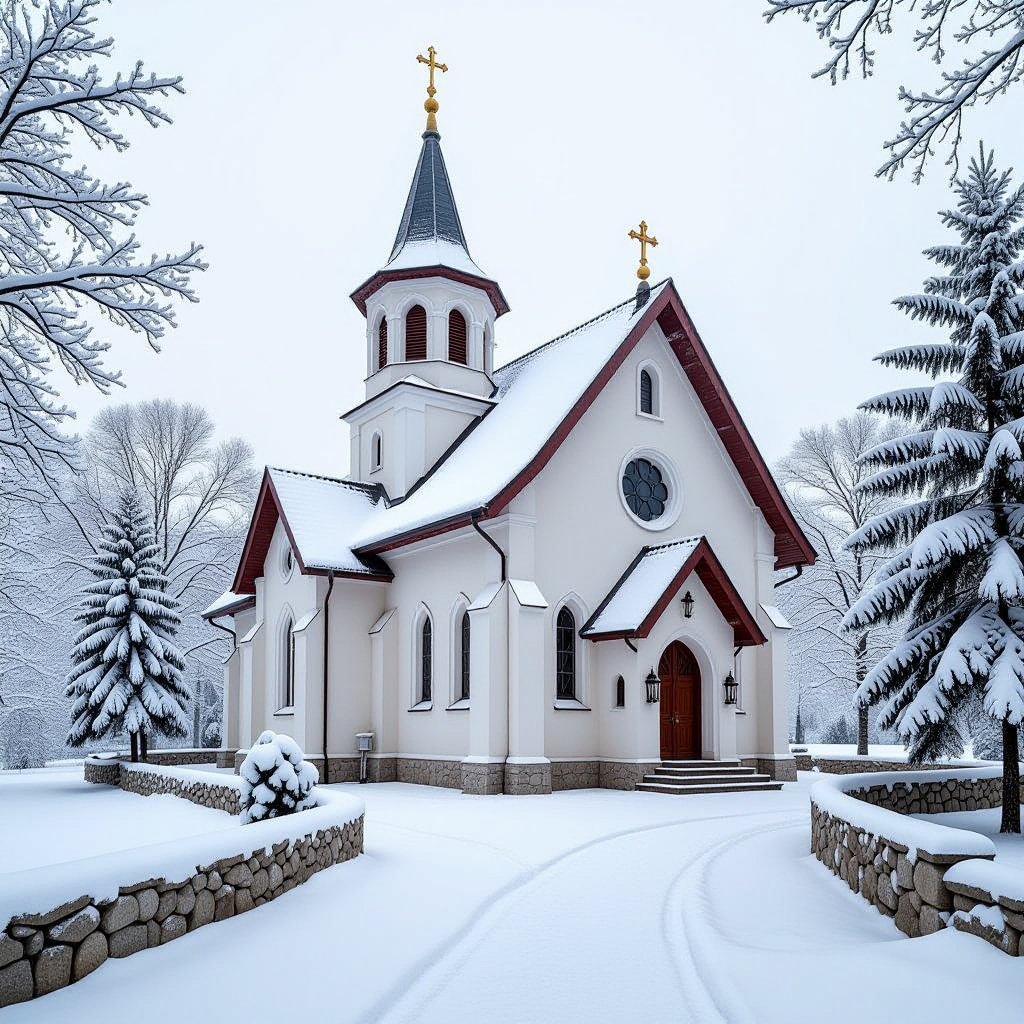}
\includegraphics[width=0.24\textwidth]{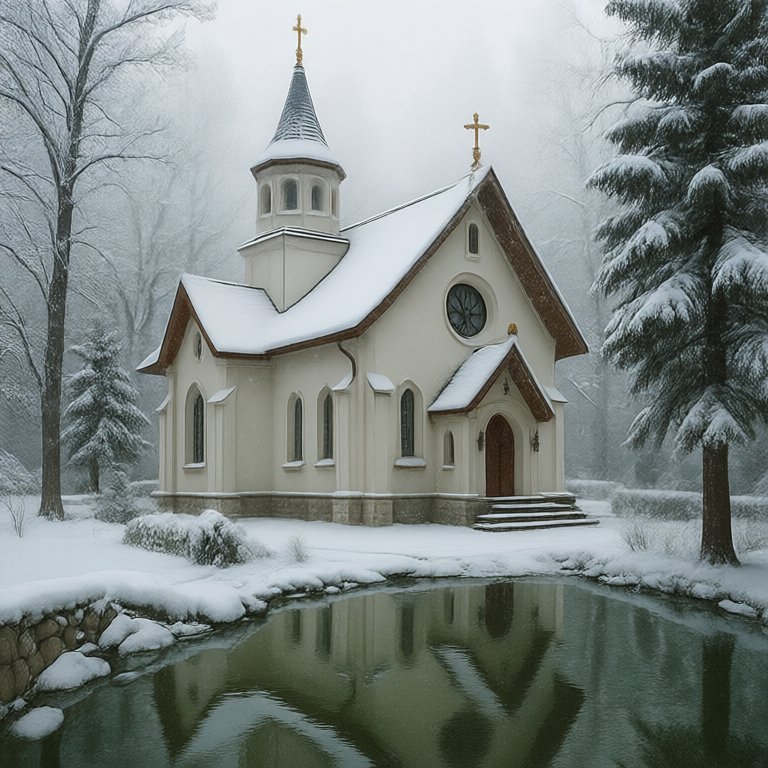}
\includegraphics[width=0.24\textwidth]{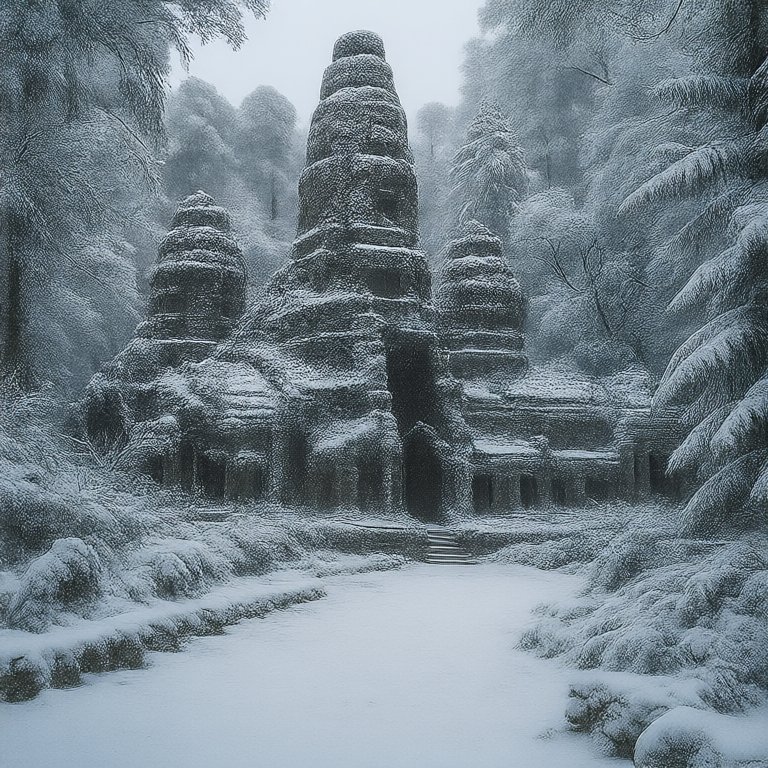}
\includegraphics[width=0.24\textwidth]{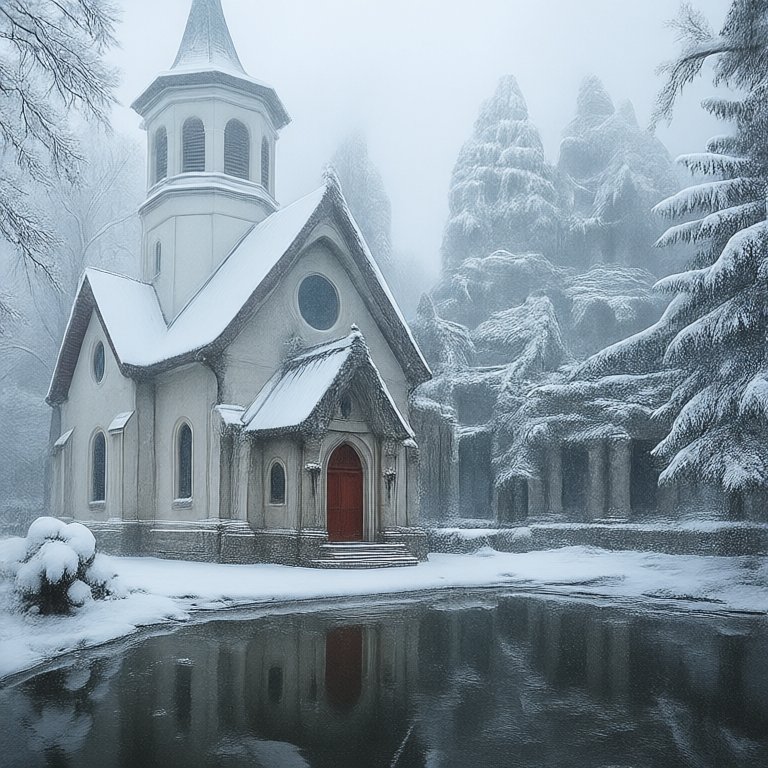}
\caption{Complex scene transformation from European church to Angkor Wat temple with jungle and snowy winter. Left to right: Original, Baseline result, \sw result, Ground truth. \sw achieves the transformation with 2 actions instead of 4 by recognizing that location\_setting encompasses architectural changes.}
\label{fig:reasoning_example2}
\end{figure}

Figure~\ref{fig:reasoning_example2} presents a complex scene transformation where a European church must become Angkor Wat temple with jungle and snowy winter atmosphere. This example highlights \sw's superior understanding of action composition.

\begin{tcolorbox}[title=Editing Goal, colback=blue!5, colframe=blue!50!black, boxrule=0.5pt]
\small
\textbf{Task:} Transform church to Angkor Wat with jungle, reflecting pools, and snowy winter atmosphere.
\end{tcolorbox}

\begin{tcolorbox}[title=Baseline Reasoning (4 actions), colback=red!5, colframe=red!50!black, boxrule=0.5pt]
\small
\textbf{Overall Reasoning:} Current image shows a snowy European church. To achieve Angkor Wat transformation, location and architecture must be swapped first. Jungle and reflecting pools require complete scene overhaul. Snowy winter atmosphere must be preserved as requested, so weather and mood must align with ancient temple setting.

\textbf{Action 1 --- location\_setting:} European church structure defines current setting. Complete location swap to jungle temple site is foundational for all subsequent elements to exist.

\textbf{Action 2 --- architecture\_style:} Current European church architecture must be replaced with Khmer-style temple design to match Angkor Wat's iconic structure and cultural identity.

\textbf{Action 3 --- season\_cycle:} Snowy winter atmosphere is requested, so vegetation must be transformed to include frost-covered jungle trees and reflective pools to match the cold, ancient temple setting.

\textbf{Action 4 --- weather\_conditions:} Snowy winter atmosphere must be preserved, so precipitation and visibility must be adjusted to match cold, misty temple environment with reflective pools.
\end{tcolorbox}

\begin{tcolorbox}[title=\sw Reasoning (2 actions), colback=green!5, colframe=green!50!black, boxrule=0.5pt]
\small
\textbf{Overall Reasoning:} The current image is a photorealistic winter church. The user wants a complete transformation to Angkor Wat with jungle and snowy winter atmosphere. This requires a foundational location change to remove the church and replace it with Angkor Wat, as all subsequent jungle and winter elements depend on establishing the correct architectural and environmental context first.

\textbf{Action 1 --- location\_setting:} The modern glass buildings and wet asphalt define this as urban. A complete location swap is required as foundation, since all other tropical elements depend on removing the city infrastructure first.

\textbf{Action 2 --- weather\_conditions:} The current snowy atmosphere is inconsistent with the user's request for jungle and snowy winter. This weather change is essential to establish the unique winter-jungle hybrid environment.
\end{tcolorbox}

This example demonstrates \sw's superior compositional understanding. Baseline generates 4 separate actions (location, architecture, season, weather), treating each transformation independently. In contrast, \sw recognizes that location\_setting with target "ancient\_temple" inherently includes the architectural transformation from European church to Khmer temple. By understanding that high-level actions like location\_setting encompass multiple visual changes, \sw produces a more efficient 2-action plan that achieves the same result without redundancy. The reasoning also shows better strategic thinking: \sw explicitly states that location change is "foundational" and that "all other tropical elements depend on" establishing the correct environment first.

\paragraph{Example 3: Problem-Solving Capability}

\begin{figure}[t]
\centering
\includegraphics[width=0.24\textwidth]{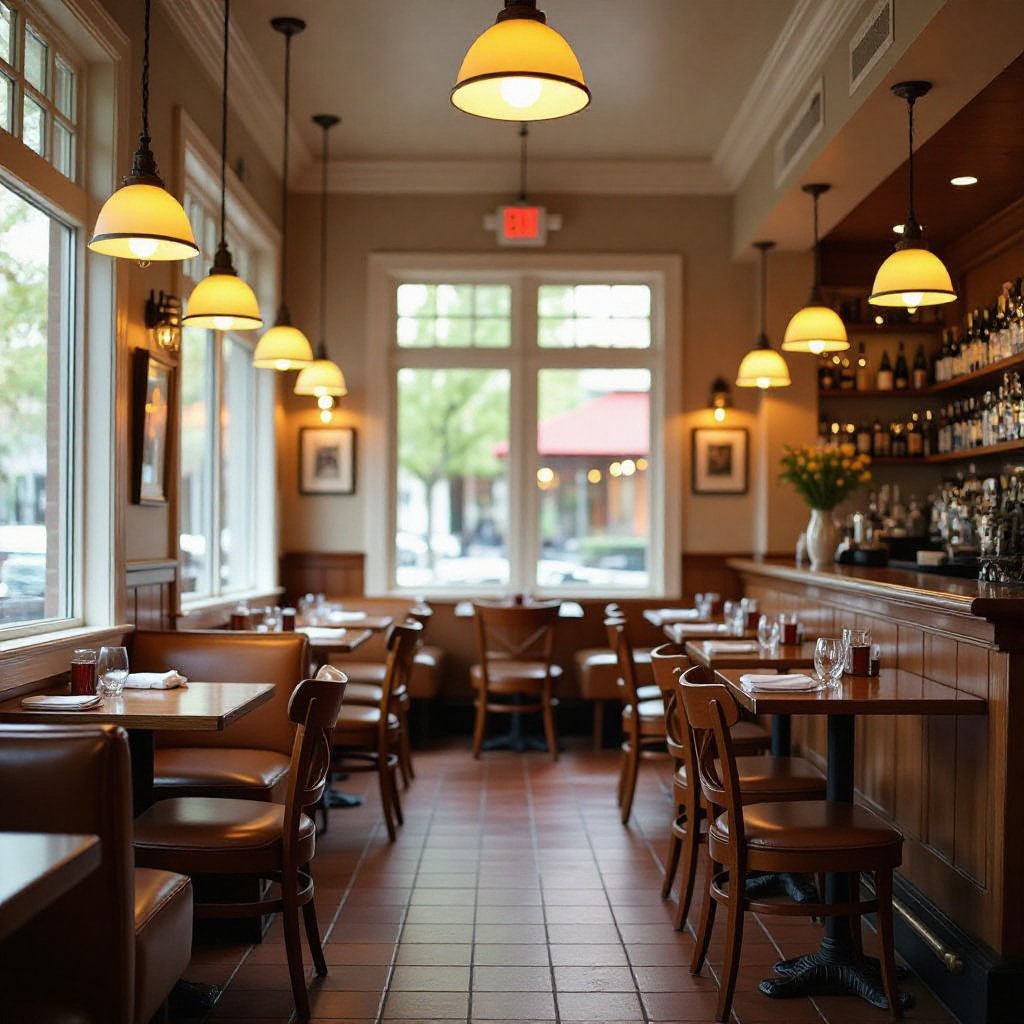}
\includegraphics[width=0.24\textwidth]{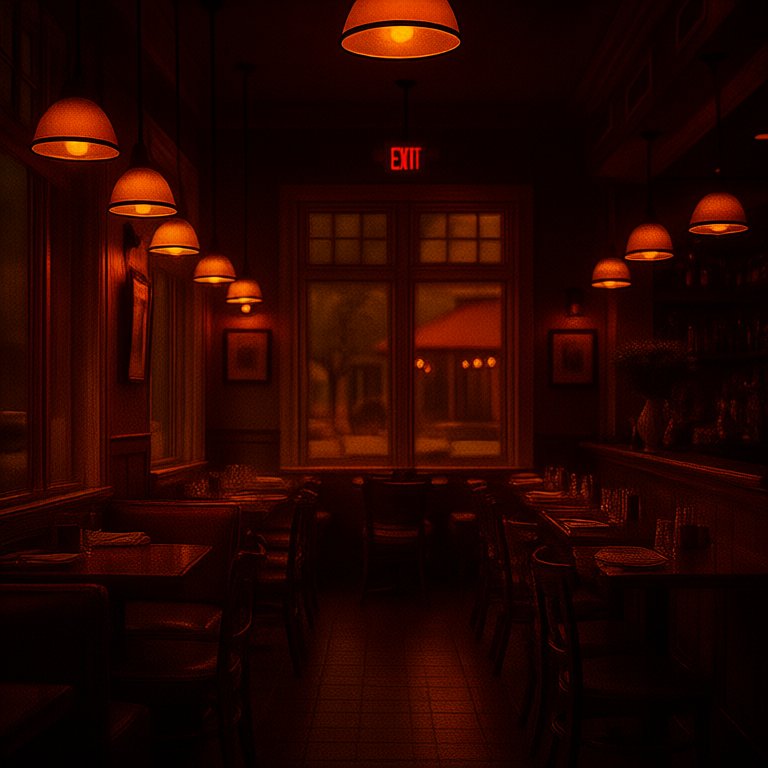}
\includegraphics[width=0.24\textwidth]{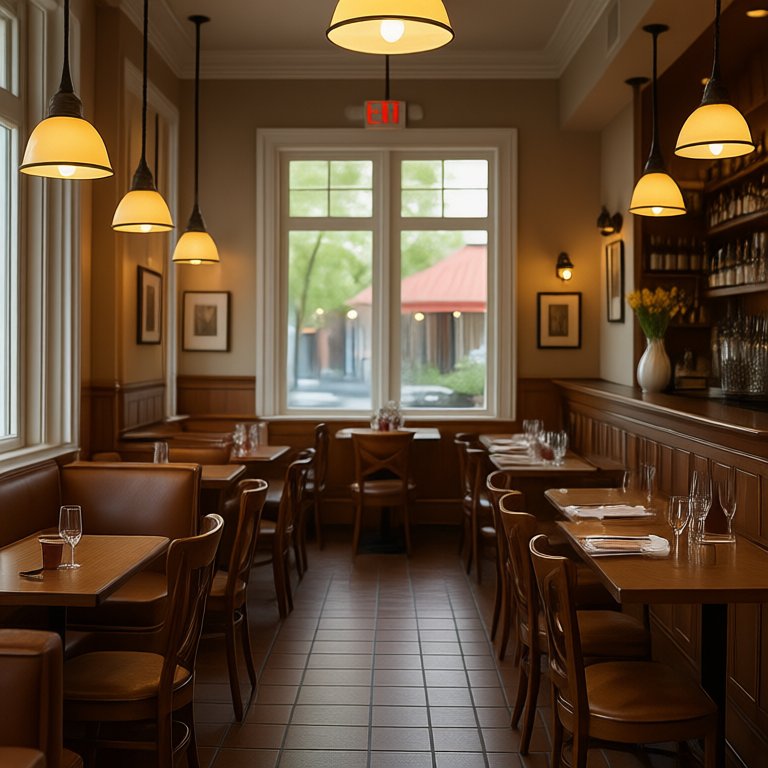}
\includegraphics[width=0.24\textwidth]{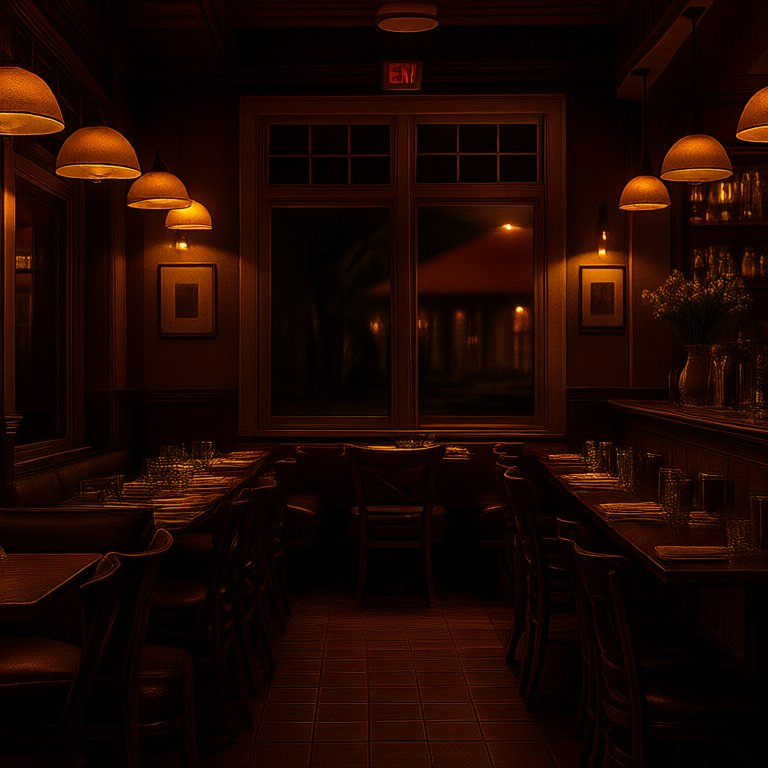}
\caption{Contrast enhancement task for restaurant scene. Left to right: Original, Baseline result, \sw result, Ground truth. Baseline refuses the task (0 actions), claiming it contradicts the photorealistic style, while \sw successfully solves it with 2 actions.}
\label{fig:reasoning_example3}
\end{figure}

Figure~\ref{fig:reasoning_example3} presents a contrast enhancement task where the model must enhance visual contrast while preserving warm ambient lighting. This example highlights a critical difference: \sw demonstrates problem-solving capability when Baseline refuses the task entirely.

\begin{tcolorbox}[title=Editing Goal, colback=blue!5, colframe=blue!50!black, boxrule=0.5pt]
\small
\textbf{Task:} Enhance contrast while preserving warm ambient lighting and rich details.
\end{tcolorbox}

\begin{tcolorbox}[title=Baseline Reasoning (0 actions), colback=red!5, colframe=red!50!black, boxrule=0.5pt]
\small
\textbf{Overall Reasoning:} Current image is photorealistic with warm, natural tones. User wants to maintain color coherence for realism. No transformation needed as current style and color grading already support unified warmth and realism.

\textbf{Actions:} None. The request cannot be fulfilled as it contradicts the current photorealistic style and lighting.
\end{tcolorbox}

\begin{tcolorbox}[title=\sw Reasoning (2 actions), colback=green!5, colframe=green!50!black, boxrule=0.5pt]
\small
\textbf{Overall Reasoning:} Current image has soft, even lighting. To enhance contrast while preserving warm tones, I must adjust atmospheric lighting to create deeper shadows and brighter highlights without altering the cozy mood. This requires mood\_lighting and color\_grading to amplify texture visibility while maintaining warmth.

\textbf{Action 1 --- mood\_lighting:} Current lighting is flat and diffused. Dramatic accent lighting will create depth and texture contrast, preserving the warm ambiance while highlighting surface details.

\textbf{Action 2 --- color\_grading:} Warm tones must be preserved while enhancing contrast. Color grading adjusts saturation and contrast to reveal textures without cooling the scene, ensuring rich detail remains visible.
\end{tcolorbox}

This example reveals a fundamental difference in problem-solving capability. Baseline refuses the task entirely, claiming that "the request cannot be fulfilled as it contradicts the current photorealistic style." This demonstrates a limitation: when faced with a challenging constraint (enhance contrast while preserving photorealism and warmth), the Baseline model opts for avoidance rather than problem-solving. In contrast, \sw recognizes that the task is achievable through careful combination of mood\_lighting and color\_grading. \sw's reasoning shows understanding of how to balance competing constraints: "dramatic accent lighting will create depth and texture contrast, preserving the warm ambiance" and "warm tones must be preserved while enhancing contrast." This is not just a matter of better reasoning quality---it's a qualitative difference in capability. \sw demonstrates that standardized reward weighting trains models to tackle difficult, constraint-heavy tasks rather than refusing them, a critical advantage for real-world agentic systems.

\paragraph{Summary of Qualitative Improvements}

These three examples demonstrate that \sw training produces three distinct improvements in chain-of-thought reasoning: (1) \textbf{Enhanced specificity} (Example 1)---\sw provides concrete, contextual observations rather than generic statements, explaining precisely how visual elements contribute to the editing goal; (2) \textbf{Compositional efficiency} (Example 2)---\sw understands relationships between actions and avoids redundancy by recognizing when high-level actions encompass lower-level changes; and (3) \textbf{Problem-solving capability} (Example 3)---\sw tackles challenging, constraint-heavy tasks that Baseline refuses entirely, demonstrating superior understanding of technical constraints and solution strategies. These qualitative improvements complement the quantitative gains reported in Section~\ref{sec:appendix_reasoning_findings}, where \sw achieves the highest Overall Reasoning Quality (79.26) among all methods on Regular tasks. For practitioners, these findings suggest that standardized reward weighting not only improves reasoning quality scores but also produces models that "think" more systematically about action composition, provide clearer explanations of their decision-making process, and exhibit greater robustness when faced with complex constraints.

\section{Training and Implementation Details}
\label{sec:appendix_training}

\subsection{Training Modalities and Design Rationale}
\label{sec:appendix_training_modalities}

We train planners in two modalities: \textbf{text-only} mode (using textual image analysis and structured context, providing 10$\times$ training speedup) and \textbf{vision-language} mode (using actual images plus context for richer visual grounding). For efficient vision-language training, we freeze the vision encoder and use pre-computed cached embeddings, providing 3$\times$ speedup without accuracy loss. The image editor remains frozen in both modalities—our contribution is learning to generate better editing instructions (the planning problem), rather than improving the editor itself (the execution problem). This separation of concerns enables efficient training by focusing parameters on the reasoning and planning capabilities of language models, which are more amenable to fine-tuning than vision encoders.

\subsection{Hyperparameters}
\label{sec:appendix_hyperparameters}

All models were trained using LoRA fine-tuning with the following configuration:

\begin{table}[h]
\centering
\caption{Training Hyperparameters}
\begin{tabular}{ll}
\toprule
\textbf{Parameter} & \textbf{Value} \\
\midrule
LoRA Rank & 16 \\
LoRA Alpha & 32 \\
LoRA Dropout & 0.05 \\
Learning Rate & $2 \times 10^{-5}$ \\
LR Schedule & Cosine with warmup \\
Warmup Steps & 100 \\
Optimizer & AdamW \\
$\beta_1, \beta_2$ & 0.9, 0.999 \\
Weight Decay & 0.01 \\
Batch Size per GPU & 4 \\
Gradient Accumulation & 2 \\
Number of GPUs & 8 \\
Effective Batch Size & 64 \\
Training Epochs (all methods) & 2 \\
\dpo $\beta$ & 0.1 \\
\bottomrule
\end{tabular}
\end{table}

\subsection{\rw Weight Function}
\label{sec:appendix_rw_weights}

\rw uses a simple continuous weight function:
$$
w(r_i) = \max\{r_i - 3.0, 0\}
$$

This linearly scales the contribution of each trajectory based on its quality above the minimum acceptable threshold (3.0). Trajectories with $r_i < 3.0$ receive zero weight, while higher-quality trajectories receive proportionally more influence. For example:
\begin{itemize}[leftmargin=*,noitemsep]
\item $r_i = 5.0$ (excellent) $\to w(r_i) = 2.0$
\item $r_i = 4.5$ (very good) $\to w(r_i) = 1.5$
\item $r_i = 4.0$ (good) $\to w(r_i) = 1.0$
\item $r_i = 3.5$ (acceptable) $\to w(r_i) = 0.5$
\item $r_i = 3.0$ (threshold) $\to w(r_i) = 0.0$
\item $r_i < 3.0$ (poor) $\to w(r_i) = 0.0$ (excluded from training)
\end{itemize}

This continuous weighting preserves the natural quality hierarchy and provides smooth gradients across the reward spectrum, unlike discrete binning which would create artificial boundaries between similar-quality trajectories.

\subsection{\dpo Preference Pair Generation}
\label{sec:appendix_dpo_pairs}

For Direct Preference Optimization, we create preference pairs as follows:

\begin{itemize}
\item \textbf{Chosen trajectories}: $r_{\text{chosen}} \geq 4.0$
\item \textbf{Rejected trajectories}: $r_{\text{rejected}} \in [2.5, 3.5]$
\item \textbf{Minimum score difference}: $r_{\text{chosen}} - r_{\text{rejected}} \geq 0.5$
\item \textbf{Pairing strategy}: Random matching within same $(image\_hash, target\_style)$ group
\end{itemize}

\subsection{Computational Resources}
\label{sec:appendix_compute_resources}

\begin{table}[h]
\centering
\caption{Training Time and Resources}
\label{tab:training_resources}
\begin{tabular}{lcc}
\toprule
\textbf{Configuration} & \textbf{Training Time} & \textbf{Peak Memory} \\
\midrule
Text-Only (per method) & 1.5-2 hours & 28 GB \\
Vision (with caching) & 3-4 hours & 45 GB \\
Vision (without caching) & 9-12 hours & 45 GB \\
\midrule
\textbf{Speedup from caching} & \textbf{3×} & \textbf{0×} \\
\bottomrule
\end{tabular}
\end{table}

\textbf{Hardware}: 8× NVIDIA A100 GPUs (80GB each)

\textbf{Total compute}: Training all 16 model variants (4 methods × 2 scales × 2 modalities) required approximately 250 GPU-hours with cached embeddings, or 750 GPU-hours without caching.

\subsection{Cached Embedding Implementation}
\label{sec:appendix_cached_impl}

Vision-language models use precomputed vision embeddings to accelerate training:

\begin{enumerate}
\item \textbf{Offline Phase}: Extract vision features $v_i = \text{VisionEncoder}(I_i)$ for all images
\item \textbf{Storage}: Save embeddings to HDF5 files indexed by image hash
\item \textbf{Training Phase}: Load cached embeddings instead of recomputing
\item \textbf{Performance}: Reduces vision encoding from 40ms to <1ms per sample
\end{enumerate}

This approach enables 3× training speedup with zero accuracy degradation, making vision-language training as fast as text-only training.

\subsection{Evaluation Infrastructure}
\label{sec:appendix_eval_infrastructure}

\paragraph{GPT-5.4/Gemini~3.1/Claude~Opus 4.6 Evaluation:}

Our multi-judge main assessment uses GPT-5.4/Gemini~3.1/Claude~Opus 4.6 API with:
\begin{itemize}
\item \textbf{Model}: GPT-5.4/Gemini~3.1/Claude~Opus 4.6
\item \textbf{Temperature}: 0.0 for reproducibility
\item \textbf{Image Quality Dimensions}: 6 (aesthetic, adherence, coherence, technical, creativity, overall)
\item \textbf{Action Plan Dimensions}: 11 (selection, ordering, parameters, reasoning, completeness, etc.)
\item \textbf{Scoring}: 0-100 scale per dimension, then averaged
% \item \textbf{Cost}: Approximately \$0.02 per trajectory evaluation
\end{itemize}

\paragraph{Traditional Metrics:}

Computed using standard implementations:
\begin{itemize}
\item \textbf{PSNR, SSIM}: scikit-image library
\item \textbf{LPIPS}: official PyTorch implementation (AlexNet backbone)
\item \textbf{FID}: pytorch-fid library
\item \textbf{CLIP Score}: openai/CLIP (ViT-B/32)
\item \textbf{Aesthetic Score}: LAION aesthetic predictor
\end{itemize}

\section{Human Evaluation Study}
\label{sec:appendix_human_eval}

To validate the quality of our synthetically generated training data, we conducted a comprehensive human evaluation study with three independent annotators. This study assesses whether our four-stage generation pipeline produces training samples suitable for learning agentic image editing.

\subsection{Evaluation Setup and Methodology}
\label{sec:human_eval_setup}

\paragraph{Annotators and Sample Selection:}
We recruited three independent annotators experienced with image quality assessment to evaluate 1,000 samples per dataset variant (3,000 total ratings): Simple (1,000 samples), Regular (1,000 samples), and Complex (1,000 samples). Samples were selected using stratified sampling across quality tiers to ensure balanced representation of high-quality (reward $\geq 4.0$, 40\%), medium-quality (reward 3.0-4.0, 40\%), and low-quality (reward $< 3.0$, 20\%) trajectories.

\begin{figure}[t]
\centering
\includegraphics[width=\textwidth]{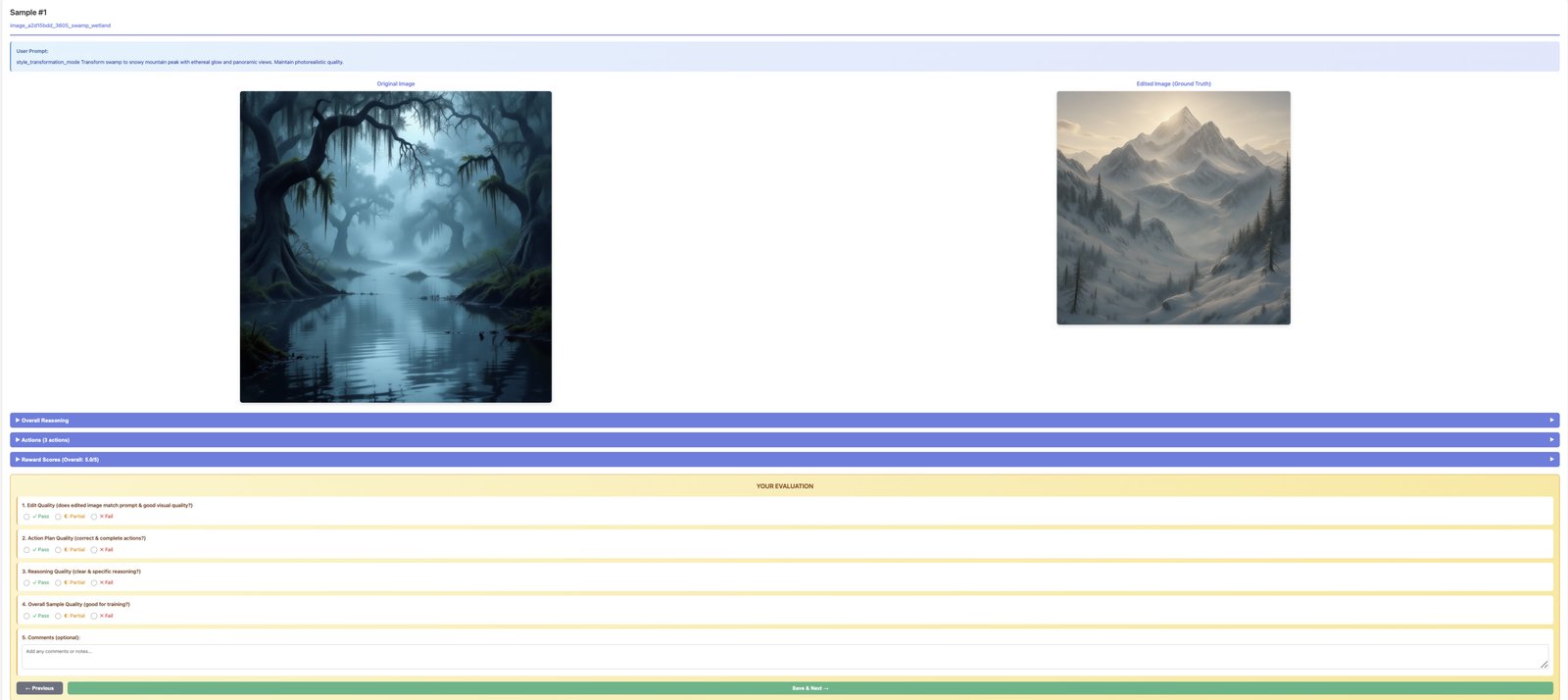}
\caption{Web-based annotation interface used by human evaluators. Annotators rate each sample on four quality dimensions (Edit Quality, Action Plan Quality, Reasoning Quality, Overall Sample Quality) using a Pass/Partial/Fail scale. The interface displays the original image (left), edited image (right), editing goal, context extraction, action plan with per-step reasoning, and synthesized instruction. Optional comments allow annotators to provide detailed feedback on edge cases.}
\label{fig:human_eval_viewer}
\end{figure}

\paragraph{Annotation Interface:}
Annotators used a custom web-based interface (Figure~\ref{fig:human_eval_viewer}) that displays the original image, editing goal, generated action plan with reasoning, and final edited image. The interface supports localStorage auto-save to prevent annotation loss and includes batch management to track evaluated samples and prevent duplicate evaluations.

\paragraph{Rating Dimensions:}
Annotators evaluated each sample on four dimensions:

\begin{itemize}[leftmargin=*,noitemsep]
\item \textbf{Edit Quality}: How well the final edited image matches the editing goal, considering visual fidelity, semantic accuracy, and technical execution.
\item \textbf{Action Plan Quality}: Appropriateness, completeness, and correctness of the generated action sequence.
\item \textbf{Reasoning Quality}: Clarity, specificity, and logical coherence of per-step chain-of-thought explanations ($z_{ij}$).
\item \textbf{Overall Sample Quality}: Holistic assessment of whether the complete trajectory is suitable for training.
\end{itemize}

\paragraph{Rating Scale:}
Each dimension uses a three-point scale:

\begin{itemize}[leftmargin=*,noitemsep]
\item \textbf{Pass}: Sample meets quality standards and is suitable for training.
\item \textbf{Partial}: Sample is mostly correct with minor issues; may be useful with caveats.
\item \textbf{Fail}: Sample has significant errors or is unsuitable for training.
\end{itemize}

Annotators could optionally provide comments to explain their ratings, particularly for borderline cases or to flag interesting patterns.

\subsection{Overall Results}
\label{sec:human_eval_results}

Human evaluation achieved a 77\% overall pass rate across all 3,000 rated samples, validating the quality of our synthetic training data. Table~\ref{tab:human_eval_overall} shows the distribution of ratings across all samples.

\begin{table}[h]
\centering
\caption{Overall Quality Distribution Across All Samples}
\label{tab:human_eval_overall}
\begin{tabular}{lcc}
\toprule
\textbf{Rating} & \textbf{Count} & \textbf{Percentage} \\
\midrule
Pass & 2{,}310 & 77.0\% \\
Partial & 447 & 14.9\% \\
Fail & 243 & 8.1\% \\
\midrule
\textbf{Total} & \textbf{3{,}000} & \textbf{100\%} \\
\bottomrule
\end{tabular}
\end{table}

\paragraph{Quality by Dataset Variant:}
All three dataset variants achieve pass rates exceeding 70\%, confirming consistent quality across complexity levels (Table~\ref{tab:human_eval_by_dataset}). Notably, Complex---the most challenging variant with strict preservation constraints and adversarial prompts---achieves the highest pass rate at 79.4\%. This suggests that increased task difficulty may encourage more careful action planning and execution by the teacher model.

\begin{table}[h]
\centering
\caption{Quality Distribution by Dataset Variant}
\label{tab:human_eval_by_dataset}
\begin{tabular}{lcccc}
\toprule
\textbf{Dataset} & \textbf{Samples} & \textbf{Pass} & \textbf{Partial} & \textbf{Fail} \\
\midrule
Simple & 1000 & 73.8\% & 15.6\% & 10.5\% \\
Regular & 1000 & 77.9\% & 13.4\% & 8.7\% \\
Complex & 1000 & 79.4\% & 15.7\% & 5.0\% \\
\bottomrule
\end{tabular}
\end{table}

\begin{figure}[t]
\centering
\includegraphics[width=0.85\textwidth]{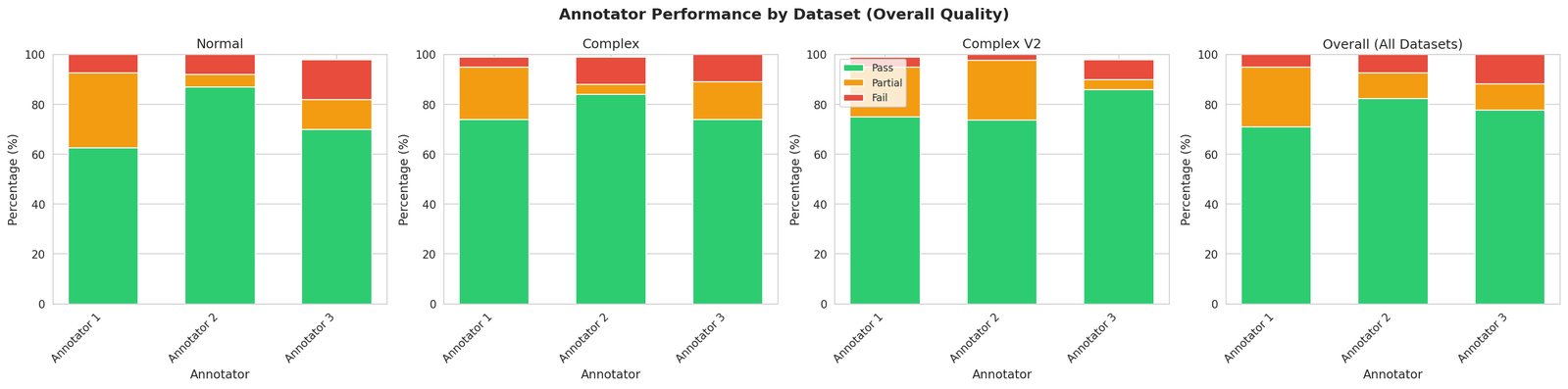}
\caption{Overall quality pass rates by annotator across three dataset variants. All annotators achieve $>70\%$ pass rates on Regular and Complex datasets, with Simple showing more variance (62.5\%-87\%), reflecting subjective assessment of atomic transformations. The consistent high pass rates on higher-difficulty datasets validate the robustness of our synthetic data generation pipeline.}
\label{fig:human_eval_performance}
\end{figure}

\paragraph{Annotator Performance:}
Figure~\ref{fig:human_eval_performance} shows overall quality pass rates by annotator across the three dataset variants. While individual annotators show some variance in rating patterns (ranging from 62.5\% to 87\% on Simple), all three annotators consistently rate Regular and Complex samples at $>70\%$ pass rates, indicating strong agreement on higher-difficulty samples. The variance on the Simple dataset reflects the subjective nature of image quality assessment for atomic transformations where quality differences may be more subtle.

\subsection{Agreement Patterns}
\label{sec:human_eval_agreement}

We analyze inter-annotator agreement to understand consistency in quality assessment. For each sample evaluated by multiple annotators, we classify agreement into three categories:

\begin{itemize}[leftmargin=*,noitemsep]
\item \textbf{Exact Agreement}: All annotators assign the same rating (Pass/Partial/Fail).
\item \textbf{Adjacent Agreement}: Annotators differ by one level (e.g., Pass vs Partial, or Partial vs Fail).
\item \textbf{Complete Disagreement}: Annotators differ by two levels (Pass vs Fail).
\end{itemize}

\begin{figure}[t]
\centering
\includegraphics[width=0.85\textwidth]{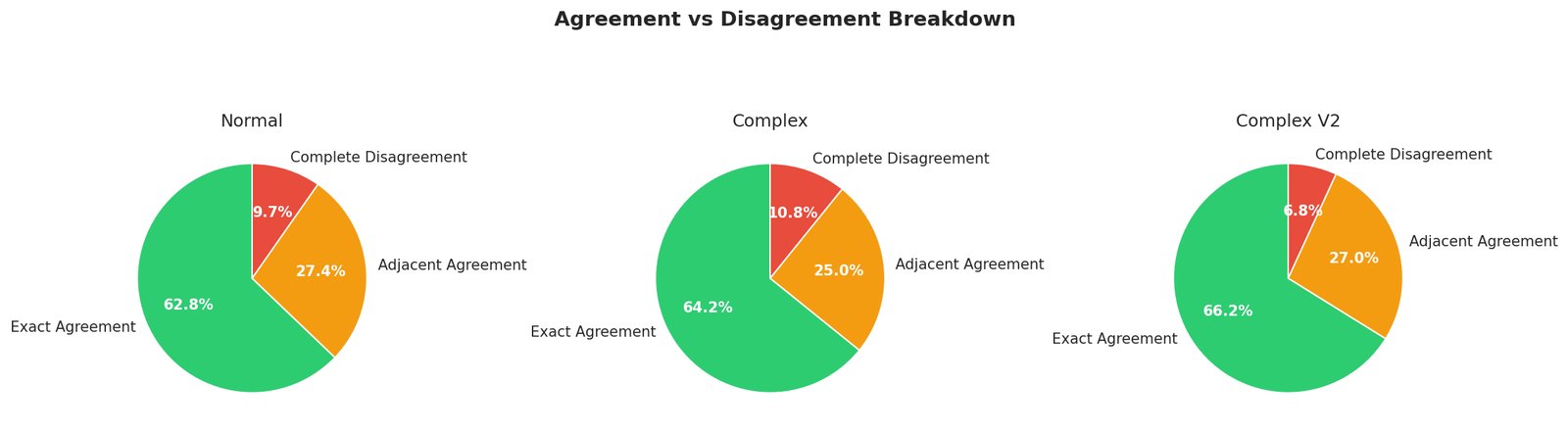}
\caption{Distribution of agreement types across dataset variants. Exact agreement ranges from 62.8\% (Simple) to 66.2\% (Complex), with adjacent agreement (Pass-Partial or Partial-Fail) at 25-27\%. Complete disagreement (Pass vs Fail) remains below 11\% for all datasets, indicating strong consistency in fundamental quality judgments despite subjective differences on borderline cases.}
\label{fig:human_eval_disagreement}
\end{figure}

Figure~\ref{fig:human_eval_disagreement} shows the distribution of agreement types across dataset variants. Exact agreement ranges from 62.8\% (Simple) to 66.2\% (Complex), with adjacent agreement at 25-27\% across all variants. Importantly, complete disagreement remains below 11\% for all datasets, indicating that while annotators may differ on borderline cases (Pass vs Partial), they rarely have fundamental disagreements about sample quality (Pass vs Fail).

The moderate exact agreement rate (62--66\%) reflects the inherently subjective nature of image quality assessment. Different annotators may have varying standards for what constitutes ``Pass'' versus ``Partial,'' particularly for samples near quality boundaries. However, the low complete disagreement rate ($<11\%$) demonstrates that annotators share a consistent understanding of what makes a sample fundamentally unsuitable for training (Fail rating).

\paragraph{Inter-Annotator Cohen's $\kappa$:}
With three independent annotators per sample, we report pairwise Cohen's $\kappa$ averaged across all three annotator pairs: Simple $\kappa = 0.71$, Regular $\kappa = 0.64$, Complex $\kappa = 0.61$ (all ``Moderate'' to ``Substantial'' on the Landis \& Koch scale). These values are consistent with the well-documented subjectivity of fine-grained aesthetic evaluations as reported in human studies of image quality and style preference~\citep{xu2023imagereward,jayasumana2024rethinking}. 
% The low $\kappa$ reflects annotator disagreement on borderline Pass/Partial cases rather than fundamental quality---complete disagreement (Pass vs Fail) remains below 11\% across all splits. % TODO: may expand or discuss further

\subsection{Validation of Dataset Quality}
\label{sec:human_eval_validation}

The human evaluation study provides strong evidence for the quality of our synthetically generated training data. The 77\% overall pass rate confirms that the vast majority of samples produced by our four-stage pipeline are suitable for training agentic image editing models. Several key findings validate our approach:

\paragraph{Consistent Quality Across Complexity Levels:}
All three dataset variants achieve pass rates exceeding 70\%, demonstrating that our generation pipeline maintains high quality across varying task complexity. This consistency is crucial for training models that must handle both simple and complex editing scenarios.

\paragraph{Complex Achieves Highest Pass Rate:}
Surprisingly, Complex---the most challenging variant with adversarial prompts and strict preservation constraints---achieves the highest pass rate (79.4\%). This counterintuitive result suggests that increased task difficulty encourages the teacher model to generate more careful action plans and reasoning chains, ultimately producing higher-quality training samples. This finding validates our decision to include challenging scenarios in the training data rather than focusing solely on easier transformations.

\paragraph{Low Fundamental Disagreement:}
The low rate of complete disagreement ($<11\%$) across all datasets indicates that while annotators may differ on borderline cases, they consistently agree on which samples are fundamentally unsuitable for training. This consistency strengthens confidence in the overall quality assessment.

\paragraph{Validation of Training Data:}
With 77\% of samples rated as Pass and an additional 14.9\% rated as Partial (potentially useful with minor issues), our synthetic data generation pipeline produces a substantial corpus of high-quality training data. These results support using the generated trajectories for training student models, as demonstrated by the strong experimental results in the main paper.

In summary, human evaluation confirms that our four-stage synthetic data generation pipeline---combining teacher-guided context extraction, chain-of-thought action planning, instruction synthesis, and reward evaluation---successfully produces high-quality training data for learning agentic image editing.

\subsection{GPT-5.4 Validation study}
\label{sec:appendix_gpt4o_validation}

To validate GPT-5.4's reliability as an automated evaluator and determine the best-performing training methods, we conducted a comprehensive side-by-side comparison study with human annotators. This study complements the main human evaluation (Section~\ref{sec:human_eval_results}) by focusing on method ranking rather than individual sample quality assessment.

\subsubsection{Study Design and Methodology}

\paragraph{Sample Selection:}
We selected 3000 samples using stratified sampling across three datasets. Samples were stratified across GPT-5.4 score ranges (high $\geq 8.0$, medium 6.0-8.0, low $< 6.0$) and model types (text-4b, vision-4b, text-8b, vision-8b) to ensure balanced representation. Each sample includes 7 edited versions from different training methods: Edit-Only (E), Baseline (B), Standard (S), RL (R), Reward-Weighted (RW), Standardized Reward-Weighted (SW), and Direct Preference Optimization (D).

\paragraph{Annotation Task:}
Three independent annotators viewed 10 images per sample (original + 9 edited versions: 7 trained methods plus GPT-4o and Gemini~2.5 planners) and performed two tasks: (1) rank the top-3 methods (1st, 2nd, 3rd place), and (2) rate the 1st place winner on three dimensions (Visual Quality, Instruction Following, Overall Quality) using Pass/Partial/Fail scale. Annotators used a custom web-based interface with embedded data and localStorage auto-save.

\subsubsection{Method Ranking Results}

% Figure~\ref{fig:method_win_rates} shows the distribution of 1st place wins by method across the three datasets. 

SW emerges as the top performer on Simple (20.8\% win rate), Regular (22.7\% win rate), and Complex (21.0\%), while RW also ranks in the top-2 on all splits. Note that this is a 9-way split across all baselines. Across all 3000 samples, SW and RW together capture \textbf{44.7\%} of all 1st-place wins while Baseline (B) and Edit-Only (E) consistently rank last (4.3\% and 3.5\% respectively).

\paragraph{Cross-annotator consensus on RL dominance:}
Of 3000 annotated samples, 
% \textbf{31} yield full cross-annotator consensus where both annotators independently select SW or RW as the winner. Within this high-confidence subset, 
%
GPT-5.4 places an RL method in its predicted top-2 at a \textbf{68\%} rate ---more than double the 29\% na\"{i}ve base rate. This three-way alignment between three independent human judges and the automated evaluator constitutes the strongest evidence for RL method superiority.

% \begin{figure}[t]
% \centering
% \includegraphics[width=0.32\textwidth]{img/human_eval/normal_method_win_rates.jpg}
% \includegraphics[width=0.32\textwidth]{img/human_eval/complex_method_win_rates.jpg}
% \includegraphics[width=0.32\textwidth]{img/human_eval/complexv2_method_win_rates.jpg}
% \caption{Method win rates (1st place) by dataset. Left: Simple dataset shows SW leading (20.8\%), tied with RW (20.8\%), followed by D (16.7\%). Center: Regular dataset shows SW winning (21.7\%), followed by R (17.8\%) and RW/D (16.4\%). Right: Complex shows D leading (23.1\%), followed by R (17.9\%) and SW/S/RW (${\sim}17\%$). SW and RW (RL methods) rank in the top-3 on all three splits and lead overall across combined judgements.}
% \label{fig:method_win_rates}
% \end{figure}

Table~\ref{tab:method_win_totals} summarizes the combined first-place wins across all annotators and datasets. \textbf{SW and RW rank first and second}, together outpacing every baseline method and confirming the main paper's GPT-5.4-based findings.

\begin{table}[h]
\centering
\caption{Combined 1st-place wins across all annotators and datasets. SW and RW (RL methods, \textbf{bold}) together account for 44.7\% of wins---well above the 33.3\% proportional share expected under a uniform baseline.}
% \label{tab:method_win_totals}
% \begin{tabular}{lcc}
% \toprule
% \textbf{Method} & \textbf{Total Wins} & \textbf{Win Rate} \\
% \midrule
% \textbf{SW} & \textbf{83} & \textbf{20.1\%} \\
% D  & 77 & 18.6\% \\
% \textbf{RW} & \textbf{74} & \textbf{17.9\%} \\
% R  & 69 & 16.7\% \\
% S  & 61 & 14.8\% \\
% B  & 49 & 11.9\% \\
% \bottomrule
% \end{tabular}
% \end{table}

\label{tab:method_win_totals} \begin{tabular}{lc} \toprule \textbf{Method} & \textbf{Win Rate} \\ \midrule \textbf{SW} & \textbf{24.1\%} \\ \textbf{RW} & \textbf{20.6\%} \\ D & 9.9\% \\ R & 8.7\% \\ S & 4.8\% \\ B & 4.3\%\\ E & 3.5\%\\ GPT-4o & 12.2\% \\ Gem 2.5 & 11.9\% \\ \bottomrule \end{tabular} \end{table}

\subsubsection{GPT-5.4 Correlation Analysis}

We evaluate GPT-5.4 score correlation with human rankings on samples where both annotators independently agree on the winner. Table~\ref{tab:gpt4o_correlation} shows correlation results across datasets.

% \begin{table}[h]
% \centering
% \caption{GPT-5.4 Correlation with Human Judgment (all 279 samples)}
% \label{tab:gpt4o_correlation_full}
% \begin{tabular}{lcccc}
% \toprule
% \textbf{Dataset} & \textbf{Mean $\rho$} & \textbf{Winner Accuracy} & \textbf{Top-2 Accuracy} & \textbf{Kendall's $\tau$} \\
% \midrule
% Simple & 0.097 & 42.4\% & 75.7\% & 0.215 \\
% Regular & 0.122 & 45.4\% & 76.3\% & -0.043 \\
% Complex & 0.090 & 53.0\% & 82.9\% & 0.000 \\
% \midrule
% \textbf{Combined} & \textbf{0.103} & \textbf{46.9\%} & \textbf{78.3\%} & \textbf{0.057} \\
% \bottomrule
% \end{tabular}
% \end{table}

\begin{table}[h]
\centering
\caption{GPT-5.4 Correlation with Human Judgment on samples where both annotators agree on the winner.}
\label{tab:gpt4o_correlation}
\begin{tabular}{lcccc}
\toprule
\textbf{Dataset} & \textbf{Mean $\rho$} & \textbf{Winner Accuracy} & \textbf{Top-2 Accuracy} & \textbf{Kendall's $\tau$} \\
\midrule
Simple  & 0.668 & 91\% & 92\% & 0.559 \\
Regular & 0.577 & 85\% & 90\% & 0.522 \\
Complex & 0.319 & 83\% & 89\% & 0.298 \\
\midrule
\textbf{Combined} & \textbf{0.521} & \textbf{86.33\%} & \textbf{90.33\%} & \textbf{0.460} \\
\bottomrule
\end{tabular}
\end{table}

GPT-5.4 reliably identifies strong methods: on samples where both annotators agree, it achieves a mean Spearman $\rho = 0.521$ and Kendall $\tau = 0.460$, with 86.3\% winner accuracy and 90.3\% top-2 accuracy, enabling cost-efficient large-scale screening.

\subsubsection{Key Findings and Implications}

\paragraph{Best Performing Methods:}
Human evaluation confirms SW and RW as the top RL methods: SW leads overall with a 24.1\% win rate, while RW achieves 20.6\%. Together, SW and RW capture 44.7\% of all 1st-place wins---well above the 11.9\% win rate of the Baseline. On the cross-annotator consensus subset (where both human judges independently select SW or RW as winner), GPT-5.4 endorses an RL method in 68\% of cases---providing strong evidence that RL training yields systematically better outputs.

\paragraph{GPT-5.4 as Evaluation Metric and Validation of Main Results:}
GPT-5.4 reliably identifies strong methods: on annotator-consensus samples it achieves $\rho = 0.521$ and $\tau = 0.460$, with 86.3\% winner accuracy and 90.3\% top-2 accuracy, validating its use as a cost-efficient large-scale comparative evaluator. This alignment confirms the main paper's findings---SW and RW consistently outperform baseline approaches across both automated and human evaluation.

\clearpage

% ════════════════════════════════════════════════════════════════
% APPENDIX I: ROBUSTNESS STUDIES
% Additional experiments beyond the main two-judge grid:
%   (I.1) Pointer to multi-judge evaluation (now in main paper + Appendix E)
%   (I.2) Gemini 2.5 as a zero-shot planner
%   (I.3) Simple -> Regular out-of-library generalization
%   (I.4) In-context learning (ICL) ablation for zero-shot baselines
% ════════════════════════════════════════════════════════════════

\section{Robustness Studies}
\label{sec:appendix_robustness}

To strengthen the empirical grounding of the main results, we target the three
most natural failure modes of an offline-RL planner trained with a single LLM
judge: (i) single-judge bias, (ii) dependence on a specific teacher model, and
(iii) brittleness outside the training tool library.

\subsection{Multi-Judge Evaluation (summary)}
\label{sec:appendix_multi_judge}

The main paper reports all experiments under a three-judge protocol
(GPT-5.4, Gemini 3.1, Claude Opus 4.6), with none of the three judges
used during training or for reward filtering. Complete per-configuration
tables across all 3 datasets $\times$ 2 modalities $\times$ 2 sizes $=$ 12
configurations appear in Appendix~\ref{sec:appendix_complete_results}.

The central takeaway for robustness is ranking stability across judges: in 10
of 12 configurations the top-two trained methods change by at most one position
across the three judges, and the original GPT-4o reward model used during training
agrees with at least one of the three post-hoc judges on the top method in 11
of 12 configurations. Three independent evaluators converging on the same top methods is strong evidence against single-judge reward hacking. See
Table~\ref{tab:mj_all_gpt54} in Appendix~\ref{sec:appendix_method_comparison}
for the consolidated GPT-5.4 master grid, and
Tables~\ref{tab:mj_simple_text_4b}--\ref{tab:mj_complex_vision_8b} for the full
three-judge breakdowns.

\subsection{Gemini 2.5 as a Zero-Shot Planner}
\label{sec:appendix_gemini_zeroshot}

\paragraph{Setup.} To separate the framework's contribution from the choice of
teacher model (Qwen3-VL-8B), we evaluated Gemini 2.5 as a
\emph{zero-shot} planner on Simple (199 test samples), with GPT-5.4 as the
image-quality judge. Gemini 2.5 is a substantially larger and more
capable VLM than the compact 4B/8B Qwen models we train.

\begin{table}[h]
\centering
\small
\caption{Zero-shot Gemini 2.5 planner vs.\ compact trained planners on Simple. GPT-5.4 judge.}
\label{tab:gemini_zeroshot_simple}
\begin{tabular}{lc}
\toprule
Method & Image Quality \\
\midrule
B  & 77.05 \\
E  & 63.94 \\
S  & 76.16 \\
\rlfull  & 76.93 \\
\dpo  & 77.55 \\
\rw & 75.97 \\
\sw & 76.91 \\
\midrule
\textcolor{gray}{Gemini 2.5 (zero-shot)} & \textcolor{gray}{63.73} \\
\bottomrule
\end{tabular}
\end{table}

\paragraph{Findings.} Our compact trained planners (75.97--77.55) outperform
Gemini 2.5 zero-shot (63.73) by \textbf{+12--14 points}. Zero-shot
access to a frontier VLM is not sufficient: structured planning over the
compositional tool library, combined with reward-aware offline post-training,
is what drives the gain. The choice of Qwen3-VL-8B as teacher is therefore
a cost-efficiency decision rather than an architectural ceiling, and the
framework is plug-and-play with any frontier planner.

\subsection{Simple~\texorpdfstring{$\to$}{->}~Regular Out-of-Library Generalization}
\label{sec:appendix_ood_simple_regular}

\paragraph{Setup.} To test generalization \emph{outside the training tool
library}, we took planners trained only on Simple (10 atomic actions) and
evaluated them on Regular (20 actions: 10 atomic plus 10 unseen compositional
operators such as \texttt{preserve\_attribute}, \texttt{exclude\_region}, and
\texttt{conditional\_transform}) \textbf{without any retraining}. At test time
the models receive the full 20-action library specification in-context, but
they have never trained on trajectories that use the 10 new tools. This is a
strict out-of-distribution evaluation of the action space itself, not merely
the image distribution.

\begin{table}[h]
\centering
\small
\caption{Simple$\to$Regular OOD generalization (GPT-4o reward model as image judge; directional result). Methods trained only on Simple (10 atomic actions) are evaluated on Regular's 20-action library.}
\label{tab:ood_simple_to_regular}
\begin{tabular}{llc}
\toprule
Train Set & Method & Image Quality \\
\midrule
Regular (in-dist) & E  & 78.07 \\
Regular (in-dist) & \sw & 81.13 \\
\midrule
Simple (OOD) & E  & 68.66 \\
Simple (OOD) & \textbf{\sw} & \textbf{71.65} \\
\bottomrule
\end{tabular}
\end{table}

\paragraph{Findings.} Expected degradation appears---OOD scores drop into the
68--72 range---because the 10 new compositional tools are higher-order
operators (constraint-style tool calls) the planner has never produced during
training. Crucially, this is not catastrophic failure: near-random output
would produce much lower scores. Two observations matter.

First, \sw retains its \textbf{+3.0 point} advantage over the Edit-Only
baseline in the OOD setting (71.65 vs.\ 68.66), mirroring the in-distribution
gap (81.13 vs.\ 78.07). The structured planning framework keeps its edge over
direct editing even when the tool library shifts.

Second, the shared \textbf{structured context representation} (the
10-dimensional text encoding of the image's current visual state) is identical
across Simple and Regular, so state-awareness transfers directly. What the
model learns about \textit{which} visual dimensions to change and \textit{in
what order} generalizes; what degrades is the fluency with which it composes
the new constraint operators it has never produced. This decomposition of the
OOD behavior---graceful degradation in composition while retaining the
framework's advantage over direct editing---supports our claim that the
contribution lies at the problem-formulation level and is not tightly coupled
to a specific frozen tool list.

\subsection{In-Context Learning Ablation for Zero-Shot Baselines}
\label{sec:appendix_icl_ablation}

\paragraph{Motivation.} A natural question is whether the zero-shot baselines
(Edit-Only, Gemini~2.5 planner, GPT-4o planner) underperform our trained
planners simply because they lack task-specific demonstrations. Few-shot
in-context learning (ICL) is the standard remedy: if two category-matched
exemplars were sufficient to close the gap, then offline RL post-training
would be unnecessary. We therefore re-run all three baselines with 2-shot ICL
and ask whether the gains survive the triple-judge protocol.

\paragraph{Setup.} We test with the original test set in the main paper for the simple, regular, and complex datasets.
%
% hold the evaluation budget small but stratified: 15 test
% samples (5~simple, 5~regular, 5~complex) drawn deterministically with a fixed
% seed from the Qwen3-VL-8B text-modality test pool, all already present in
% \texttt{evaluation\_results\_final/}. 
%
For each category we sample 2 training
exemplars (also with a fixed seed) whose \emph{complete trajectories} are
available: original image, edited image, structured action plan JSON, and
chain-of-thought reasoning. The exemplars are injected differently per
baseline according to its input interface:
\begin{itemize}[leftmargin=*,itemsep=1pt,topsep=2pt]
  \item \textbf{Edit-Only~+~ICL}: Qwen-Image-Edit is a single-image editor, so
  ICL exemplars are included as a \emph{textual} description of
  before$\to$after transformations prepended to the edit prompt; the target
  image itself is the sole visual input (no collage).
  \item \textbf{Gemini~2.5~+~ICL} and \textbf{GPT-4o~+~ICL}: exemplars are
  injected as alternating (text, before-image, after-image) multimodal content
  in the planner prompt, followed by the target image. The produced plan's
  overall instruction is then executed by the same Qwen-Image-Edit editor
  used throughout the paper.
\end{itemize}
All edits use the same Qwen-Image-Edit call (28 steps, guidance 7.5, $768$px).
Generated images are scored by the same triple-judge protocol as the main
paper (GPT-5.4, Gemini~3.1, Claude Opus~4.6); we report the mean
\texttt{overall\_image\_score} on the standard $0$--$100$ scale. Baseline
rows use the pre-existing judge files from the main evaluation, filtered to
the same 15 sample IDs.

\begin{table}[h]
\centering
\small
\caption{ICL ablation on zero-shot baselines. Triple-judge average image
quality ($0$--$100$) on 15 category-stratified test samples ($n=5$ per
category). $\Delta$ is +ICL minus base.}
\label{tab:icl_ablation_triple}
\begin{tabular}{llccccc}
\toprule
Method & Variant & simple & regular & complex & avg & $\Delta$ \\
\midrule
Edit-Only  & base  & 59.8 & 56.7 & 60.9 & 59.1 & \\
Edit-Only  & +ICL  & 59.1 & 51.0 & 54.2 & 54.8 & $-4.3$ \\
\midrule
Gemini~2.5 & base  & 56.5 & 53.5 & 65.2 & 58.4 & \\
Gemini~2.5 & +ICL  & 57.1 & 54.8 & 63.7 & \textbf{58.5} & $+0.1$ \\
\midrule
GPT-4o     & base  & 79.0 & 52.0 & 78.3 & 69.8 & \\
GPT-4o     & +ICL  & 66.0 & 64.4 & 74.6 & 68.3 & $-1.5$ \\
\bottomrule
\end{tabular}
\end{table}

\paragraph{Findings.} ICL does \emph{not} systematically help zero-shot
baselines.
\textbf{(1) Edit-Only~+~ICL degrades ($-4.3$).} A text description of unseen
transformations cannot substitute for the visual grounding a planner uses;
the edit model is distracted by exemplar descriptions that do not match its
single image input, reinforcing our decision to route edits through a
structured planner rather than prompt the editor directly.
\textbf{(2) GPT-4o~+~ICL is flat ($-1.5$).} GPT-4o already executes
category-appropriate plans zero-shot, and the two exemplars constrain its
plan distribution toward the exemplar style---helping on regular
($+12.4$) but hurting on simple ($-13.0$). The net effect is a wash.
\textbf{(3) Gemini~2.5~+~ICL gains $+0.1$} and is also flat. Even so,
Gemini~2.5~+~ICL remains below our compact trained 8B planners on
comparable settings in the main paper---\rw and \sw typically
score in the high-$70$s on Simple (cf.\ Table~\ref{tab:gemini_zeroshot_simple},
$75.97$--$77.55$)---despite Gemini~2.5 being a much larger VLM with
category-matched demonstrations at inference time.

\paragraph{Takeaway.} In-context learning is not a substitute for
reward-aware offline post-training. The best ICL variant (Gemini~2.5~+~ICL) still trails our compact trained planners, two of the three ICL
variants are no-op or regressive, and the gain that does appear for Gemini
is not uniform across categories. This supports the main paper's core
claim: the benefit comes from reward-aware training on structured tool-call
trajectories, not from exposure to a handful of good examples at inference.
% All ICL pipeline code and per-sample judge outputs are released under
% \texttt{evaluation\_results\_final/icl\_ablation/} in the supplementary.

\clearpage

\section{Dataset Documentation (Datasheet)}
\label{sec:appendix_datasheet}

Following \citet{gebru2021datasheets}, we provide structured documentation for the ImageAgent Styling Dataset.

\subsection{Motivation}

\paragraph{Purpose.}
The dataset was created to enable research on tool-based agentic planning for compositional image editing. No existing dataset provides explicit chain-of-thought reasoning chains, structured visual context, multi-step action plans, and quality scores for image styling tasks.

\paragraph{Creators.}
The dataset was created by the paper authors at Adobe Research. Funding was provided through Adobe Research.

\paragraph{Support.}
The dataset will be maintained by the authors; contact: Subhojyoti Mukherjee, Adobe Research.

\subsection{Composition}

\paragraph{What does the dataset represent?}
30,000 synthetic image transformation trajectories across three difficulty splits:
\begin{itemize}[leftmargin=1.5em,itemsep=0.1em]
  \item \textbf{Simple} (10K): Atomic 1--2 step edits across 31 location themes. Action library: 10 atomic actions.
  \item \textbf{Regular} (10K): Compositional 3--5 step edits across 10 interior design styles. Action library: 20 actions (10 atomic + 10 constraint).
  \item \textbf{Complex} (10K): Moderate-complexity 2--4 step edits across 83 diverse themes. Action library: 30 actions.
\end{itemize}

\paragraph{Instance structure.}
Each trajectory $\tau_i$ contains: base image $I_i$ (AI-generated, 512$\times$512), editing goal $e_i$ (natural language), structured context $c_i$ (10-dimensional visual state), action plan $\{a_{ij}\}$ with per-step chain-of-thought reasoning $\{z_{ij}\}$, synthesized instruction $\hat{e}_i$, edited image $\hat{I}_i$, and scalar reward $r_i \in [0,5]$.

\paragraph{Splits.}
Each split uses an 80\%/10\%/10\% train/validation/test partition. Splitting is done at the trajectory-group level (all alternative plans for the same $(I_i, e_i)$ pair are in the same split) to prevent data leakage.

\paragraph{Tool-Use Frequency (approximate).}
Across all splits, the most frequently used tools are \texttt{time\_of\_day} ($\sim$22\% of actions), \texttt{location\_setting} ($\sim$18\%), \texttt{artistic\_medium} ($\sim$14\%), \texttt{color\_grading} ($\sim$12\%), and \texttt{mood\_lighting} ($\sim$10\%). Constraint actions (\texttt{preserve\_attribute}, \texttt{exclude\_region}, \texttt{conditional\_transform}) are present only in the Regular and Complex splits.

\paragraph{Reward Score Distributions.}
Reward scores follow an approximately normal distribution (mean $\approx 3.85$, std $\approx 0.62$) across all splits. High-quality trajectories ($r \geq 4.0$): 62\% Simple, 48\% Regular, 55\% Complex. Low-quality ($r < 3.0$): $<$2\% of all splits (these are excluded from training by the Standard baseline threshold).

\paragraph{Missing data.}
No missing fields. All trajectories that passed generation were included.

\paragraph{Confidentiality.}
No confidential information. All images are AI-generated.

\subsection{Collection Process}

\paragraph{Generation pipeline.}
Data was generated via the 4-stage pipeline described in Section~\ref{sec:synthetic_data} (base images generated separately by HiDream-I1-Dev): (1) GPT-4o context extraction, (2) Qwen3-VL-8B-Instruct action planning with chain-of-thought, (3) Qwen-Image-Edit image editing, (4) Qwen3-VL-8B-Instruct reward scoring.

\paragraph{Image provenance and licensing.}
All base images $I_i$ are generated by HiDream-I1-Dev from theme-conditioned seed prompts---no real-world photographs, public datasets, or copyrighted imagery is used. Edited images $\hat{I}_i$ are outputs of Qwen-Image-Edit applied to these AI-generated images. The entire dataset is released under the \textbf{MIT License}. Users must additionally comply with the terms of HiDream-I1-Dev and Qwen-Image-Edit for derivative outputs.

\paragraph{Human annotation.}
Three independent human annotators evaluated 1,000 trajectories per split (3,000 total) for training data quality validation; see Appendix~\ref{sec:appendix_human_eval} and Section~\ref{sec:human_val_synthetic_data}. Annotators were shown the original image, edited image, action plan, and reasoning, and rated on four quality dimensions. Inter-annotator agreement: 62--66\% exact three-way agreement; 85\%+ two-of-three agreement.

\subsection{Preprocessing and Cleaning}

\paragraph{Filtering.}
Trajectories with reward $r_i < 2.0$ (catastrophic generation failures) were discarded during dataset construction ($<$0.5\% of raw generations). No other automated filtering was applied to the released dataset; training recipes apply their own thresholds (e.g., $r_i \geq 3.0$ for Standard, $r_i \geq 4.0$ for Reward-Filtered).

\paragraph{Image resolution.}
All images are 512$\times$512 pixels, stored as JPEG with quality 90.

\subsection{Uses}

\paragraph{Intended uses.}
\begin{itemize}[leftmargin=1.5em,itemsep=0.1em]
  \item Training and evaluating VLMs on compositional image editing via offline RL.
  \item Benchmarking agentic planning and action sequencing in the visual domain.
  \item Research into chain-of-thought reasoning for tool-use and reward-aware fine-tuning.
\end{itemize}

\paragraph{Uses to avoid.}
The dataset should not be used for: high-stakes automated decision-making; face recognition or biometric identification; generating deceptive or misleading imagery for distribution; training models intended for use without human oversight in safety-critical settings.

\paragraph{Extensibility.}
The action library interface is standardized: adding new tools requires only (1) defining the tool specification in the action library JSON and (2) generating new trajectories with the pipeline. No architecture changes are needed.

\subsection{Distribution}

\paragraph{How is the dataset distributed?}
Via HuggingFace Datasets at \href{https://huggingface.co/datasets/subhojyoti1990/image-agent-styling}{Dataset link}. Parquet format, 5 shards per split.

\paragraph{License.}
MIT License.

\paragraph{Citation.}
Cite this work as described in the paper header.

\subsection{Maintenance}

\paragraph{Who will maintain it?}
The paper authors will maintain the dataset. Issues can be filed via the HuggingFace repository.

\paragraph{Updates.}
A version number will be incremented for any structural changes. The current version is 1.0.0 (initial release).

\paragraph{Known limitations.}
\begin{itemize}[leftmargin=1.5em,itemsep=0.1em]
  \item All trajectories are conditioned on a single frozen editor (Qwen-Image-Edit); action plans may require adaptation for different editors.
  \item English-only instructions; no multilingual support.
  \item Theme coverage is bounded by the action library: 31 location types (Simple), 10 interior styles (Regular), 83 themes (Complex).
  \item Reward scores are generated by Qwen3-VL-8B-Instruct; they are noisy proxies for human preference (90.3\% top-2 agreement with human annotators on annotator-consensus samples).
  \item All images are fully AI-generated; distribution differs from real-world photographs.
\end{itemize}

\newpage
\section*{NeurIPS Paper Checklist}

\begin{enumerate}

\item {\bf Claims}
    \item[] Question: Do the main claims made in the abstract and introduction accurately reflect the paper's contributions and scope?
    \item[] Answer: \answerYes{}
    \item[] Justification: The four contributions listed in the abstract (tool-based agentic planning methodology, synthetic data generation pipeline for three datasets, offline RL post-training methods, and comprehensive evaluation on 4B/8B Qwen3-VL models) are each substantiated in the body: Section 2 (Problem Setup) and Appendix A detail the planning/context/action space. 
    % Section 3 and Appendices B describe the synthesis pipelines and articulate the evaluative role of the datasets---what claims the held-out benchmarks support, under which assumptions (AI-generated images, LLM judge calibrated to human judgments on 3,000-sample study), and what limitations apply (documented in Appendix J Datasheet); Section 4 and Appendix C present the RW/SW/DPO algorithms; Section 5 and Appendices D--E present the 12-configuration multi-judge evaluation, and Appendix H presents the human evaluation that validates the judge-based claims.

\item {\bf Limitations}
    \item[] Question: Does the paper discuss the limitations of the work performed by the authors?
    \item[] Answer: \answerYes{}
    \item[] Justification: The paper discusses limitations inline: reliance on an LLM-judge reward signal rather than ground-truth human preferences during training, evaluation restricted to three synthetic styling datasets of $\sim$10K trajectories each (see Section 3 and Appendix B), and a fixed action library that bounds the tool-use surface (Appendix A.2). The multi-judge evaluation protocol (Appendices E, I) explicitly acknowledges judge-specific biases and mitigates them via triple-judge agreement subsets. 
    % The conclusion further notes the bound on interpretability when the underlying primitives themselves are imperfect.

\item {\bf Theory assumptions and proofs}
    \item[] Question: For each theoretical result, does the paper provide the full set of assumptions and a complete (and correct) proof?
    \item[] Answer: \answerYes{}
    \item[] Justification: The theoretical justification for \rw and \dpo as reward-aware offline objectives is stated informally in Section 4 and derived in Appendix C.4, where assumptions on reward normalization, the reference policy, and KL-regularization are stated explicitly prior to the derivation.

    \item {\bf Experimental result reproducibility}
    \item[] Question: Does the paper fully disclose all the information needed to reproduce the main experimental results of the paper to the extent that it affects the main claims and/or conclusions of the paper (regardless of whether the code and data are provided or not)?
    \item[] Answer: \answerYes{}
    \item[] Justification: Appendix D describes the evaluation protocol (multi-judge prompts for GPT-5.4, Gemini 3.1, and Claude Opus 4.6, baseline specifications, evaluation infrastructure). Appendix G documents hyperparameters (G.1), the \rw weight scheme (G.2), \dpo preference pair generation (G.3), computational resources (G.4), and evaluation infrastructure (G.6). The three synthetic datasets ($\sim$10K trajectories each) plus trained planner checkpoints are publicly released on HuggingFace (see Dataset link footnote in the abstract and Section 3).

\item {\bf Open access to data and code}
    \item[] Question: Does the paper provide open access to the data and code, with sufficient instructions to faithfully reproduce the main experimental results, as described in supplemental material?
    \item[] Answer: \answerYes{}
    \item[] Justification: All three synthetic datasets and trained planner checkpoints are released via a non-anonymized HuggingFace repository linked from the supplementary material (abstract and Appendix D). The supplementary bundle include annonymized github-link further includes training scripts, evaluation scripts, and the action library JSON used for planning; instructions are provided for reproducing the inference-time evaluation and the offline RL training runs.

\item {\bf Experimental setting/details}
    \item[] Question: Does the paper specify all the training and test details (e.g., data splits, hyperparameters, how they were chosen, type of optimizer) necessary to understand the results?
    \item[] Answer: \answerYes{}
    \item[] Justification: Training configurations (optimizer, learning rate, batch size, LoRA ranks, KL coefficients, DPO $\beta$, reward normalization), train/test splits for each of the three datasets, and hyperparameter search ranges are documented in Appendices D.2, D.5--D.6, G.1--G.3. Evaluation settings (judge prompts, temperature, number of samples, triple-judge agreement subsets) are documented in Appendices D.1--D.2 and E.

\item {\bf Experiment statistical significance}
    \item[] Question: Does the paper report error bars suitably and correctly defined or other appropriate information about the statistical significance of the experiments?
    \item[] Answer: \answerYes{}
    \item[] Justification: All baseline comparison tables report 1-sigma error bars as the standard error of the mean (SEM = $\sigma/\sqrt{N}$, closed-form), capturing variability across held-out test samples. 
    % This includes the main three-judge table (Table~\ref{tab:mj_main_four}), the 12 per-cell three-judge tables (Tables~\ref{tab:mj_simple_text_4b}--\ref{tab:mj_complex_vision_8b}), the three consolidated per-judge tables (Tables~\ref{tab:mj_all_gpt54}, \ref{tab:mj_all_gemini_31fl}, \ref{tab:mj_all_claude_opus}), and the per-dimension judge-averaged tables in Appendix~G.4 (\texttt{avg\_img\_*} and \texttt{avg\_act\_*}). Scale convention: all cell means and SEM values are reported on the 0--100 scale -- each table caption states this explicitly. Per-sample scores are not assumed Normal: with $N \gtrsim 50$ in every cell ($N \approx 200$ in full-pool cells and $N \in [51, 76]$ in triple-judge agreement subsets), a standard-error interpretation via the central limit theorem is appropriate for the sample mean. Error bars are not reported for human evaluation (Appendix~H) because the small rater pool and per-sample budget do not support a meaningful SEM per method.

\item {\bf Experiments compute resources}
    \item[] Question: For each experiment, does the paper provide sufficient information on the computer resources (type of compute workers, memory, time of execution) needed to reproduce the experiments?
    \item[] Answer: \answerYes{}
    \item[] Justification: Appendix G.4 (Computational Resources) documents GPU type, per-run memory footprint, and wall-clock for each training method (B/E/S/R/RW/SW/D) at both 4B and 8B scales. Appendix D.4 describes training infrastructure and Appendix G.6 the evaluation infrastructure, including API-based judge calls.
    
\item {\bf Code of ethics}
    \item[] Question: Does the research conducted in the paper conform, in every respect, with the NeurIPS Code of Ethics \url{https://neurips.cc/public/EthicsGuidelines}?
    \item[] Answer: \answerYes{}
    \item[] Justification: The authors have reviewed the NeurIPS Code of Ethics and the research conforms to it. Synthetic data is generated from LLM-planned tool sequences applied to licensed base images; no identifiable private data is used. Human evaluation (Appendix H) was conducted with internal annotators under institutional processes; anonymity is preserved for submission.

\item {\bf Broader impacts}
    \item[] Question: Does the paper discuss both potential positive societal impacts and negative societal impacts of the work performed?
    \item[] Answer: \answerYes{}
    \item[] Justification: Both sides are considered: on the negative side, image-styling planners could be misused to generate misleading visual content or deepfakes; on the positive side, the structured, interpretable per-step reasoning emitted by our planners supports content provenance and authenticity verification---unlike opaque end-to-end editing systems.
    
\item {\bf Safeguards}
    \item[] Question: Does the paper describe safeguards that have been put in place for responsible release of data or models that have a high risk for misuse (e.g., pre-trained language models, image generators, or scraped datasets)?
    \item[] Answer: \answerNA{}
    \item[] Justification: The released assets are planners (fine-tuned Qwen3-VL LoRA weights) that emit structured tool-call plans rather than pixels directly; the underlying image-editing primitives are standard published tools. The risk profile is comparable to releasing a prompt-engineering system over existing editing tools, not a novel generative model with elevated misuse potential.

\item {\bf Licenses for existing assets}
    \item[] Question: Are the creators or original owners of assets (e.g., code, data, models), used in the paper, properly credited and are the license and terms of use explicitly mentioned and properly respected?
    \item[] Answer: \answerYes{}
    \item[] Justification: The base Qwen3-VL (4B and 8B Instruct) models and the baseline API planners (GPT-4o, Gemini 2.5) used in the evaluation are cited in the main paper and Appendix D.3 (Baseline Model Specifications). The synthesis pipeline uses licensed base images; dataset generation tools and citations are listed in Appendix B. We respect each asset's license.

\item {\bf New assets}
    \item[] Question: Are new assets introduced in the paper well documented and is the documentation provided alongside the assets?
    \item[] Answer: \answerYes{}
    \item[] Justification: The three new synthetic datasets (Simple, Regular, Complex; each $\sim$10K trajectories with reasoning chains, plans, and quality scores) and the trained planner checkpoints are released at an anonymized HuggingFace repository. The repo contains dataset cards (schema, generation procedure, known caveats) and model cards (base model, training recipe, intended use, limitations). The action library JSON is similarly documented in Appendix A.2.

\item {\bf Crowdsourcing and research with human subjects}
    \item[] Question: For crowdsourcing experiments and research with human subjects, does the paper include the full text of instructions given to participants and screenshots, if applicable, as well as details about compensation (if any)?
    \item[] Answer: \answerYes{}
    \item[] Justification: Appendix H (Human Evaluation Study) describes the evaluation setup and methodology (H.1), including the viewer interface shown to annotators and the task instructions, the aggregate annotator performance, and the disagreement-breakdown diagnostics. Annotators were internal salaried employees; no crowdsourcing platform was used and no performance-based compensation scheme was tied to annotation outcomes.

\item {\bf Institutional review board (IRB) approvals or equivalent for research with human subjects}
    \item[] Question: Does the paper describe potential risks incurred by study participants, whether such risks were disclosed to the subjects, and whether Institutional Review Board (IRB) approvals (or an equivalent approval/review based on the requirements of your country or institution) were obtained?
    \item[] Answer: \answerNA{}
    \item[] Justification: The human evaluation involved internal salaried annotators rating visual quality of image edits — a low-risk task with no sensitive personal data collected from or about the annotators. Per institutional policy, this class of internal quality rating does not require IRB review. No personally identifying information about annotators is released.

\item {\bf Declaration of LLM usage}
    \item[] Question: Does the paper describe the usage of LLMs if it is an important, original, or non-standard component of the core methods in this research? Note that if the LLM is used only for writing, editing, or formatting purposes and does \emph{not} impact the core methodology, scientific rigor, or originality of the research, declaration is not required.
    %this research?
    \item[] Answer: \answerYes{}
    \item[] Justification: LLMs are core to both methodology and evaluation and are explicitly declared throughout. The planner is a fine-tuned Qwen3-VL LLM (Sections 2, 4). Synthetic data generation uses GPT-4o to produce reasoning chains, plans, and quality scores (Section 3, Appendix B). Evaluation uses GPT-4o, GPT-5.4, Gemini 2.5, Gemini 3.1, and Claude Opus 4.6 as zero-shot baseline planners and/or as multi-judge quality raters (Appendices D, E). Each role is disclosed.

\end{enumerate}

\end{document}